\pdfoutput=1
\documentclass[11pt]{article}

\usepackage[preprint]{acl}

\usepackage{times}
\usepackage{latexsym}
\usepackage[dvipsnames]{xcolor}
\usepackage{enumitem}

\usepackage{amsfonts, amsmath, amssymb, amsthm}
\usepackage{dsfont}

\usepackage[T1]{fontenc}
\usepackage[utf8]{inputenc}

\usepackage{microtype}

\usepackage{inconsolata}

\usepackage{graphicx}

\usepackage{booktabs, multirow, multicol}

\usepackage[capitalize,noabbrev]{cleveref}

\theoremstyle{plain}
\newtheorem{theorem}{Theorem}[section]
\newtheorem{proposition}[theorem]{Proposition}

\theoremstyle{definition}
\newtheorem{definition}[theorem]{Definition}

\theoremstyle{remark}
\newtheorem{remark}[theorem]{Remark}

\usepackage{amsmath,amsfonts,bm}

\def\eqref#1{equation~\ref{#1}}
\def\1{\bm{1}}

\def\vtheta{{\bm{\theta}}}

\def\vd{{\bm{d}}}

\def\vu{{\bm{u}}}

\def\vx{{\bm{x}}}

\DeclareMathAlphabet{\mathsfit}{\encodingdefault}{\sfdefault}{m}{sl}
\SetMathAlphabet{\mathsfit}{bold}{\encodingdefault}{\sfdefault}{bx}{n}

\def\vtheta{\bm{\theta}}
\def\vxigma{\bm{\sigma}}

\def\vell{\bm{\ell}}

\title{Investigating the Multilingual Calibration Effects of Language Model Instruction-Tuning}

\author{
    \textbf{Jerry Huang}$^{\clubsuit\spadesuit\heartsuit}$\thanks{Work done as a visiting student at the University of Tokyo.}\, \textbf{Peng Lu}$^{\spadesuit}$\, \textbf{Qiuhao Zeng}$^{\diamondsuit\nabla}$\\ 
    \textbf{Yusuke Iwasawa}$^{\heartsuit}$\, \textbf{Yutaka Matsuo}$^{\heartsuit}$ \,
    \textbf{Sarath Chandar}$^{\clubsuit\maltese\blacksquare}$ \\
    \textbf{Edison Marrese-Taylor}$^{\heartsuit\bigstar}$\thanks{Equal supervision.} \, 
    \textbf{Irene Li}$^{\heartsuit\dagger}$\footnotemark[2] 
    \\
        $^\clubsuit$Mila - Quebec AI Institute \ $^\spadesuit$Universit\'{e} de Montr\'{e}al \ $^\heartsuit$The University of Tokyo \
        $^\diamondsuit$Western University \
    \\
        $^\nabla$Vector Institute \
        $^\maltese$Polytechnique Montr\'{e}al \ 
        $^\blacksquare$CIFAR AI Chair \
        $^\bigstar$AIST \ 
    \\
    \small{
       \textbf{$^\dagger$Corresponding Author:} \href{mailto:irene.li@weblab.t.u-tokyo.ac.jp}{\texttt{irene.li@weblab.t.u-tokyo.ac.jp}}
    }
}

\begin{document}
\maketitle
\begin{abstract}
Ensuring that deep learning models are well-calibrated in terms of their predictive uncertainty is essential in maintaining their trustworthiness and reliability, yet despite increasing advances in foundation model research, the relationship between such large language models (LLMs) and their calibration remains an open area of research. In this work, we look at a critical gap in the calibration of LLMs within multilingual settings, in an attempt to better understand how the data scarcity can potentially lead to different calibration effects and how commonly used techniques can apply in these settings. Our analysis on two multilingual benchmarks, over 29 and 42 languages respectively, reveals that even in low-resource languages, model confidence can increase significantly after instruction-tuning on high-resource language SFT datasets. However, improvements in accuracy are marginal or non-existent, resulting in mis-calibration, highlighting a critical shortcoming of standard SFT for multilingual languages. Furthermore, we observe that the use of label smoothing to be a reasonable method alleviate this concern, again without any need for low-resource SFT data, maintaining better calibration across all languages. Overall, this highlights the importance of multilingual considerations for both training and tuning LLMs in order to improve their reliability and fairness in downstream use.
% This document is a supplement to the general instructions for *ACL authors. It contains instructions for using the \LaTeX{} style files for ACL conferences.
% The document itself conforms to its own specifications, and is therefore an example of what your manuscript should look like.
% These instructions should be used both for papers submitted for review and for final versions of accepted papers.
\end{abstract}

\section{Introduction}

Tremendous progress has been made in building models that follow natural language instructions~\citep{it_human_feedback, prompting_zsg, it_scaling} through the use of LLMs pre-trained on large amounts of data as well as high-quality datasets that enable them to learn to interact in a human-like manner~\citep{promptsource, supernli, tulu}. However, such models have demonstrated a propensity for over-confidence in their predictions~\citep{contextual_calibration, calibration_qa, llms_confidence}, eliciting concerns over their use in more high-stakes decision-making scenarios. Such observations are not new with respect to neural networks, which have consistently been shown to suffer from over-confident predictions and over-estimate the likelihood of their correctness~\citep{Calibration_NN, inception, when_ls_work, ECE, revisiting_calibration}. To improve this, methods such as temperature scaling~\citep{Calibration_NN} and label smoothing~\citep{when_ls_work} have been proposed as solutions with varying effectiveness, spurring additional work in ensuring that predictions and confidence remain matching~\citep{focal_loss, focal_loss_calibration, penalizing_confidence, margin_based_label_smoothing}.

Investigations into the calibration of LLMs remains limited~\citep{contextual_calibration}, but studies have shown them to behave similarly to prototypical networks in terms of calibration~\citep{huang2025calibrated}. However, LLMs remain unique in their application, one in particular is their ability to be used on different languages that can vary resource scarcity. This raises an interesting question: 
\begin{center}
    \textit{
        Do solutions to calibrating LLMs act on multiple languages simultaneously?
    }
\end{center}
In this work, we provide a preliminary empirical answer to this question. Our work and contributions are summarized as follows:
\begin{enumerate}[leftmargin=15pt,parsep=3pt,itemsep=0pt,label=\emph{\alph*)}]
    \item We evaluate a series of evaluations on multiple multilingual classification tasks, evaluating whether or not the downstream calibration of LLMs remains unchanged between languages. % We evaluate on a set of questions translated by experts into various languages, testing how the calibration of LLMs across these questions can vary between languages. 
    \item Interestingly, all languages exhibit an interesting characteristic: instruction-tuned models are less calibrated than their base counterparts. Yet an additional observation comes on those not used for instruction-tuning, where confidence increases without improvements in accuracy, leading to greater overconfidence.
    \item Using label smoothing only on the instruction-tuning data, we observe both better calibration and minimal drops in accuracy compared to models not using smoothing.
\end{enumerate}
Overall, our work demonstrates a simple yet effective solution for ensuring better calibration for multilingual language models, potentially opening a path towards better methods for ensuring their robustness and reliability for downstream use.

\section{Related Work}

\paragraph{Uncertainty Calibration.} Uncertainty calibration~\citep{statistical_calibration, murphy1972, rss1983} attempts to match the prediction probabilities yielded for different inputs to the expected accuracy on these inputs. Widely used metrics for measuring such properties include the expected calibration error (ECE)~\citep{ECE}. Additional metrics that have been proposed include the Root Mean Square~\citep{RMS_CE} and Static/Adaptive Calibration Errors (SCE/ACE)~\citep{SCE_ACE}, offering complementary perspectives that enable a more comprehensive assessment of uncertainty alignment. 

Nevertheless, calibration of LLMs remains underexplored, particularly from a statistical perspective. \citet{contextual_calibration} first show a general lack of calibration in models, with a simple solution being to prompt models using additional content-free in-context samples (i.e. examples with input "N/A") and calibration parameters. \citet{huang2025calibrated} meanwhile show that instruction-tuning itself can lead to a significant loss in calibration from a base model, with a proposed solution being the use of label smoothing during training.

\paragraph{Multilingual LLMs.} Multilingual NLP have greatly evolved beyond the initial English-centric paradigm to address the linguistic diversity of our world. Recent LLMs~\citep{llama3, qwen2.5, openai_gpt-4_2023, gemma2} have demonstrated remarkable multilingual capabilities by leveraging massive pre-training datasets spanning dozens to hundreds of languages. However, research indicates persistent challenges in these systems, particularly cases where models internally process non-English inputs through English-like representations, and consistent performance gaps between high-resource and low-resource languages~\citep{englishcentric}.

\section{Methodology}\label{sec:setup}

\paragraph{Models and Setup.} To investigate our hypothesis, we train and test the \texttt{Mistral-7B}~\citep{mistral}, \texttt{Llama3.1-8B}~\citep{llama3} and \texttt{Gemma2-2B}~\citep{gemma2} models. We tune on the \texttt{Alpaca}~\citep{alpaca}, \texttt{Tulu3Mixture}~\citep{tulu} and \texttt{OpenHermes}~\citep{OpenHermes} datasets, using recommendations from \citet{tulu}. We employ the AdamW optimizer using a grid search over learning rates $\mathtt{\{5e}$-$\mathtt{6, 1e}$-$\mathtt{5, 5e}$-$\mathtt{5, 1e}$-$\mathtt{4\}}$, linear warm-up over the first $\mathsf{2}$\% of training, a batch size of 128 and dropout $\mathsf{0.1}$. % to facilitate stable training and prevent over-fitting.

\paragraph{Tasks.} To evaluate calibration fairly across multiple languages, we use the \textbf{\texttt{MMLU-ProX}} dataset~\citep{mmluprox} and \textbf{\texttt{GlobalMMLU}}~\citep{globalmmlu}, two comprehensive benchmark covering diverse sets of languages (29 and 42 respectively), built upon the English-only MMLU dataset~\citep{mmlu}. Each language version consists of 12\textsf{k} and 14\textsf{k} identical questions, enabling more direct cross-linguistic comparisons. For each question, an input is given and the model must select between $K$ different candidate answers (up to 10 for \textbf{\texttt{MMLU-ProX}} and 4 for \textbf{\texttt{GlobalMMLU}}). More specifically, the perplexities over candidate generations is computed and normalized to form a probability distribution over the different options, which is then used for classification.

\begin{figure}[ht!]
    \centering
    \resizebox{0.85\linewidth}{!}{
        \includegraphics{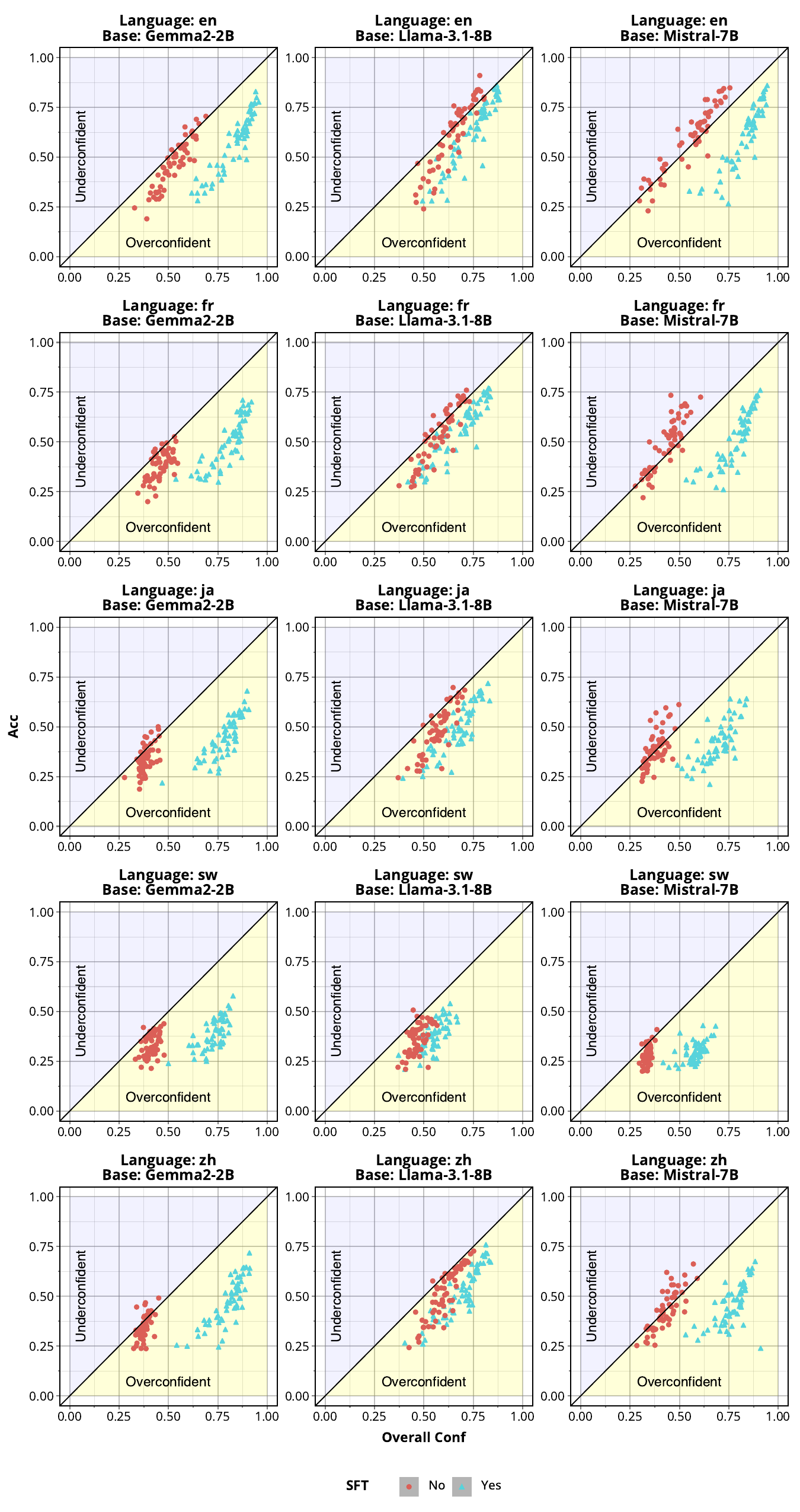}
    }
    \caption{Comparison of \textbf{\texttt{Base}} models (\textcolor{red}{red}) and instruction-tuned (\textcolor{Turquoise}{blue}) models on various languages on \textbf{\texttt{GlobalMMLU}}. Deviation from the straight line indicates under or over-confidence; models show increasing overconfidence across all languages.}
    \label{fig:globalmmlu-base}
    \vspace{-\baselineskip}
\end{figure}

\section{Results}

\subsection{Calibration Across Languages}

The first question we attempt to answer is whether or not LLM calibration is only relevant for the languages that appear in LLM training data. For this purpose, we conduct an initial validation by using a number of open models and their instruction-tuned variants and evaluating on a number of languages. This is illustrated in \cref{fig:globalmmlu-base}. Interestingly, we observe a consistent phenomenon across all languages: instruction-tuning leads to worsening calibration across all languages, even those which are unlikely to appear in the tuning datasets.

\subsection{Accounting For Over-Confidence}\label{sec:ls}

With over-confidence becoming an issue after tuning, this enables us to pose a further question: \textit{How can we reduce the risk of over-confidence without performance degradations?} To this end, we can look towards label smoothing as a solution.

\paragraph{Label smoothing} (LS) has previously been demonstrated to be a promising paradigm in settings to prevent models from becoming over-confident~\citep{inception, when_ls_work} or when noise exists in the provided labels~\citep{smoothing_noisy_labels, smoothing_noisy_labels2, LABO}. Consider a model parameterized by $\vtheta$ to output a conditional distribution $P(\cdot|\vx;\vtheta)$ over a label set. 
Models are usually trained by minimizing a cross-entropy (CE) loss on a dataset $\mathcal{D}=\{\vx_n, y_n\}_{n=1}^N$ sampled from an unknown distribution $p(\vx, y)$,
% \begin{equation}
    \begin{align*}
        \mathcal{L}^{\text{CE}}_{\mathcal{D}}(\vtheta)
        &=-\frac{1}{N}\sum_{i=1}^N\sum_{k=1}^{K}\delta_{y_n}^{\gamma_k}\log P(\gamma_k|\vx;\vtheta) \\
        &\approx -{\mathbb{E}}_{p(\vx, y)}\left[\sum_{k=1}^Kp(\gamma_k|\vx)\log P(\gamma_k|\vx;\vtheta)\right] \\
        &= -{\mathbb{E}}_{p(\vx, y)}[\mathrm{KL}\left[\vxigma(\vx)\|\widehat{\vxigma}(\vx;\vtheta)]\right] + c \\
        &= \mathcal{L}^{\text{CE}}_{p(\vx, y)}(\bm{\theta)},
    \end{align*}
% \end{equation}
where $\delta_i^j$ is the Kronecker delta and $\widehat{\vxigma}(\vx;\vtheta)\in[0, 1]^K$ is the output distribution. Label smoothing mixes the original distribution with a discrete uniform distribution $\mathcal{U} = \left[1/K\right]^K\in \mathbb{R}^K$ using a smoothing rate $\beta\in \left[0, 1\right]$. Thus, label smoothing can be understood to regularize towards a uniform distribution over the output labels, preventing over-fitting by encouraging a higher entropy and can therefore become particularly effective for countering overconfidence. This claim is illustrated in \cref{sec:ls-proof}. Furthermore, this can be viewed as a constrained optimization problem that encourages equality among the logits for each class to ensure that over-confidence is penalized (\cref{sec:constraint_proof}).

\begin{table*}[ht!]
    \caption{Result of instruction-tuning various LLMs with and without label smoothing. Using smoothing can reduce calibration error (ECE and RMS) for all model/dataset combinations, with minimal decrease in accuracy. 
    % Furthermore, smoothing maintains a higher entropy, indicating that the model retains in general greater uncertainty over its predictive distribution. 
    The better performing setting is \textbf{bolded} and further \textbf{\textit{italicized}} when the gap is statistically significant (using a student's $t$-test).}
    \label{tab:ls}
    \centering
    \resizebox{\linewidth}{!}{
        \begin{tabular}{ccc|cccc|cccc}
        \toprule
        \multirow{2}{*}{\textbf{Model}}& \multirow{2}{*}{\textbf{SFT Dataset}} & \multirow{2}{*}{\textbf{Smoothing}} & \multicolumn{4}{c|}{\textbf{\texttt{MMLU-ProX}}} & \multicolumn{4}{c}{\textbf{\texttt{GlobalMMLU}}} \\
        & & & \textbf{Accuracy} & \textbf{Entropy} & \textbf{ECE} & \textbf{RMS} & \textbf{Accuracy} & \textbf{Entropy} & \textbf{ECE} & \textbf{RMS} \\
        \midrule
        \multirow{5.5}{*}{\textbf{\texttt{Gemma2-2B}}}  
            & \multicolumn{2}{c|}{\textit{\texttt{Base Model}}} & 0.137 & 1.361 & 0.040 & 0.064 & 0.326 & 1.787 & 0.051 & 0.057 \\
            \cmidrule{2-11}
            & \multirow{2}{*}{\textbf{\texttt{OpenHermes}}} & 0.0 & \textbf{0.253} & 0.945 & 0.171 & 0.126 & \textbf{0.372} & 0.949 & 0.317 & 0.071 \\
            & & 0.1 & \textbf{0.253} & 1.324 & \textbf{\textit{0.087}} & \textbf{\textit{0.099}} & 0.370 & 1.673 & \textbf{\textit{0.044}} & \textbf{\textit{0.054}} \\
            \cmidrule{2-11}
            & \multirow{2}{*}{\textbf{\texttt{Tulu3Mixture}}} & 0.0 & \textbf{0.223} & 1.185 & 0.100 & 0.097 & 0.359 & 1.247 & 0.185 & 0.064 \\
            & & 0.1 & 0.222 & 1.246 & \textbf{\textit{0.059}} & \textbf{\textit{0.087}} & \textbf{0.360} & 1.717 & \textbf{\textit{0.043}} & \textbf{\textit{0.049}} \\
        \midrule
        \multirow{5.5}{*}{\textbf{\texttt{Mistral-7B}}} 
            & \multicolumn{2}{c|}{\textit{\texttt{Base Model}}} & 0.238 & 1.498 & 0.150 & 0.118 & 0.354 & 1.855 & 0.041 & 0.041 \\
            \cmidrule{2-11}
            & \multirow{2}{*}{\textbf{\texttt{OpenHermes}}} & 0.0 & \textbf{0.275} & 1.275 & 0.132 & 0.116 & 0.372 & 1.561 & 0.053 & 0.036 \\
            & & 0.1 & 0.273 & 1.360 & \textbf{\textit{0.113}} & \textbf{0.113} & \textbf{0.374} & 1.699 & \textbf{\textit{0.043}} & \textbf{\textit{0.030}} \\
            \cmidrule{2-11}
            & \multirow{2}{*}{\textbf{\texttt{Tulu3Mixture}}} & 0.0 & 0.312 & 1.160 & 0.129 & 0.123 & 0.391 & 1.467 & 0.066 & 0.041 \\
            & & 0.1 & \textbf{\textit{0.319}} & 1.297 & \textbf{\textit{0.107}} & \textbf{\textit{0.117}} & \textbf{0.394} & 1.568 & \textbf{\textit{0.046}} & \textbf{\textit{0.035}} \\
        \midrule
        \multirow{5.5}{*}{\textbf{\texttt{LLama-3.1-8B}}} & \multicolumn{2}{c|}{\textbf{\texttt{Base Model}}} & 0.192 & 0.391 & 0.143 & 0.119 & 0.431 & 1.593 & 0.023 & 0.031 \\
            \cmidrule{2-11}
            & \multirow{2}{*}{\textbf{\texttt{OpenHermes}}} & 0.0 & 0.259 & 0.435 & 0.159 & 0.131 & \textbf{0.439} & 1.345 & 0.071 & 0.041 \\
            & & 0.1 & \textbf{\textit{0.350}} & 0.831 & \textbf{\textit{0.153}} & \textbf{\textit{0.123}} & 0.438 & 1.521 & \textbf{\textit{0.032}} & \textbf{\textit{0.030}} \\
            \cmidrule{2-11}
            & \multirow{2}{*}{\textbf{\texttt{Tulu3Mixture}}} & 0.0 & 0.296 & 0.583 & 0.163 & 0.127 & 0.440 & 1.363 & 0.073 & 0.041 \\
            & & 0.1 & \textbf{\textit{0.315}} & 0.772 & \textbf{\textit{0.148}} & \textbf{0.126} & \textbf{0.441} & 1.489 & \textbf{\textit{0.038}} & \textbf{\textit{0.034}} \\
        \bottomrule
        \end{tabular}
    }
    \vspace{-0.5\baselineskip}
\end{table*}

\paragraph{Can LS help with Multilingual Calibration?} A natural question is whether or not LS is sufficient to help with multilingual calibration as a whole.%, i.e. can it help maintain better confidence calibration without negative impacts on model accuracy. 
\cref{tab:ls} shows that models using smoothing become on average much better calibrated across both tasks, while any decrease in accuracy remains minimal ($\leq$ 0.005 accuracy points). This is further consistent across individual languages (\cref{sec:complete_results}), highlighting the versatility it has for maintaining more robust and reliable language models. However, high amounts of smoothing can also lead to some decreases in accuracy, highlighting a need to properly account for such a hyperparameter, either in a fixed or adaptive manner.

\begin{figure}[h!]
    \centering
    \resizebox{\linewidth}{!}{\includegraphics[width=\linewidth]{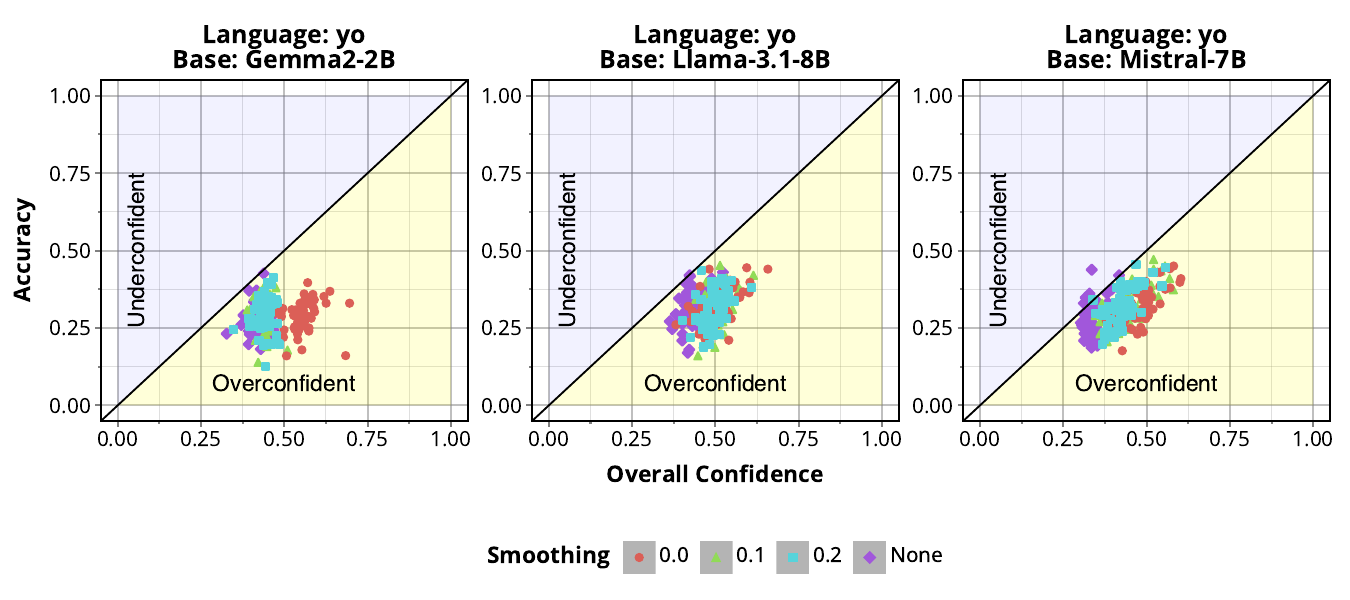}}
    \resizebox{\linewidth}{!}{\includegraphics[width=\linewidth]{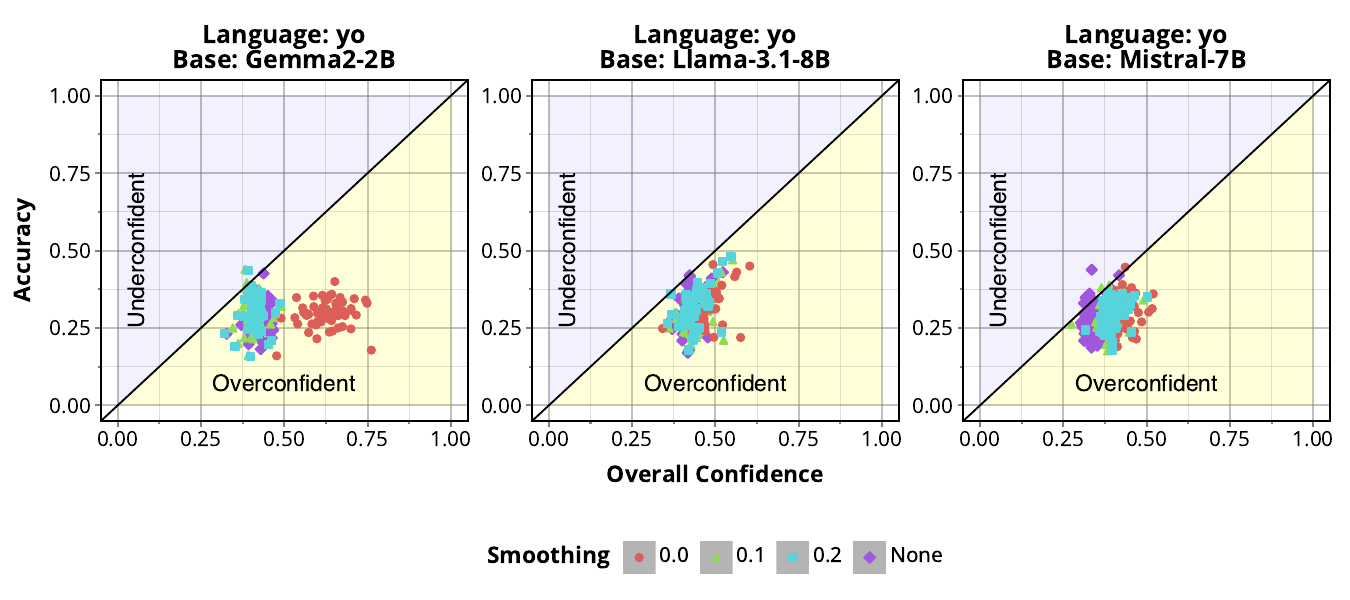}}
    \vspace{-\baselineskip}
    \caption{Reliability diagrams for the \texttt{yo} split of \textbf{\texttt{GlobalMMLU}} after instruction-tuning on \textbf{\texttt{Tulu3Mixture}} (top) and \textbf{\texttt{OpenHermes}} (bottom).}
    \vspace{-\baselineskip}
    \label{fig:globalmmlu-yo}
\end{figure}

\cref{fig:globalmmlu-yo} offers a specific illustration on the Yoruba language, which is not present within tuning data; while accuracy does not change (no vertical shift), confidence increases (horizontal shift towards the right), leading to greater mis-calibration. Label smoothing visibly mitigates this over-confidence without any change in accuracy.

\section{Discussion} 
\paragraph{Why does SFT Lead to Mis-Calibration?} \citet{calibrated_finetuning} offer a perspective from the lens of out-of-distribution (OOD) generalization. To better understand this, the SFT data can be interpreted as consisting of an in-distribution (ID) dataset whereas the downstream dataset on which generalization and calibration are tested constitutes an OOD dataset. Under such a setting, simultaneously maintaining accuracy and calibration of the final classifier (in the case of an auto-regressive LLM, this is the language modeling head) has a direct relationship to the diversity of the feature embeddings. In particular, they show that there exists a dependence of the bound on the minimal singular value of the covariance matrix, indicating that as the set of learnt feature embeddings (the embedding of the prompt in this context) becomes less mutually dependent, both calibration error and classification error can be minimized. Prior works have shown that fine-tuning can significantly reduce the diversity of such features~\citep{finetuning_cripple, finetuning_distort, platonic_representations}, hinting why SFT can significantly degrade calibration.

\paragraph{On Difference Confidence Measures for LLMs.} In additional to statistical properties of models, LLMs have the additional axis of being able to be directly prompted for their internal confidence measures, called \textit{verbalized confidence}~\citep{DBLP:conf/naacl/GengCWKNG24}. However, such approaches often rely on the LLM producing an internalized measure of confidence~\citep{DBLP:conf/acl/TaoYDXC0GSD24}, for example a raw value $c\in[0, 1]$, for which it remains poorly understood whether or not such responses are true measures of the model confidence or whether or not such values can be trusted. overall, this motivates further needs for better understanding confidence/uncertainty calibration in LLMs, both from statistical as well as user-facing perspectives.

\section{Conclusion}

Through controlled tuning experiments with various LLMs, we reveal a surprising finding where LLMs become significantly uncalibrated on low-resource language predictions even without having observed them during this tuning. In particular, models demonstrate an innate tendency to become overconfident, even without improvements in performance, a particular setting which enables the use of label smoothing as a generalized solution. In sum, our research reveals that language models can observe adverse downstream effects in subtle manners unobservable through traditional evaluation, and despite the possibility of solutions for mitigating these concerns, it highlights a need to better consider how datasets are used for training and how to adequately evaluate models. 

\clearpage

\section*{Limitations}

\subsection*{Lack of Explicit Solutions} As an empirical study, our work only investigates the existence of different calibration phenomena depending on the input language. As such, we do not provide a whole-sale solution to calibrating models on any arbitrary language.

\section*{Ethical Considerations}

This paper proposes a method to improve calibration in large-vocabulary language models. We anticipate minimal societal impact or ethical concerns that may stem from the findings of this work.

\section*{Acknowledgments}
Jerry Huang was supported by the Natural Sciences and Engineering Research Council of Canada (NSERC) Canada Graduate Scholarships (reference number 589326) as well as Japan Society for the Promotion of Science (JSPS) Postdoctoral Fellowship for Research in Japan (reference number SP25408). Sarath Chandar is supported by a Canada CIFAR AI Chair, the Canada Research Chair in Lifelong Machine Learning and a NSERC Discovery Grant. The authors thank Yufei Cui for discussions that prompted the initial direction of this work. This work was made possible in part thanks to computational resources from Calcul Qu\'{e}bec\footnote{\url{https://www.calculquebec.ca/}}, the Digital Research Alliance of Canada (DRAC)\footnote{\url{https://alliancecan.ca/en}} and the AI Bridging Cloud Infrastructure (ABCI)\footnote{\url{https://abci.ai/ja/}}.

\nocite{*}
\bibliography{custom}

\appendix

\onecolumn

\section{Extended Related Works}

\subsection{Uncertainty Calibration}

In a $K$-way classification setting, let $\mathcal{X} \in \mathbb{R}^{D}$ and $\mathcal{Y} \in \{\gamma_k\}_{k=1}^K$ indicate the input and label space, respectively. Let $f$ be a classifier and $f\left(\widehat{y}|\vx\right) = \widehat{c}$ be its confidence, i.e., the maximum of probabilities among $K$ dimensions corresponding to its prediction $\widehat{y}$. A model is \textit{perfectly-calibrated} when 
\begin{align}
    P\left(\widehat{y} = y | \widehat{c} = c\right) = c \ \ \forall c \in [0,1].
\end{align}
Model calibration can be expressed as $\mathbb{E}\left[\left|{P}\left(\widehat{y} = y | \widehat{c} = c\right) - c\right|\right]$. 

\paragraph{Expected Calibration Error.} Widely used calibration metrics include the expected calibration error (ECE)~\citep{ECE}, which divides the confidence scores of $N$ samples into $M$ uniform confidence bins $\{B_{m}\}_{m=1}^{M}$ and takes a weighted sum over the bin-wise errors. 
\begin{align}
   \text{ECE} = \sum_{m=1}^{M} \frac{\left|B_m\right|}{N}\left|\mathrm{acc}\left(B_{m}\right)-\mathrm{conf}\left(B_{m}\right)\right|. 
\end{align}

\paragraph{Multiclass \& Static Calibration Error.}
Static Calibration Error (SCE), which is a simple extension of Expected Calibration Error to every probability in the multiclass setting. SCE bins predictions separately for each class probability, computes the calibration error within the bin, and averages across bins:
\begin{align}
    \text{SCE} = \frac{1}{K}\sum_{k=1}^K\sum_{m=1}^{M} \frac{\left|B_{mk}\right|}{N}\left|\mathrm{acc}\left(m, k\right)-\mathrm{conf}\left(m, k\right)\right|. 
\end{align}
Here, $\mathrm{acc}(m, k)$ and $\mathrm{conf}(m, k)$ are the accuracy and confidence of bin $m$ for class label $k$, respectively; $\left|B_{mk}\right|$ is the number of predictions in bin $b$ for class label $k$.

\paragraph{Adaptivity \& Adaptive Calibration Error.} Adaptive calibration ranges are motivated by the bias-variance tradeoff in the choice of ranges, suggesting that in order to get the best estimate of the overall calibration error the metric should focus on the regions where the predictions are made (and focus less on regions with few predictions). This leads to the Adaptive Calibration Error (ACE), which uses an adaptive scheme which spaces the bin intervals so that each contains an equal number of predictions.

In detail, ACE takes as input the predictions $P$ (usually out of a softmax), correct labels, and a number of ranges $R$.
\begin{align}
    \text{ACE} = \frac{1}{K\cdot R}\sum_{k=1}^K\sum_{r=1}^{R} \left|\mathrm{acc}\left(r, k\right)-\mathrm{conf}\left(r, k\right)\right|. 
\end{align}
Here, $\mathrm{acc}(r, k)$ and $\mathrm{conf}(r, k)$ are the accuracy and confidence of adaptive calibration range $r$ for class label $k$. Calibration range $r$ defined by the $\left\lfloor N/R\right\rfloor$-th index of the sorted and thresholded predictions.

\paragraph{RMS and MAD Calibration Error.} The Root Mean Square Calibration Error measures the square root of the expected squared difference between confidence and accuracy at a confidence level. A similar formulation which less severely penalizes large confidence-accuracy deviations is the Mean Absolute Value (MAD), which is a lower bound of the RMS Calibration Error. 

To empirically estimate these mis-calibration measures, the $N$ samples of are partitioned again into $M$ bins. Here, bins are not equally spaced since the distribution of confidence values is not uniform but dynamic. Concretely, the RMS Calibration Error is estimated with the numerically stable formula
\begin{align}
    \text{RMSCE}=\sqrt{\sum_{m=1}^M \frac{\left|B_m\right|}{N} \left( \frac{1}{\left|B_m\right|}\sum_{k\in B_i}\mathds{1}\left(y_k = \widehat{y}_k\right) - \frac{1}{\left|B_m\right|}\sum_{k\in B_m} c_k \right)^2}.
\end{align}
%Estimating the MAD Calibration Error is similar.
Along similar lines, the MAD Calibration Error --- which is an improper scoring rule due to its use of absolute differences rather than squared differences --- is estimated with
\begin{align}
\text{MAD}= \sum_{m=1}^M \frac{\left|B_m\right|}{N} \left| \frac{1}{\left|B_m\right|}\sum_{k\in B_m}\mathds{1}\left(y_k = \widehat{y}_k\right) - \frac{1}{\left|B_m\right|}\sum_{k\in B_m} c_k \right|.
\end{align}

% \section{Additional Experimental Details}

% \paragraph{Training Environment.} All models were trained on a server with 8 NVIDIA H100 80GB GPUs with HBM3.

\section{Proofs}\label{sec:proofs}

\subsection{Proof of Label Smoothing Claim}\label{sec:ls-proof}

\begin{remark}
    Label smoothing can be understood to regularize towards a uniform distribution over the output labels, preventing over-fitting by encouraging a higher entropy and can therefore become particularly effective for countering overconfidence.
\end{remark}

\begin{proof}
Re-using our notation from \cref{sec:ls}, consider a smoothing rate $\beta\in \left[0, 1\right]$ and the cross-entropy loss. If label smoothing is used, the loss then becomes
\begin{equation}\label{eq:ls-loss}
    \begin{split}
        % \mathcal{L}^{\text{LS}}_{\mathcal{D}}(\vtheta;\beta)
        \mathcal{L}^{\text{LS}}_{\mathcal{D}}(\vtheta)
        &=-\frac{1}{N}\sum_{i=1}^N\left[\sum_{k=1}^{K}\left[(1-\beta)\delta_{y_n}^{\gamma_k}+\frac{\beta}{K}\right]\log P(\gamma_k|\vx;\vtheta)\right] \\
        &=(1-\beta)\mathcal{L}^{\text{CE}}_{\mathcal{D}}(\vtheta) + \frac{\beta}{K}\sum_{i=1}^N\mathrm{KL}[\vu\|\widehat{\vxigma}(\vx_n;\vtheta)]+c \\
        &\approx-{\mathbb{E}}_{p(\vx, y)}\left[\mathrm{KL}\left[(1-\beta)\vxigma(\vx)+\beta\vu\|\widehat{\vxigma}(\vx;\vtheta)\right]\right]+c \\
        &= \mathcal{L}^{\text{LS}}_{p(\vx, y)}(\vtheta).
    \end{split}
\end{equation}
\end{proof}

\subsection{Proof of Constraint Claim}\label{sec:constraint_proof}

To better understand the underlying effects of LS, one can repose its effects from a constraint optimization perspective, where the constraints are imposed from regularization penalties~\citep{nonlinear_programming}. First, define
\begin{definition}\label{def:logit_distance}
    The \textbf{logit distance} vector for $\vx$, $\bm{d}(\vx)$, is
    \begin{equation}\label{eq:logit_distance}
        \bm{d}(\vx) = \left[\max_{1\leq i\leq K}\vell(\vx)_i - \vell(\vx)_k\right]_{k=1}^K\in\mathbb{R}^K.
    \end{equation}
\end{definition}
\noindent
One way of ensuring that a model does not over-estimate a specific class is to enforce this as a hard constraint, which results in equal logits among all classes and a $\mathrm{softmax}$ output of $\bm{o}=f(\vx;\vtheta)=[1/K]^K$. As such, it is often preferable to enforce this as a soft-penalty function $\mathcal{P}: \mathbb{R}^K\to\mathbb{R}$ into the objective function minimized during training. Recalling \cref{eq:ls-loss}, we can relate this soft-penalty to the additional KL-divergence introduced by the label smoothing objective.
\begin{proposition}\label{prop:consraint}
    A linear penalty (or a Lagrangian term) for the hard constraint $\bm{d}(\vx) = \bm{0}$ is bounded from above and below by $\mathrm{KL}\left(\vu\|\widehat{\vxigma}\left(\vx;\vtheta\right)\right)$, up to additive constants
    \begin{equation}
        \mathrm{KL}[\vu\|\widehat{\vxigma}\left(\vx;\vtheta\right)]-\log K \leq \sum_{i=1}^K\frac{\bm{d}\left(\vx\right)_i}{K}\leq \mathrm{KL}\left[\vu\|\widehat{\vxigma}\left(\vx;\vtheta\right)\right].
    \end{equation}
\end{proposition}
\noindent
The proof (in \cref{sec:constraint_proof}) indicates that label smoothing approximately minimizes, for a linear penalty, the constraint $\bm{d}(\vx)=\bm{0}$, encouraging equality among the logits for each class to ensure that over-confidence is penalized.

\begin{proof}
Given the KL divergence
\[
\mathrm{KL}\left[\vu\|\widehat{\vxigma}\left(\vx;\vtheta\right)\right] = -\frac{1}{K}\sum_{k=1}^K\log{P\left(\gamma_i|\vx;\vtheta\right)} + \text{const}
\]
we have that
\begin{equation}
    \begin{split}
        \mathrm{KL}\left[\vu\|\widehat{\vxigma}\left(\vx;\vtheta\right)\right] 
        =-\frac{1}{K}\sum_{k=1}^K\log\left(\frac{e^{\vell\left(\vx;\vtheta\right)_i}}{\sum_{j=1}^Ke^{\vell\left(\vx;\vtheta\right)_j}}\right) + c
        =-\frac{1}{K}\sum_{k=1}^K\log\left(\sum_{j=1}^Ke^{\vell\left(\vx;\vtheta\right)_j}-\vell\left(\vx;\vtheta\right)_i\right) + c
    \end{split}
\end{equation}
Considering the property of the LogSumExp (LSE) function, it follows that
\[
\max_j\phantom{0}\vell\left(\vx;\vtheta\right)_j\leq \log\sum_{j=1}^K e^{\vell\left(\vx;\vtheta\right)_j}\leq \max_j\phantom{0}\vell\left(\vx;\vtheta\right)_j+\log\left(K\right)
\]
and
\begin{equation}
    \mathrm{KL}\left[\vu\|\widehat{\vxigma}\left(\vx;\vtheta\right)\right]-\log K 
    \leq -\frac{1}{K}\sum_{k=1}^K \left(\max_j\phantom{0}\vell\left(\vx;\vtheta\right)_j-\vell\left(\vx;\vtheta\right)_k\right) \\
    \leq \mathrm{KL}\left[\vu\|\widehat{\vxigma}\left(\vx;\vtheta\right)\right]
\end{equation}
and given the definition of $\vd\left(\vx\right)$, then the additional penalty $\mathrm{KL}\left[\vu\|\widehat{\vxigma}\left(\vx;\vtheta\right)\right]$ imposed by LS in addition to the standard cross-entropy loss $\mathcal{L}^{\text{CE}}$ is approximately optimizing a linear penalty (or a Lagrangian) for the constraint
\[\vd\left(\vx\right)=\bm{0}\] to encourage equality of the logits.
\end{proof}

\clearpage
\section{Complete Results}
\label{sec:complete_results}

\begin{table*}[h!]
    \caption{Result of instruction-tuning various LLMs with and without label smoothing. We observe that using smoothing can reduce calibration error (ECE and RMS) for all model/dataset combinations, with minimal decrease in accuracy. Furthermore, smoothing maintains a higher entropy, indicating that the model retains in general greater uncertainty over its predictive distribution.}
    \centering
    \resizebox{\linewidth}{!}{
\begin{tabular}{ccc|cccc|cccc}
        \toprule
        \multirow{2}{*}{\textbf{Model}}& \multirow{2}{*}{\textbf{SFT Dataset}} & \multirow{2}{*}{\textbf{Smoothing}} & \multicolumn{4}{c|}{\textbf{\texttt{MMLU-ProX}}} & \multicolumn{4}{c}{\textbf{\texttt{GlobalMMLU}}} \\
        & & & \textbf{Accuracy} & \textbf{Entropy} & \textbf{ECE} & \textbf{RMS} & \textbf{Accuracy} & \textbf{Entropy} & \textbf{ECE} & \textbf{RMS} \\
        \midrule
        \midrule
        \textbf{\texttt{Gemma2-2B}} & \multicolumn{2}{c|}{\multirow{3}{*}{\textbf{\texttt{Base Model}}}} & 0.137 & 1.361 & 0.040 & 0.064 & 0.326 & 1.787 & 0.051 & 0.057 \\
        \textbf{\texttt{Mistral-7B}} & & & 0.238 & 1.498 & 0.150 & 0.118 & 0.354 & 1.855 & 0.041 & 0.041 \\
        \textbf{\texttt{Llama-3.1-8B}} & & & 0.192 & 0.391 & 0.143 & 0.119 & 0.431 & 1.593 & 0.023 & 0.031 \\
        \midrule
        \multirow{6.75}{*}{\textbf{\texttt{Gemma2-2B}}} 
            & \multirow{2}{*}{\textbf{\texttt{Alpaca}}} & 0.0 & 0.194 & 0.878 & 0.102 & 0.109 & 0.370 & 1.111 & 0.242 & 0.068 \\
            & & 0.1 & 0.209 & 0.956 & 0.088 & 0.103 & 0.368 & 1.302 & 0.170 & 0.057 \\
            \cmidrule{2-11}
            & \multirow{2}{*}{\textbf{\texttt{OpenHermes}}} & 0.0 & 0.253 & 0.945 & 0.171 & 0.126 & 0.372 & 0.949 & 0.317 & 0.071 \\
\bottomrule
        \end{tabular}
    }
    \label{tab:ls-full-agg}
\end{table*}

\newpage
\twocolumn
\subsection{Individual Languages}

The following tables contain the individual language results on both \texttt{GlobalMMLU} and \texttt{MMLU-ProX} datasets.

\subsubsection{GlobalMMLU}

\begin{table}[h!]
\caption{Results on the \texttt{GlobalMMLU} subset for the \texttt{am} language by model, SFT dataset, and label smoothing.}
\label{tab:ls-mmluprox-ja}
\resizebox{\linewidth}{!}{\begin{tabular}{llc|cccc}
\toprule
\textbf{SFT Dataset} & \textbf{Base} & \textbf{Smoothing} & \textbf{Accuracy} & \textbf{Entropy} & \textbf{ECE} & \textbf{RMS} \\
\midrule
\multirow{3}{*}{\textbf{\texttt{Base Model}}} & \textbf{\texttt{Gemma2-2B}} & None & 0.268 & 1.886 & 0.071 & 0.091 \\
& \textbf{\texttt{Mistral-7B}} & None & 0.256 & 1.956 & 0.072 & 0.087 \\
& \textbf{\texttt{Llama-3.1-8B}} & None & 0.292 & 1.784 & 0.055 & 0.085 \\
\midrule \multirow{9.75}{*}{\textbf{\texttt{Alpaca}}} & \multirow{3}{*}{\textbf{\texttt{Gemma2-2B}}} & 0.0 & 0.287 & 1.484 & 0.138 & 0.145 \\
& & 0.1 & 0.284 & 1.538 & 0.120 & 0.138 \\
& & 0.2 & 0.285 & 1.607 & 0.093 & 0.128 \\
\cmidrule{2-7} & \multirow{3}{*}{\textbf{\texttt{Mistral-7B}}} & 0.0 & 0.249 & 1.547 & 0.144 & 0.149 \\
\bottomrule
\end{tabular}}
\end{table}

\begin{table}[h!]\caption{Results on the \texttt{GlobalMMLU} subset for the \texttt{ar} language by model, SFT dataset, and label smoothing.}
\resizebox{\linewidth}{!}{\begin{tabular}{llc|cccc}
\toprule
\textbf{SFT Dataset} & \textbf{Base} & \textbf{Smoothing} & \textbf{Accuracy} & \textbf{Entropy} & \textbf{ECE} & \textbf{RMS} \\
\midrule
\multirow{3}{*}{\textbf{\texttt{Base Model}}} & \textbf{\texttt{Gemma2-2B}} & None & 0.329 & 1.828 & 0.055 & 0.080 \\
& \textbf{\texttt{Mistral-7B}} & None & 0.305 & 1.907 & 0.060 & 0.083 \\
& \textbf{\texttt{Llama-3.1-8B}} & None & 0.453 & 1.604 & 0.025 & 0.067 \\
\midrule \multirow{9.75}{*}{\textbf{\texttt{Alpaca}}} & \multirow{3}{*}{\textbf{\texttt{Gemma2-2B}}} & 0.0 & 0.371 & 1.069 & 0.288 & 0.172 \\
& & 0.1 & 0.367 & 1.274 & 0.207 & 0.149 \\
& & 0.2 & 0.369 & 1.302 & 0.196 & 0.146 \\
\cmidrule{2-7} & \multirow{3}{*}{\textbf{\texttt{Mistral-7B}}} & 0.0 & 0.305 & 1.205 & 0.331 & 0.185 \\
\bottomrule
\end{tabular}}
\end{table}

\begin{table}[h!]\caption{Results on the \texttt{GlobalMMLU} subset for the \texttt{bn} language by model, SFT dataset, and label smoothing.}
\resizebox{\linewidth}{!}{\begin{tabular}{llc|cccc}
\toprule
\textbf{SFT Dataset} & \textbf{Base} & \textbf{Smoothing} & \textbf{Accuracy} & \textbf{Entropy} & \textbf{ECE} & \textbf{RMS} \\
\midrule
\multirow{3}{*}{\textbf{\texttt{Base Model}}} & \textbf{\texttt{Gemma2-2B}} & None & 0.301 & 1.824 & 0.069 & 0.087 \\
& \textbf{\texttt{Mistral-7B}} & None & 0.277 & 1.923 & 0.070 & 0.087 \\
& \textbf{\texttt{Llama-3.1-8B}} & None & 0.375 & 1.690 & 0.027 & 0.074 \\
\midrule \multirow{9.75}{*}{\textbf{\texttt{Alpaca}}} & \multirow{3}{*}{\textbf{\texttt{Gemma2-2B}}} & 0.0 & 0.331 & 1.210 & 0.227 & 0.160 \\
& & 0.1 & 0.335 & 1.413 & 0.144 & 0.140 \\
& & 0.2 & 0.329 & 1.439 & 0.130 & 0.137 \\
\cmidrule{2-7}
& \multirow{3}{*}{\textbf{\texttt{Mistral-7B}}} & 0.0 & 0.278 & 1.210 & 0.310 & 0.187 \\
\bottomrule
\end{tabular}}
\end{table}

\begin{table}[h!]\caption{Results on the \texttt{GlobalMMLU} subset for the \texttt{cs} language by model, SFT dataset, and label smoothing.}
\resizebox{\linewidth}{!}{\begin{tabular}{llc|cccc}
\toprule
\textbf{SFT Dataset} & \textbf{Base} & \textbf{Smoothing} & \textbf{Accuracy} & \textbf{Entropy} & \textbf{ECE} & \textbf{RMS} \\
\midrule
\multirow{3}{*}{\textbf{\texttt{Base Model}}} & \textbf{\texttt{Gemma2-2B}} & None & 0.353 & 1.788 & 0.039 & 0.076 \\
& \textbf{\texttt{Mistral-7B}} & None & 0.425 & 1.817 & 0.027 & 0.071 \\
& \textbf{\texttt{Llama-3.1-8B}} & None & 0.479 & 1.523 & 0.024 & 0.065 \\
\midrule \multirow{9.75}{*}{\textbf{\texttt{Alpaca}}} & \multirow{3}{*}{\textbf{\texttt{Gemma2-2B}}} & 0.0 & 0.393 & 1.015 & 0.253 & 0.162 \\
& & 0.1 & 0.393 & 1.260 & 0.163 & 0.138 \\
& & 0.2 & 0.394 & 1.286 & 0.154 & 0.136 \\
\cmidrule{2-7} & \multirow{3}{*}{\textbf{\texttt{Mistral-7B}}} & 0.0 & 0.419 & 1.097 & 0.196 & 0.143 \\
\bottomrule
\end{tabular}}
\end{table}

\begin{table}[h!]\caption{Results on the \texttt{GlobalMMLU} subset for the \texttt{de} language by model, SFT dataset, and label smoothing.}
\resizebox{\linewidth}{!}{\begin{tabular}{llc|cccc}
\toprule
\textbf{SFT Dataset} & \textbf{Base} & \textbf{Smoothing} & \textbf{Accuracy} & \textbf{Entropy} & \textbf{ECE} & \textbf{RMS} \\
\midrule
\multirow{3}{*}{\textbf{\texttt{Base Model}}} & \textbf{\texttt{Gemma2-2B}} & None & 0.367 & 1.744 & 0.044 & 0.078 \\
& \textbf{\texttt{Mistral-7B}} & None & 0.445 & 1.773 & 0.022 & 0.064 \\
& \textbf{\texttt{Llama-3.1-8B}} & None & 0.515 & 1.440 & 0.022 & 0.067 \\
\midrule \multirow{9.75}{*}{\textbf{\texttt{Alpaca}}} & \multirow{3}{*}{\textbf{\texttt{Gemma2-2B}}} & 0.0 & 0.428 & 0.884 & 0.264 & 0.163 \\
& & 0.1 & 0.426 & 1.112 & 0.204 & 0.146 \\
& & 0.2 & 0.425 & 1.138 & 0.204 & 0.143 \\
\cmidrule{2-7} & \multirow{3}{*}{\textbf{\texttt{Mistral-7B}}} & 0.0 & 0.448 & 1.003 & 0.204 & 0.145 \\
\bottomrule
\end{tabular}}
\end{table}

\begin{table}[h!]\caption{Results on the \texttt{GlobalMMLU} subset for the \texttt{el} language by model, SFT dataset, and label smoothing.}
\resizebox{\linewidth}{!}{\begin{tabular}{llc|cccc}
\toprule
\textbf{SFT Dataset} & \textbf{Base} & \textbf{Smoothing} & \textbf{Accuracy} & \textbf{Entropy} & \textbf{ECE} & \textbf{RMS} \\
\midrule
\multirow{3}{*}{\textbf{\texttt{Base Model}}} & \textbf{\texttt{Gemma2-2B}} & None & 0.329 & 1.811 & 0.048 & 0.080 \\
& \textbf{\texttt{Mistral-7B}} & None & 0.305 & 1.936 & 0.034 & 0.069 \\
& \textbf{\texttt{Llama-3.1-8B}} & None & 0.442 & 1.571 & 0.027 & 0.070 \\
\midrule \multirow{9.75}{*}{\textbf{\texttt{Alpaca}}} & \multirow{3}{*}{\textbf{\texttt{Gemma2-2B}}} & 0.0 & 0.372 & 0.989 & 0.307 & 0.174 \\
& & 0.1 & 0.368 & 1.209 & 0.209 & 0.154 \\
& & 0.2 & 0.365 & 1.221 & 0.201 & 0.153 \\
\cmidrule{2-7} & \multirow{3}{*}{\textbf{\texttt{Mistral-7B}}} & 0.0 & 0.306 & 1.224 & 0.291 & 0.182 \\
\bottomrule
\end{tabular}}
\end{table}

\begin{table}[h!]\caption{Results on the \texttt{GlobalMMLU} subset for the \texttt{en} language by model, SFT dataset, and label smoothing.}
\resizebox{\linewidth}{!}{\begin{tabular}{llc|cccc}
\toprule
\textbf{SFT Dataset} & \textbf{Base} & \textbf{Smoothing} & \textbf{Accuracy} & \textbf{Entropy} & \textbf{ECE} & \textbf{RMS} \\
\midrule
\multirow{3}{*}{\textbf{\texttt{Base Model}}} & \textbf{\texttt{Gemma2-2B}} & None & 0.461 & 1.552 & 0.032 & 0.069 \\
& \textbf{\texttt{Mistral-7B}} & None & 0.576 & 1.488 & 0.022 & 0.066 \\
& \textbf{\texttt{Llama-3.1-8B}} & None & 0.603 & 1.260 & 0.020 & 0.065 \\
\midrule \multirow{9.75}{*}{\textbf{\texttt{Alpaca}}} & \multirow{3}{*}{\textbf{\texttt{Gemma2-2B}}} & 0.0 & 0.511 & 0.768 & 0.249 & 0.156 \\
& & 0.1 & 0.510 & 1.100 & 0.139 & 0.125 \\
& & 0.2 & 0.508 & 1.100 & 0.144 & 0.125 \\
\cmidrule{2-7} & \multirow{3}{*}{\textbf{\texttt{Mistral-7B}}} & 0.0 & 0.544 & 0.828 & 0.234 & 0.151 \\
\bottomrule
\end{tabular}}
\end{table}

\begin{table}[h!]\caption{Results on the \texttt{GlobalMMLU} subset for the \texttt{es} language by model, SFT dataset, and label smoothing.}
\resizebox{\linewidth}{!}{\begin{tabular}{llc|cccc}
\toprule
\textbf{SFT Dataset} & \textbf{Base} & \textbf{Smoothing} & \textbf{Accuracy} & \textbf{Entropy} & \textbf{ECE} & \textbf{RMS} \\
\midrule
\multirow{3}{*}{\textbf{\texttt{Base Model}}} & \textbf{\texttt{Gemma2-2B}} & None & 0.381 & 1.723 & 0.042 & 0.077 \\
& \textbf{\texttt{Mistral-7B}} & None & 0.489 & 1.700 & 0.021 & 0.066 \\
& \textbf{\texttt{Llama-3.1-8B}} & None & 0.525 & 1.410 & 0.023 & 0.064 \\
\midrule \multirow{9.75}{*}{\textbf{\texttt{Alpaca}}} & \multirow{3}{*}{\textbf{\texttt{Gemma2-2B}}} & 0.0 & 0.442 & 0.817 & 0.276 & 0.167 \\
& & 0.1 & 0.444 & 1.108 & 0.193 & 0.143 \\
& & 0.2 & 0.442 & 1.140 & 0.181 & 0.140 \\
\cmidrule{2-7} & \multirow{3}{*}{\textbf{\texttt{Mistral-7B}}} & 0.0 & 0.465 & 0.967 & 0.213 & 0.145 \\
\bottomrule
\end{tabular}}
\end{table}

\begin{table}[h!]\caption{Results on the \texttt{GlobalMMLU} subset for the \texttt{fa} language by model, SFT dataset, and label smoothing.}
\resizebox{\linewidth}{!}{\begin{tabular}{llc|cccc}
\toprule
\textbf{SFT Dataset} & \textbf{Base} & \textbf{Smoothing} & \textbf{Accuracy} & \textbf{Entropy} & \textbf{ECE} & \textbf{RMS} \\
\midrule
\multirow{3}{*}{\textbf{\texttt{Base Model}}} & \textbf{\texttt{Gemma2-2B}} & None & 0.320 & 1.851 & 0.049 & 0.078 \\
& \textbf{\texttt{Mistral-7B}} & None & 0.298 & 1.914 & 0.057 & 0.081 \\
& \textbf{\texttt{Llama-3.1-8B}} & None & 0.446 & 1.605 & 0.026 & 0.066 \\
\midrule \multirow{9.75}{*}{\textbf{\texttt{Alpaca}}} & \multirow{3}{*}{\textbf{\texttt{Gemma2-2B}}} & 0.0 & 0.365 & 1.120 & 0.256 & 0.163 \\
& & 0.1 & 0.366 & 1.303 & 0.178 & 0.146 \\
& & 0.2 & 0.364 & 1.334 & 0.173 & 0.143 \\
\cmidrule{2-7} & \multirow{3}{*}{\textbf{\texttt{Mistral-7B}}} & 0.0 & 0.300 & 1.236 & 0.285 & 0.180 \\
\bottomrule
\end{tabular}}
\end{table}

\begin{table}[h!]\caption{Results on the \texttt{GlobalMMLU} subset for the \texttt{fil} language by model, SFT dataset, and label smoothing.}
\resizebox{\linewidth}{!}{\begin{tabular}{llc|cccc}
\toprule
\textbf{SFT Dataset} & \textbf{Base} & \textbf{Smoothing} & \textbf{Accuracy} & \textbf{Entropy} & \textbf{ECE} & \textbf{RMS} \\
\midrule
\multirow{3}{*}{\textbf{\texttt{Base Model}}} & \textbf{\texttt{Gemma2-2B}} & None & 0.373 & 1.735 & 0.046 & 0.079 \\
& \textbf{\texttt{Mistral-7B}} & None & 0.393 & 1.820 & 0.019 & 0.063 \\
& \textbf{\texttt{Llama-3.1-8B}} & None & 0.454 & 1.558 & 0.028 & 0.070 \\
\midrule \multirow{9.75}{*}{\textbf{\texttt{Alpaca}}} & \multirow{3}{*}{\textbf{\texttt{Gemma2-2B}}} & 0.0 & 0.413 & 0.984 & 0.265 & 0.162 \\
& & 0.1 & 0.414 & 1.250 & 0.165 & 0.136 \\
& & 0.2 & 0.412 & 1.279 & 0.148 & 0.134 \\
\cmidrule{2-7} & \multirow{3}{*}{\textbf{\texttt{Mistral-7B}}} & 0.0 & 0.356 & 1.164 & 0.308 & 0.177 \\
\bottomrule
\end{tabular}}
\end{table}

\begin{table}[h!]\caption{Results on the \texttt{GlobalMMLU} subset for the \texttt{fr} language by model, SFT dataset, and label smoothing.}
\resizebox{\linewidth}{!}{\begin{tabular}{llc|cccc}
\toprule
\textbf{SFT Dataset} & \textbf{Base} & \textbf{Smoothing} & \textbf{Accuracy} & \textbf{Entropy} & \textbf{ECE} & \textbf{RMS} \\
\midrule
\multirow{3}{*}{\textbf{\texttt{Base Model}}} & \textbf{\texttt{Gemma2-2B}} & None & 0.376 & 1.728 & 0.043 & 0.077 \\
& \textbf{\texttt{Mistral-7B}} & None & 0.478 & 1.757 & 0.025 & 0.069 \\
& \textbf{\texttt{Llama-3.1-8B}} & None & 0.525 & 1.446 & 0.022 & 0.066 \\
\midrule \multirow{9.75}{*}{\textbf{\texttt{Alpaca}}} & \multirow{3}{*}{\textbf{\texttt{Gemma2-2B}}} & 0.0 & 0.435 & 0.828 & 0.301 & 0.172 \\
& & 0.1 & 0.430 & 1.097 & 0.225 & 0.149 \\
& & 0.2 & 0.433 & 1.136 & 0.192 & 0.144 \\
\cmidrule{2-7} & \multirow{3}{*}{\textbf{\texttt{Mistral-7B}}} & 0.0 & 0.459 & 0.984 & 0.226 & 0.147 \\
\bottomrule
\end{tabular}}
\end{table}

\begin{table}[h!]\caption{Results on the \texttt{GlobalMMLU} subset for the \texttt{ha} language by model, SFT dataset, and label smoothing.}
\resizebox{\linewidth}{!}{\begin{tabular}{llc|cccc}
\toprule
\textbf{SFT Dataset} & \textbf{Base} & \textbf{Smoothing} & \textbf{Accuracy} & \textbf{Entropy} & \textbf{ECE} & \textbf{RMS} \\
\midrule
\multirow{3}{*}{\textbf{\texttt{Base Model}}} & \textbf{\texttt{Gemma2-2B}} & None & 0.288 & 1.817 & 0.050 & 0.083 \\
& \textbf{\texttt{Mistral-7B}} & None & 0.261 & 1.929 & 0.086 & 0.092 \\
& \textbf{\texttt{Llama-3.1-8B}} & None & 0.347 & 1.759 & 0.037 & 0.075 \\
\midrule \multirow{9.75}{*}{\textbf{\texttt{Alpaca}}} & \multirow{3}{*}{\textbf{\texttt{Gemma2-2B}}} & 0.0 & 0.304 & 1.371 & 0.210 & 0.158 \\
& & 0.1 & 0.303 & 1.463 & 0.184 & 0.147 \\
& & 0.2 & 0.299 & 1.518 & 0.155 & 0.145 \\
\cmidrule{2-7} & \multirow{3}{*}{\textbf{\texttt{Mistral-7B}}} & 0.0 & 0.271 & 1.327 & 0.249 & 0.171 \\
\bottomrule
\end{tabular}}
\end{table}

\begin{table}[h!]\caption{Results on the \texttt{GlobalMMLU} subset for the \texttt{he} language by model, SFT dataset, and label smoothing.}
\resizebox{\linewidth}{!}{\begin{tabular}{llc|cccc}
\toprule
\textbf{SFT Dataset} & \textbf{Base} & \textbf{Smoothing} & \textbf{Accuracy} & \textbf{Entropy} & \textbf{ECE} & \textbf{RMS} \\
\midrule
\multirow{3}{*}{\textbf{\texttt{Base Model}}} & \textbf{\texttt{Gemma2-2B}} & None & 0.309 & 1.791 & 0.062 & 0.088 \\
& \textbf{\texttt{Mistral-7B}} & None & 0.280 & 1.935 & 0.066 & 0.086 \\
& \textbf{\texttt{Llama-3.1-8B}} & None & 0.392 & 1.676 & 0.028 & 0.072 \\
\midrule \multirow{9.75}{*}{\textbf{\texttt{Alpaca}}} & \multirow{3}{*}{\textbf{\texttt{Gemma2-2B}}} & 0.0 & 0.352 & 1.044 & 0.308 & 0.177 \\
& & 0.1 & 0.352 & 1.302 & 0.191 & 0.150 \\
& & 0.2 & 0.346 & 1.319 & 0.180 & 0.148 \\
\cmidrule{2-7} & \multirow{3}{*}{\textbf{\texttt{Mistral-7B}}} & 0.0 & 0.281 & 1.122 & 0.397 & 0.200 \\
\bottomrule
\end{tabular}}
\end{table}

\begin{table}[h!]\caption{Results on the \texttt{GlobalMMLU} subset for the \texttt{hi} language by model, SFT dataset, and label smoothing.}
\resizebox{\linewidth}{!}{\begin{tabular}{llc|cccc}
\toprule
\textbf{SFT Dataset} & \textbf{Base} & \textbf{Smoothing} & \textbf{Accuracy} & \textbf{Entropy} & \textbf{ECE} & \textbf{RMS} \\
\midrule
\multirow{3}{*}{\textbf{\texttt{Base Model}}} & \textbf{\texttt{Gemma2-2B}} & None & 0.312 & 1.817 & 0.055 & 0.085 \\
& \textbf{\texttt{Mistral-7B}} & None & 0.287 & 1.912 & 0.067 & 0.084 \\
& \textbf{\texttt{Llama-3.1-8B}} & None & 0.433 & 1.701 & 0.019 & 0.064 \\
\midrule \multirow{9.75}{*}{\textbf{\texttt{Alpaca}}} & \multirow{3}{*}{\textbf{\texttt{Gemma2-2B}}} & 0.0 & 0.363 & 1.036 & 0.304 & 0.175 \\
& & 0.1 & 0.355 & 1.280 & 0.208 & 0.153 \\
& & 0.2 & 0.357 & 1.299 & 0.192 & 0.150 \\
\cmidrule{2-7} & \multirow{3}{*}{\textbf{\texttt{Mistral-7B}}} & 0.0 & 0.290 & 1.183 & 0.329 & 0.188 \\
\bottomrule
\end{tabular}}
\end{table}

\begin{table}[h!]\caption{Results on the \texttt{GlobalMMLU} subset for the \texttt{id} language by model, SFT dataset, and label smoothing.}
\resizebox{\linewidth}{!}{\begin{tabular}{llc|cccc}
\toprule
\textbf{SFT Dataset} & \textbf{Base} & \textbf{Smoothing} & \textbf{Accuracy} & \textbf{Entropy} & \textbf{ECE} & \textbf{RMS} \\
\midrule
\multirow{3}{*}{\textbf{\texttt{Base Model}}} & \textbf{\texttt{Gemma2-2B}} & None & 0.351 & 1.815 & 0.041 & 0.076 \\
& \textbf{\texttt{Mistral-7B}} & None & 0.397 & 1.819 & 0.029 & 0.068 \\
& \textbf{\texttt{Llama-3.1-8B}} & None & 0.492 & 1.548 & 0.023 & 0.065 \\
\midrule \multirow{9.75}{*}{\textbf{\texttt{Alpaca}}} & \multirow{3}{*}{\textbf{\texttt{Gemma2-2B}}} & 0.0 & 0.417 & 1.049 & 0.253 & 0.160 \\
& & 0.1 & 0.411 & 1.255 & 0.182 & 0.141 \\
& & 0.2 & 0.411 & 1.284 & 0.168 & 0.137 \\
\cmidrule{2-7} & \multirow{3}{*}{\textbf{\texttt{Mistral-7B}}} & 0.0 & 0.387 & 1.166 & 0.272 & 0.169 \\
\bottomrule
\end{tabular}}
\end{table}

\begin{table}[h!]\caption{Results on the \texttt{GlobalMMLU} subset for the \texttt{ig} language by model, SFT dataset, and label smoothing.}
\resizebox{\linewidth}{!}{\begin{tabular}{llc|cccc}
\toprule
\textbf{SFT Dataset} & \textbf{Base} & \textbf{Smoothing} & \textbf{Accuracy} & \textbf{Entropy} & \textbf{ECE} & \textbf{RMS} \\
\midrule
\multirow{3}{*}{\textbf{\texttt{Base Model}}} & \textbf{\texttt{Gemma2-2B}} & None & 0.289 & 1.774 & 0.068 & 0.093 \\
& \textbf{\texttt{Mistral-7B}} & None & 0.267 & 1.927 & 0.080 & 0.090 \\
& \textbf{\texttt{Llama-3.1-8B}} & None & 0.356 & 1.732 & 0.033 & 0.076 \\
\midrule \multirow{9.75}{*}{\textbf{\texttt{Alpaca}}} & \multirow{3}{*}{\textbf{\texttt{Gemma2-2B}}} & 0.0 & 0.310 & 1.300 & 0.229 & 0.164 \\
& & 0.1 & 0.311 & 1.422 & 0.185 & 0.152 \\
& & 0.2 & 0.307 & 1.480 & 0.151 & 0.143 \\
\cmidrule{2-7} & \multirow{3}{*}{\textbf{\texttt{Mistral-7B}}} & 0.0 & 0.271 & 1.533 & 0.279 & 0.175 \\
\bottomrule
\end{tabular}}
\end{table}

\begin{table}[h!]\caption{Results on the \texttt{GlobalMMLU} subset for the \texttt{it} language by model, SFT dataset, and label smoothing.}
\resizebox{\linewidth}{!}{\begin{tabular}{llc|cccc}
\toprule
\textbf{SFT Dataset} & \textbf{Base} & \textbf{Smoothing} & \textbf{Accuracy} & \textbf{Entropy} & \textbf{ECE} & \textbf{RMS} \\
\midrule
\multirow{3}{*}{\textbf{\texttt{Base Model}}} & \textbf{\texttt{Gemma2-2B}} & None & 0.344 & 1.668 & 0.043 & 0.082 \\
& \textbf{\texttt{Mistral-7B}} & None & 0.463 & 1.764 & 0.018 & 0.065 \\
& \textbf{\texttt{Llama-3.1-8B}} & None & 0.521 & 1.461 & 0.020 & 0.062 \\
\midrule \multirow{9.75}{*}{\textbf{\texttt{Alpaca}}} & \multirow{3}{*}{\textbf{\texttt{Gemma2-2B}}} & 0.0 & 0.428 & 0.881 & 0.276 & 0.166 \\
& & 0.1 & 0.423 & 1.166 & 0.173 & 0.140 \\
& & 0.2 & 0.426 & 1.195 & 0.167 & 0.137 \\
\cmidrule{2-7} & \multirow{3}{*}{\textbf{\texttt{Mistral-7B}}} & 0.0 & 0.445 & 1.022 & 0.268 & 0.164 \\
\bottomrule
\end{tabular}}
\end{table}

\begin{table}[h!]\caption{Results on the \texttt{GlobalMMLU} subset for the \texttt{ja} language by model, SFT dataset, and label smoothing.}
\resizebox{\linewidth}{!}{\begin{tabular}{llc|cccc}
\toprule
\textbf{SFT Dataset} & \textbf{Base} & \textbf{Smoothing} & \textbf{Accuracy} & \textbf{Entropy} & \textbf{ECE} & \textbf{RMS} \\
\midrule
\multirow{3}{*}{\textbf{\texttt{Base Model}}} & \textbf{\texttt{Gemma2-2B}} & None & 0.330 & 1.865 & 0.039 & 0.074 \\
& \textbf{\texttt{Mistral-7B}} & None & 0.373 & 1.867 & 0.017 & 0.065 \\
& \textbf{\texttt{Llama-3.1-8B}} & None & 0.461 & 1.459 & 0.031 & 0.076 \\
\midrule \multirow{9.75}{*}{\textbf{\texttt{Alpaca}}} & \multirow{3}{*}{\textbf{\texttt{Gemma2-2B}}} & 0.0 & 0.388 & 1.174 & 0.190 & 0.150 \\
& & 0.1 & 0.387 & 1.264 & 0.172 & 0.142 \\
& & 0.2 & 0.388 & 1.293 & 0.163 & 0.139 \\
\cmidrule{2-7} & \multirow{3}{*}{\textbf{\texttt{Mistral-7B}}} & 0.0 & 0.364 & 1.163 & 0.307 & 0.175 \\
\bottomrule
\end{tabular}}
\end{table}

\begin{table}[h!]\caption{Results on the \texttt{GlobalMMLU} subset for the \texttt{ko} language by model, SFT dataset, and label smoothing.}
\resizebox{\linewidth}{!}{\begin{tabular}{llc|cccc}
\toprule
\textbf{SFT Dataset} & \textbf{Base} & \textbf{Smoothing} & \textbf{Accuracy} & \textbf{Entropy} & \textbf{ECE} & \textbf{RMS} \\
\midrule
\multirow{3}{*}{\textbf{\texttt{Base Model}}} & \textbf{\texttt{Gemma2-2B}} & None & 0.323 & 1.846 & 0.054 & 0.080 \\
& \textbf{\texttt{Mistral-7B}} & None & 0.385 & 1.836 & 0.022 & 0.065 \\
& \textbf{\texttt{Llama-3.1-8B}} & None & 0.447 & 1.462 & 0.033 & 0.077 \\
\midrule \multirow{9.75}{*}{\textbf{\texttt{Alpaca}}} & \multirow{3}{*}{\textbf{\texttt{Gemma2-2B}}} & 0.0 & 0.368 & 1.058 & 0.276 & 0.168 \\
& & 0.1 & 0.367 & 1.226 & 0.202 & 0.154 \\
& & 0.2 & 0.365 & 1.257 & 0.188 & 0.150 \\
\cmidrule{2-7} & \multirow{3}{*}{\textbf{\texttt{Mistral-7B}}} & 0.0 & 0.362 & 1.136 & 0.254 & 0.162 \\
\bottomrule
\end{tabular}}
\end{table}

\begin{table}[h!]\caption{Results on the \texttt{GlobalMMLU} subset for the \texttt{ky} language by model, SFT dataset, and label smoothing.}
\resizebox{\linewidth}{!}{\begin{tabular}{llc|cccc}
\toprule
\textbf{SFT Dataset} & \textbf{Base} & \textbf{Smoothing} & \textbf{Accuracy} & \textbf{Entropy} & \textbf{ECE} & \textbf{RMS} \\
\midrule
\multirow{3}{*}{\textbf{\texttt{Base Model}}} & \textbf{\texttt{Gemma2-2B}} & None & 0.288 & 1.834 & 0.067 & 0.088 \\
& \textbf{\texttt{Mistral-7B}} & None & 0.274 & 1.936 & 0.066 & 0.083 \\
& \textbf{\texttt{Llama-3.1-8B}} & None & 0.362 & 1.556 & 0.078 & 0.101 \\
\midrule \multirow{9.75}{*}{\textbf{\texttt{Alpaca}}} & \multirow{3}{*}{\textbf{\texttt{Gemma2-2B}}} & 0.0 & 0.307 & 1.383 & 0.159 & 0.149 \\
& & 0.1 & 0.309 & 1.494 & 0.124 & 0.138 \\
& & 0.2 & 0.302 & 1.546 & 0.112 & 0.132 \\
\cmidrule{2-7} & \multirow{3}{*}{\textbf{\texttt{Mistral-7B}}} & 0.0 & 0.280 & 1.585 & 0.298 & 0.179 \\
\bottomrule
\end{tabular}}
\end{table}

\begin{table}[h!]\caption{Results on the \texttt{GlobalMMLU} subset for the \texttt{lt} language by model, SFT dataset, and label smoothing.}
\resizebox{\linewidth}{!}{\begin{tabular}{llc|cccc}
\toprule
\textbf{SFT Dataset} & \textbf{Base} & \textbf{Smoothing} & \textbf{Accuracy} & \textbf{Entropy} & \textbf{ECE} & \textbf{RMS} \\
\midrule
\multirow{3}{*}{\textbf{\texttt{Base Model}}} & \textbf{\texttt{Gemma2-2B}} & None & 0.317 & 1.792 & 0.057 & 0.083 \\
& \textbf{\texttt{Mistral-7B}} & None & 0.295 & 1.919 & 0.060 & 0.082 \\
& \textbf{\texttt{Llama-3.1-8B}} & None & 0.389 & 1.636 & 0.026 & 0.076 \\
\midrule \multirow{9.75}{*}{\textbf{\texttt{Alpaca}}} & \multirow{3}{*}{\textbf{\texttt{Gemma2-2B}}} & 0.0 & 0.356 & 1.159 & 0.233 & 0.159 \\
& & 0.1 & 0.354 & 1.356 & 0.151 & 0.139 \\
& & 0.2 & 0.352 & 1.404 & 0.131 & 0.135 \\
\cmidrule{2-7} & \multirow{3}{*}{\textbf{\texttt{Mistral-7B}}} & 0.0 & 0.295 & 1.171 & 0.344 & 0.190 \\
\bottomrule
\end{tabular}}
\end{table}

\begin{table}[h!]\caption{Results on the \texttt{GlobalMMLU} subset for the \texttt{mg} language by model, SFT dataset, and label smoothing.}
\resizebox{\linewidth}{!}{\begin{tabular}{llc|cccc}
\toprule
\textbf{SFT Dataset} & \textbf{Base} & \textbf{Smoothing} & \textbf{Accuracy} & \textbf{Entropy} & \textbf{ECE} & \textbf{RMS} \\
\midrule
\multirow{3}{*}{\textbf{\texttt{Base Model}}} & \textbf{\texttt{Gemma2-2B}} & None & 0.289 & 1.792 & 0.073 & 0.093 \\
& \textbf{\texttt{Mistral-7B}} & None & 0.263 & 1.927 & 0.090 & 0.094 \\
& \textbf{\texttt{Llama-3.1-8B}} & None & 0.334 & 1.785 & 0.041 & 0.074 \\
\midrule \multirow{9.75}{*}{\textbf{\texttt{Alpaca}}} & \multirow{3}{*}{\textbf{\texttt{Gemma2-2B}}} & 0.0 & 0.311 & 1.416 & 0.176 & 0.151 \\
& & 0.1 & 0.305 & 1.489 & 0.149 & 0.145 \\
& & 0.2 & 0.301 & 1.557 & 0.120 & 0.138 \\
\cmidrule{2-7} & \multirow{3}{*}{\textbf{\texttt{Mistral-7B}}} & 0.0 & 0.274 & 1.355 & 0.244 & 0.168 \\
\bottomrule
\end{tabular}}
\end{table}

\begin{table}[h!]\caption{Results on the \texttt{GlobalMMLU} subset for the \texttt{ms} language by model, SFT dataset, and label smoothing.}
\resizebox{\linewidth}{!}{\begin{tabular}{llc|cccc}
\toprule
\textbf{SFT Dataset} & \textbf{Base} & \textbf{Smoothing} & \textbf{Accuracy} & \textbf{Entropy} & \textbf{ECE} & \textbf{RMS} \\
\midrule
\multirow{3}{*}{\textbf{\texttt{Base Model}}} & \textbf{\texttt{Gemma2-2B}} & None & 0.341 & 1.820 & 0.045 & 0.076 \\
& \textbf{\texttt{Mistral-7B}} & None & 0.377 & 1.847 & 0.023 & 0.063 \\
& \textbf{\texttt{Llama-3.1-8B}} & None & 0.467 & 1.597 & 0.021 & 0.063 \\
\midrule \multirow{9.75}{*}{\textbf{\texttt{Alpaca}}} & \multirow{3}{*}{\textbf{\texttt{Gemma2-2B}}} & 0.0 & 0.407 & 1.087 & 0.245 & 0.158 \\
& & 0.1 & 0.400 & 1.295 & 0.164 & 0.139 \\
& & 0.2 & 0.400 & 1.326 & 0.151 & 0.135 \\
\cmidrule{2-7} & \multirow{3}{*}{\textbf{\texttt{Mistral-7B}}} & 0.0 & 0.362 & 1.222 & 0.267 & 0.170 \\
\bottomrule
\end{tabular}}
\end{table}

\begin{table}[h!]\caption{Results on the \texttt{GlobalMMLU} subset for the \texttt{ne} language by model, SFT dataset, and label smoothing.}
\resizebox{\linewidth}{!}{\begin{tabular}{llc|cccc}
\toprule
\textbf{SFT Dataset} & \textbf{Base} & \textbf{Smoothing} & \textbf{Accuracy} & \textbf{Entropy} & \textbf{ECE} & \textbf{RMS} \\
\midrule
\multirow{3}{*}{\textbf{\texttt{Base Model}}} & \textbf{\texttt{Gemma2-2B}} & None & 0.296 & 1.795 & 0.067 & 0.090 \\
& \textbf{\texttt{Mistral-7B}} & None & 0.265 & 1.917 & 0.085 & 0.092 \\
& \textbf{\texttt{Llama-3.1-8B}} & None & 0.391 & 1.752 & 0.031 & 0.069 \\
\midrule \multirow{9.75}{*}{\textbf{\texttt{Alpaca}}} & \multirow{3}{*}{\textbf{\texttt{Gemma2-2B}}} & 0.0 & 0.332 & 1.133 & 0.272 & 0.170 \\
& & 0.1 & 0.328 & 1.306 & 0.199 & 0.155 \\
& & 0.2 & 0.323 & 1.339 & 0.187 & 0.151 \\
\cmidrule{2-7} & \multirow{3}{*}{\textbf{\texttt{Mistral-7B}}} & 0.0 & 0.273 & 1.166 & 0.345 & 0.192 \\
\bottomrule
\end{tabular}}
\end{table}

\begin{table}[h!]\caption{Results on the \texttt{GlobalMMLU} subset for the \texttt{nl} language by model, SFT dataset, and label smoothing.}
\resizebox{\linewidth}{!}{\begin{tabular}{llc|cccc}
\toprule
\textbf{SFT Dataset} & \textbf{Base} & \textbf{Smoothing} & \textbf{Accuracy} & \textbf{Entropy} & \textbf{ECE} & \textbf{RMS} \\
\midrule
\multirow{3}{*}{\textbf{\texttt{Base Model}}} & \textbf{\texttt{Gemma2-2B}} & None & 0.372 & 1.706 & 0.044 & 0.081 \\
& \textbf{\texttt{Mistral-7B}} & None & 0.441 & 1.771 & 0.024 & 0.066 \\
& \textbf{\texttt{Llama-3.1-8B}} & None & 0.502 & 1.484 & 0.021 & 0.066 \\
\midrule \multirow{9.75}{*}{\textbf{\texttt{Alpaca}}} & \multirow{3}{*}{\textbf{\texttt{Gemma2-2B}}} & 0.0 & 0.418 & 0.856 & 0.299 & 0.171 \\
& & 0.1 & 0.418 & 1.109 & 0.207 & 0.149 \\
& & 0.2 & 0.413 & 1.136 & 0.191 & 0.146 \\
\cmidrule{2-7} & \multirow{3}{*}{\textbf{\texttt{Mistral-7B}}} & 0.0 & 0.433 & 1.054 & 0.252 & 0.162 \\
\bottomrule
\end{tabular}}
\end{table}

\begin{table}[h!]\caption{Results on the \texttt{GlobalMMLU} subset for the \texttt{ny} language by model, SFT dataset, and label smoothing.}
\resizebox{\linewidth}{!}{\begin{tabular}{llc|cccc}
\toprule
\textbf{SFT Dataset} & \textbf{Base} & \textbf{Smoothing} & \textbf{Accuracy} & \textbf{Entropy} & \textbf{ECE} & \textbf{RMS} \\
\midrule
\multirow{3}{*}{\textbf{\texttt{Base Model}}} & \textbf{\texttt{Gemma2-2B}} & None & 0.286 & 1.798 & 0.061 & 0.087 \\
& \textbf{\texttt{Mistral-7B}} & None & 0.273 & 1.931 & 0.073 & 0.088 \\
& \textbf{\texttt{Llama-3.1-8B}} & None & 0.324 & 1.829 & 0.035 & 0.073 \\
\midrule \multirow{9.75}{*}{\textbf{\texttt{Alpaca}}} & \multirow{3}{*}{\textbf{\texttt{Gemma2-2B}}} & 0.0 & 0.301 & 1.422 & 0.155 & 0.146 \\
& & 0.1 & 0.299 & 1.477 & 0.141 & 0.138 \\
& & 0.2 & 0.297 & 1.563 & 0.099 & 0.127 \\
\cmidrule{2-7} & \multirow{3}{*}{\textbf{\texttt{Mistral-7B}}} & 0.0 & 0.275 & 1.199 & 0.328 & 0.186 \\
\bottomrule
\end{tabular}}
\end{table}

\begin{table}[h!]\caption{Results on the \texttt{GlobalMMLU} subset for the \texttt{pl} language by model, SFT dataset, and label smoothing.}
\resizebox{\linewidth}{!}{\begin{tabular}{llc|cccc}
\toprule
\textbf{SFT Dataset} & \textbf{Base} & \textbf{Smoothing} & \textbf{Accuracy} & \textbf{Entropy} & \textbf{ECE} & \textbf{RMS} \\
\midrule
\multirow{3}{*}{\textbf{\texttt{Base Model}}} & \textbf{\texttt{Gemma2-2B}} & None & 0.340 & 1.772 & 0.057 & 0.084 \\
& \textbf{\texttt{Mistral-7B}} & None & 0.421 & 1.816 & 0.022 & 0.067 \\
& \textbf{\texttt{Llama-3.1-8B}} & None & 0.466 & 1.542 & 0.027 & 0.068 \\
\midrule \multirow{9.75}{*}{\textbf{\texttt{Alpaca}}} & \multirow{3}{*}{\textbf{\texttt{Gemma2-2B}}} & 0.0 & 0.393 & 0.945 & 0.301 & 0.174 \\
& & 0.1 & 0.392 & 1.234 & 0.185 & 0.144 \\
& & 0.2 & 0.392 & 1.257 & 0.175 & 0.141 \\
\cmidrule{2-7} & \multirow{3}{*}{\textbf{\texttt{Mistral-7B}}} & 0.0 & 0.414 & 1.103 & 0.263 & 0.162 \\
\bottomrule
\end{tabular}}
\end{table}

\begin{table}[h!]\caption{Results on the \texttt{GlobalMMLU} subset for the \texttt{pt} language by model, SFT dataset, and label smoothing.}
\resizebox{\linewidth}{!}{\begin{tabular}{llc|cccc}
\toprule
\textbf{SFT Dataset} & \textbf{Base} & \textbf{Smoothing} & \textbf{Accuracy} & \textbf{Entropy} & \textbf{ECE} & \textbf{RMS} \\
\midrule
\multirow{3}{*}{\textbf{\texttt{Base Model}}} & \textbf{\texttt{Gemma2-2B}} & None & 0.365 & 1.716 & 0.038 & 0.078 \\
& \textbf{\texttt{Mistral-7B}} & None & 0.471 & 1.744 & 0.023 & 0.067 \\
& \textbf{\texttt{Llama-3.1-8B}} & None & 0.522 & 1.434 & 0.025 & 0.064 \\
\midrule \multirow{9.75}{*}{\textbf{\texttt{Alpaca}}} & \multirow{3}{*}{\textbf{\texttt{Gemma2-2B}}} & 0.0 & 0.436 & 0.881 & 0.271 & 0.163 \\
& & 0.1 & 0.435 & 1.152 & 0.187 & 0.141 \\
& & 0.2 & 0.432 & 1.188 & 0.169 & 0.137 \\
\cmidrule{2-7} & \multirow{3}{*}{\textbf{\texttt{Mistral-7B}}} & 0.0 & 0.448 & 1.022 & 0.293 & 0.170 \\
\bottomrule
\end{tabular}}
\end{table}

\begin{table}[h!]\caption{Results on the \texttt{GlobalMMLU} subset for the \texttt{ro} language by model, SFT dataset, and label smoothing.}
\resizebox{\linewidth}{!}{\begin{tabular}{llc|cccc}
\toprule
\textbf{SFT Dataset} & \textbf{Base} & \textbf{Smoothing} & \textbf{Accuracy} & \textbf{Entropy} & \textbf{ECE} & \textbf{RMS} \\
\midrule
\multirow{3}{*}{\textbf{\texttt{Base Model}}} & \textbf{\texttt{Gemma2-2B}} & None & 0.353 & 1.757 & 0.040 & 0.076 \\
& \textbf{\texttt{Mistral-7B}} & None & 0.429 & 1.805 & 0.019 & 0.067 \\
& \textbf{\texttt{Llama-3.1-8B}} & None & 0.480 & 1.516 & 0.028 & 0.066 \\
\midrule \multirow{9.75}{*}{\textbf{\texttt{Alpaca}}} & \multirow{3}{*}{\textbf{\texttt{Gemma2-2B}}} & 0.0 & 0.405 & 0.943 & 0.275 & 0.165 \\
& & 0.1 & 0.405 & 1.220 & 0.177 & 0.141 \\
& & 0.2 & 0.401 & 1.242 & 0.170 & 0.139 \\
\cmidrule{2-7} & \multirow{3}{*}{\textbf{\texttt{Mistral-7B}}} & 0.0 & 0.416 & 1.084 & 0.194 & 0.145 \\
\bottomrule
\end{tabular}}
\end{table}

\begin{table}[h!]\caption{Results on the \texttt{GlobalMMLU} subset for the \texttt{ru} language by model, SFT dataset, and label smoothing.}
\resizebox{\linewidth}{!}{\begin{tabular}{llc|cccc}
\toprule
\textbf{SFT Dataset} & \textbf{Base} & \textbf{Smoothing} & \textbf{Accuracy} & \textbf{Entropy} & \textbf{ECE} & \textbf{RMS} \\
\midrule
\multirow{3}{*}{\textbf{\texttt{Base Model}}} & \textbf{\texttt{Gemma2-2B}} & None & 0.342 & 1.753 & 0.048 & 0.076 \\
& \textbf{\texttt{Mistral-7B}} & None & 0.435 & 1.779 & 0.023 & 0.070 \\
& \textbf{\texttt{Llama-3.1-8B}} & None & 0.488 & 1.458 & 0.019 & 0.067 \\
\midrule \multirow{9.75}{*}{\textbf{\texttt{Alpaca}}} & \multirow{3}{*}{\textbf{\texttt{Gemma2-2B}}} & 0.0 & 0.399 & 0.941 & 0.270 & 0.168 \\
& & 0.1 & 0.390 & 1.213 & 0.186 & 0.146 \\
& & 0.2 & 0.393 & 1.246 & 0.174 & 0.143 \\
\cmidrule{2-7} & \multirow{3}{*}{\textbf{\texttt{Mistral-7B}}} & 0.0 & 0.427 & 1.016 & 0.251 & 0.156 \\
\bottomrule
\end{tabular}}
\end{table}

\begin{table}[h!]\caption{Results on the \texttt{GlobalMMLU} subset for the \texttt{si} language by model, SFT dataset, and label smoothing.}
\resizebox{\linewidth}{!}{\begin{tabular}{llc|cccc}
\toprule
\textbf{SFT Dataset} & \textbf{Base} & \textbf{Smoothing} & \textbf{Accuracy} & \textbf{Entropy} & \textbf{ECE} & \textbf{RMS} \\
\midrule
\multirow{3}{*}{\textbf{\texttt{Base Model}}} & \textbf{\texttt{Gemma2-2B}} & None & 0.284 & 1.860 & 0.067 & 0.088 \\
& \textbf{\texttt{Mistral-7B}} & None & 0.250 & 1.926 & 0.113 & 0.107 \\
& \textbf{\texttt{Llama-3.1-8B}} & None & 0.342 & 1.760 & 0.034 & 0.073 \\
\midrule \multirow{9.75}{*}{\textbf{\texttt{Alpaca}}} & \multirow{3}{*}{\textbf{\texttt{Gemma2-2B}}} & 0.0 & 0.297 & 1.378 & 0.196 & 0.157 \\
& & 0.1 & 0.293 & 1.474 & 0.162 & 0.148 \\
& & 0.2 & 0.293 & 1.521 & 0.145 & 0.144 \\
\cmidrule{2-7} & \multirow{3}{*}{\textbf{\texttt{Mistral-7B}}} & 0.0 & 0.256 & 1.266 & 0.280 & 0.185 \\
\bottomrule
\end{tabular}}
\end{table}

\begin{table}[h!]\caption{Results on the \texttt{GlobalMMLU} subset for the \texttt{sn} language by model, SFT dataset, and label smoothing.}
\resizebox{\linewidth}{!}{\begin{tabular}{llc|cccc}
\toprule
\textbf{SFT Dataset} & \textbf{Base} & \textbf{Smoothing} & \textbf{Accuracy} & \textbf{Entropy} & \textbf{ECE} & \textbf{RMS} \\
\midrule
\multirow{3}{*}{\textbf{\texttt{Base Model}}} & \textbf{\texttt{Gemma2-2B}} & None & 0.285 & 1.821 & 0.065 & 0.087 \\
& \textbf{\texttt{Mistral-7B}} & None & 0.281 & 1.928 & 0.063 & 0.081 \\
& \textbf{\texttt{Llama-3.1-8B}} & None & 0.328 & 1.799 & 0.039 & 0.072 \\
\midrule \multirow{9.75}{*}{\textbf{\texttt{Alpaca}}} & \multirow{3}{*}{\textbf{\texttt{Gemma2-2B}}} & 0.0 & 0.307 & 1.453 & 0.149 & 0.142 \\
& & 0.1 & 0.310 & 1.503 & 0.138 & 0.137 \\
& & 0.2 & 0.307 & 1.575 & 0.094 & 0.127 \\
\cmidrule{2-7} & \multirow{3}{*}{\textbf{\texttt{Mistral-7B}}} & 0.0 & 0.284 & 1.228 & 0.348 & 0.188 \\
\bottomrule
\end{tabular}}
\end{table}

\begin{table}[h!]\caption{Results on the \texttt{GlobalMMLU} subset for the \texttt{so} language by model, SFT dataset, and label smoothing.}
\resizebox{\linewidth}{!}{\begin{tabular}{llc|cccc}
\toprule
\textbf{SFT Dataset} & \textbf{Base} & \textbf{Smoothing} & \textbf{Accuracy} & \textbf{Entropy} & \textbf{ECE} & \textbf{RMS} \\
\midrule
\multirow{3}{*}{\textbf{\texttt{Base Model}}} & \textbf{\texttt{Gemma2-2B}} & None & 0.280 & 1.873 & 0.070 & 0.089 \\
& \textbf{\texttt{Mistral-7B}} & None & 0.279 & 1.946 & 0.054 & 0.080 \\
& \textbf{\texttt{Llama-3.1-8B}} & None & 0.317 & 1.781 & 0.041 & 0.076 \\
\midrule \multirow{9.75}{*}{\textbf{\texttt{Alpaca}}} & \multirow{3}{*}{\textbf{\texttt{Gemma2-2B}}} & 0.0 & 0.294 & 1.493 & 0.121 & 0.141 \\
& & 0.1 & 0.297 & 1.556 & 0.101 & 0.130 \\
& & 0.2 & 0.290 & 1.620 & 0.080 & 0.120 \\
\cmidrule{2-7} & \multirow{3}{*}{\textbf{\texttt{Mistral-7B}}} & 0.0 & 0.272 & 1.367 & 0.236 & 0.169 \\
\bottomrule
\end{tabular}}
\end{table}

\begin{table}[h!]\caption{Results on the \texttt{GlobalMMLU} subset for the \texttt{sr} language by model, SFT dataset, and label smoothing.}
\resizebox{\linewidth}{!}{\begin{tabular}{llc|cccc}
\toprule
\textbf{SFT Dataset} & \textbf{Base} & \textbf{Smoothing} & \textbf{Accuracy} & \textbf{Entropy} & \textbf{ECE} & \textbf{RMS} \\
\midrule
\multirow{3}{*}{\textbf{\texttt{Base Model}}} & 
\textbf{\texttt{Gemma2-2B}} & None & 0.324 & 1.717 & 0.063 & 0.093 \\
& \textbf{\texttt{Mistral-7B}} & None & 0.404 & 1.826 & 0.013 & 0.062 \\
& \textbf{\texttt{Llama-3.1-8B}} & None & 0.432 & 1.539 & 0.035 & 0.075 \\
\midrule \multirow{9.75}{*}{\textbf{\texttt{Alpaca}}} & \multirow{3}{*}{\textbf{\texttt{Gemma2-2B}}} & 0.0 & 0.360 & 1.048 & 0.292 & 0.172 \\
& & 0.1 & 0.357 & 1.273 & 0.192 & 0.150 \\
& & 0.2 & 0.358 & 1.304 & 0.182 & 0.148 \\
\cmidrule{2-7} & \multirow{3}{*}{\textbf{\texttt{Mistral-7B}}} & 0.0 & 0.401 & 1.103 & 0.293 & 0.173 \\
\bottomrule
\end{tabular}}
\end{table}

\begin{table}[h!]\caption{Results on the \texttt{GlobalMMLU} subset for the \texttt{sv} language by model, SFT dataset, and label smoothing.}
\resizebox{\linewidth}{!}{\begin{tabular}{llc|cccc}
\toprule
\textbf{SFT Dataset} & \textbf{Base} & \textbf{Smoothing} & \textbf{Accuracy} & \textbf{Entropy} & \textbf{ECE} & \textbf{RMS} \\
\midrule
\multirow{3}{*}{\textbf{\texttt{Base Model}}} & \textbf{\texttt{Gemma2-2B}} & None & 0.359 & 1.730 & 0.051 & 0.082 \\
& \textbf{\texttt{Mistral-7B}} & None & 0.428 & 1.779 & 0.026 & 0.068 \\
& \textbf{\texttt{Llama-3.1-8B}} & None & 0.497 & 1.525 & 0.016 & 0.067 \\
\midrule \multirow{9.75}{*}{\textbf{\texttt{Alpaca}}} & \multirow{3}{*}{\textbf{\texttt{Gemma2-2B}}} & 0.0 & 0.414 & 0.929 & 0.282 & 0.167 \\
& & 0.1 & 0.406 & 1.154 & 0.197 & 0.147 \\
& & 0.2 & 0.407 & 1.185 & 0.184 & 0.143 \\
\cmidrule{2-7} & \multirow{3}{*}{\textbf{\texttt{Mistral-7B}}} & 0.0 & 0.425 & 1.046 & 0.219 & 0.149 \\
\bottomrule
\end{tabular}}
\end{table}

\begin{table}[h!]\caption{Results on the \texttt{GlobalMMLU} subset for the \texttt{sw} language by model, SFT dataset, and label smoothing.}
\resizebox{\linewidth}{!}{\begin{tabular}{llc|cccc}
\toprule
\textbf{SFT Dataset} & \textbf{Base} & \textbf{Smoothing} & \textbf{Accuracy} & \textbf{Entropy} & \textbf{ECE} & \textbf{RMS} \\
\midrule
\multirow{3}{*}{\textbf{\texttt{Base Model}}} & \textbf{\texttt{Gemma2-2B}} & None & 0.323 & 1.816 & 0.045 & 0.078 \\
& \textbf{\texttt{Mistral-7B}} & None & 0.280 & 1.943 & 0.056 & 0.080 \\
& \textbf{\texttt{Llama-3.1-8B}} & None & 0.356 & 1.731 & 0.039 & 0.075 \\
\midrule \multirow{9.75}{*}{\textbf{\texttt{Alpaca}}} & \multirow{3}{*}{\textbf{\texttt{Gemma2-2B}}} & 0.0 & 0.353 & 1.203 & 0.218 & 0.157 \\
& & 0.1 & 0.356 & 1.347 & 0.187 & 0.145 \\
& & 0.2 & 0.352 & 1.388 & 0.171 & 0.141 \\
\cmidrule{2-7} & \multirow{3}{*}{\textbf{\texttt{Mistral-7B}}} & 0.0 & 0.280 & 1.241 & 0.295 & 0.182 \\
\bottomrule
\end{tabular}}
\end{table}

\begin{table}[h!]\caption{Results on the \texttt{GlobalMMLU} subset for the \texttt{te} language by model, SFT dataset, and label smoothing.}
\resizebox{\linewidth}{!}{\begin{tabular}{llc|cccc}
\toprule
\textbf{SFT Dataset} & \textbf{Base} & \textbf{Smoothing} & \textbf{Accuracy} & \textbf{Entropy} & \textbf{ECE} & \textbf{RMS} \\
\midrule
\multirow{3}{*}{\textbf{\texttt{Base Model}}} & \textbf{\texttt{Gemma2-2B}} & None & 0.288 & 1.823 & 0.082 & 0.094 \\
& \textbf{\texttt{Mistral-7B}} & None & 0.250 & 1.920 & 0.111 & 0.105 \\
& \textbf{\texttt{Llama-3.1-8B}} & None & 0.356 & 1.758 & 0.042 & 0.075 \\
\midrule \multirow{9.75}{*}{\textbf{\texttt{Alpaca}}} & \multirow{3}{*}{\textbf{\texttt{Gemma2-2B}}} & 0.0 & 0.319 & 1.211 & 0.261 & 0.168 \\
& & 0.1 & 0.321 & 1.409 & 0.160 & 0.144 \\
& & 0.2 & 0.318 & 1.424 & 0.149 & 0.142 \\
\cmidrule{2-7} & \multirow{3}{*}{\textbf{\texttt{Mistral-7B}}} & 0.0 & 0.254 & 1.229 & 0.291 & 0.187 \\
\bottomrule
\end{tabular}}
\end{table}

\begin{table}[h!]\caption{Results on the \texttt{GlobalMMLU} subset for the \texttt{tr} language by model, SFT dataset, and label smoothing.}
\resizebox{\linewidth}{!}{\begin{tabular}{llc|cccc}
\toprule
\textbf{SFT Dataset} & \textbf{Base} & \textbf{Smoothing} & \textbf{Accuracy} & \textbf{Entropy} & \textbf{ECE} & \textbf{RMS} \\
\midrule
\multirow{3}{*}{\textbf{\texttt{Base Model}}} & \textbf{\texttt{Gemma2-2B}} & None & 0.328 & 1.830 & 0.052 & 0.078 \\
& \textbf{\texttt{Mistral-7B}} & None & 0.341 & 1.885 & 0.032 & 0.069 \\
& \textbf{\texttt{Llama-3.1-8B}} & None & 0.452 & 1.539 & 0.028 & 0.071 \\
\midrule \multirow{9.75}{*}{\textbf{\texttt{Alpaca}}} & \multirow{3}{*}{\textbf{\texttt{Gemma2-2B}}} & 0.0 & 0.373 & 1.141 & 0.214 & 0.155 \\
& & 0.1 & 0.372 & 1.332 & 0.154 & 0.137 \\
& & 0.2 & 0.371 & 1.363 & 0.137 & 0.133 \\
\cmidrule{2-7} & \multirow{3}{*}{\textbf{\texttt{Mistral-7B}}} & 0.0 & 0.337 & 1.229 & 0.297 & 0.176 \\
\bottomrule
\end{tabular}}
\end{table}

\begin{table}[h!]\caption{Results on the \texttt{GlobalMMLU} subset for the \texttt{uk} language by model, SFT dataset, and label smoothing.}
\resizebox{\linewidth}{!}{\begin{tabular}{llc|cccc}
\toprule
\textbf{SFT Dataset} & \textbf{Base} & \textbf{Smoothing} & \textbf{Accuracy} & \textbf{Entropy} & \textbf{ECE} & \textbf{RMS} \\
\midrule
\multirow{3}{*}{\textbf{\texttt{Base Model}}} & \textbf{\texttt{Gemma2-2B}} & None & 0.331 & 1.758 & 0.051 & 0.082 \\
& \textbf{\texttt{Mistral-7B}} & None & 0.423 & 1.794 & 0.021 & 0.065 \\
& \textbf{\texttt{Llama-3.1-8B}} & None & 0.460 & 1.499 & 0.022 & 0.070 \\
\midrule \multirow{9.75}{*}{\textbf{\texttt{Alpaca}}} & \multirow{3}{*}{\textbf{\texttt{Gemma2-2B}}} & 0.0 & 0.385 & 0.963 & 0.297 & 0.171 \\
& & 0.1 & 0.378 & 1.237 & 0.186 & 0.147 \\
& & 0.2 & 0.378 & 1.265 & 0.173 & 0.143 \\
\cmidrule{2-7} & \multirow{3}{*}{\textbf{\texttt{Mistral-7B}}} & 0.0 & 0.414 & 1.057 & 0.299 & 0.172 \\
\bottomrule
\end{tabular}}
\end{table}

\begin{table}[h!]\caption{Results on the \texttt{GlobalMMLU} subset for the \texttt{vi} language by model, SFT dataset, and label smoothing.}
\resizebox{\linewidth}{!}{\begin{tabular}{llc|cccc}
\toprule
\textbf{SFT Dataset} & \textbf{Base} & \textbf{Smoothing} & \textbf{Accuracy} & \textbf{Entropy} & \textbf{ECE} & \textbf{RMS} \\
\midrule
\multirow{3}{*}{\textbf{\texttt{Base Model}}} & \textbf{\texttt{Gemma2-2B}} & None & 0.299 & 1.752 & 0.044 & 0.081 \\
& \textbf{\texttt{Mistral-7B}} & None & 0.365 & 1.870 & 0.020 & 0.067 \\
& \textbf{\texttt{Llama-3.1-8B}} & None & 0.473 & 1.523 & 0.020 & 0.067 \\
\midrule \multirow{9.75}{*}{\textbf{\texttt{Alpaca}}} & \multirow{3}{*}{\textbf{\texttt{Gemma2-2B}}} & 0.0 & 0.384 & 1.036 & 0.282 & 0.169 \\
& & 0.1 & 0.387 & 1.274 & 0.179 & 0.146 \\
& & 0.2 & 0.382 & 1.258 & 0.186 & 0.147 \\
\cmidrule{2-7} & \multirow{3}{*}{\textbf{\texttt{Mistral-7B}}} & 0.0 & 0.346 & 1.212 & 0.283 & 0.171 \\
\bottomrule
\end{tabular}}
\end{table}

\begin{table}[h!]\caption{Results on the \texttt{GlobalMMLU} subset for the \texttt{yo} language by model, SFT dataset, and label smoothing.}
\resizebox{\linewidth}{!}{\begin{tabular}{llc|cccc}
\toprule
\textbf{SFT Dataset} & \textbf{Base} & \textbf{Smoothing} & \textbf{Accuracy} & \textbf{Entropy} & \textbf{ECE} & \textbf{RMS} \\
\midrule
\multirow{3}{*}{\textbf{\texttt{Base Model}}} & \textbf{\texttt{Gemma2-2B}} & None & 0.276 & 1.796 & 0.068 & 0.093 \\
& \textbf{\texttt{Mistral-7B}} & None & 0.265 & 1.945 & 0.069 & 0.087 \\
& \textbf{\texttt{Llama-3.1-8B}} & None & 0.308 & 1.809 & 0.052 & 0.079 \\
\midrule \multirow{9.75}{*}{\textbf{\texttt{Alpaca}}} & \multirow{3}{*}{\textbf{\texttt{Gemma2-2B}}} & 0.0 & 0.289 & 1.450 & 0.141 & 0.146 \\
& & 0.1 & 0.286 & 1.485 & 0.149 & 0.141 \\
& & 0.2 & 0.288 & 1.556 & 0.105 & 0.132 \\
\cmidrule{2-7} & \multirow{3}{*}{\textbf{\texttt{Mistral-7B}}} & 0.0 & 0.266 & 1.434 & 0.176 & 0.158 \\
\bottomrule
\end{tabular}}
\end{table}

\begin{table}[h!]\caption{Results on the \texttt{GlobalMMLU} subset for the \texttt{zh} language by model, SFT dataset, and label smoothing.}
\resizebox{\linewidth}{!}{\begin{tabular}{llc|cccc}
\toprule
\textbf{SFT Dataset} & \textbf{Base} & \textbf{Smoothing} & \textbf{Accuracy} & \textbf{Entropy} & \textbf{ECE} & \textbf{RMS} \\
\midrule
\multirow{3}{*}{\textbf{\texttt{Base Model}}} & \textbf{\texttt{Gemma2-2B}} & None & 0.343 & 1.881 & 0.027 & 0.069 \\
& \textbf{\texttt{Mistral-7B}} & None & 0.410 & 1.781 & 0.021 & 0.066 \\
& \textbf{\texttt{Llama-3.1-8B}} & None & 0.491 & 1.378 & 0.036 & 0.077 \\
\midrule \multirow{9.75}{*}{\textbf{\texttt{Alpaca}}} & \multirow{3}{*}{\textbf{\texttt{Gemma2-2B}}} & 0.0 & 0.408 & 1.126 & 0.216 & 0.151 \\
& & 0.1 & 0.408 & 1.244 & 0.169 & 0.139 \\
& & 0.2 & 0.407 & 1.278 & 0.157 & 0.137 \\
\cmidrule{2-7} & \multirow{3}{*}{\textbf{\texttt{Mistral-7B}}} & 0.0 & 0.400 & 1.042 & 0.299 & 0.172 \\
\bottomrule
\end{tabular}}
\end{table}

\clearpage
\subsubsection{MMLU-ProX}

\begin{table}[h!]
\caption{Results on the \texttt{MMLU-ProX} subset for the \texttt{af} language by model, SFT dataset, and label smoothing.}
\resizebox{\linewidth}{!}{\begin{tabular}{llc|cccc}
\toprule
\textbf{SFT Dataset} & \textbf{Base} & \textbf{Smoothing} & \textbf{Accuracy} & \textbf{Entropy} & \textbf{ECE} & \textbf{RMS} \\
\midrule
\multirow{3}{*}{\textbf{\texttt{Base Model}}} & \textbf{\texttt{Gemma2-2B}} & None & 0.115 & 3.098 & 0.085 & 0.095 \\
& \textbf{\texttt{Mistral-7B}} & None & 0.176 & 2.997 & 0.066 & 0.083 \\
& \textbf{\texttt{Llama-3.1-8B}} & None & 0.122 & 1.566 & 0.227 & 0.163 \\
\midrule
\multirow{9.75}{*}{\textbf{\texttt{Alpaca}}} & \multirow{3}{*}{\textbf{\texttt{Gemma2-2B}}} & 0.0 & 0.159 & 2.866 & 0.091 & 0.110 \\
& & 0.1 & 0.164 & 2.826 & 0.059 & 0.090 \\
& & 0.2 & 0.158 & 2.637 & 0.063 & 0.091 \\
\cmidrule{2-7} & \multirow{3}{*}{\textbf{\texttt{Mistral-7B}}} & 0.0 & 0.182 & 2.114 & 0.125 & 0.150 \\
\bottomrule
\end{tabular}}
\end{table}

\begin{table}[h!]
\caption{Results on the \texttt{MMLU-ProX} subset for the \texttt{ar} language by model, SFT dataset, and label smoothing.}
\resizebox{\linewidth}{!}{\begin{tabular}{llc|cccc}
\toprule
\textbf{SFT Dataset} & \textbf{Base} & \textbf{Smoothing} & \textbf{Accuracy} & \textbf{Entropy} & \textbf{ECE} & \textbf{RMS} \\
\midrule
\multirow{3}{*}{\textbf{\texttt{Base Model}}} & \textbf{\texttt{Gemma2-2B}} & None & 0.123 & 3.147 & 0.060 & 0.083 \\
& \textbf{\texttt{Mistral-7B}} & None & 0.158 & 3.042 & 0.074 & 0.088 \\
& \textbf{\texttt{Llama-3.1-8B}} & None & 0.126 & 1.530 & 0.251 & 0.167 \\
\midrule
\multirow{9.75}{*}{\textbf{\texttt{Alpaca}}} & \multirow{3}{*}{\textbf{\texttt{Gemma2-2B}}} & 0.0 & 0.150 & 2.787 & 0.089 & 0.105 \\
& & 0.1 & 0.151 & 2.911 & 0.077 & 0.095 \\
& & 0.2 & 0.148 & 2.926 & 0.076 & 0.094 \\
\cmidrule{2-7} & \multirow{3}{*}{\textbf{\texttt{Mistral-7B}}} & 0.0 & 0.156 & 2.329 & 0.081 & 0.127 \\
\bottomrule
\end{tabular}}
\end{table}

\begin{table}[h!]
\caption{Results on the \texttt{MMLU-ProX} subset for the \texttt{bn} language by model, SFT dataset, and label smoothing.}
\resizebox{\linewidth}{!}{\begin{tabular}{llc|cccc}
\toprule
\textbf{SFT Dataset} & \textbf{Base} & \textbf{Smoothing} & \textbf{Accuracy} & \textbf{Entropy} & \textbf{ECE} & \textbf{RMS} \\
\midrule
\multirow{3}{*}{\textbf{\texttt{Base Model}}} & \textbf{\texttt{Gemma2-2B}} & None & 0.099 & 3.156 & 0.090 & 0.099 \\
& \textbf{\texttt{Mistral-7B}} & None & 0.137 & 3.029 & 0.097 & 0.100 \\
& \textbf{\texttt{Llama-3.1-8B}} & None & 0.121 & 1.610 & 0.197 & 0.154 \\
\midrule
\multirow{9.75}{*}{\textbf{\texttt{Alpaca}}} & \multirow{3}{*}{\textbf{\texttt{Gemma2-2B}}} & 0.0 & 0.139 & 2.839 & 0.088 & 0.102 \\
& & 0.1 & 0.135 & 2.941 & 0.079 & 0.094 \\
& & 0.2 & 0.129 & 2.922 & 0.094 & 0.101 \\
\cmidrule{2-7} & \multirow{3}{*}{\textbf{\texttt{Mistral-7B}}} & 0.0 & 0.140 & 2.415 & 0.081 & 0.129 \\
\bottomrule
\end{tabular}}
\end{table}

\begin{table}[h!]
\caption{Results on the \texttt{MMLU-ProX} subset for the \texttt{cs} language by model, SFT dataset, and label smoothing.}
\resizebox{\linewidth}{!}{\begin{tabular}{llc|cccc}
\toprule
\textbf{SFT Dataset} & \textbf{Base} & \textbf{Smoothing} & \textbf{Accuracy} & \textbf{Entropy} & \textbf{ECE} & \textbf{RMS} \\
\midrule
\multirow{3}{*}{\textbf{\texttt{Base Model}}} & \textbf{\texttt{Gemma2-2B}} & None & 0.123 & 3.079 & 0.084 & 0.094 \\
& \textbf{\texttt{Mistral-7B}} & None & 0.206 & 2.996 & 0.046 & 0.075 \\
& \textbf{\texttt{Llama-3.1-8B}} & None & 0.132 & 1.539 & 0.243 & 0.165 \\
\midrule
\multirow{9.75}{*}{\textbf{\texttt{Alpaca}}} & \multirow{3}{*}{\textbf{\texttt{Gemma2-2B}}} & 0.0 & 0.167 & 2.637 & 0.072 & 0.113 \\
& & 0.1 & 0.171 & 2.834 & 0.049 & 0.085 \\
& & 0.2 & 0.168 & 2.822 & 0.047 & 0.085 \\
\cmidrule{2-7} & \multirow{3}{*}{\textbf{\texttt{Mistral-7B}}} & 0.0 & 0.211 & 2.089 & 0.111 & 0.140 \\
\bottomrule
\end{tabular}}
\end{table}

\begin{table}[h!]
\caption{Results on the \texttt{MMLU-ProX} subset for the \texttt{de} language by model, SFT dataset, and label smoothing.}
\resizebox{\linewidth}{!}{\begin{tabular}{llc|cccc}
\toprule
\textbf{SFT Dataset} & \textbf{Base} & \textbf{Smoothing} & \textbf{Accuracy} & \textbf{Entropy} & \textbf{ECE} & \textbf{RMS} \\
\midrule
\multirow{3}{*}{\textbf{\texttt{Base Model}}} & \textbf{\texttt{Gemma2-2B}} & None & 0.123 & 3.109 & 0.078 & 0.090 \\
& \textbf{\texttt{Mistral-7B}} & None & 0.210 & 2.979 & 0.048 & 0.076 \\
& \textbf{\texttt{Llama-3.1-8B}} & None & 0.144 & 1.603 & 0.228 & 0.163 \\
\midrule
\multirow{9.75}{*}{\textbf{\texttt{Alpaca}}} & \multirow{3}{*}{\textbf{\texttt{Gemma2-2B}}} & 0.0 & 0.173 & 2.646 & 0.073 & 0.106 \\
& & 0.1 & 0.174 & 2.825 & 0.056 & 0.088 \\
& & 0.2 & 0.166 & 2.840 & 0.060 & 0.088 \\
\cmidrule{2-7} & \multirow{3}{*}{\textbf{\texttt{Mistral-7B}}} & 0.0 & 0.213 & 2.030 & 0.120 & 0.142 \\
\bottomrule
\end{tabular}}
\end{table}

\begin{table}[h!]
\caption{Results on the \texttt{MMLU-ProX} subset for the \texttt{en} language by model, SFT dataset, and label smoothing.}
\resizebox{\linewidth}{!}{\begin{tabular}{llc|cccc}
\toprule
\textbf{SFT Dataset} & \textbf{Base} & \textbf{Smoothing} & \textbf{Accuracy} & \textbf{Entropy} & \textbf{ECE} & \textbf{RMS} \\
\midrule
\multirow{3}{*}{\textbf{\texttt{Base Model}}} & \textbf{\texttt{Gemma2-2B}} & None & 0.155 & 3.053 & 0.064 & 0.085 \\
& \textbf{\texttt{Mistral-7B}} & None & 0.268 & 2.870 & 0.022 & 0.063 \\
& \textbf{\texttt{Llama-3.1-8B}} & None & 0.175 & 1.571 & 0.238 & 0.161 \\
\midrule
\multirow{9.75}{*}{\textbf{\texttt{Alpaca}}} & \multirow{3}{*}{\textbf{\texttt{Gemma2-2B}}} & 0.0 & 0.200 & 2.420 & 0.093 & 0.119 \\
& & 0.1 & 0.198 & 2.694 & 0.063 & 0.096 \\
& & 0.2 & 0.199 & 2.708 & 0.057 & 0.096 \\
\cmidrule{2-7} & \multirow{3}{*}{\textbf{\texttt{Mistral-7B}}} & 0.0 & 0.254 & 1.910 & 0.116 & 0.138 \\
\bottomrule
\end{tabular}}
\end{table}

\begin{table}[h!]
\caption{Results on the \texttt{MMLU-ProX} subset for the \texttt{es} language by model, SFT dataset, and label smoothing.}
\resizebox{\linewidth}{!}{\begin{tabular}{llc|cccc}
\toprule
\textbf{SFT Dataset} & \textbf{Base} & \textbf{Smoothing} & \textbf{Accuracy} & \textbf{Entropy} & \textbf{ECE} & \textbf{RMS} \\
\midrule
\multirow{3}{*}{\textbf{\texttt{Base Model}}} & \textbf{\texttt{Gemma2-2B}} & None & 0.128 & 3.115 & 0.071 & 0.086 \\
& \textbf{\texttt{Mistral-7B}} & None & 0.229 & 2.948 & 0.034 & 0.069 \\
& \textbf{\texttt{Llama-3.1-8B}} & None & 0.141 & 1.600 & 0.219 & 0.159 \\
\midrule
\multirow{9.75}{*}{\textbf{\texttt{Alpaca}}} & \multirow{3}{*}{\textbf{\texttt{Gemma2-2B}}} & 0.0 & 0.182 & 2.574 & 0.075 & 0.112 \\
& & 0.1 & 0.180 & 2.779 & 0.060 & 0.095 \\
& & 0.2 & 0.172 & 2.788 & 0.067 & 0.095 \\
\cmidrule{2-7} & \multirow{3}{*}{\textbf{\texttt{Mistral-7B}}} & 0.0 & 0.217 & 1.997 & 0.124 & 0.143 \\
\bottomrule
\end{tabular}}
\end{table}

\begin{table}[h!]
\caption{Results on the \texttt{MMLU-ProX} subset for the \texttt{fr} language by model, SFT dataset, and label smoothing.}
\resizebox{\linewidth}{!}{\begin{tabular}{llc|cccc}
\toprule
\textbf{SFT Dataset} & \textbf{Base} & \textbf{Smoothing} & \textbf{Accuracy} & \textbf{Entropy} & \textbf{ECE} & \textbf{RMS} \\
\midrule
\multirow{3}{*}{\textbf{\texttt{Base Model}}} & \textbf{\texttt{Gemma2-2B}} & None & 0.116 & 3.129 & 0.077 & 0.090 \\
& \textbf{\texttt{Mistral-7B}} & None & 0.221 & 2.955 & 0.042 & 0.073 \\
& \textbf{\texttt{Llama-3.1-8B}} & None & 0.153 & 1.672 & 0.178 & 0.145 \\
\midrule
\multirow{9.75}{*}{\textbf{\texttt{Alpaca}}} & \multirow{3}{*}{\textbf{\texttt{Gemma2-2B}}} & 0.0 & 0.169 & 2.628 & 0.083 & 0.112 \\
& & 0.1 & 0.171 & 2.804 & 0.066 & 0.095 \\
& & 0.2 & 0.168 & 2.821 & 0.066 & 0.096 \\
\cmidrule{2-7} & \multirow{3}{*}{\textbf{\texttt{Mistral-7B}}} & 0.0 & 0.219 & 2.017 & 0.106 & 0.138 \\
\bottomrule
\end{tabular}}
\end{table}

\begin{table}[h!]
\caption{Results on the \texttt{MMLU-ProX} subset for the \texttt{hi} language by model, SFT dataset, and label smoothing.}
\resizebox{\linewidth}{!}{\begin{tabular}{llc|cccc}
\toprule
\textbf{SFT Dataset} & \textbf{Base} & \textbf{Smoothing} & \textbf{Accuracy} & \textbf{Entropy} & \textbf{ECE} & \textbf{RMS} \\
\midrule
\multirow{3}{*}{\textbf{\texttt{Base Model}}} & \textbf{\texttt{Gemma2-2B}} & None & 0.102 & 3.155 & 0.090 & 0.099 \\
& \textbf{\texttt{Mistral-7B}} & None & 0.148 & 3.006 & 0.089 & 0.096 \\
& \textbf{\texttt{Llama-3.1-8B}} & None & 0.121 & 1.544 & 0.256 & 0.169 \\
\midrule
\multirow{9.75}{*}{\textbf{\texttt{Alpaca}}} & \multirow{3}{*}{\textbf{\texttt{Gemma2-2B}}} & 0.0 & 0.148 & 2.733 & 0.070 & 0.112 \\
& & 0.1 & 0.148 & 2.879 & 0.061 & 0.094 \\
& & 0.2 & 0.144 & 2.869 & 0.067 & 0.097 \\
\cmidrule{2-7} & \multirow{3}{*}{\textbf{\texttt{Mistral-7B}}} & 0.0 & 0.153 & 2.399 & 0.133 & 0.159 \\
\bottomrule
\end{tabular}}
\end{table}

\begin{table}[h!]
\caption{Results on the \texttt{MMLU-ProX} subset for the \texttt{hu} language by model, SFT dataset, and label smoothing.}
\resizebox{\linewidth}{!}{\begin{tabular}{llc|cccc}
\toprule
\textbf{SFT Dataset} & \textbf{Base} & \textbf{Smoothing} & \textbf{Accuracy} & \textbf{Entropy} & \textbf{ECE} & \textbf{RMS} \\
\midrule
\multirow{3}{*}{\textbf{\texttt{Base Model}}} & \textbf{\texttt{Gemma2-2B}} & None & 0.116 & 3.149 & 0.067 & 0.088 \\
& \textbf{\texttt{Mistral-7B}} & None & 0.176 & 2.978 & 0.092 & 0.097 \\
& \textbf{\texttt{Llama-3.1-8B}} & None & 0.129 & 1.507 & 0.244 & 0.169 \\
\midrule
\multirow{9.75}{*}{\textbf{\texttt{Alpaca}}} & \multirow{3}{*}{\textbf{\texttt{Gemma2-2B}}} & 0.0 & 0.159 & 2.724 & 0.072 & 0.104 \\
& & 0.1 & 0.164 & 2.924 & 0.050 & 0.081 \\
& & 0.2 & 0.157 & 2.927 & 0.054 & 0.083 \\
\cmidrule{2-7} & \multirow{3}{*}{\textbf{\texttt{Mistral-7B}}} & 0.0 & 0.197 & 2.250 & 0.089 & 0.131 \\
\bottomrule
\end{tabular}}
\end{table}

\begin{table}[h!]
\caption{Results on the \texttt{MMLU-ProX} subset for the \texttt{id} language by model, SFT dataset, and label smoothing.}
\resizebox{\linewidth}{!}{\begin{tabular}{llc|cccc}
\toprule
\textbf{SFT Dataset} & \textbf{Base} & \textbf{Smoothing} & \textbf{Accuracy} & \textbf{Entropy} & \textbf{ECE} & \textbf{RMS} \\
\midrule
\multirow{3}{*}{\textbf{\texttt{Base Model}}} & \textbf{\texttt{Gemma2-2B}} & None & 0.127 & 3.113 & 0.075 & 0.088 \\
& \textbf{\texttt{Mistral-7B}} & None & 0.200 & 2.997 & 0.046 & 0.076 \\
& \textbf{\texttt{Llama-3.1-8B}} & None & 0.143 & 1.709 & 0.185 & 0.149 \\
\midrule
\multirow{9.75}{*}{\textbf{\texttt{Alpaca}}} & \multirow{3}{*}{\textbf{\texttt{Gemma2-2B}}} & 0.0 & 0.170 & 2.675 & 0.068 & 0.105 \\
& & 0.1 & 0.164 & 2.816 & 0.063 & 0.092 \\
& & 0.2 & 0.166 & 2.818 & 0.063 & 0.092 \\
\cmidrule{2-7} & \multirow{3}{*}{\textbf{\texttt{Mistral-7B}}} & 0.0 & 0.197 & 2.158 & 0.112 & 0.142 \\
\bottomrule
\end{tabular}}
\end{table}

\begin{table}[h!]
\caption{Results on the \texttt{MMLU-ProX} subset for the \texttt{it} language by model, SFT dataset, and label smoothing.}
\resizebox{\linewidth}{!}{\begin{tabular}{llc|cccc}
\toprule
\textbf{SFT Dataset} & \textbf{Base} & \textbf{Smoothing} & \textbf{Accuracy} & \textbf{Entropy} & \textbf{ECE} & \textbf{RMS} \\
\midrule
\multirow{3}{*}{\textbf{\texttt{Base Model}}} & \textbf{\texttt{Gemma2-2B}} & None & 0.124 & 3.118 & 0.076 & 0.089 \\
& \textbf{\texttt{Mistral-7B}} & None & 0.219 & 2.984 & 0.039 & 0.072 \\
& \textbf{\texttt{Llama-3.1-8B}} & None & 0.129 & 1.556 & 0.302 & 0.182 \\
\midrule
\multirow{9.75}{*}{\textbf{\texttt{Alpaca}}} & \multirow{3}{*}{\textbf{\texttt{Gemma2-2B}}} & 0.0 & 0.173 & 2.599 & 0.082 & 0.112 \\
& & 0.1 & 0.169 & 2.791 & 0.063 & 0.093 \\
& & 0.2 & 0.170 & 2.815 & 0.056 & 0.090 \\
\cmidrule{2-7} & \multirow{3}{*}{\textbf{\texttt{Mistral-7B}}} & 0.0 & 0.214 & 2.053 & 0.096 & 0.132 \\
\bottomrule
\end{tabular}}
\end{table}

\begin{table}[h!]
\caption{Results on the \texttt{MMLU-ProX} subset for the \texttt{ja} language by model, SFT dataset, and label smoothing.}
\resizebox{\linewidth}{!}{\begin{tabular}{llc|cccc}
\toprule
\textbf{SFT Dataset} & \textbf{Base} & \textbf{Smoothing} & \textbf{Accuracy} & \textbf{Entropy} & \textbf{ECE} & \textbf{RMS} \\
\midrule
\multirow{3}{*}{\textbf{\texttt{Base Model}}} & \textbf{\texttt{Gemma2-2B}} & None & 0.116 & 3.194 & 0.056 & 0.079 \\
& \textbf{\texttt{Mistral-7B}} & None & 0.196 & 3.016 & 0.044 & 0.073 \\
& \textbf{\texttt{Llama-3.1-8B}} & None & 0.141 & 1.618 & 0.216 & 0.157 \\
\midrule
\multirow{9.75}{*}{\textbf{\texttt{Alpaca}}} & \multirow{3}{*}{\textbf{\texttt{Gemma2-2B}}} & 0.0 & 0.159 & 2.845 & 0.062 & 0.096 \\
& & 0.1 & 0.164 & 2.910 & 0.059 & 0.089 \\
& & 0.2 & 0.157 & 2.901 & 0.062 & 0.090 \\
\cmidrule{2-7} & \multirow{3}{*}{\textbf{\texttt{Mistral-7B}}} & 0.0 & 0.189 & 2.185 & 0.074 & 0.125 \\
\bottomrule
\end{tabular}}
\end{table}

\begin{table}[h!]
\caption{Results on the \texttt{MMLU-ProX} subset for the \texttt{ko} language by model, SFT dataset, and label smoothing.}
\resizebox{\linewidth}{!}{\begin{tabular}{llc|cccc}
\toprule
\textbf{SFT Dataset} & \textbf{Base} & \textbf{Smoothing} & \textbf{Accuracy} & \textbf{Entropy} & \textbf{ECE} & \textbf{RMS} \\
\midrule
\multirow{3}{*}{\textbf{\texttt{Base Model}}} & \textbf{\texttt{Gemma2-2B}} & None & 0.113 & 3.165 & 0.064 & 0.085 \\
& \textbf{\texttt{Mistral-7B}} & None & 0.197 & 3.053 & 0.032 & 0.067 \\
& \textbf{\texttt{Llama-3.1-8B}} & None & 0.136 & 1.658 & 0.188 & 0.152 \\
\midrule
\multirow{9.75}{*}{\textbf{\texttt{Alpaca}}} & \multirow{3}{*}{\textbf{\texttt{Gemma2-2B}}} & 0.0 & 0.159 & 2.785 & 0.066 & 0.101 \\
& & 0.1 & 0.155 & 2.912 & 0.068 & 0.093 \\
& & 0.2 & 0.154 & 2.907 & 0.063 & 0.091 \\
\cmidrule{2-7} & \multirow{3}{*}{\textbf{\texttt{Mistral-7B}}} & 0.0 & 0.189 & 2.169 & 0.115 & 0.144 \\
\bottomrule
\end{tabular}}
\end{table}

\begin{table}[h!]
\caption{Results on the \texttt{MMLU-ProX} subset for the \texttt{mr} language by model, SFT dataset, and label smoothing.}
\resizebox{\linewidth}{!}{\begin{tabular}{llc|cccc}
\toprule
\textbf{SFT Dataset} & \textbf{Base} & \textbf{Smoothing} & \textbf{Accuracy} & \textbf{Entropy} & \textbf{ECE} & \textbf{RMS} \\
\midrule
\multirow{3}{*}{\textbf{\texttt{Base Model}}} & \textbf{\texttt{Gemma2-2B}} & None & 0.110 & 3.163 & 0.074 & 0.091 \\
& \textbf{\texttt{Mistral-7B}} & None & 0.140 & 2.991 & 0.105 & 0.104 \\
& \textbf{\texttt{Llama-3.1-8B}} & None & 0.119 & 1.571 & 0.230 & 0.162 \\
\midrule
\multirow{9.75}{*}{\textbf{\texttt{Alpaca}}} & \multirow{3}{*}{\textbf{\texttt{Gemma2-2B}}} & 0.0 & 0.140 & 2.769 & 0.086 & 0.113 \\
& & 0.1 & 0.146 & 2.870 & 0.067 & 0.099 \\
& & 0.2 & 0.141 & 2.843 & 0.068 & 0.101 \\
\cmidrule{2-7} & \multirow{3}{*}{\textbf{\texttt{Mistral-7B}}} & 0.0 & 0.147 & 2.433 & 0.095 & 0.138 \\
\bottomrule
\end{tabular}}
\end{table}

\begin{table}[h!]
\caption{Results on the \texttt{MMLU-ProX} subset for the \texttt{ne} language by model, SFT dataset, and label smoothing.}
\resizebox{\linewidth}{!}{\begin{tabular}{llc|cccc}
\toprule
\textbf{SFT Dataset} & \textbf{Base} & \textbf{Smoothing} & \textbf{Accuracy} & \textbf{Entropy} & \textbf{ECE} & \textbf{RMS} \\
\midrule
\multirow{3}{*}{\textbf{\texttt{Base Model}}} & \textbf{\texttt{Gemma2-2B}} & None & 0.108 & 3.174 & 0.069 & 0.089 \\
& \textbf{\texttt{Mistral-7B}} & None & 0.141 & 2.997 & 0.105 & 0.104 \\
& \textbf{\texttt{Llama-3.1-8B}} & None & 0.123 & 1.764 & 0.128 & 0.132 \\
\midrule
\multirow{9.75}{*}{\textbf{\texttt{Alpaca}}} & \multirow{3}{*}{\textbf{\texttt{Gemma2-2B}}} & 0.0 & 0.145 & 2.764 & 0.074 & 0.110 \\
& & 0.1 & 0.145 & 2.893 & 0.064 & 0.096 \\
& & 0.2 & 0.140 & 2.866 & 0.067 & 0.096 \\
\cmidrule{2-7} & \multirow{3}{*}{\textbf{\texttt{Mistral-7B}}} & 0.0 & 0.148 & 2.429 & 0.091 & 0.134 \\
\bottomrule
\end{tabular}}
\end{table}

\begin{table}[h!]
\caption{Results on the \texttt{MMLU-ProX} subset for the \texttt{pt} language by model, SFT dataset, and label smoothing.}
\resizebox{\linewidth}{!}{\begin{tabular}{llc|cccc}
\toprule
\textbf{SFT Dataset} & \textbf{Base} & \textbf{Smoothing} & \textbf{Accuracy} & \textbf{Entropy} & \textbf{ECE} & \textbf{RMS} \\
\midrule
\multirow{3}{*}{\textbf{\texttt{Base Model}}} & \textbf{\texttt{Gemma2-2B}} & None & 0.127 & 3.145 & 0.069 & 0.086 \\
& \textbf{\texttt{Mistral-7B}} & None & 0.220 & 2.966 & 0.044 & 0.074 \\
& \textbf{\texttt{Llama-3.1-8B}} & None & 0.147 & 1.580 & 0.237 & 0.163 \\
\midrule
\multirow{9.75}{*}{\textbf{\texttt{Alpaca}}} & \multirow{3}{*}{\textbf{\texttt{Gemma2-2B}}} & 0.0 & 0.174 & 2.631 & 0.081 & 0.106 \\
& & 0.1 & 0.175 & 2.813 & 0.060 & 0.091 \\
& & 0.2 & 0.170 & 2.852 & 0.060 & 0.087 \\
\cmidrule{2-7} & \multirow{3}{*}{\textbf{\texttt{Mistral-7B}}} & 0.0 & 0.216 & 2.055 & 0.092 & 0.132 \\
\bottomrule
\end{tabular}}
\end{table}

\begin{table}[h!]
\caption{Results on the \texttt{MMLU-ProX} subset for the \texttt{rj} language by model, SFT dataset, and label smoothing.}
\resizebox{\linewidth}{!}{\begin{tabular}{llc|cccc}
\toprule
\textbf{SFT Dataset} & \textbf{Base} & \textbf{Smoothing} & \textbf{Accuracy} & \textbf{Entropy} & \textbf{ECE} & \textbf{RMS} \\
\midrule
\multirow{3}{*}{\textbf{\texttt{Base Model}}} & \textbf{\texttt{Gemma2-2B}} & None & 0.111 & 3.137 & 0.077 & 0.089 \\
& \textbf{\texttt{Mistral-7B}} & None & 0.211 & 2.983 & 0.040 & 0.072 \\
& \textbf{\texttt{Llama-3.1-8B}} & None & 0.141 & 1.614 & 0.192 & 0.152 \\
\midrule
\multirow{9.75}{*}{\textbf{\texttt{Alpaca}}} & \multirow{3}{*}{\textbf{\texttt{Gemma2-2B}}} & 0.0 & 0.165 & 2.707 & 0.069 & 0.104 \\
& & 0.1 & 0.166 & 2.862 & 0.055 & 0.087 \\
& & 0.2 & 0.162 & 2.870 & 0.062 & 0.091 \\
\cmidrule{2-7} & \multirow{3}{*}{\textbf{\texttt{Mistral-7B}}} & 0.0 & 0.213 & 2.079 & 0.145 & 0.151 \\
\bottomrule
\end{tabular}}
\end{table}

\begin{table}[h!]
\caption{Results on the \texttt{MMLU-ProX} subset for the \texttt{sr} language by model, SFT dataset, and label smoothing.}
\resizebox{\linewidth}{!}{\begin{tabular}{llc|cccc}
\toprule
\textbf{SFT Dataset} & \textbf{Base} & \textbf{Smoothing} & \textbf{Accuracy} & \textbf{Entropy} & \textbf{ECE} & \textbf{RMS} \\
\midrule
\multirow{3}{*}{\textbf{\texttt{Base Model}}} & \textbf{\texttt{Gemma2-2B}} & None & 0.116 & 3.103 & 0.085 & 0.095 \\
& \textbf{\texttt{Mistral-7B}} & None & 0.189 & 3.012 & 0.058 & 0.081 \\
& \textbf{\texttt{Llama-3.1-8B}} & None & 0.131 & 1.630 & 0.213 & 0.160 \\
\midrule
\multirow{9.75}{*}{\textbf{\texttt{Alpaca}}} & \multirow{3}{*}{\textbf{\texttt{Gemma2-2B}}} & 0.0 & 0.160 & 2.635 & 0.085 & 0.113 \\
& & 0.1 & 0.163 & 2.818 & 0.055 & 0.091 \\
& & 0.2 & 0.162 & 2.816 & 0.050 & 0.092 \\
\cmidrule{2-7} & \multirow{3}{*}{\textbf{\texttt{Mistral-7B}}} & 0.0 & 0.199 & 2.192 & 0.129 & 0.147 \\
\bottomrule
\end{tabular}}
\end{table}

\begin{table}[h!]
\caption{Results on the \texttt{MMLU-ProX} subset for the \texttt{sw} language by model, SFT dataset, and label smoothing.}
\resizebox{\linewidth}{!}{%
}
\end{table}

\begin{table}[h!]
\caption{Results on the \texttt{MMLU-ProX} subset for the \texttt{te} language by model, SFT dataset, and label smoothing.}
\resizebox{\linewidth}{!}{%
}
\end{table}

\begin{table}[h!]
\caption{Results on the \texttt{MMLU-ProX} subset for the \texttt{th} language by model, SFT dataset, and label smoothing.}
\resizebox{\linewidth}{!}{%
}
\end{table}

\begin{table}[h!]
\caption{Results on the \texttt{MMLU-ProX} subset for the \texttt{uk} language by model, SFT dataset, and label smoothing.}
\resizebox{\linewidth}{!}{%
}
\end{table}

\begin{table}[h!]
\caption{Results on the \texttt{MMLU-ProX} subset for the \texttt{ur} language by model, SFT dataset, and label smoothing.}
\resizebox{\linewidth}{!}{%
}
\end{table}

\begin{table}[h!]
\caption{Results on the \texttt{MMLU-ProX} subset for the \texttt{vi} language by model, SFT dataset, and label smoothing.}
\resizebox{\linewidth}{!}{%
}
\end{table}

\begin{table}[h!]
\caption{Results on the \texttt{MMLU-ProX} subset for the \texttt{wo} language by model, SFT dataset, and label smoothing.}
\resizebox{\linewidth}{!}{%
}
\end{table}

\begin{table}[h!]
\caption{Results on the \texttt{MMLU-ProX} subset for the \texttt{yo} language by model, SFT dataset, and label smoothing.}
\resizebox{\linewidth}{!}{%
}
\end{table}

\begin{table}[h!]
\caption{Results on the \texttt{MMLU-ProX} subset for the \texttt{zh} language by model, SFT dataset, and label smoothing.}
\resizebox{\linewidth}{!}{%
}
\end{table}

\begin{table}[h!]
\caption{Results on the \texttt{MMLU-ProX} subset for the \texttt{zu} language by model, SFT dataset, and label smoothing.}
\resizebox{\linewidth}{!}{%
}
\end{table}

\clearpage

\section{Individual Reliability Plots}

\subsection{GlobalMMLU}
\begin{figure}[h!]\centering\resizebox{\linewidth}{!}{\includegraphics[width=\linewidth]{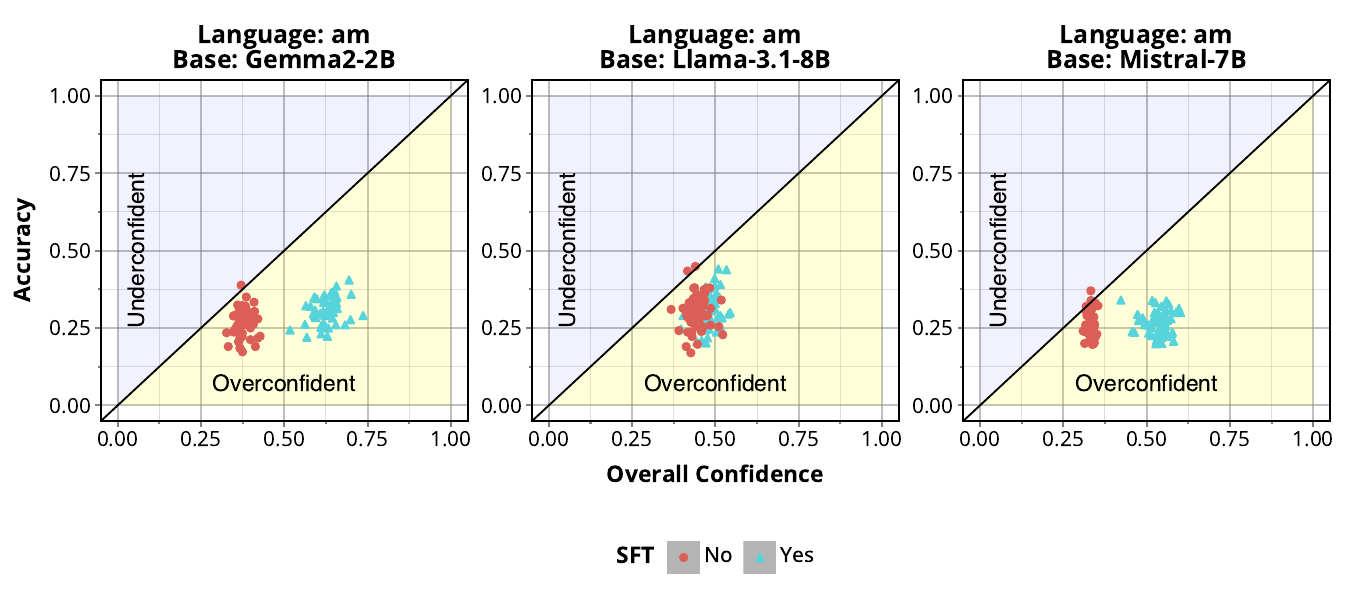}}
\caption{Reliability diagrams for the \textbf{\texttt{GlobalMMLU}} dataset for the \texttt{am} language.}\label{fig:globalmmlu-base-am}\end{figure}
\begin{figure}[h!]\centering\resizebox{\linewidth}{!}{\includegraphics[width=\linewidth]{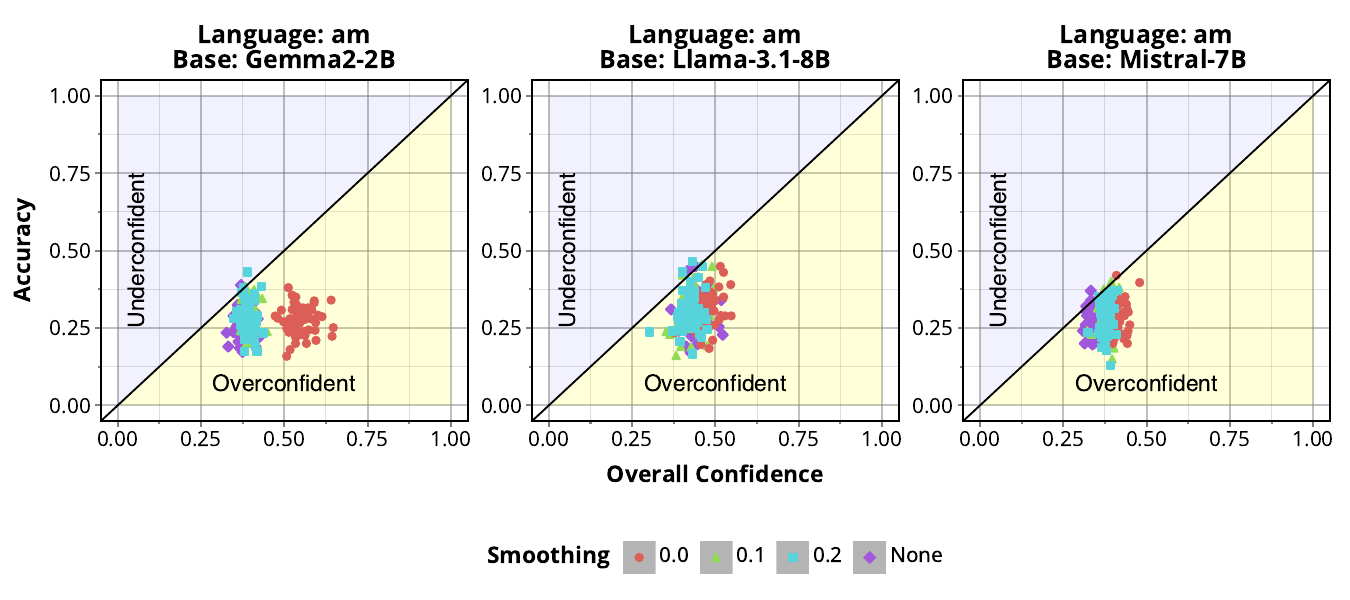}}
\caption{Reliability diagrams for the \textbf{\texttt{GlobalMMLU}} dataset for the \texttt{am} language after instruction-tuning on the \textbf{\texttt{Tulu3Mixture}} dataset.}\label{fig:globalmmlu-Tulu3Mixture-am}\end{figure}
\begin{figure}[h!]\centering\resizebox{\linewidth}{!}{\includegraphics[width=\linewidth]{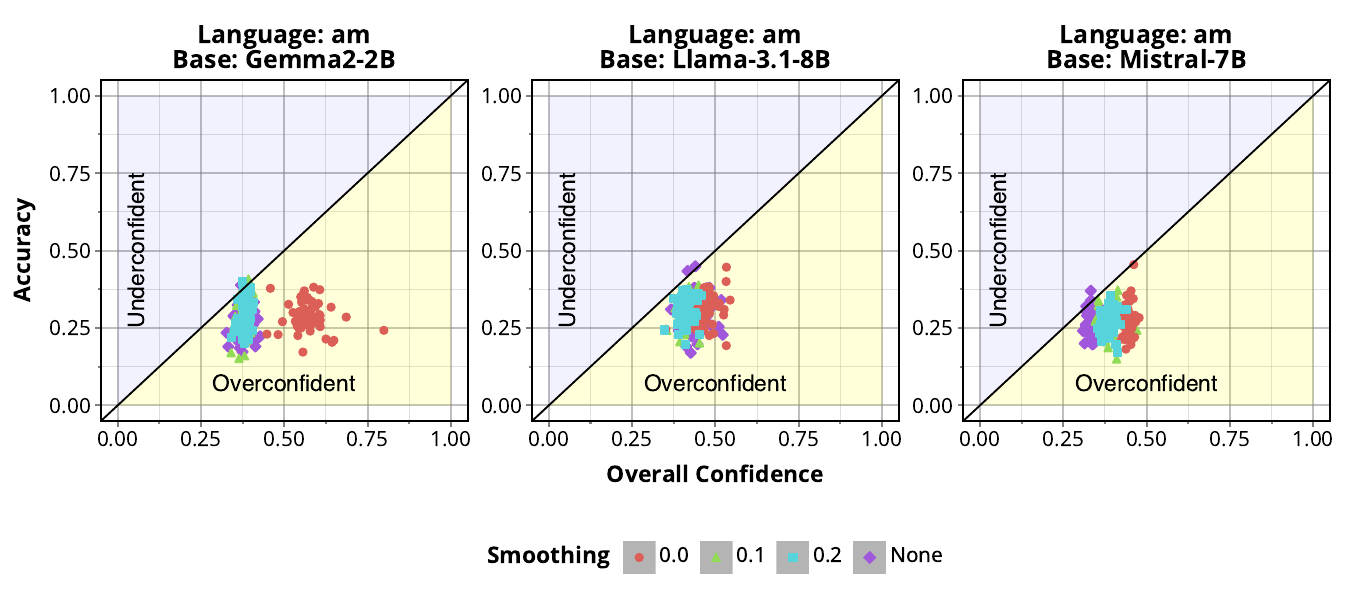}}
\caption{Reliability diagrams for the \textbf{\texttt{GlobalMMLU}} dataset for the \texttt{am} language after instruction-tuning on the \textbf{\texttt{OpenHermes}} dataset.}\label{fig:globalmmlu-OpenHermes-am}\end{figure}

\begin{figure}[h!]\centering\resizebox{\linewidth}{!}{\includegraphics[width=\linewidth]{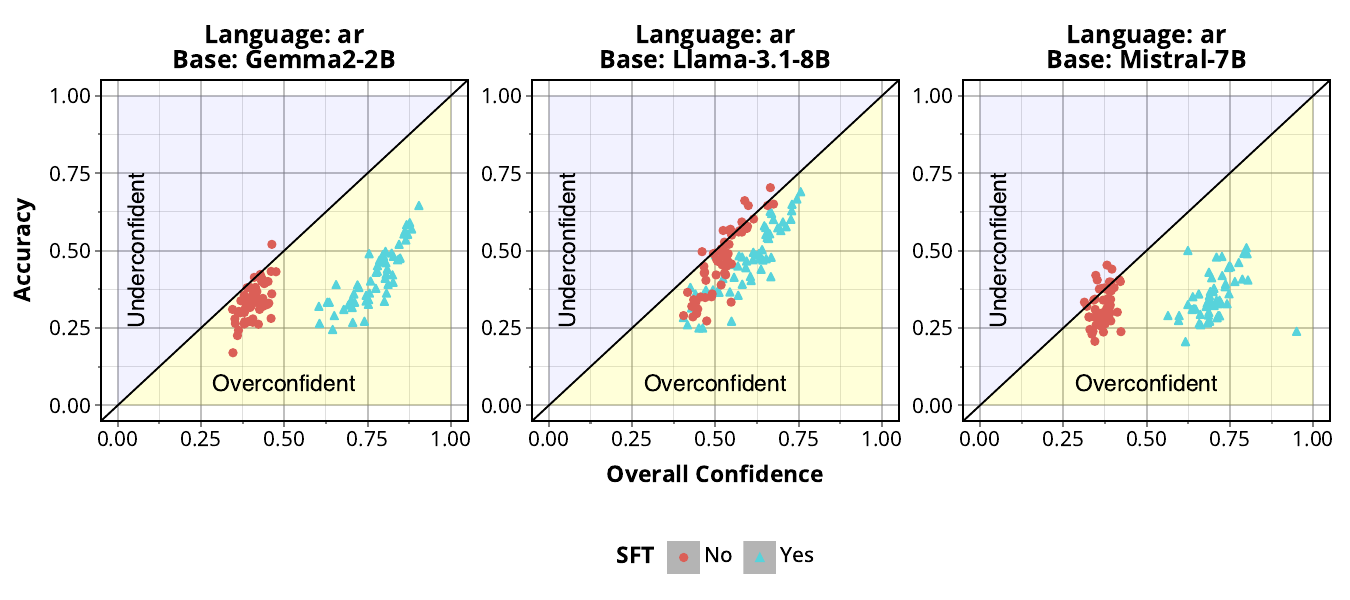}}
\caption{Reliability diagrams for the \textbf{\texttt{GlobalMMLU}} dataset for the \texttt{ar} language.}\label{fig:globalmmlu-base-ar}\end{figure}
\begin{figure}[h!]\centering\resizebox{\linewidth}{!}{\includegraphics[width=\linewidth]{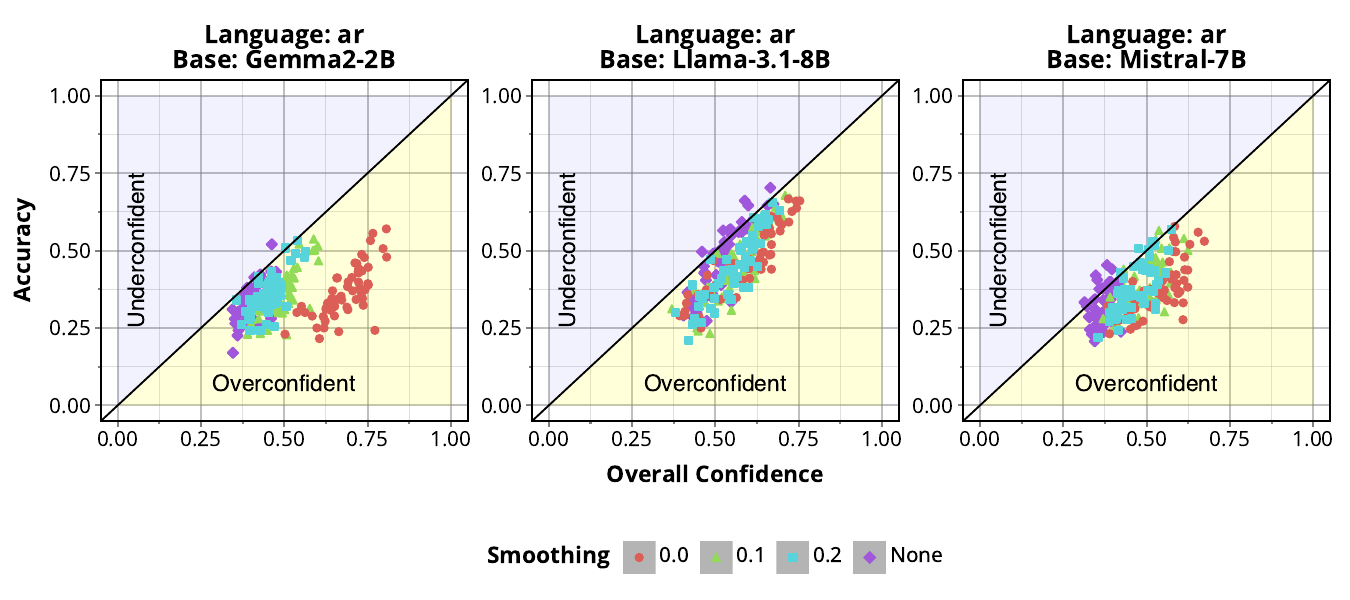}}
\caption{Reliability diagrams for the \textbf{\texttt{GlobalMMLU}} dataset for the \texttt{ar} language after instruction-tuning on the \textbf{\texttt{Tulu3Mixture}} dataset.}\label{fig:globalmmlu-Tulu3Mixture-ar}\end{figure}
\begin{figure}[h!]\centering\resizebox{\linewidth}{!}{\includegraphics[width=\linewidth]{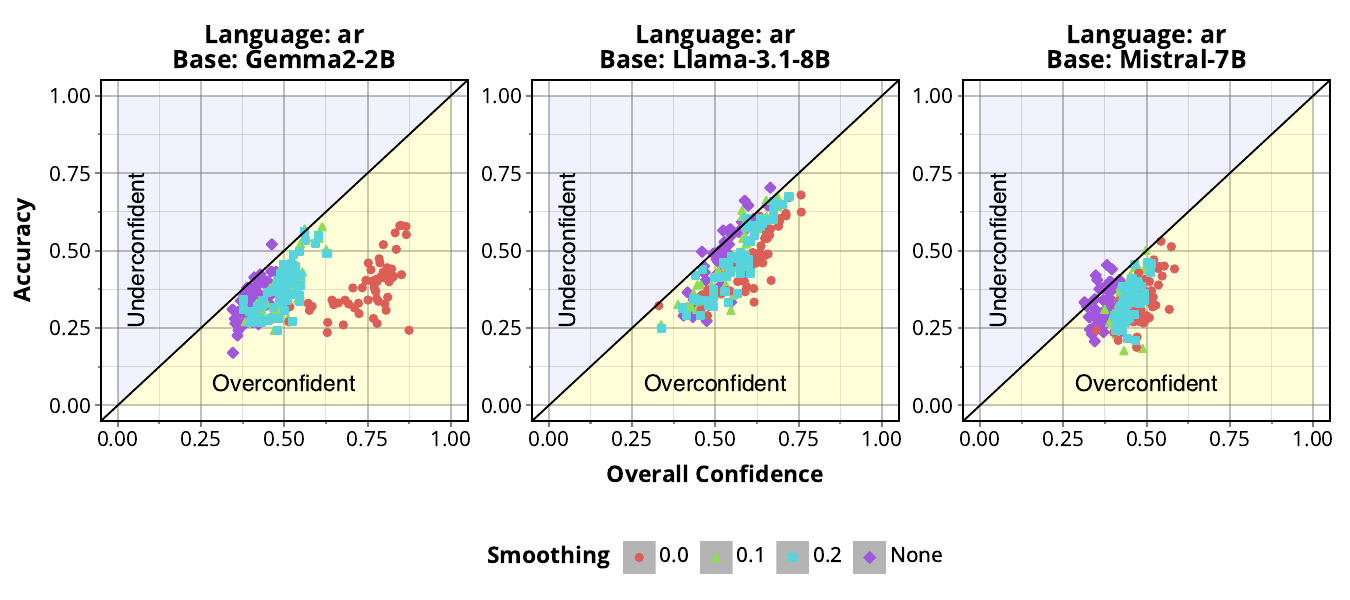}}
\caption{Reliability diagrams for the \textbf{\texttt{GlobalMMLU}} dataset for the \texttt{ar} language after instruction-tuning on the \textbf{\texttt{OpenHermes}} dataset.}\label{fig:globalmmlu-OpenHermes-ar}\end{figure}

\begin{figure}[h!]\centering\resizebox{\linewidth}{!}{\includegraphics[width=\linewidth]{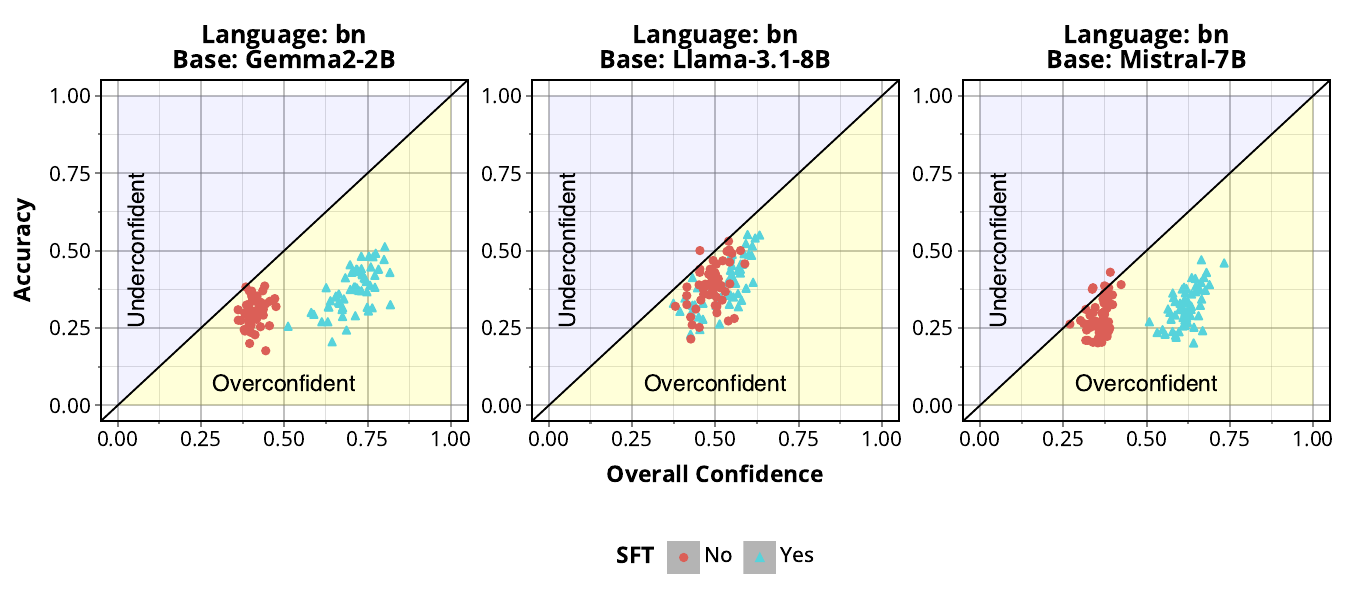}}
\caption{Reliability diagrams for the \textbf{\texttt{GlobalMMLU}} dataset for the \texttt{bn} language.}\label{fig:globalmmlu-base-bn}\end{figure}
\begin{figure}[h!]\centering\resizebox{\linewidth}{!}{\includegraphics[width=\linewidth]{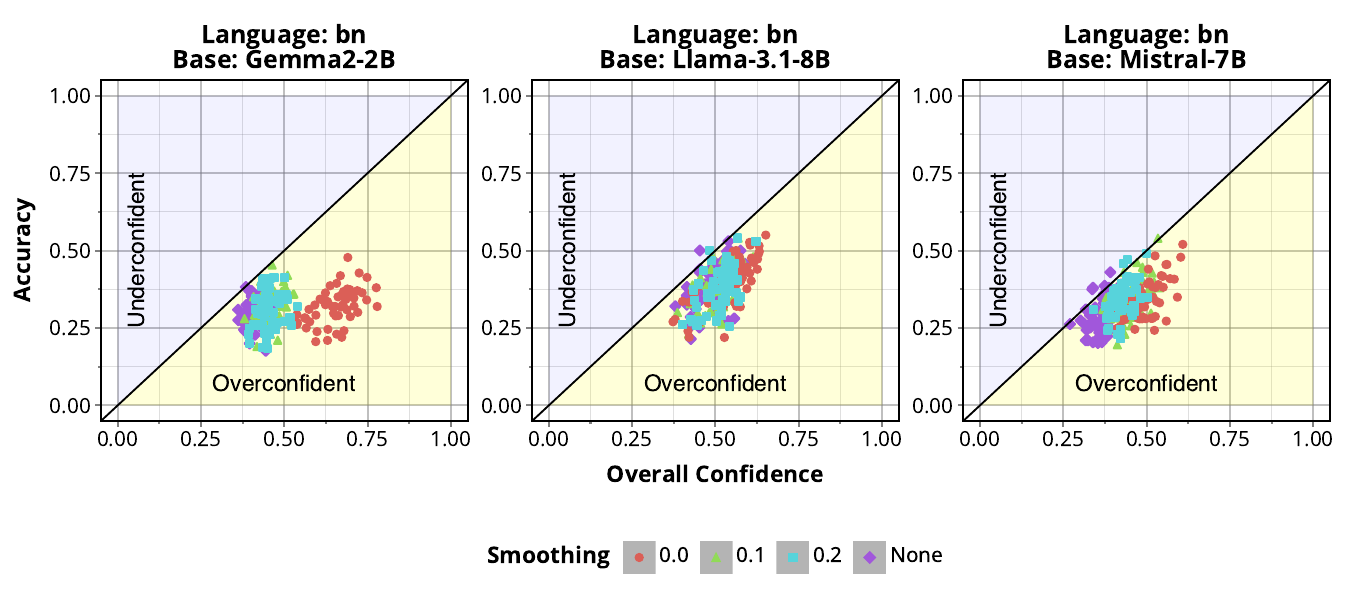}}
\caption{Reliability diagrams for the \textbf{\texttt{GlobalMMLU}} dataset for the \texttt{bn} language after instruction-tuning on the \textbf{\texttt{Tulu3Mixture}} dataset.}\label{fig:globalmmlu-Tulu3Mixture-bn}\end{figure}
\begin{figure}[h!]\centering\resizebox{\linewidth}{!}{\includegraphics[width=\linewidth]{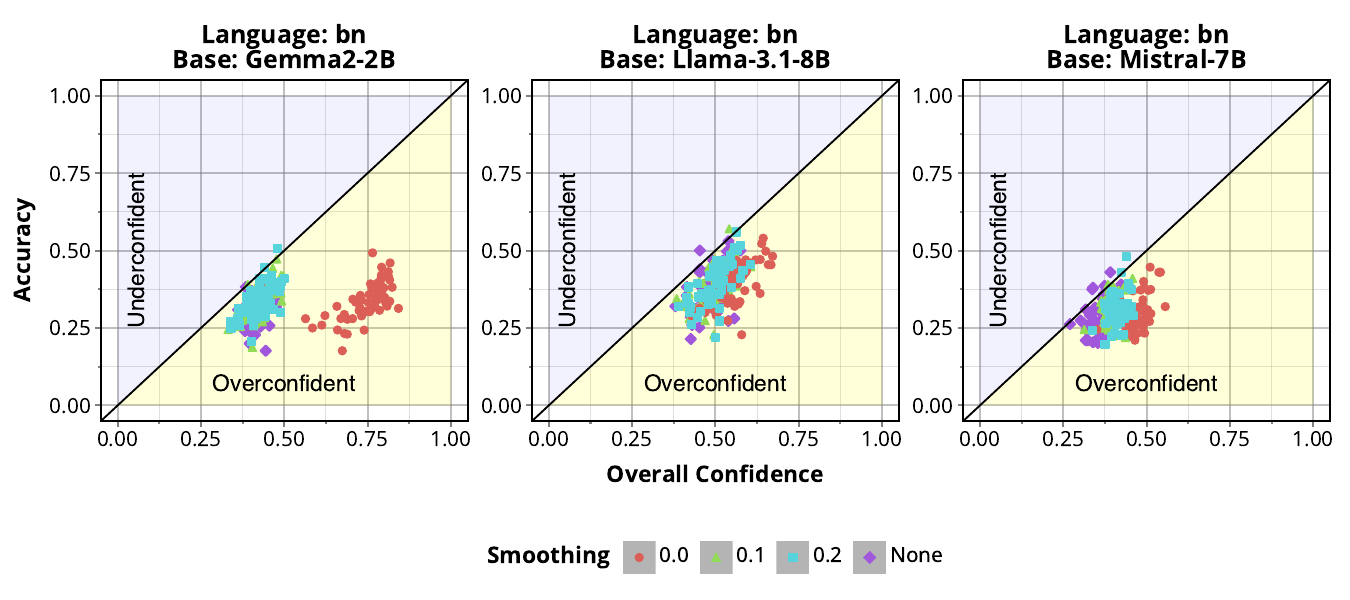}}
\caption{Reliability diagrams for the \textbf{\texttt{GlobalMMLU}} dataset for the \texttt{bn} language after instruction-tuning on the \textbf{\texttt{OpenHermes}} dataset.}\label{fig:globalmmlu-OpenHermes-bn}\end{figure}

\begin{figure}[h!]\centering\resizebox{\linewidth}{!}{\includegraphics[width=\linewidth]{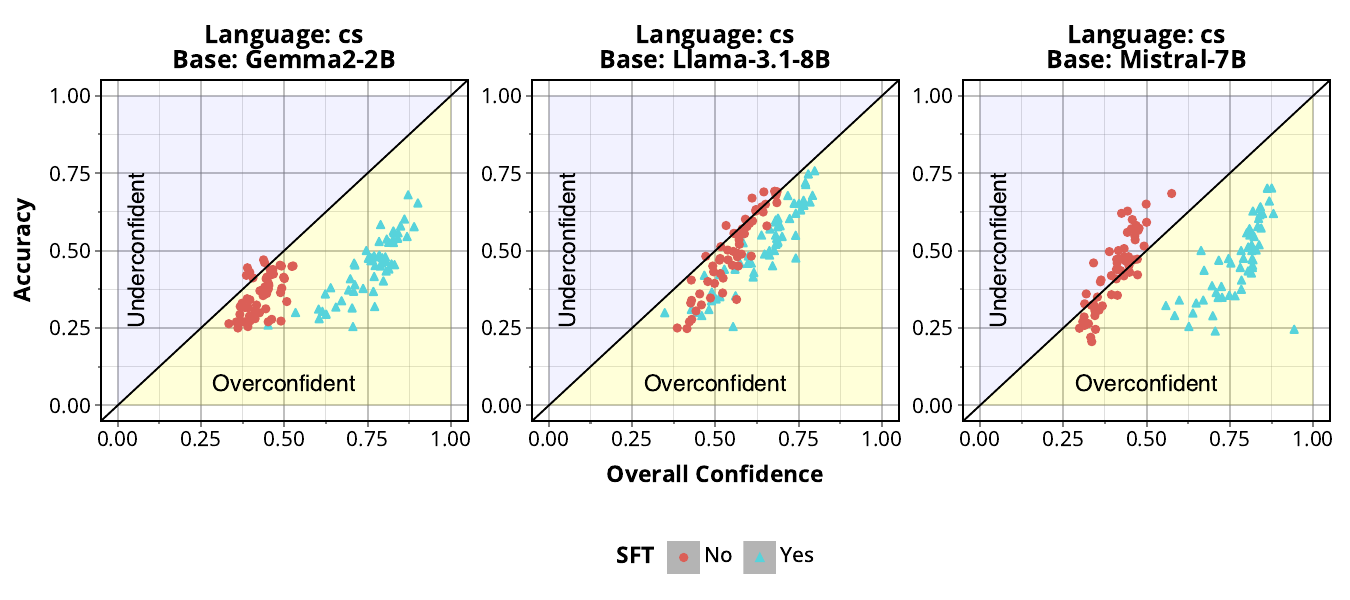}}
\caption{Reliability diagrams for the \textbf{\texttt{GlobalMMLU}} dataset for the \texttt{cs} language.}\label{fig:globalmmlu-base-cs}\end{figure}
\begin{figure}[h!]\centering\resizebox{\linewidth}{!}{\includegraphics[width=\linewidth]{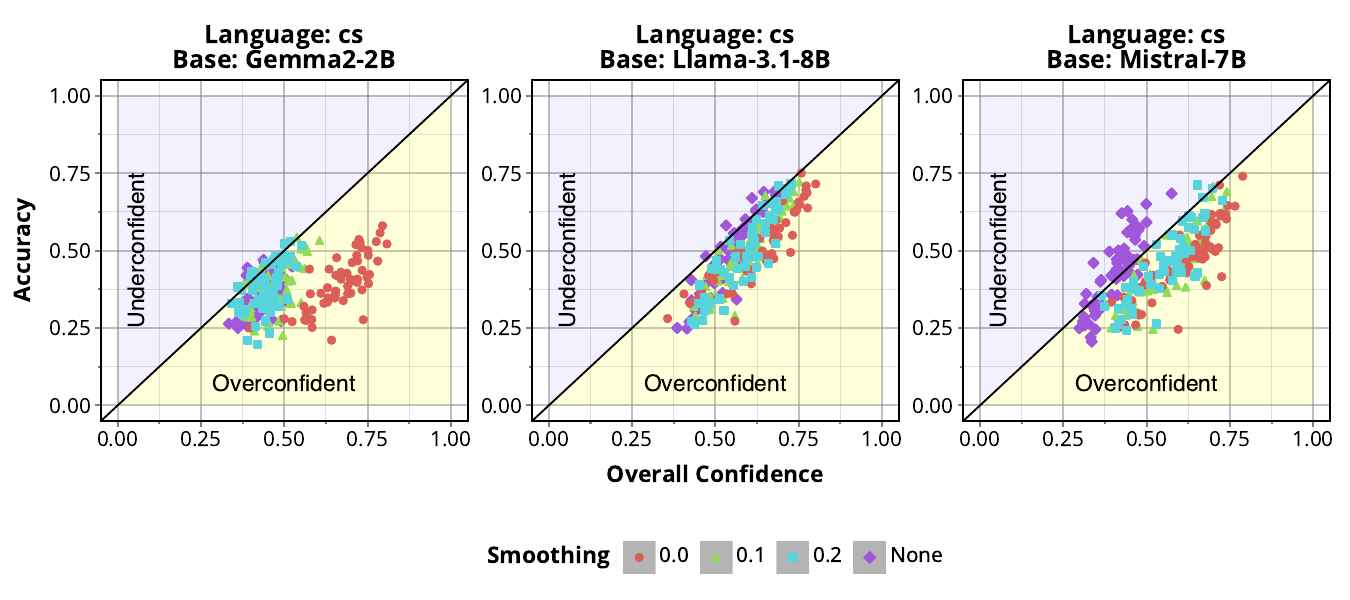}}
\caption{Reliability diagrams for the \textbf{\texttt{GlobalMMLU}} dataset for the \texttt{cs} language after instruction-tuning on the \textbf{\texttt{Tulu3Mixture}} dataset.}\label{fig:globalmmlu-Tulu3Mixture-cs}\end{figure}
\begin{figure}[h!]\centering\resizebox{\linewidth}{!}{\includegraphics[width=\linewidth]{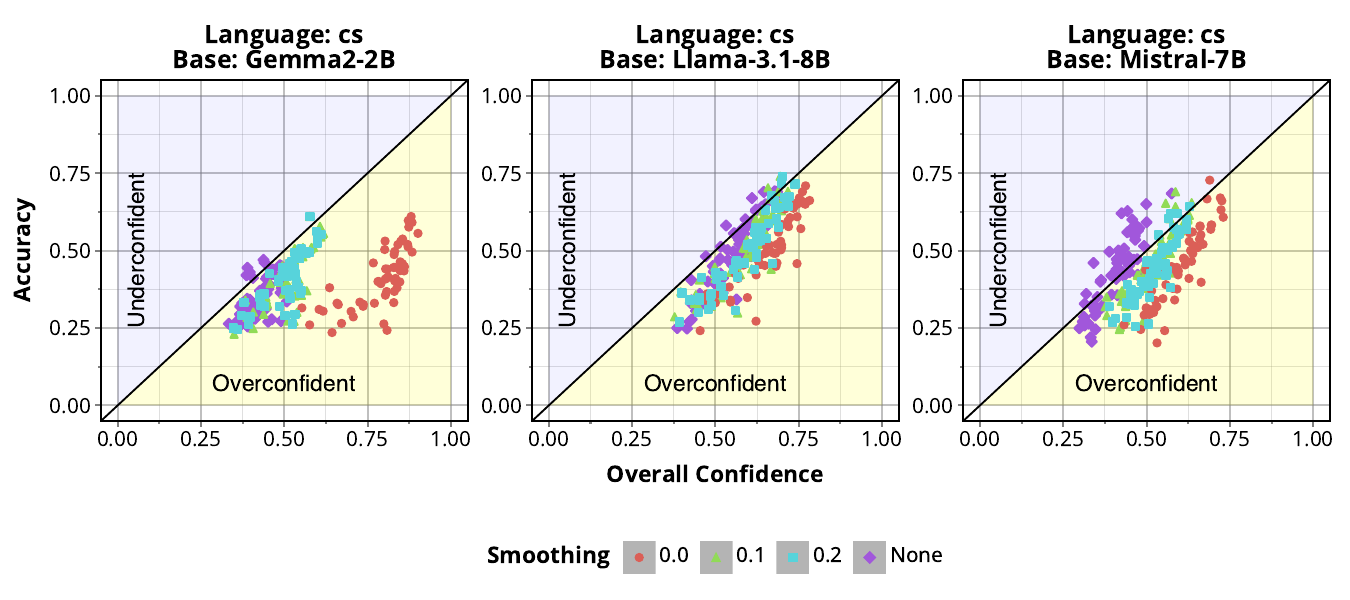}}
\caption{Reliability diagrams for the \textbf{\texttt{GlobalMMLU}} dataset for the \texttt{cs} language after instruction-tuning on the \textbf{\texttt{OpenHermes}} dataset.}\label{fig:globalmmlu-OpenHermes-cs}\end{figure}

\begin{figure}[h!]\centering\resizebox{\linewidth}{!}{\includegraphics[width=\linewidth]{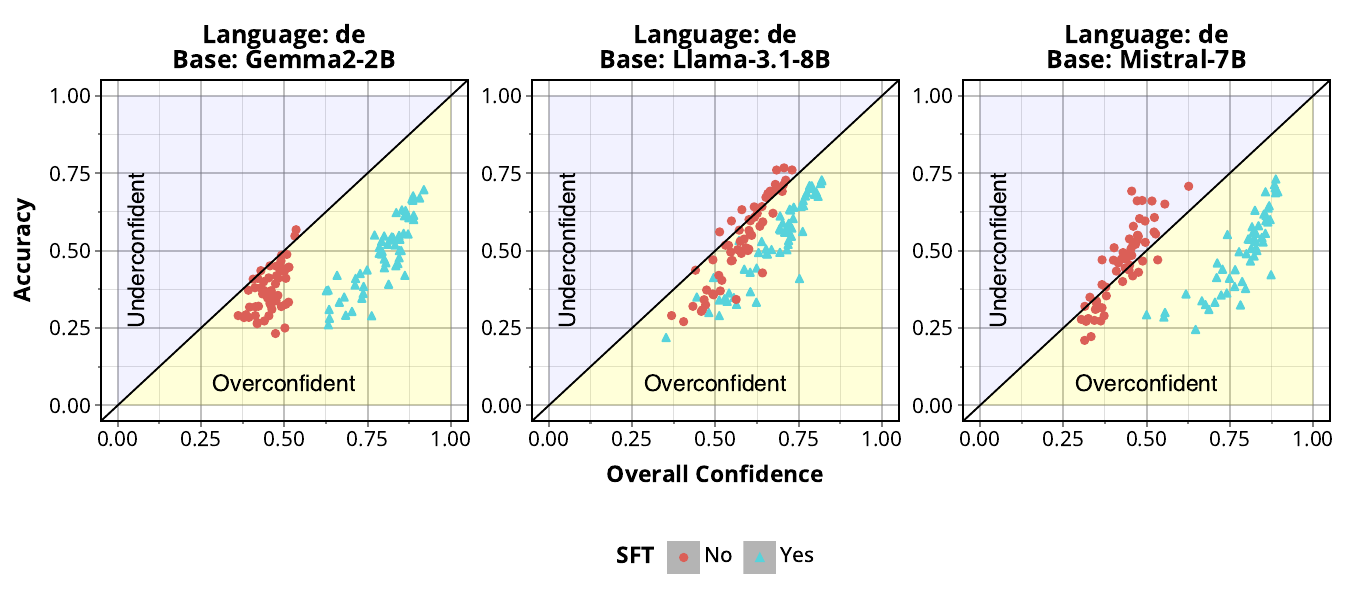}}
\caption{Reliability diagrams for the \textbf{\texttt{GlobalMMLU}} dataset for the \texttt{de} language.}\label{fig:globalmmlu-base-de}\end{figure}
\begin{figure}[h!]\centering\resizebox{\linewidth}{!}{\includegraphics[width=\linewidth]{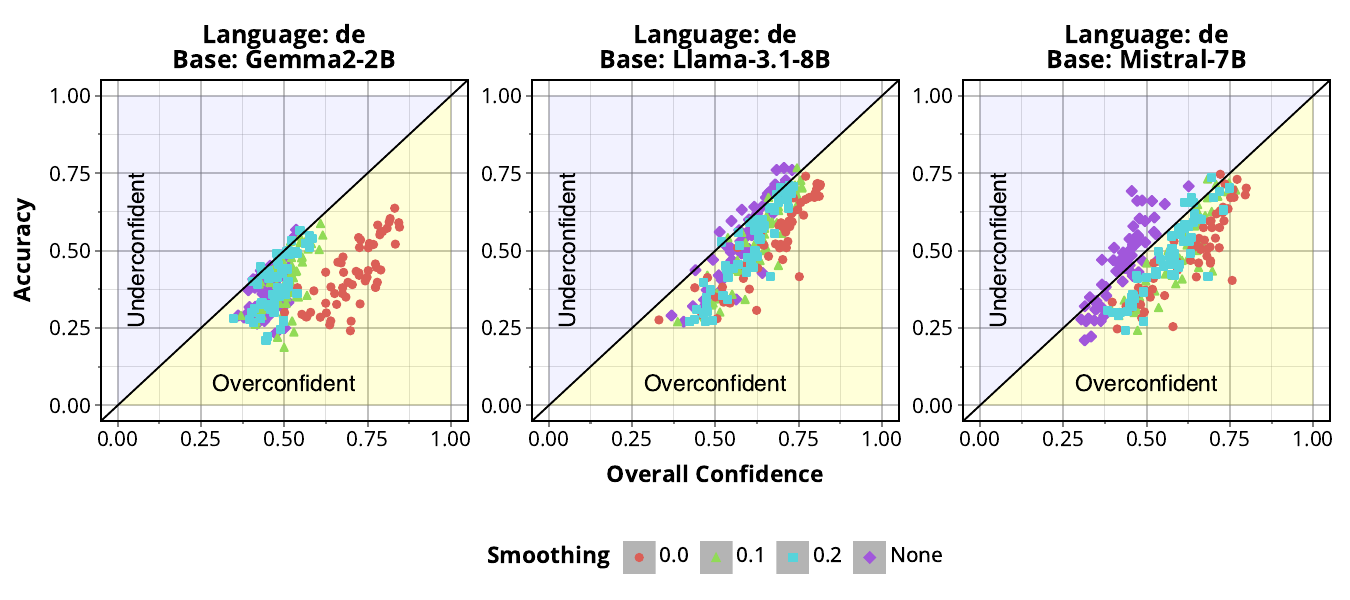}}
\caption{Reliability diagrams for the \textbf{\texttt{GlobalMMLU}} dataset for the \texttt{de} language after instruction-tuning on the \textbf{\texttt{Tulu3Mixture}} dataset.}\label{fig:globalmmlu-Tulu3Mixture-de}\end{figure}
\begin{figure}[h!]\centering\resizebox{\linewidth}{!}{\includegraphics[width=\linewidth]{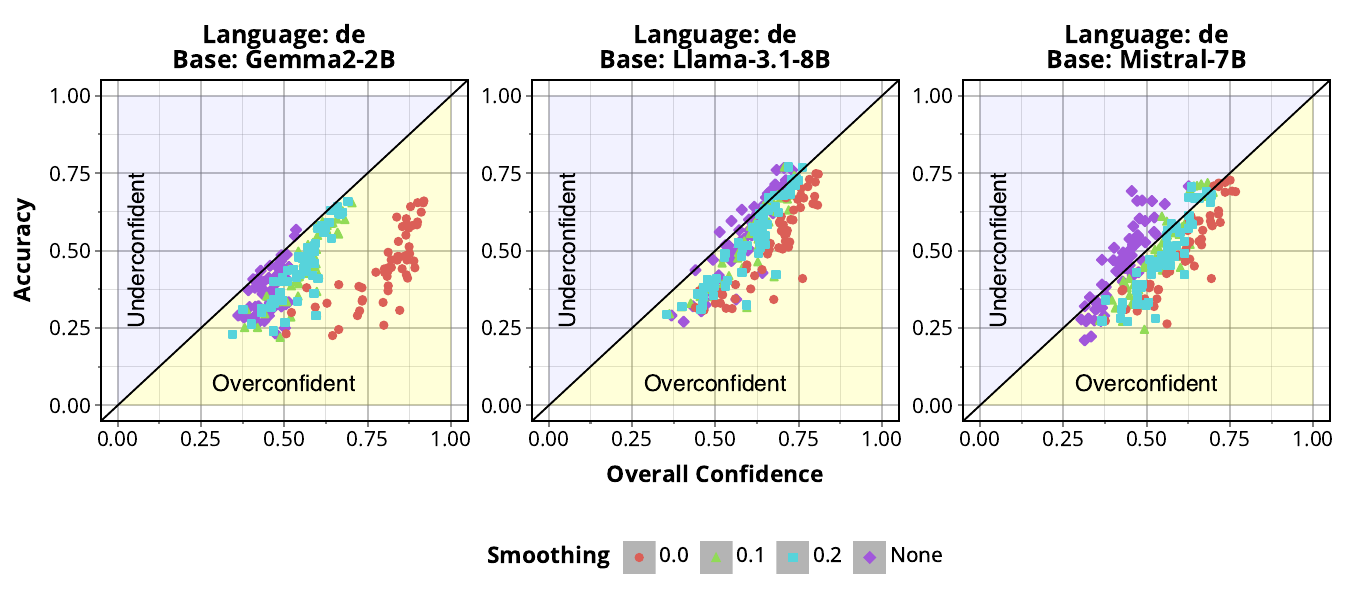}}
\caption{Reliability diagrams for the \textbf{\texttt{GlobalMMLU}} dataset for the \texttt{de} language after instruction-tuning on the \textbf{\texttt{OpenHermes}} dataset.}\label{fig:globalmmlu-OpenHermes-de}\end{figure}

\begin{figure}[h!]\centering\resizebox{\linewidth}{!}{\includegraphics[width=\linewidth]{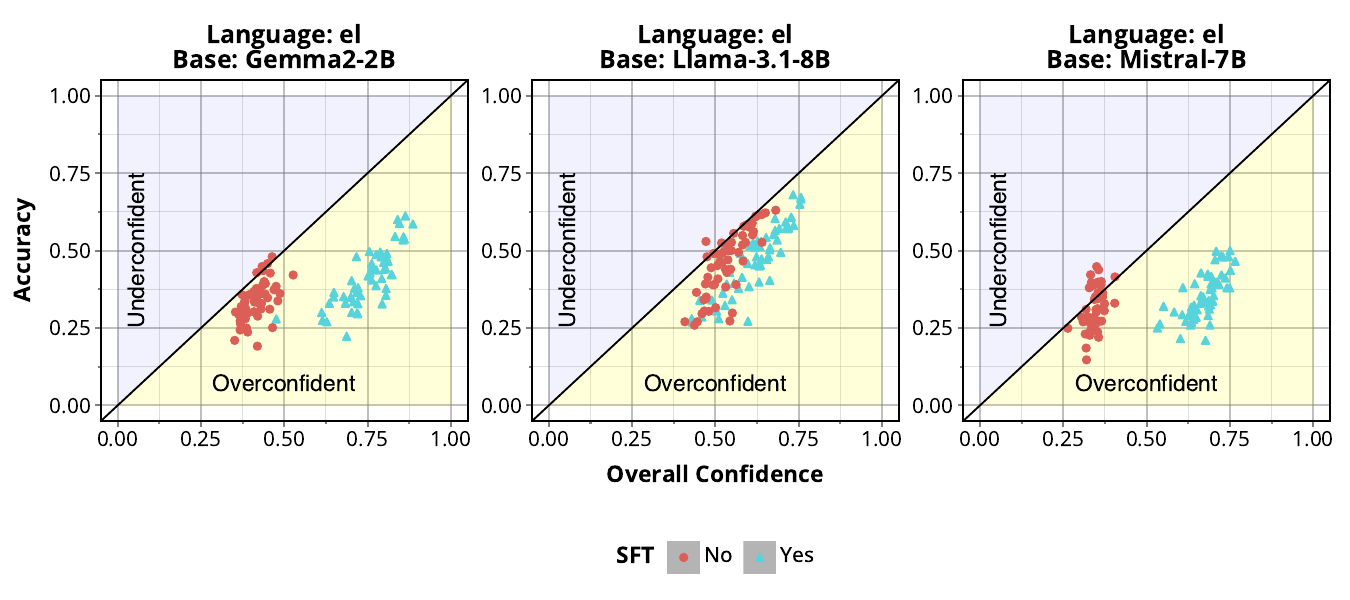}}
\caption{Reliability diagrams for the \textbf{\texttt{GlobalMMLU}} dataset for the \texttt{el} language.}\label{fig:globalmmlu-base-el}\end{figure}
\begin{figure}[h!]\centering\resizebox{\linewidth}{!}{\includegraphics[width=\linewidth]{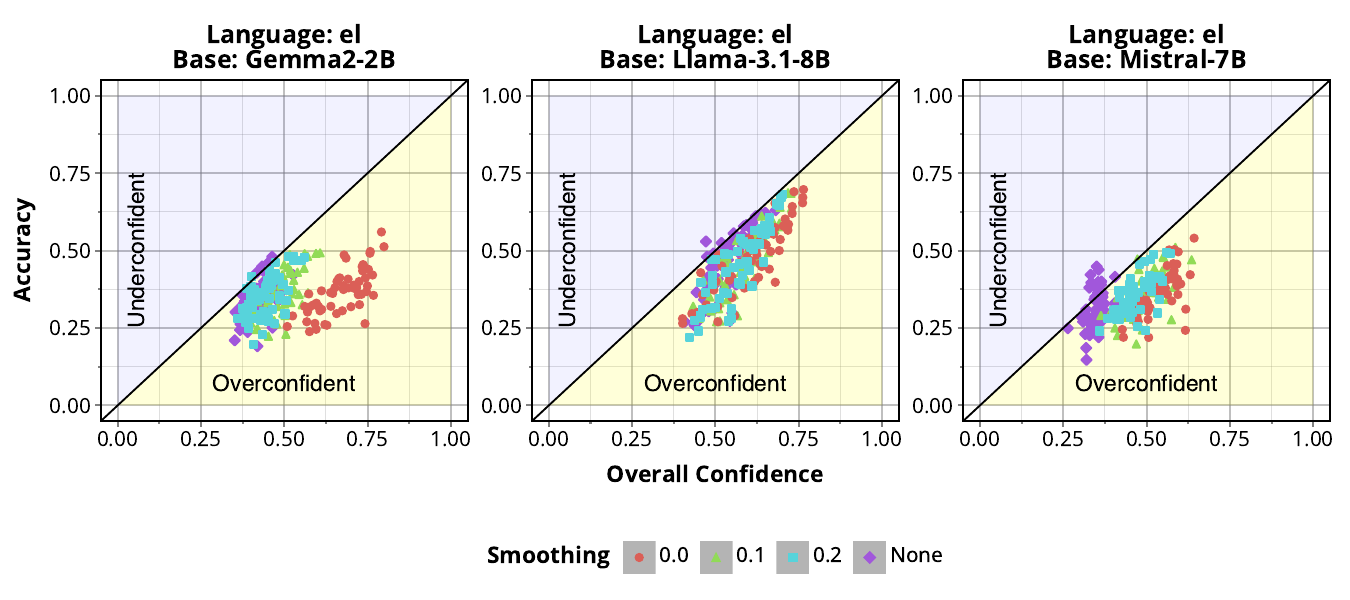}}
\caption{Reliability diagrams for the \textbf{\texttt{GlobalMMLU}} dataset for the \texttt{el} language after instruction-tuning on the \textbf{\texttt{Tulu3Mixture}} dataset.}\label{fig:globalmmlu-Tulu3Mixture-el}\end{figure}
\begin{figure}[h!]\centering\resizebox{\linewidth}{!}{\includegraphics[width=\linewidth]{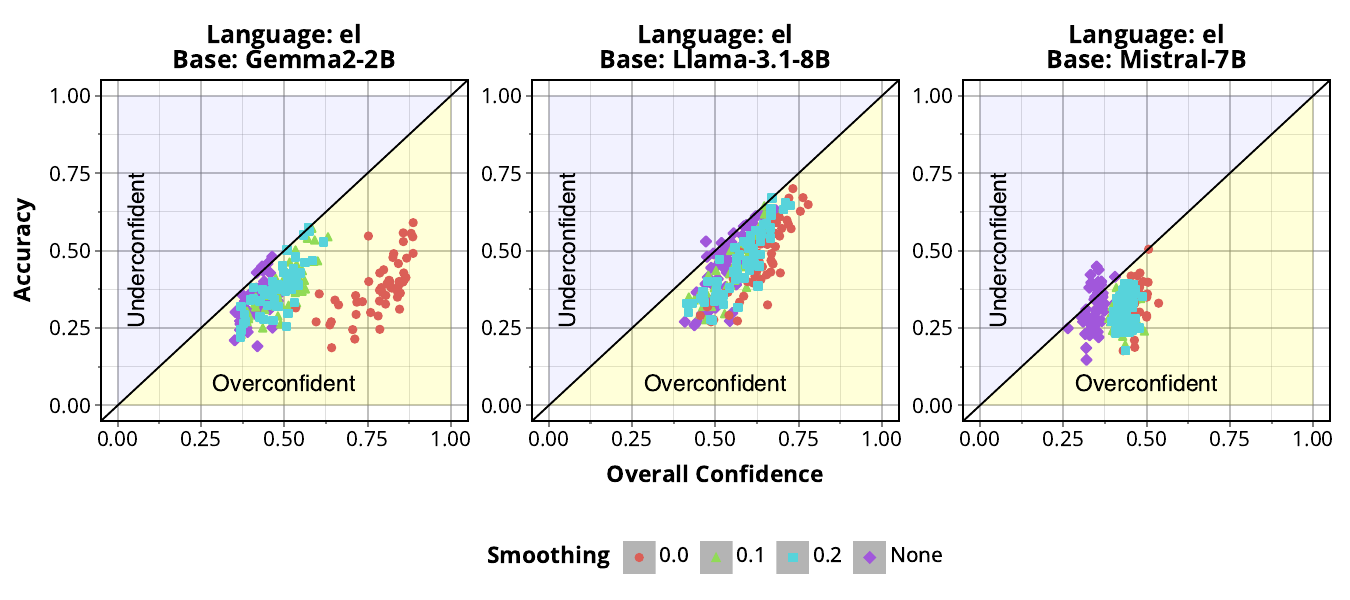}}
\caption{Reliability diagrams for the \textbf{\texttt{GlobalMMLU}} dataset for the \texttt{el} language after instruction-tuning on the \textbf{\texttt{OpenHermes}} dataset.}\label{fig:globalmmlu-OpenHermes-el}\end{figure}

\begin{figure}[h!]\centering\resizebox{\linewidth}{!}{\includegraphics[width=\linewidth]{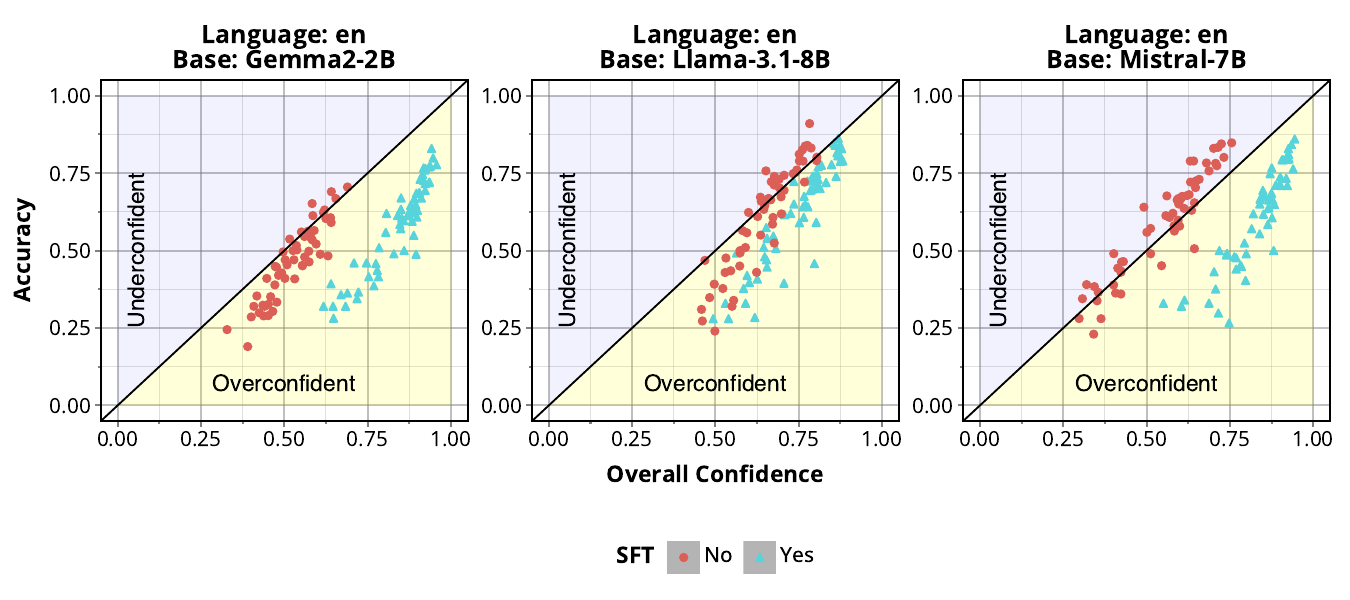}}
\caption{Reliability diagrams for the \textbf{\texttt{GlobalMMLU}} dataset for the \texttt{en} language.}\label{fig:globalmmlu-base-en}\end{figure}
\begin{figure}[h!]\centering\resizebox{\linewidth}{!}{\includegraphics[width=\linewidth]{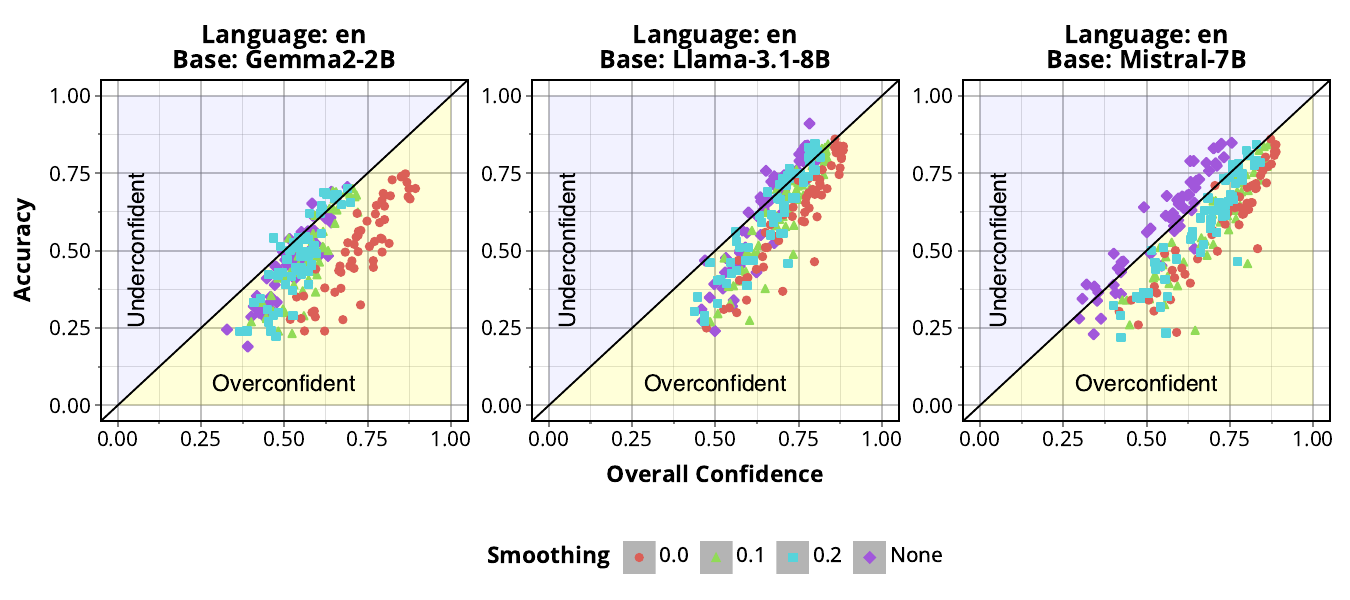}}
\caption{Reliability diagrams for the \textbf{\texttt{GlobalMMLU}} dataset for the \texttt{en} language after instruction-tuning on the \textbf{\texttt{Tulu3Mixture}} dataset.}\label{fig:globalmmlu-Tulu3Mixture-en}\end{figure}
\begin{figure}[h!]\centering\resizebox{\linewidth}{!}{\includegraphics[width=\linewidth]{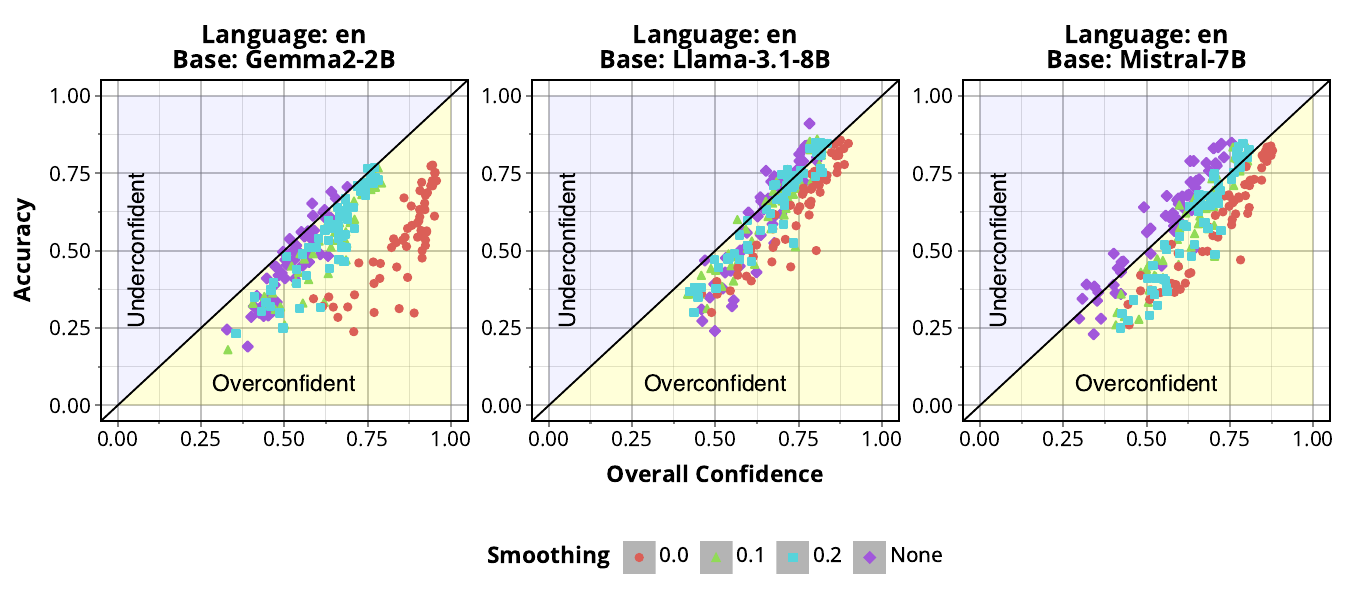}}
\caption{Reliability diagrams for the \textbf{\texttt{GlobalMMLU}} dataset for the \texttt{en} language after instruction-tuning on the \textbf{\texttt{OpenHermes}} dataset.}\label{fig:globalmmlu-OpenHermes-en}\end{figure}

\begin{figure}[h!]\centering\resizebox{\linewidth}{!}{\includegraphics[width=\linewidth]{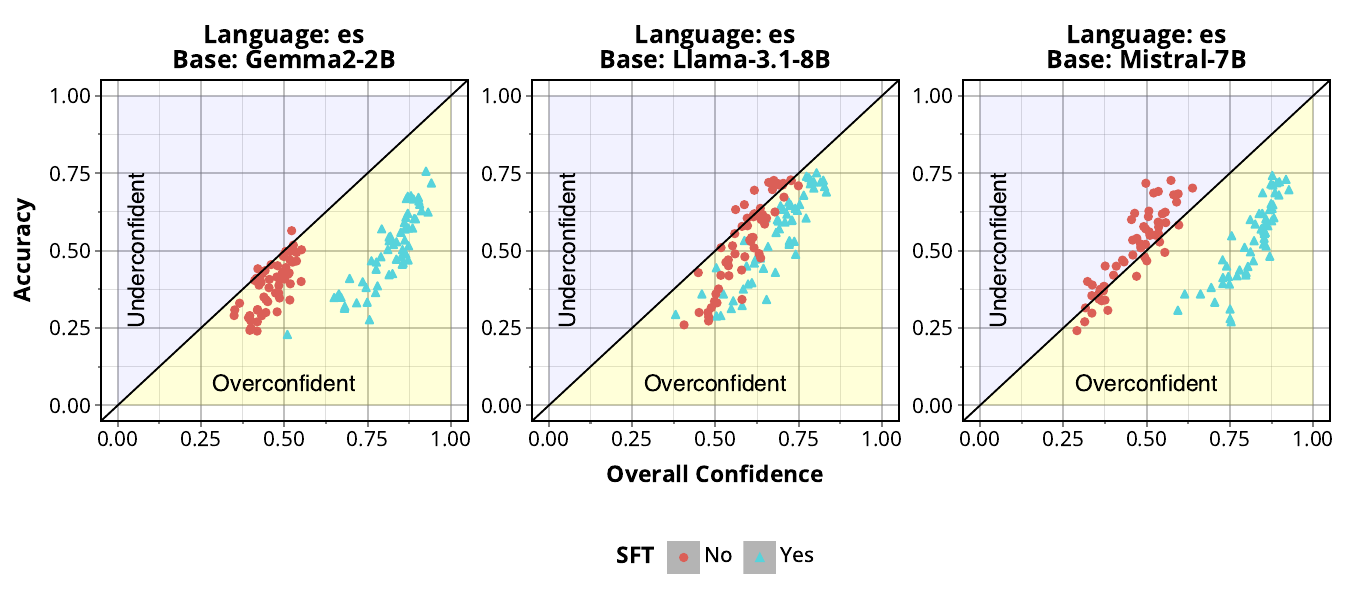}}
\caption{Reliability diagrams for the \textbf{\texttt{GlobalMMLU}} dataset for the \texttt{es} language.}\label{fig:globalmmlu-base-es}\end{figure}
\begin{figure}[h!]\centering\resizebox{\linewidth}{!}{\includegraphics[width=\linewidth]{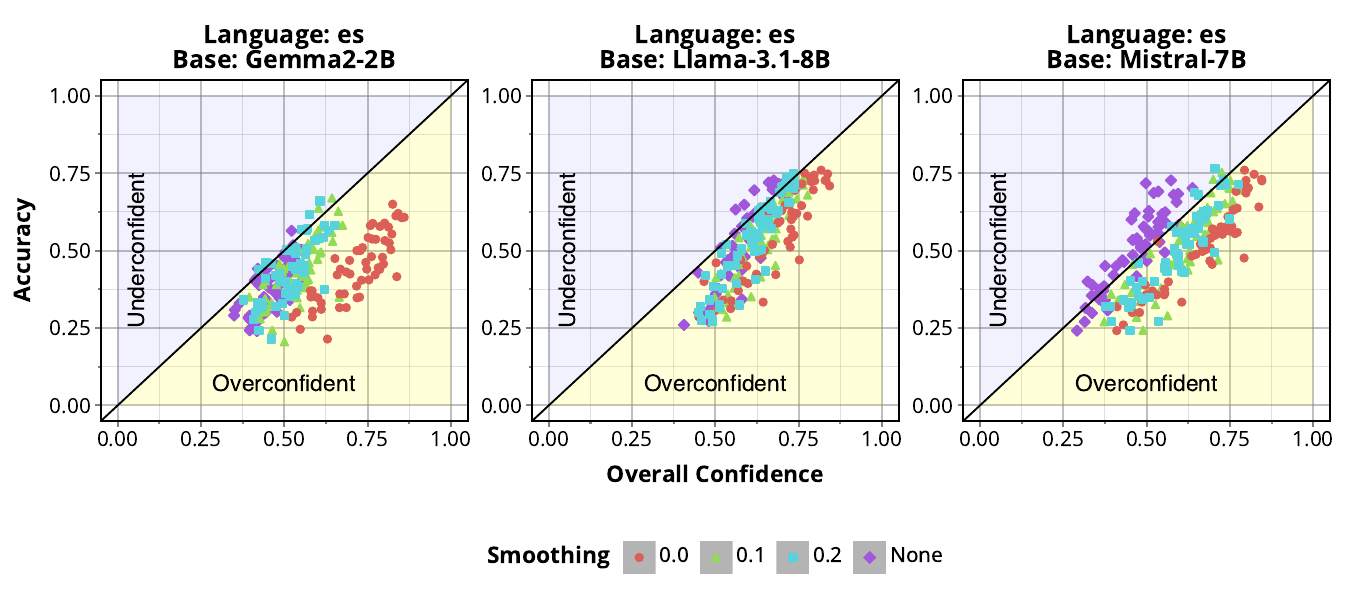}}
\caption{Reliability diagrams for the \textbf{\texttt{GlobalMMLU}} dataset for the \texttt{es} language after instruction-tuning on the \textbf{\texttt{Tulu3Mixture}} dataset.}\label{fig:globalmmlu-Tulu3Mixture-es}\end{figure}
\begin{figure}[h!]\centering\resizebox{\linewidth}{!}{\includegraphics[width=\linewidth]{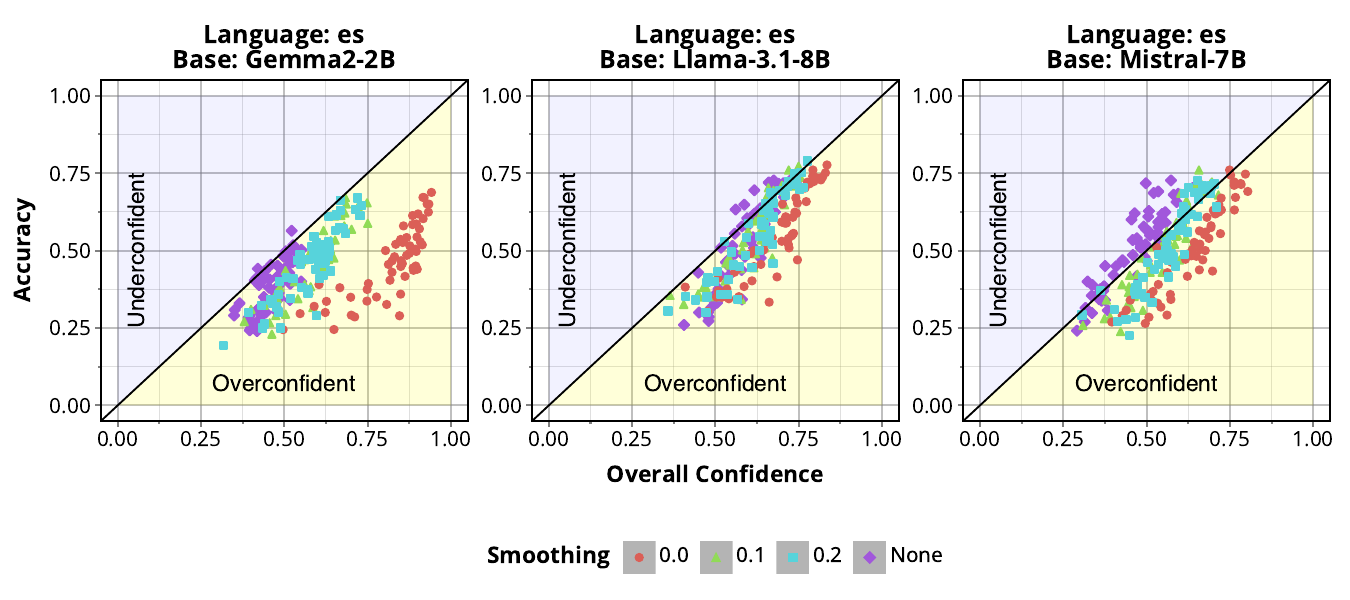}}
\caption{Reliability diagrams for the \textbf{\texttt{GlobalMMLU}} dataset for the \texttt{es} language after instruction-tuning on the \textbf{\texttt{OpenHermes}} dataset.}\label{fig:globalmmlu-OpenHermes-es}\end{figure}

\begin{figure}[h!]\centering\resizebox{\linewidth}{!}{\includegraphics[width=\linewidth]{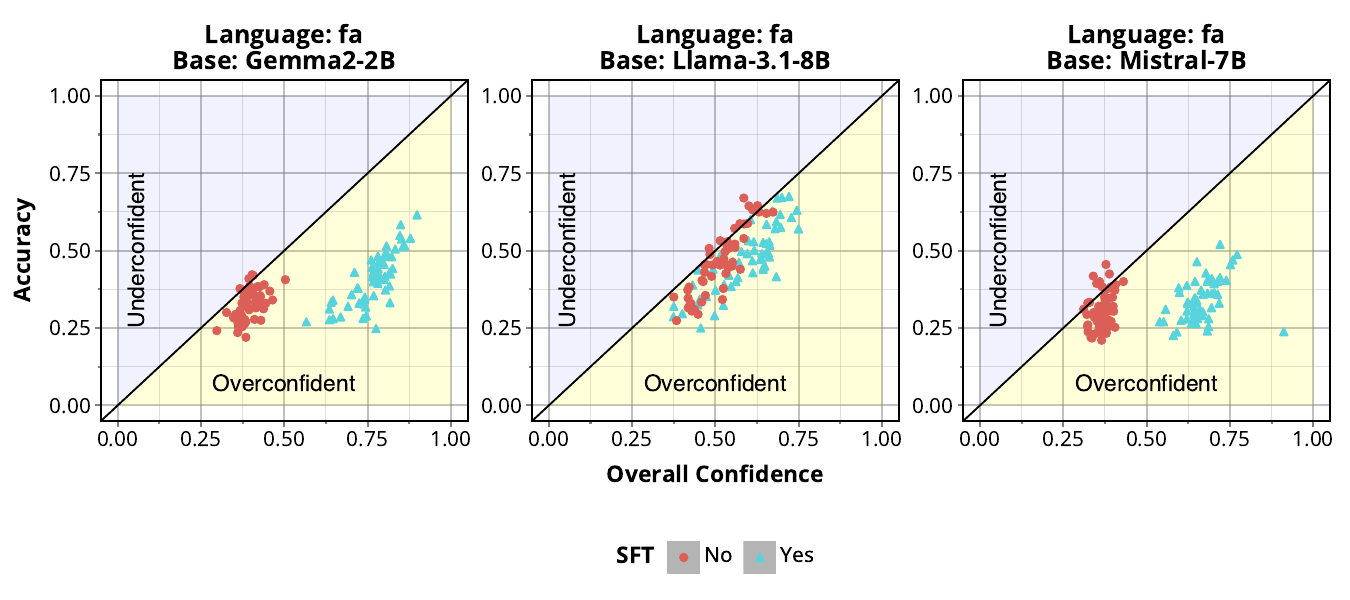}}
\caption{Reliability diagrams for the \textbf{\texttt{GlobalMMLU}} dataset for the \texttt{fa} language.}\label{fig:globalmmlu-base-fa}\end{figure}
\begin{figure}[h!]\centering\resizebox{\linewidth}{!}{\includegraphics[width=\linewidth]{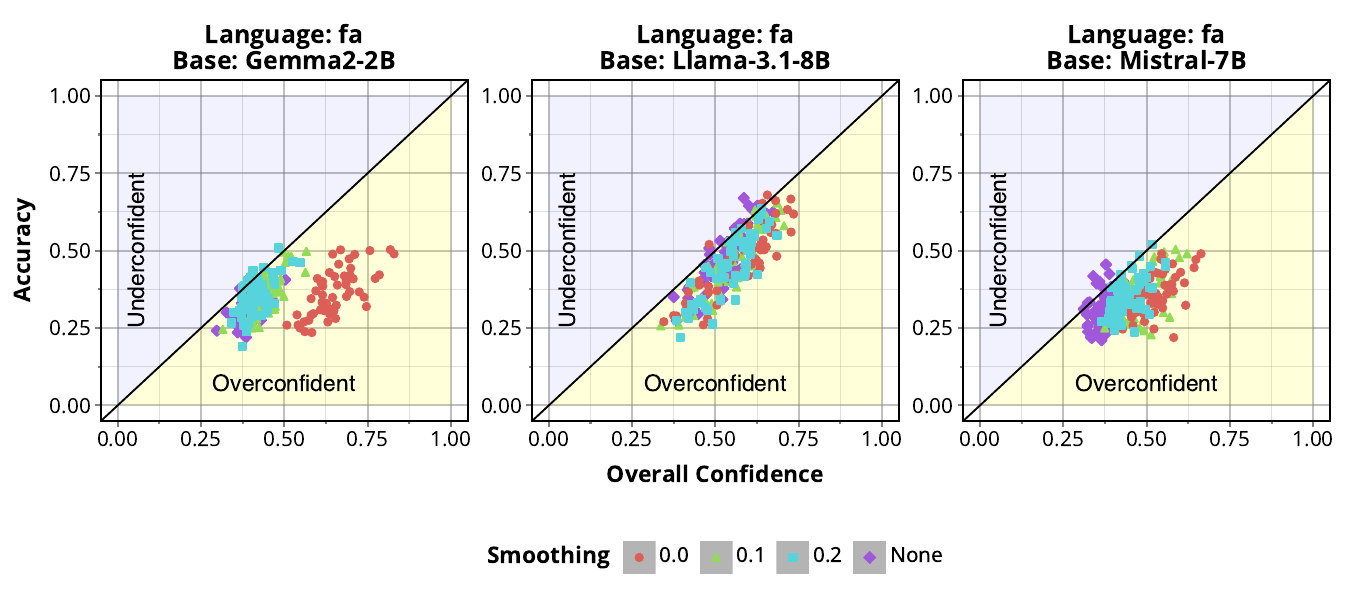}}
\caption{Reliability diagrams for the \textbf{\texttt{GlobalMMLU}} dataset for the \texttt{fa} language after instruction-tuning on the \textbf{\texttt{Tulu3Mixture}} dataset.}\label{fig:globalmmlu-Tulu3Mixture-fa}\end{figure}
\begin{figure}[h!]\centering\resizebox{\linewidth}{!}{\includegraphics[width=\linewidth]{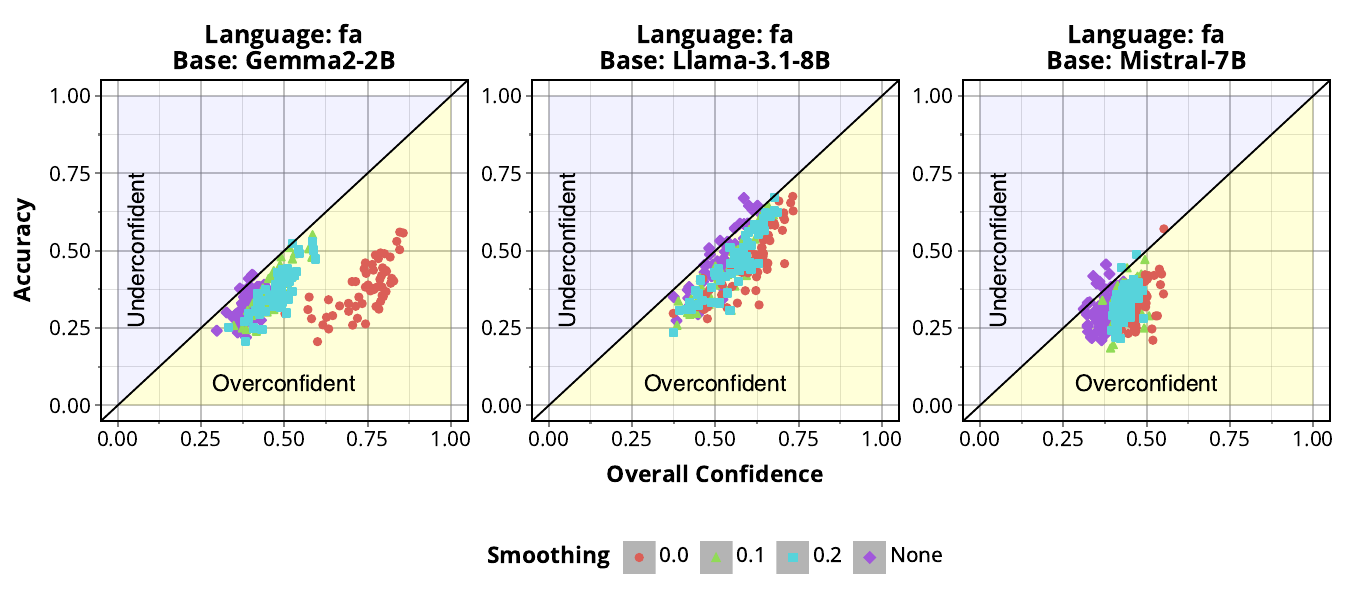}}
\caption{Reliability diagrams for the \textbf{\texttt{GlobalMMLU}} dataset for the \texttt{fa} language after instruction-tuning on the \textbf{\texttt{OpenHermes}} dataset.}\label{fig:globalmmlu-OpenHermes-fa}\end{figure}

\begin{figure}[h!]\centering\resizebox{\linewidth}{!}{\includegraphics[width=\linewidth]{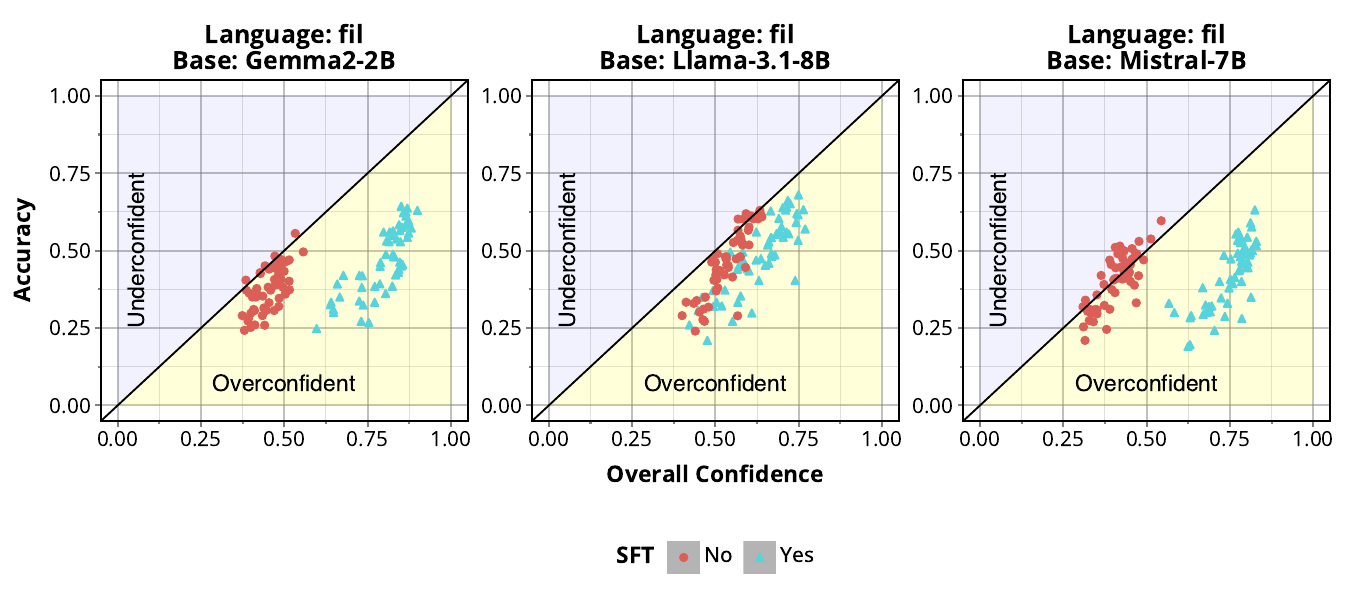}}
\caption{Reliability diagrams for the \textbf{\texttt{GlobalMMLU}} dataset for the \texttt{fil} language.}\label{fig:globalmmlu-base-fil}\end{figure}
\begin{figure}[h!]\centering\resizebox{\linewidth}{!}{\includegraphics[width=\linewidth]{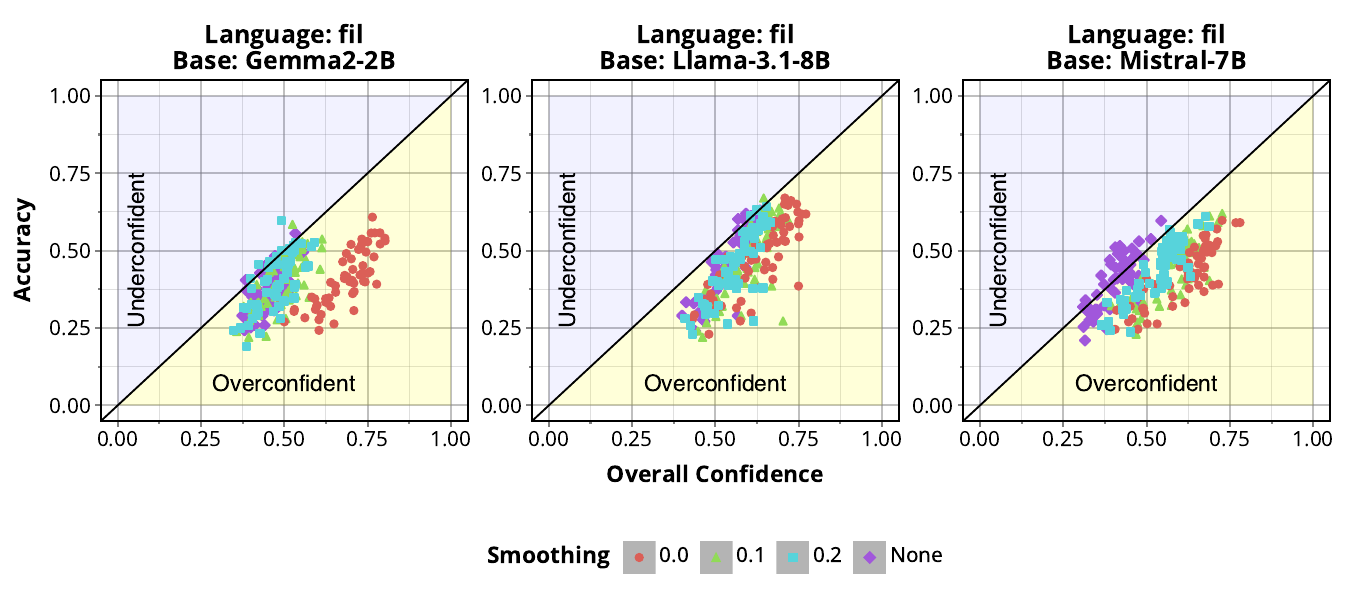}}
\caption{Reliability diagrams for the \textbf{\texttt{GlobalMMLU}} dataset for the \texttt{fil} language after instruction-tuning on the \textbf{\texttt{Tulu3Mixture}} dataset.}\label{fig:globalmmlu-Tulu3Mixture-fil}\end{figure}
\begin{figure}[h!]\centering\resizebox{\linewidth}{!}{\includegraphics[width=\linewidth]{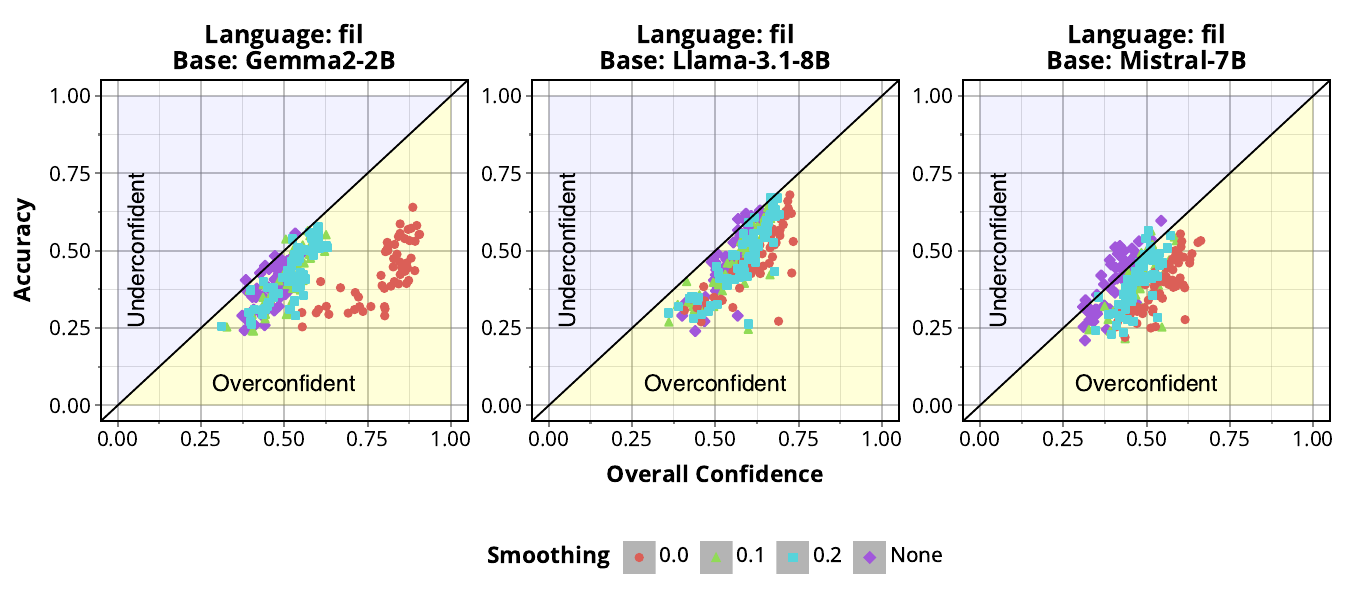}}
\caption{Reliability diagrams for the \textbf{\texttt{GlobalMMLU}} dataset for the \texttt{fil} language after instruction-tuning on the \textbf{\texttt{OpenHermes}} dataset.}\label{fig:globalmmlu-OpenHermes-fil}\end{figure}

\begin{figure}[h!]\centering\resizebox{\linewidth}{!}{\includegraphics[width=\linewidth]{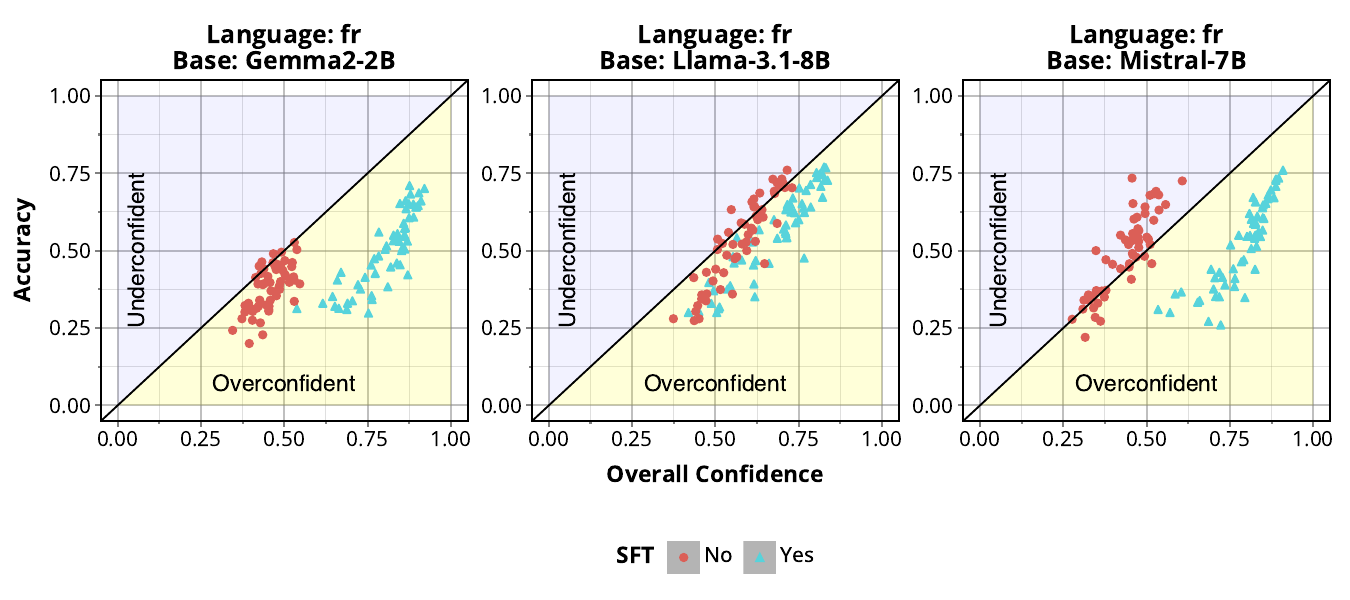}}
\caption{Reliability diagrams for the \textbf{\texttt{GlobalMMLU}} dataset for the \texttt{fr} language.}\label{fig:globalmmlu-base-fr}\end{figure}
\begin{figure}[h!]\centering\resizebox{\linewidth}{!}{\includegraphics[width=\linewidth]{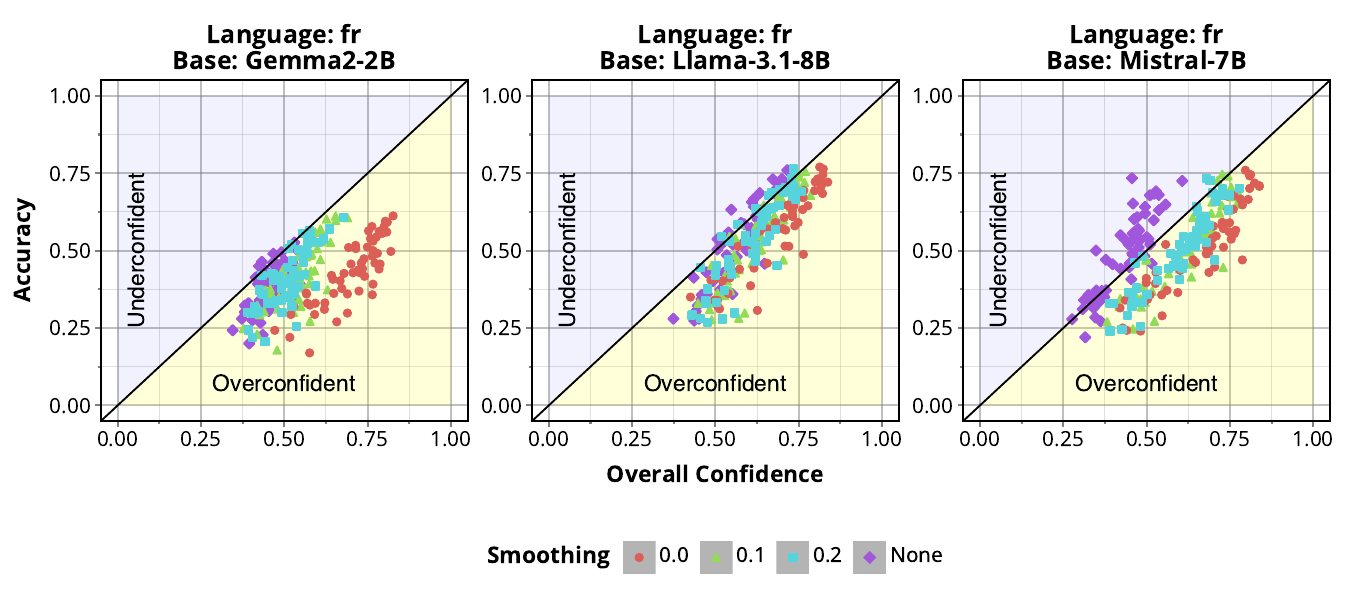}}
\caption{Reliability diagrams for the \textbf{\texttt{GlobalMMLU}} dataset for the \texttt{fr} language after instruction-tuning on the \textbf{\texttt{Tulu3Mixture}} dataset.}\label{fig:globalmmlu-Tulu3Mixture-fr}\end{figure}
\begin{figure}[h!]\centering\resizebox{\linewidth}{!}{\includegraphics[width=\linewidth]{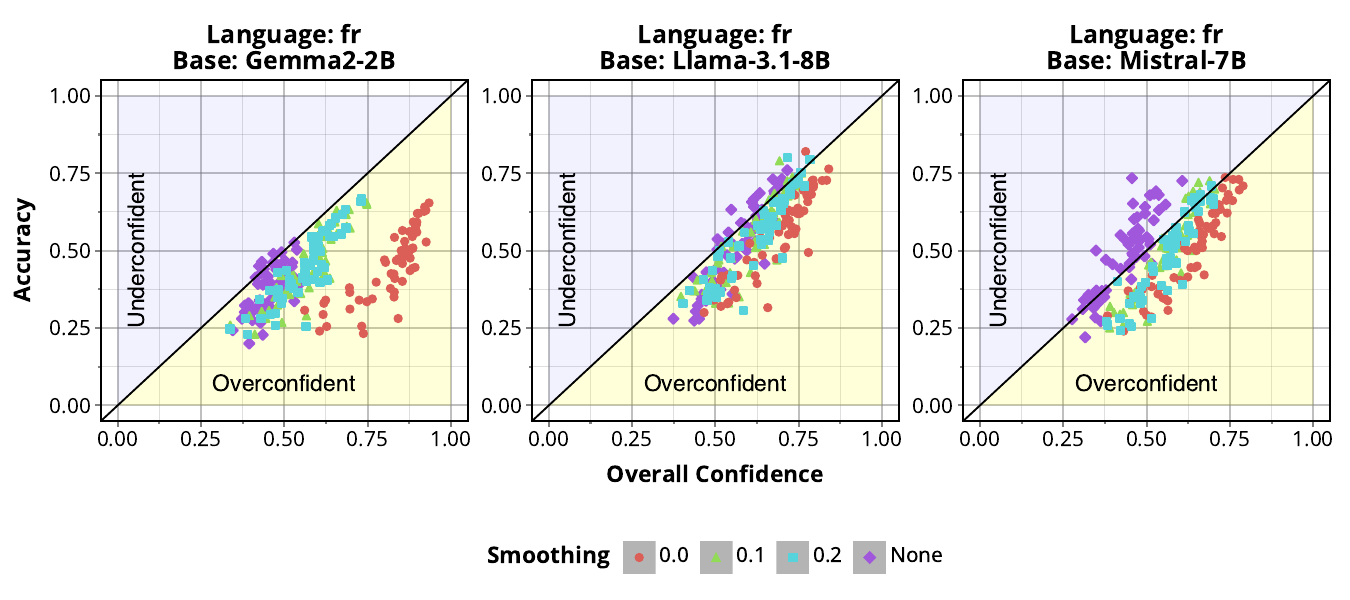}}
\caption{Reliability diagrams for the \textbf{\texttt{GlobalMMLU}} dataset for the \texttt{fr} language after instruction-tuning on the \textbf{\texttt{OpenHermes}} dataset.}\label{fig:globalmmlu-OpenHermes-fr}\end{figure}

\begin{figure}[h!]\centering\resizebox{\linewidth}{!}{\includegraphics[width=\linewidth]{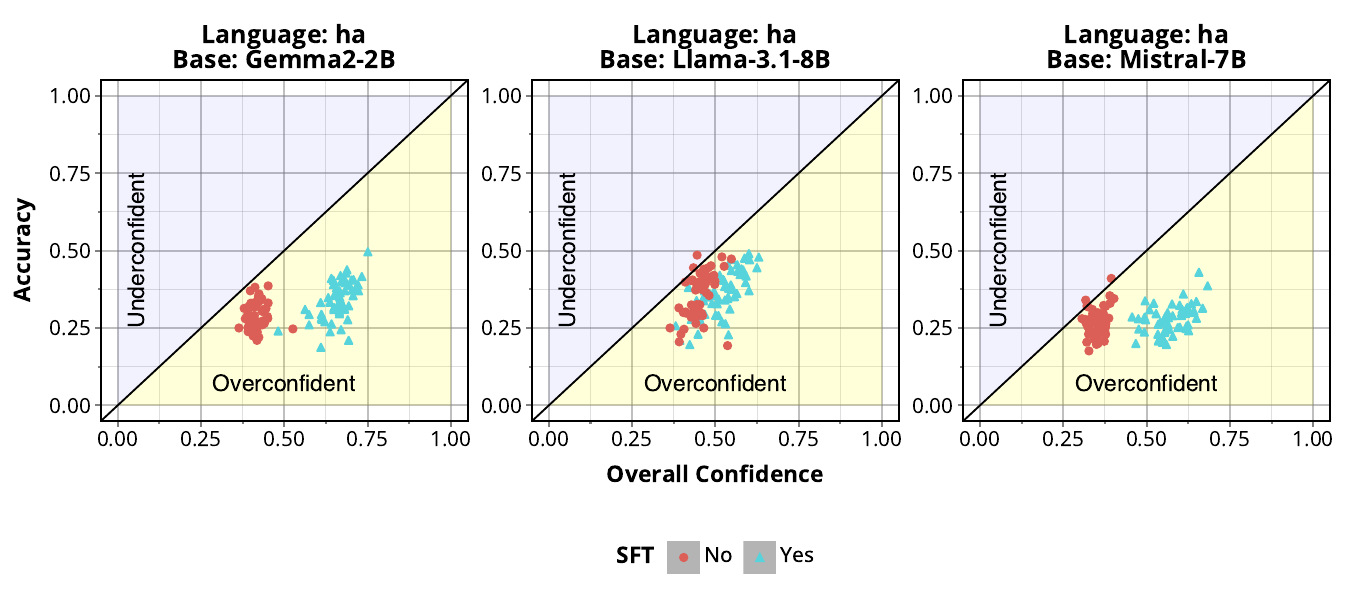}}
\caption{Reliability diagrams for the \textbf{\texttt{GlobalMMLU}} dataset for the \texttt{ha} language.}\label{fig:globalmmlu-base-ha}\end{figure}
\begin{figure}[h!]\centering\resizebox{\linewidth}{!}{\includegraphics[width=\linewidth]{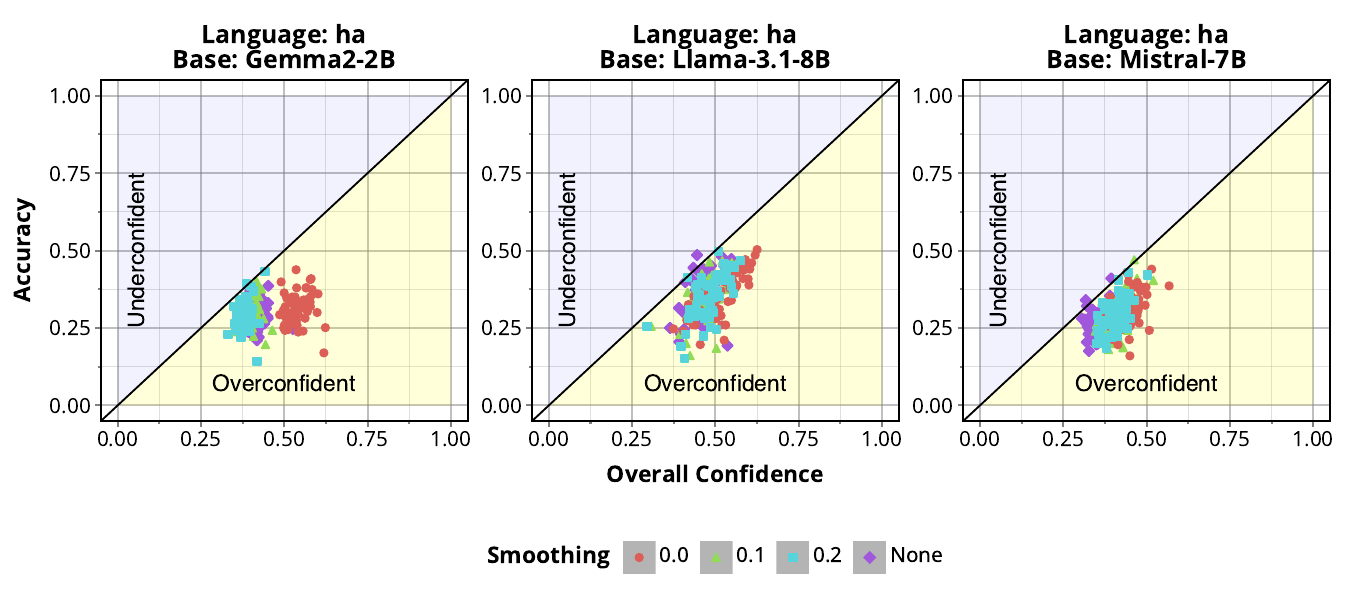}}
\caption{Reliability diagrams for the \textbf{\texttt{GlobalMMLU}} dataset for the \texttt{ha} language after instruction-tuning on the \textbf{\texttt{Tulu3Mixture}} dataset.}\label{fig:globalmmlu-Tulu3Mixture-ha}\end{figure}
\begin{figure}[h!]\centering\resizebox{\linewidth}{!}{\includegraphics[width=\linewidth]{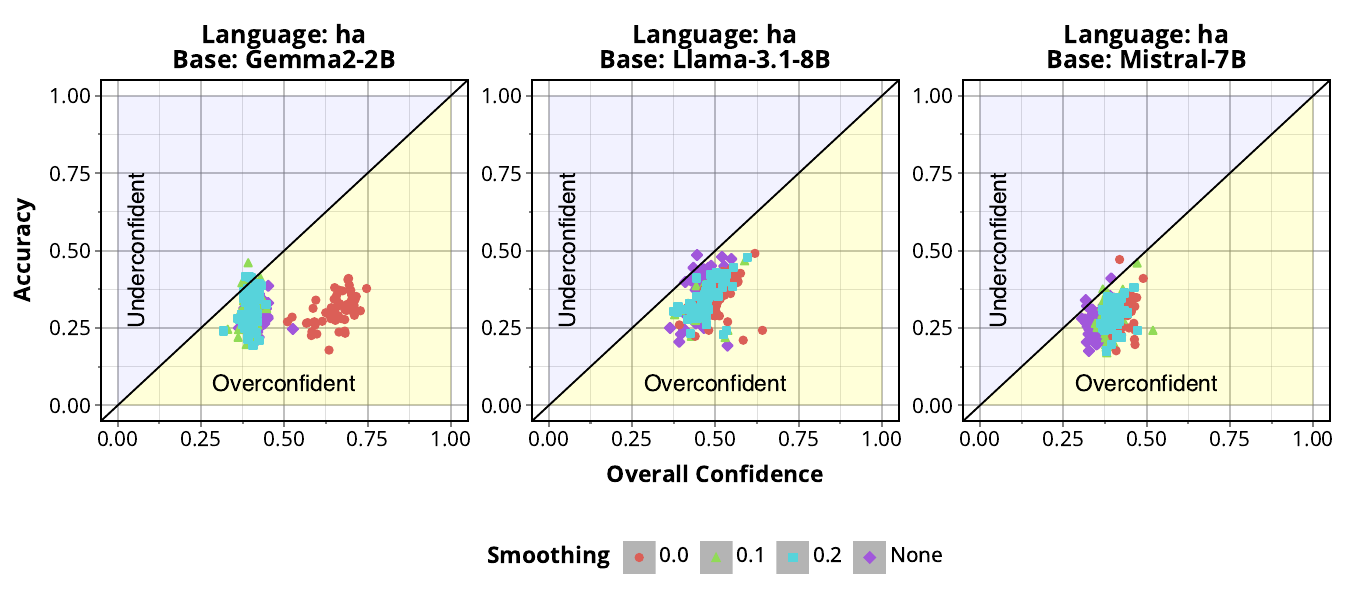}}
\caption{Reliability diagrams for the \textbf{\texttt{GlobalMMLU}} dataset for the \texttt{ha} language after instruction-tuning on the \textbf{\texttt{OpenHermes}} dataset.}\label{fig:globalmmlu-OpenHermes-ha}\end{figure}

\begin{figure}[h!]\centering\resizebox{\linewidth}{!}{\includegraphics[width=\linewidth]{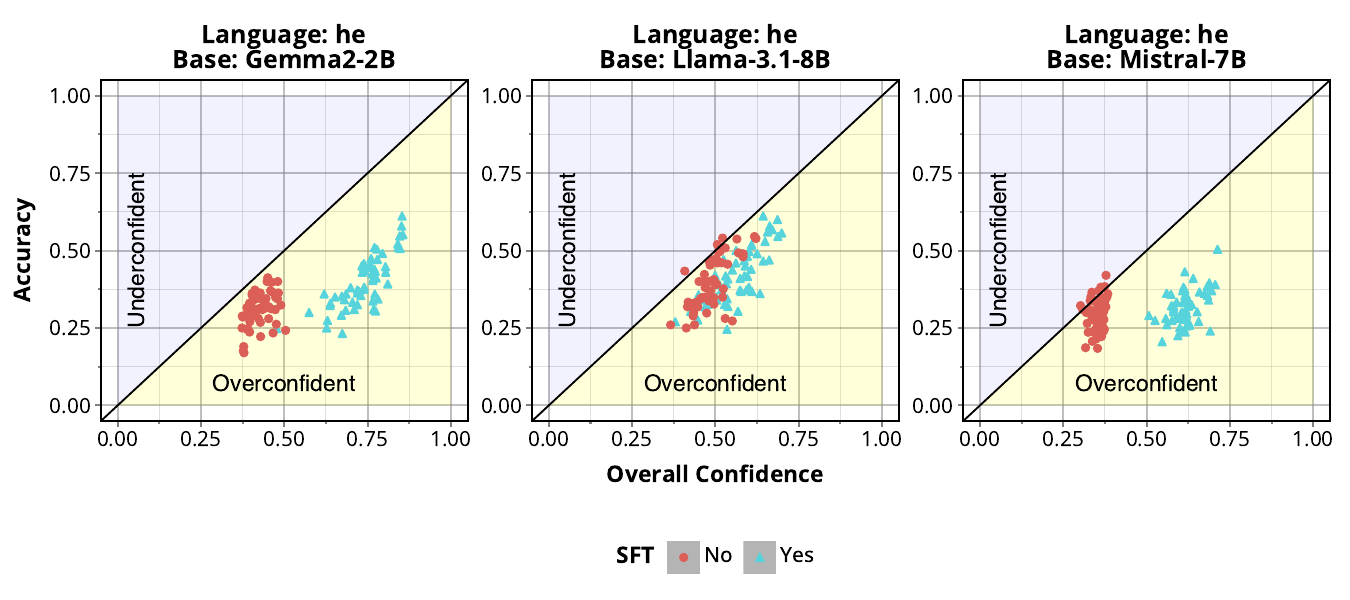}}
\caption{Reliability diagrams for the \textbf{\texttt{GlobalMMLU}} dataset for the \texttt{he} language.}\label{fig:globalmmlu-base-he}\end{figure}
\begin{figure}[h!]\centering\resizebox{\linewidth}{!}{\includegraphics[width=\linewidth]{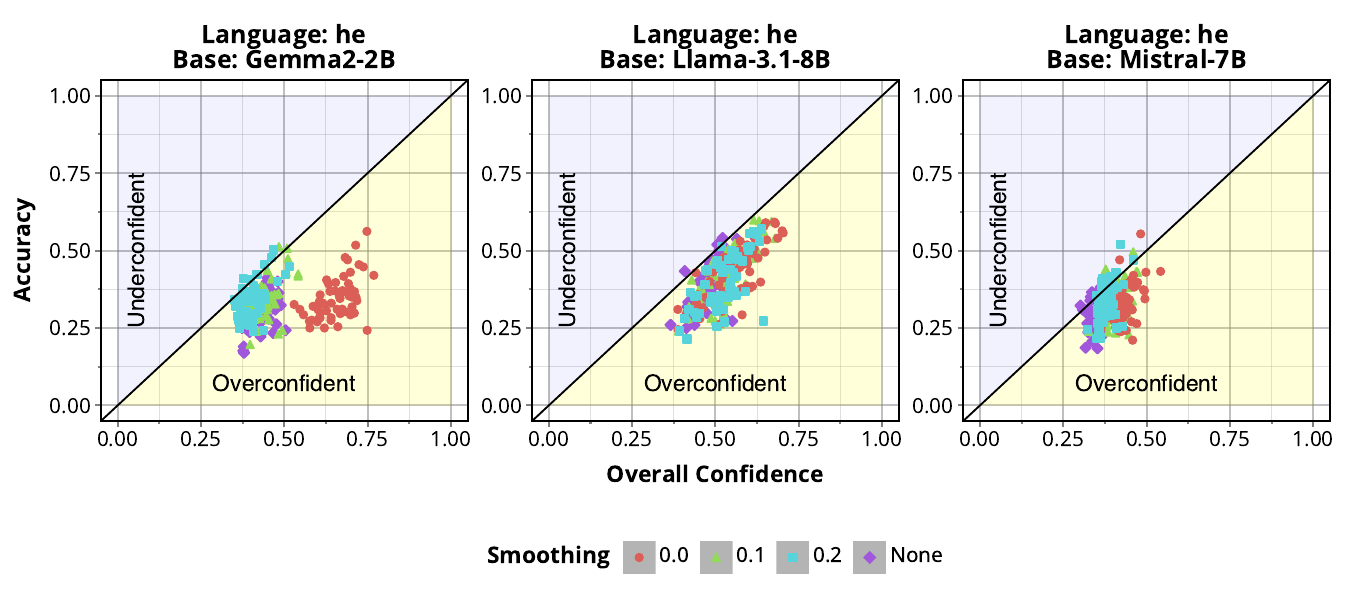}}
\caption{Reliability diagrams for the \textbf{\texttt{GlobalMMLU}} dataset for the \texttt{he} language after instruction-tuning on the \textbf{\texttt{Tulu3Mixture}} dataset.}\label{fig:globalmmlu-Tulu3Mixture-he}\end{figure}
\begin{figure}[h!]\centering\resizebox{\linewidth}{!}{\includegraphics[width=\linewidth]{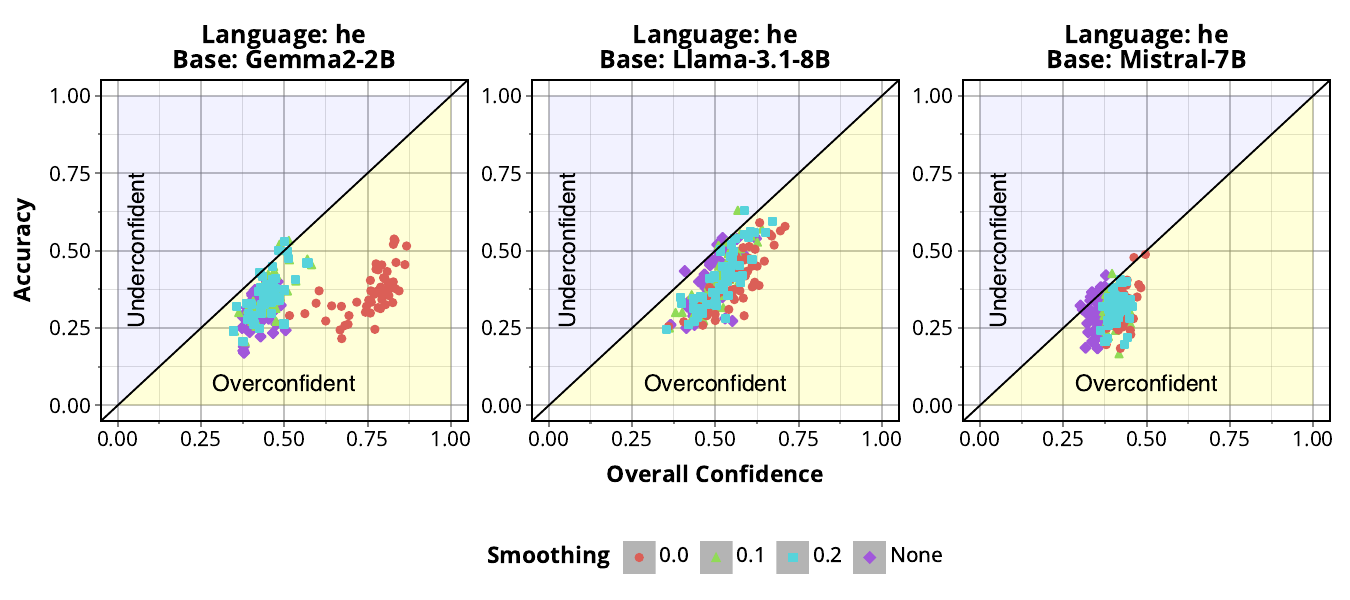}}
\caption{Reliability diagrams for the \textbf{\texttt{GlobalMMLU}} dataset for the \texttt{he} language after instruction-tuning on the \textbf{\texttt{OpenHermes}} dataset.}\label{fig:globalmmlu-OpenHermes-he}\end{figure}

\begin{figure}[h!]\centering\resizebox{\linewidth}{!}{\includegraphics[width=\linewidth]{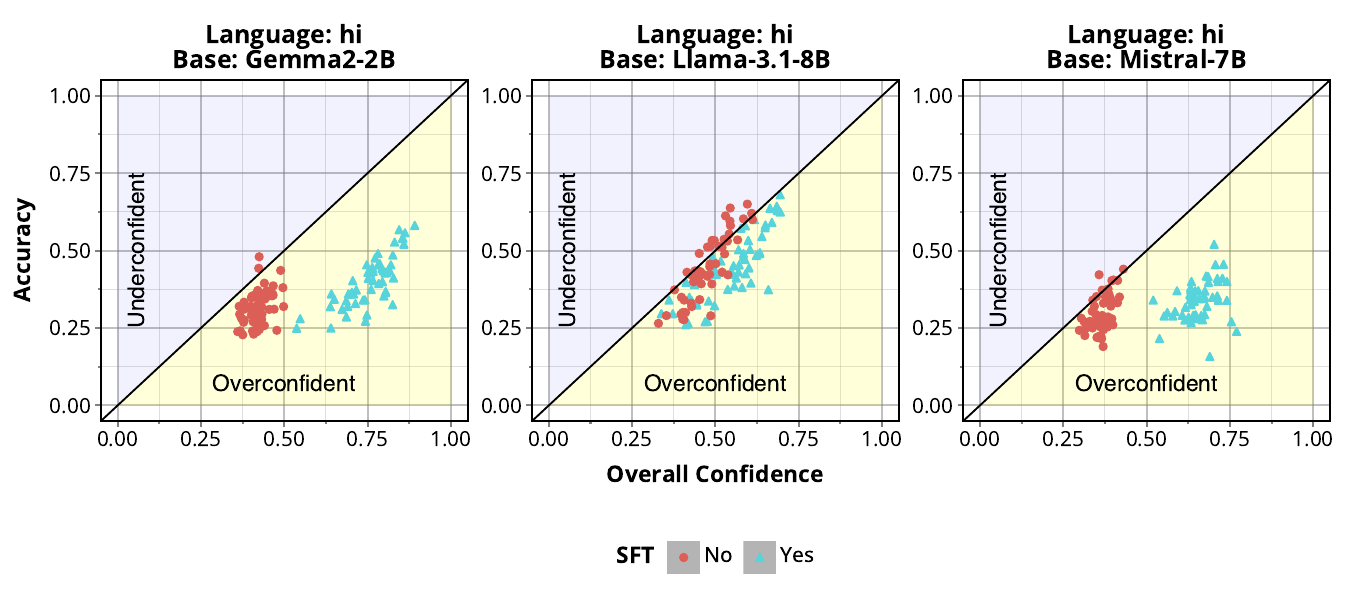}}
\caption{Reliability diagrams for the \textbf{\texttt{GlobalMMLU}} dataset for the \texttt{hi} language.}\label{fig:globalmmlu-base-hi}\end{figure}
\begin{figure}[h!]\centering\resizebox{\linewidth}{!}{\includegraphics[width=\linewidth]{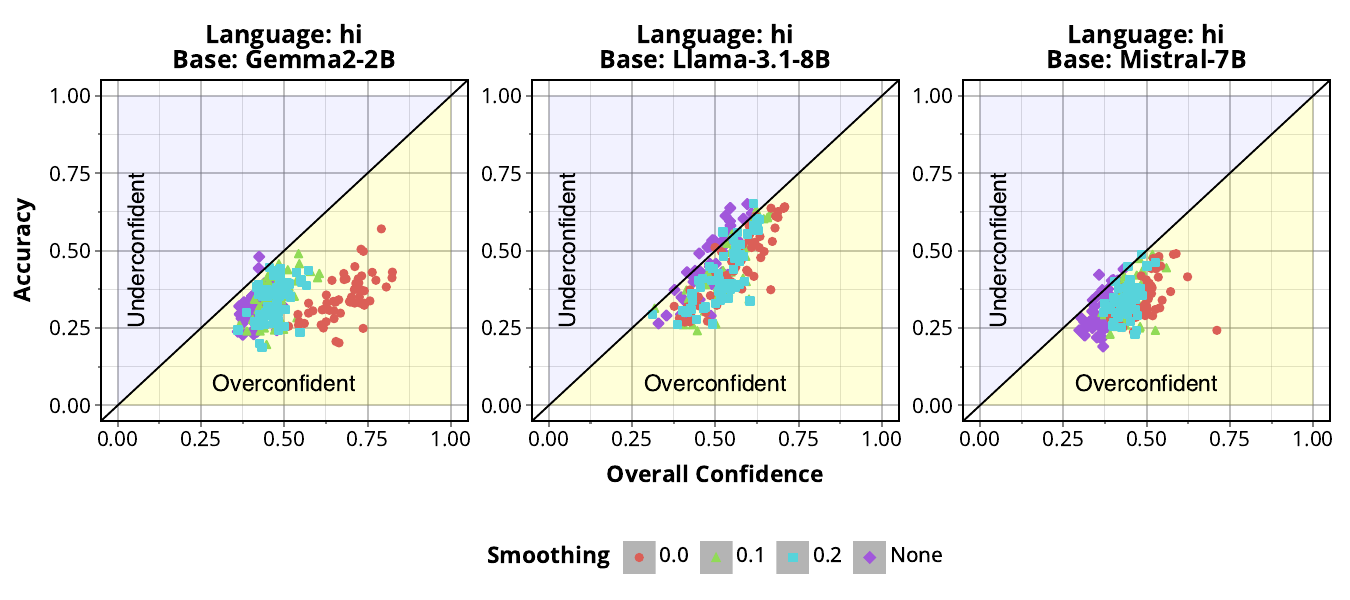}}
\caption{Reliability diagrams for the \textbf{\texttt{GlobalMMLU}} dataset for the \texttt{hi} language after instruction-tuning on the \textbf{\texttt{Tulu3Mixture}} dataset.}\label{fig:globalmmlu-Tulu3Mixture-hi}\end{figure}
\begin{figure}[h!]\centering\resizebox{\linewidth}{!}{\includegraphics[width=\linewidth]{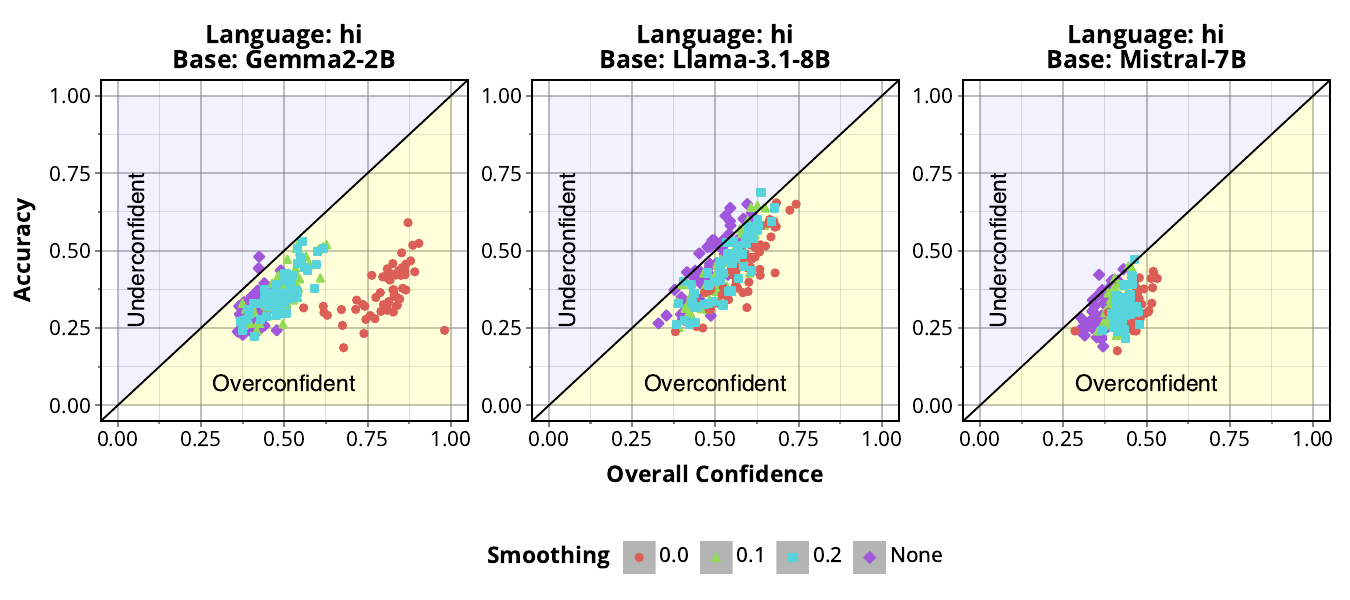}}
\caption{Reliability diagrams for the \textbf{\texttt{GlobalMMLU}} dataset for the \texttt{hi} language after instruction-tuning on the \textbf{\texttt{OpenHermes}} dataset.}\label{fig:globalmmlu-OpenHermes-hi}\end{figure}

\begin{figure}[h!]\centering\resizebox{\linewidth}{!}{\includegraphics[width=\linewidth]{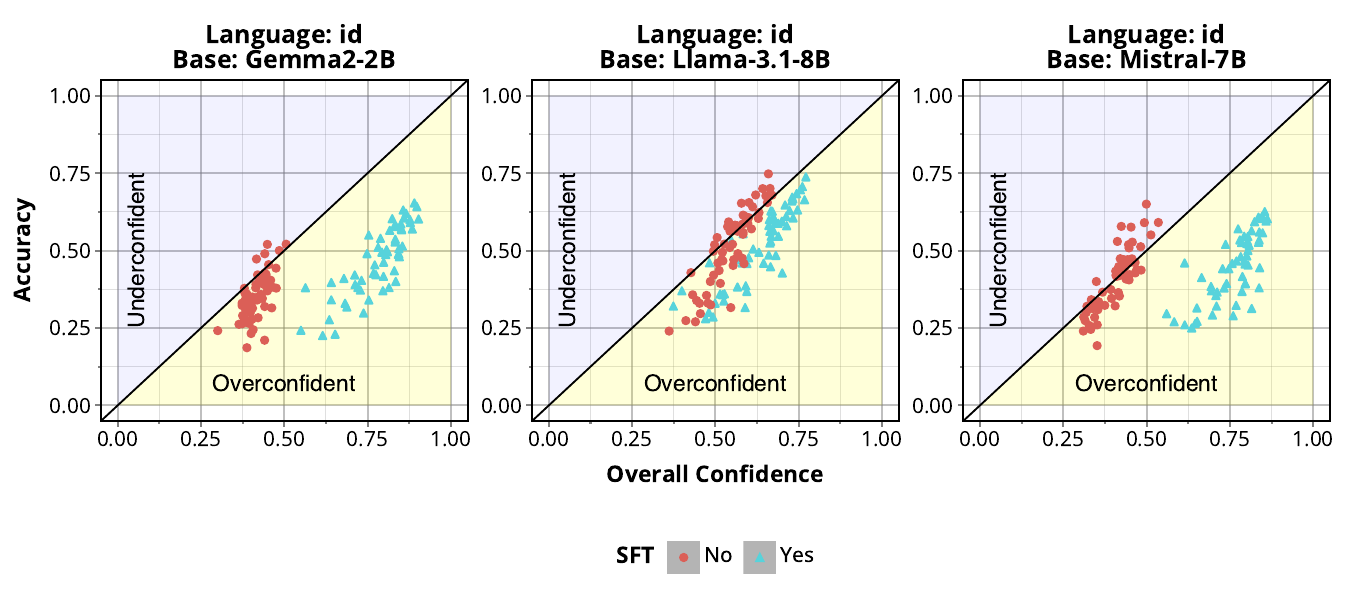}}
\caption{Reliability diagrams for the \textbf{\texttt{GlobalMMLU}} dataset for the \texttt{id} language.}\label{fig:globalmmlu-base-id}\end{figure}
\begin{figure}[h!]\centering\resizebox{\linewidth}{!}{\includegraphics[width=\linewidth]{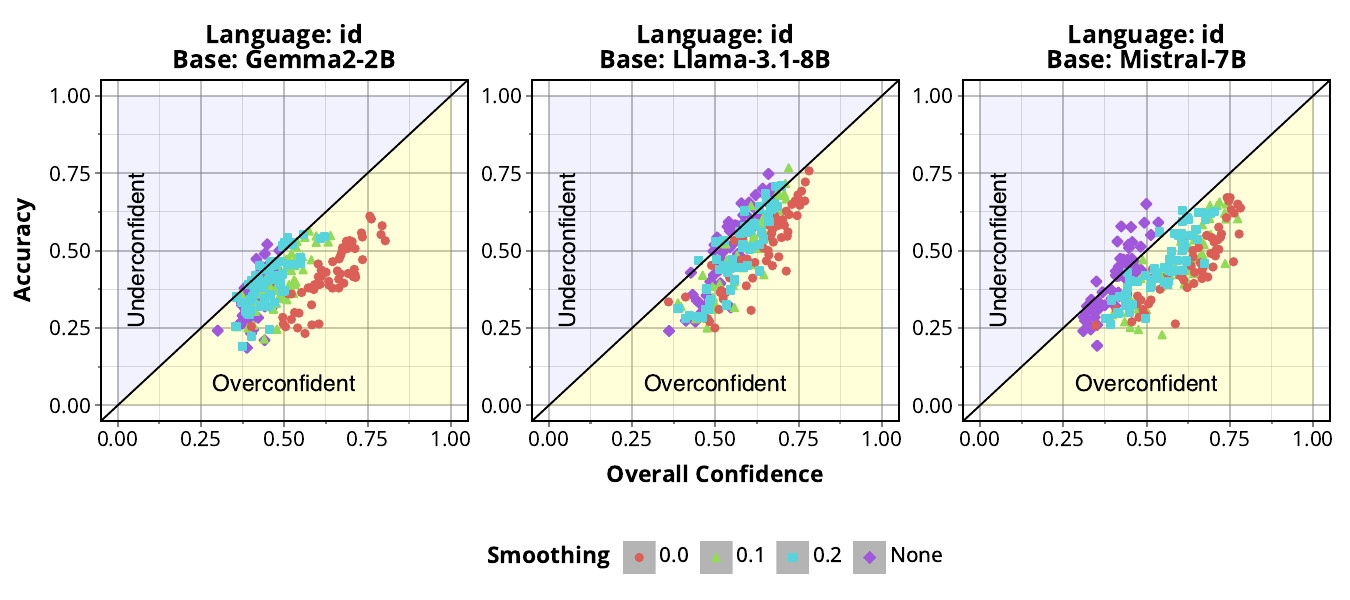}}
\caption{Reliability diagrams for the \textbf{\texttt{GlobalMMLU}} dataset for the \texttt{id} language after instruction-tuning on the \textbf{\texttt{Tulu3Mixture}} dataset.}\label{fig:globalmmlu-Tulu3Mixture-id}\end{figure}
\begin{figure}[h!]\centering\resizebox{\linewidth}{!}{\includegraphics[width=\linewidth]{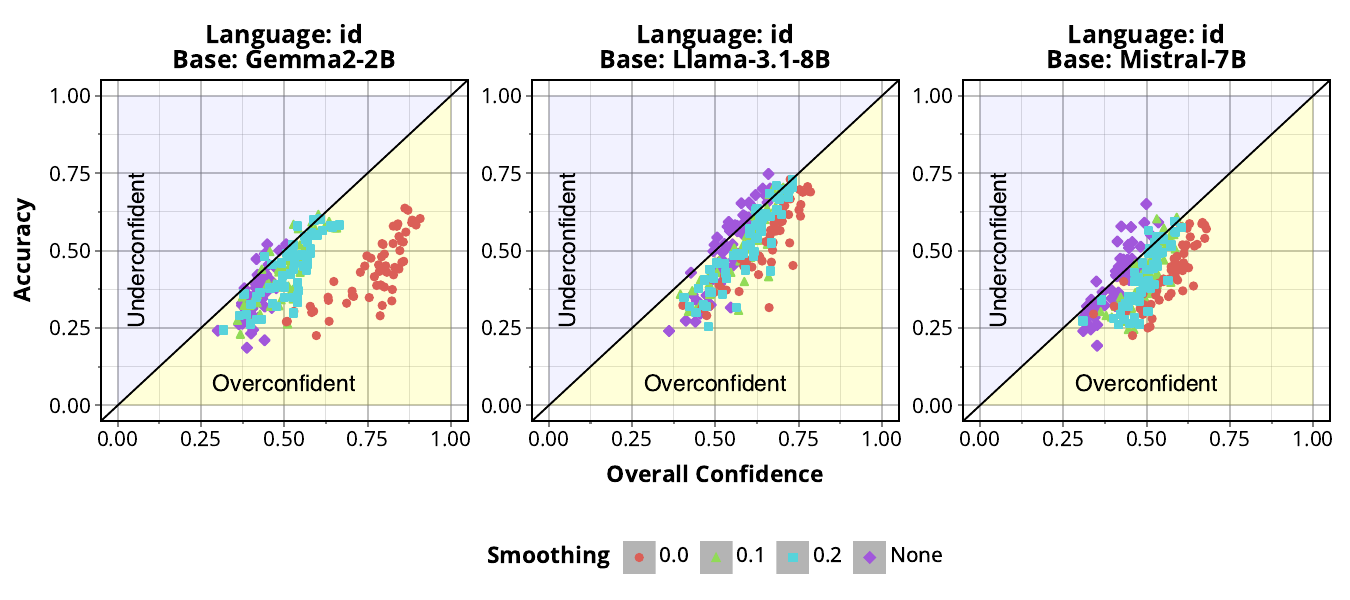}}
\caption{Reliability diagrams for the \textbf{\texttt{GlobalMMLU}} dataset for the \texttt{id} language after instruction-tuning on the \textbf{\texttt{OpenHermes}} dataset.}\label{fig:globalmmlu-OpenHermes-id}\end{figure}

\begin{figure}[h!]\centering\resizebox{\linewidth}{!}{\includegraphics[width=\linewidth]{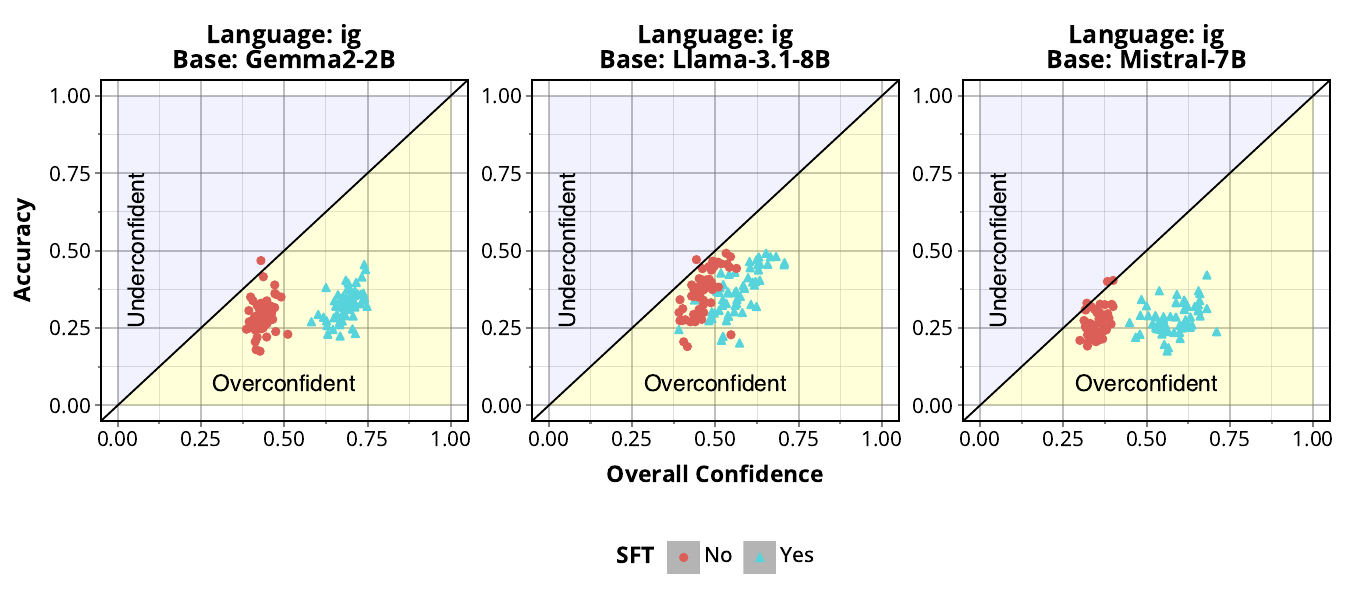}}
\caption{Reliability diagrams for the \textbf{\texttt{GlobalMMLU}} dataset for the \texttt{ig} language.}\label{fig:globalmmlu-base-ig}\end{figure}
\begin{figure}[h!]\centering\resizebox{\linewidth}{!}{\includegraphics[width=\linewidth]{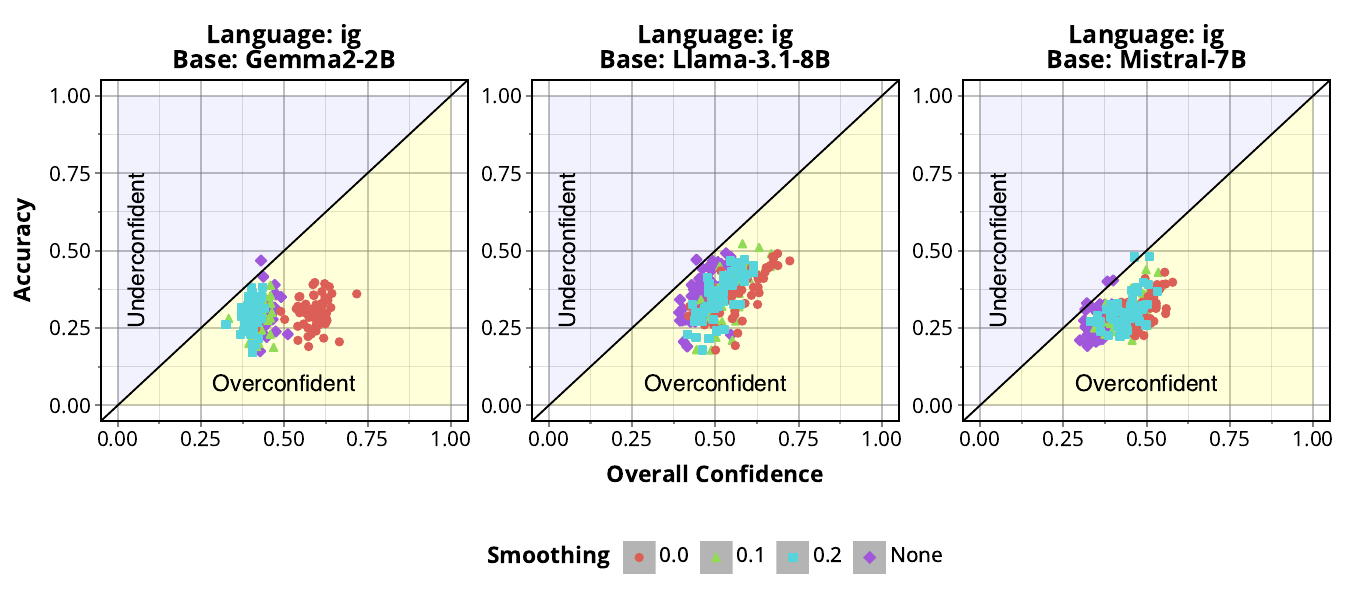}}
\caption{Reliability diagrams for the \textbf{\texttt{GlobalMMLU}} dataset for the \texttt{ig} language after instruction-tuning on the \textbf{\texttt{Tulu3Mixture}} dataset.}\label{fig:globalmmlu-Tulu3Mixture-ig}\end{figure}
\begin{figure}[h!]\centering\resizebox{\linewidth}{!}{\includegraphics[width=\linewidth]{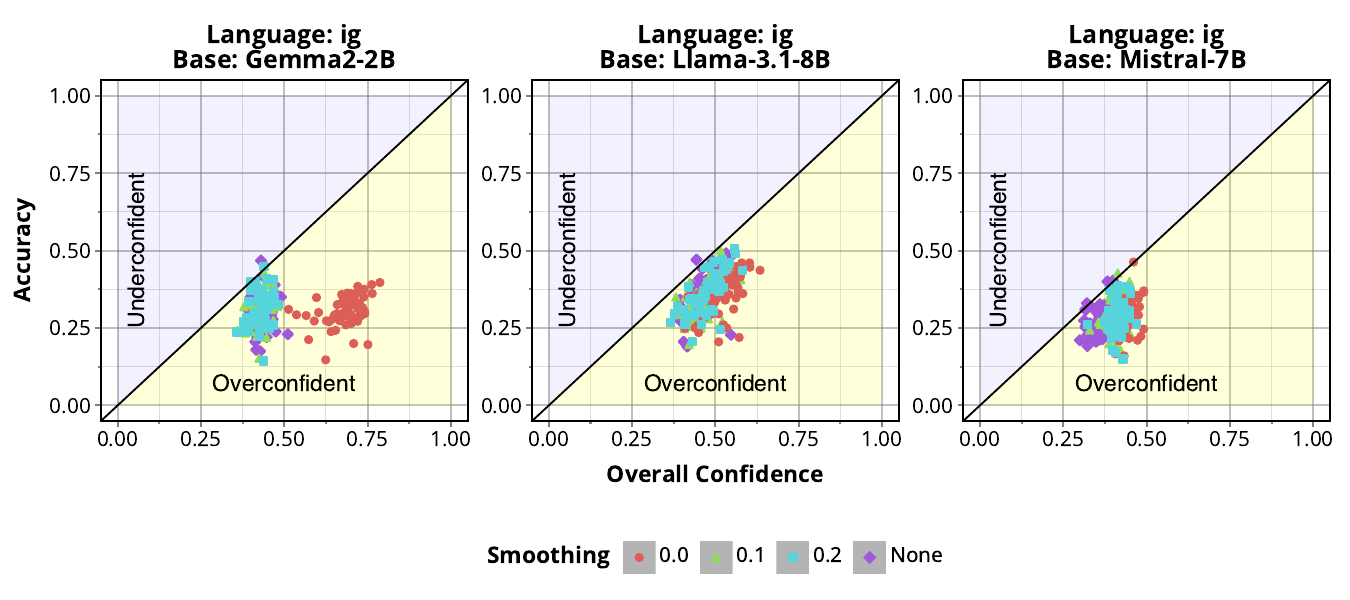}}
\caption{Reliability diagrams for the \textbf{\texttt{GlobalMMLU}} dataset for the \texttt{ig} language after instruction-tuning on the \textbf{\texttt{OpenHermes}} dataset.}\label{fig:globalmmlu-OpenHermes-ig}\end{figure}

\clearpage
\begin{figure}[h!]\centering\resizebox{\linewidth}{!}{\includegraphics[width=\linewidth]{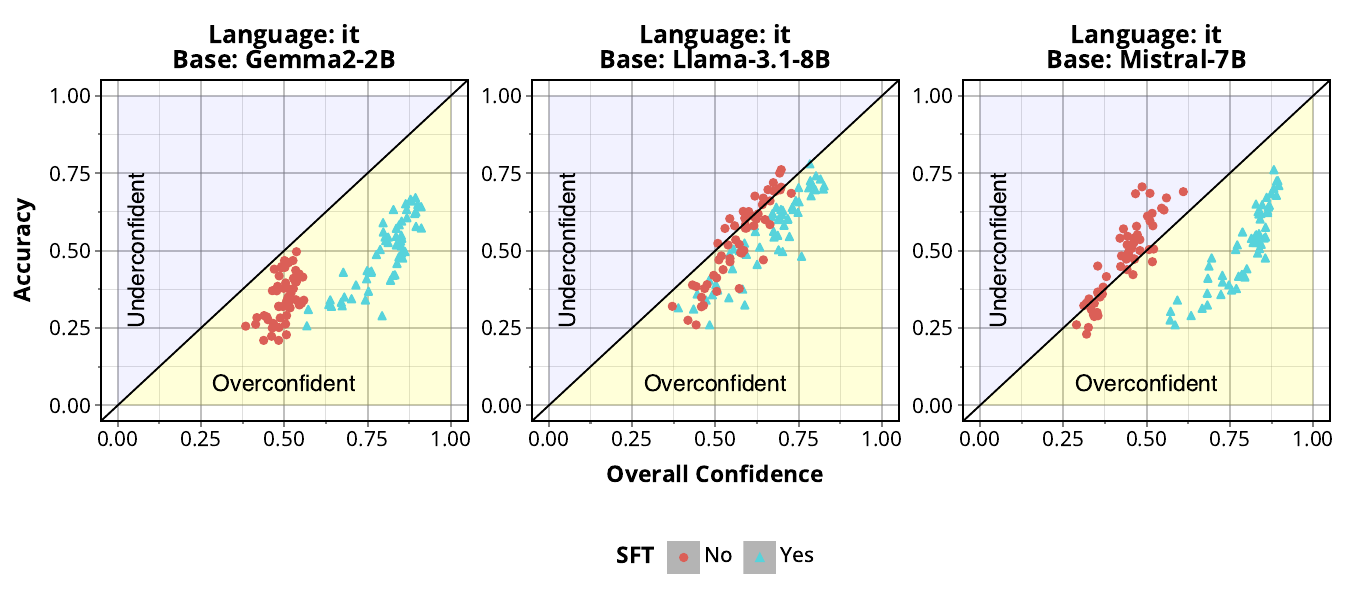}}
\caption{Reliability diagrams for the \textbf{\texttt{GlobalMMLU}} dataset for the \texttt{it} language.}\label{fig:globalmmlu-base-it}\end{figure}
\begin{figure}[h!]\centering\resizebox{\linewidth}{!}{\includegraphics[width=\linewidth]{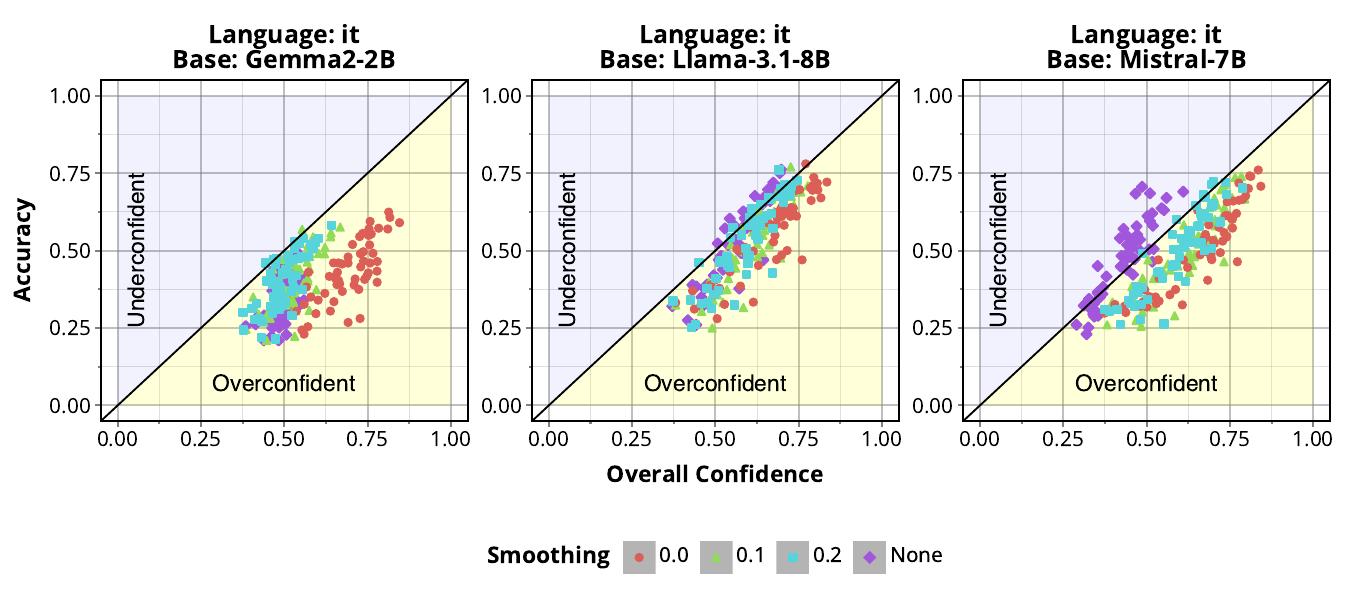}}
\caption{Reliability diagrams for the \textbf{\texttt{GlobalMMLU}} dataset for the \texttt{it} language after instruction-tuning on the \textbf{\texttt{Tulu3Mixture}} dataset.}\label{fig:globalmmlu-Tulu3Mixture-it}\end{figure}
\begin{figure}[h!]\centering\resizebox{\linewidth}{!}{\includegraphics[width=\linewidth]{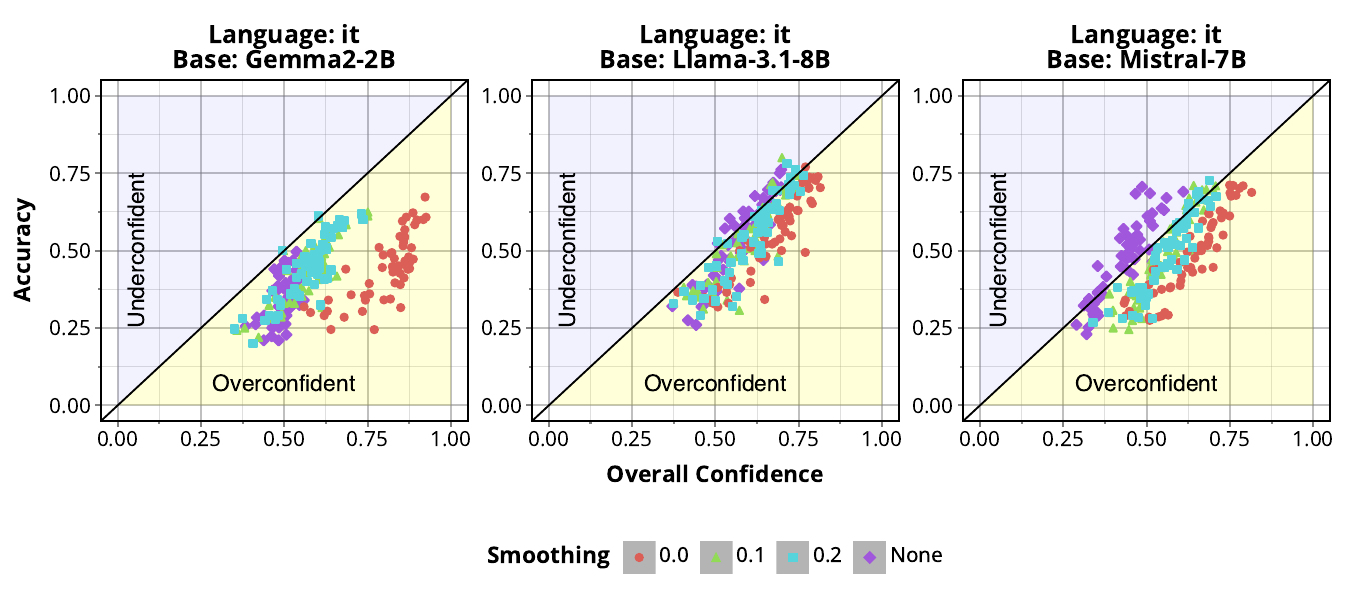}}
\caption{Reliability diagrams for the \textbf{\texttt{GlobalMMLU}} dataset for the \texttt{it} language after instruction-tuning on the \textbf{\texttt{OpenHermes}} dataset.}\label{fig:globalmmlu-OpenHermes-it}\end{figure}

\begin{figure}[h!]\centering\resizebox{\linewidth}{!}{\includegraphics[width=\linewidth]{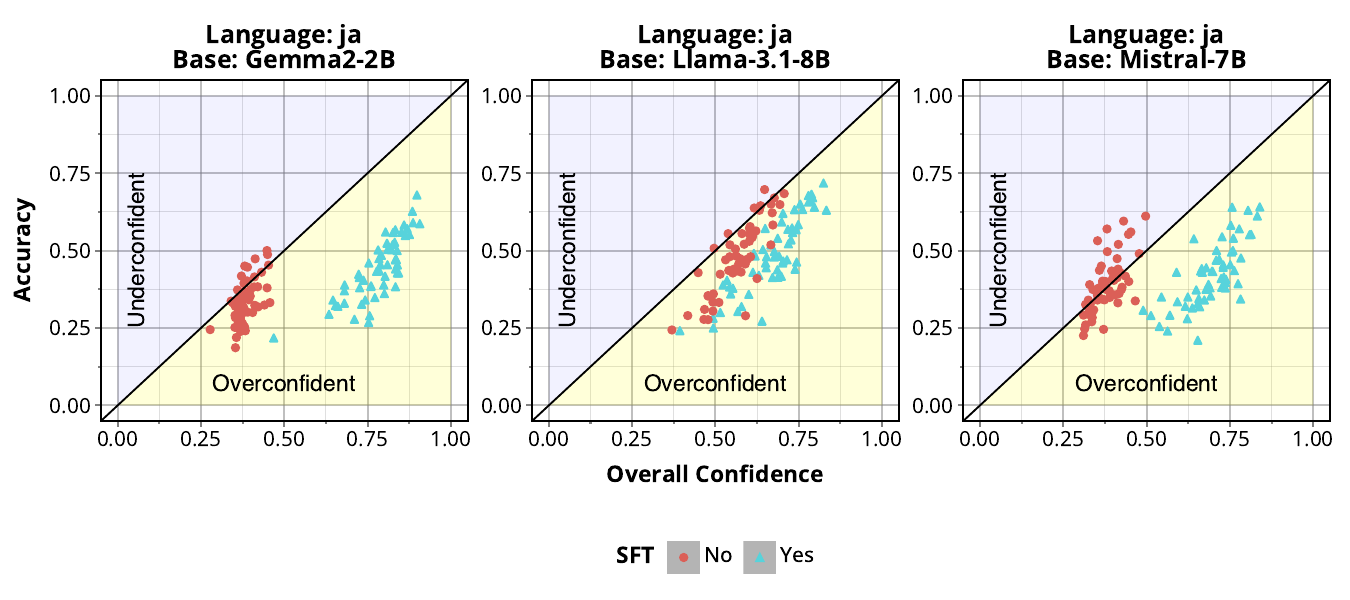}}
\caption{Reliability diagrams for the \textbf{\texttt{GlobalMMLU}} dataset for the \texttt{ja} language.}\label{fig:globalmmlu-base-ja}\end{figure}
\begin{figure}[h!]\centering\resizebox{\linewidth}{!}{\includegraphics[width=\linewidth]{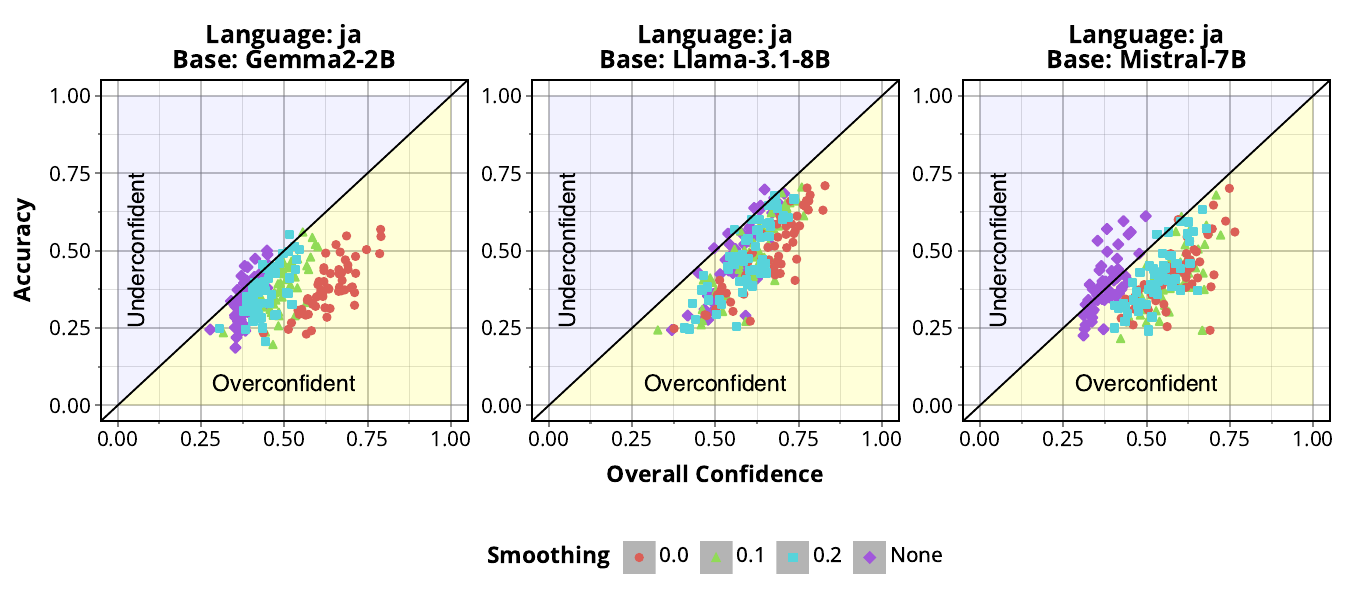}}
\caption{Reliability diagrams for the \textbf{\texttt{GlobalMMLU}} dataset for the \texttt{ja} language after instruction-tuning on the \textbf{\texttt{Tulu3Mixture}} dataset.}\label{fig:globalmmlu-Tulu3Mixture-ja}\end{figure}
\begin{figure}[h!]\centering\resizebox{\linewidth}{!}{\includegraphics[width=\linewidth]{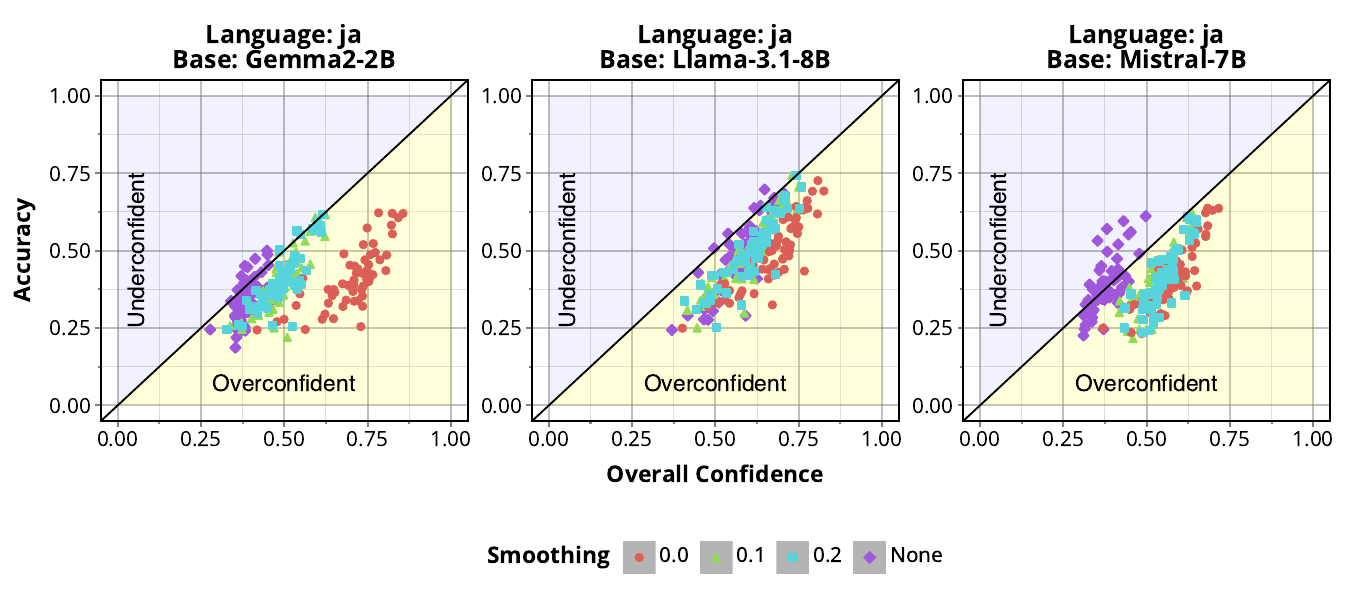}}
\caption{Reliability diagrams for the \textbf{\texttt{GlobalMMLU}} dataset for the \texttt{ja} language after instruction-tuning on the \textbf{\texttt{OpenHermes}} dataset.}\label{fig:globalmmlu-OpenHermes-ja}\end{figure}

\begin{figure}[h!]\centering\resizebox{\linewidth}{!}{\includegraphics[width=\linewidth]{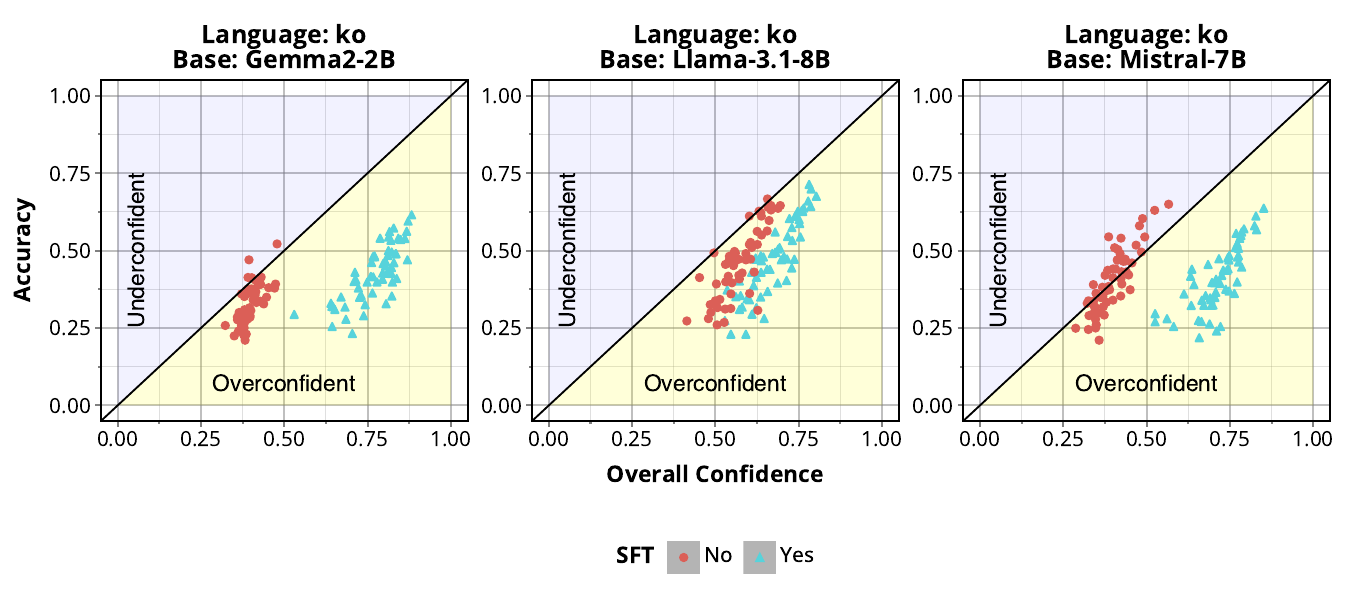}}
\caption{Reliability diagrams for the \textbf{\texttt{GlobalMMLU}} dataset for the \texttt{ko} language.}\label{fig:globalmmlu-base-ko}\end{figure}
\begin{figure}[h!]\centering\resizebox{\linewidth}{!}{\includegraphics[width=\linewidth]{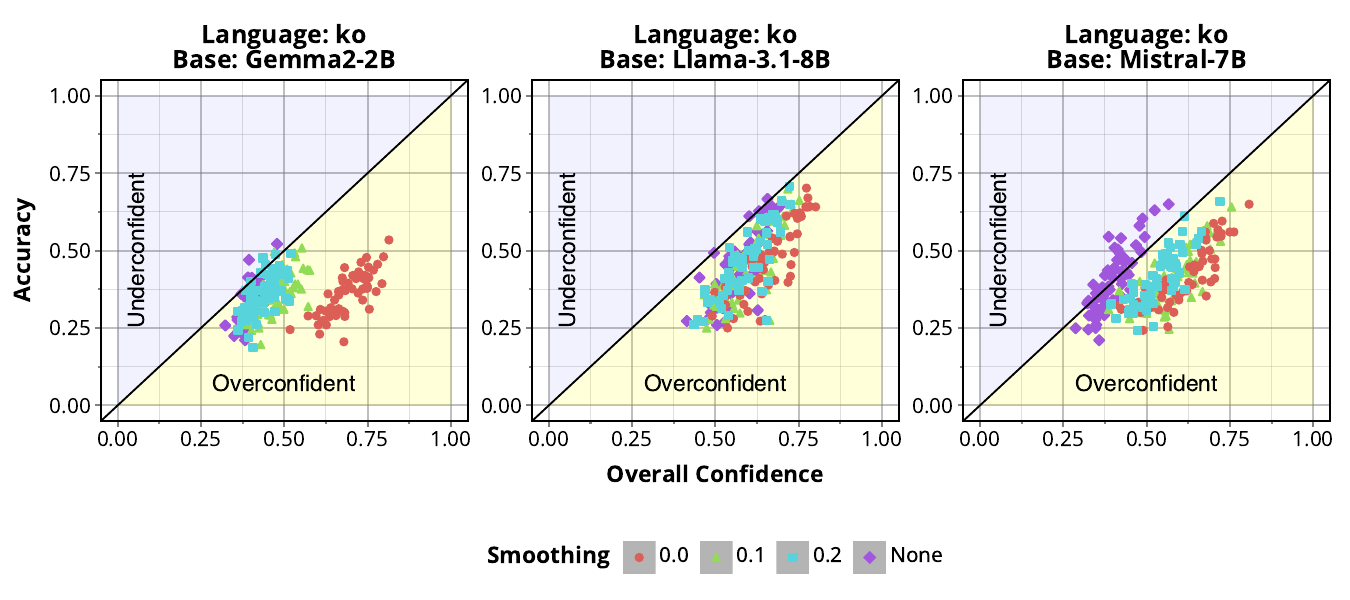}}
\caption{Reliability diagrams for the \textbf{\texttt{GlobalMMLU}} dataset for the \texttt{ko} language after instruction-tuning on the \textbf{\texttt{Tulu3Mixture}} dataset.}\label{fig:globalmmlu-Tulu3Mixture-ko}\end{figure}
\begin{figure}[h!]\centering\resizebox{\linewidth}{!}{\includegraphics[width=\linewidth]{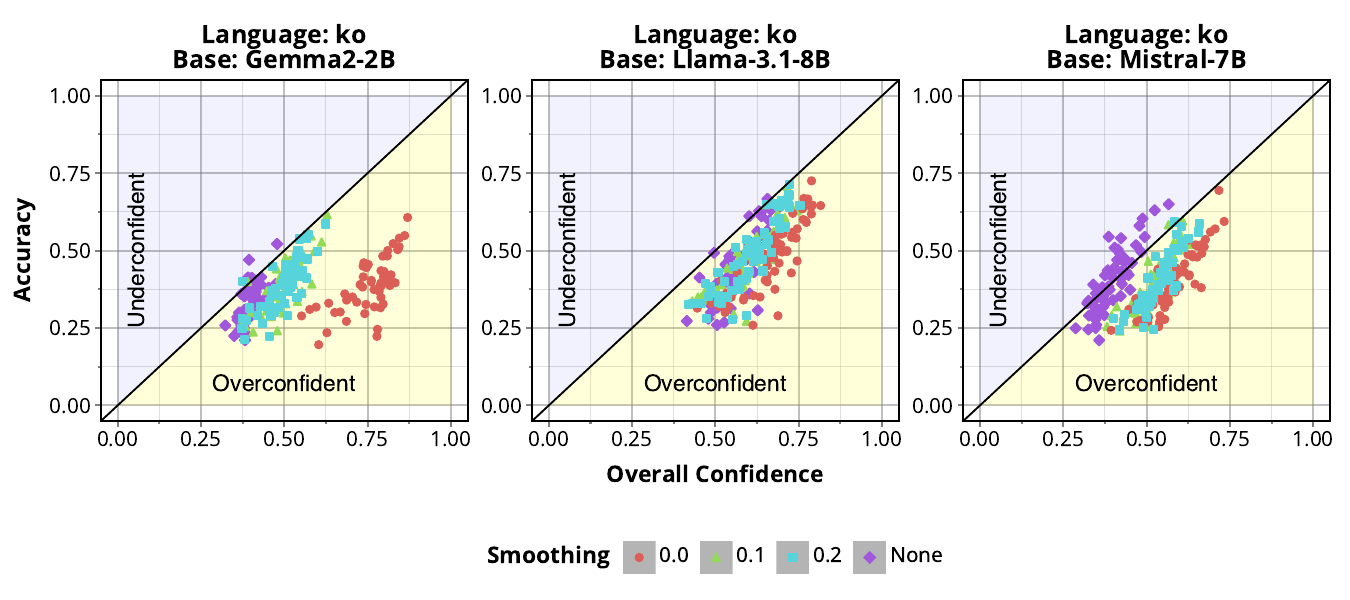}}
\caption{Reliability diagrams for the \textbf{\texttt{GlobalMMLU}} dataset for the \texttt{ko} language after instruction-tuning on the \textbf{\texttt{OpenHermes}} dataset.}\label{fig:globalmmlu-OpenHermes-ko}\end{figure}

\begin{figure}[h!]\centering\resizebox{\linewidth}{!}{\includegraphics[width=\linewidth]{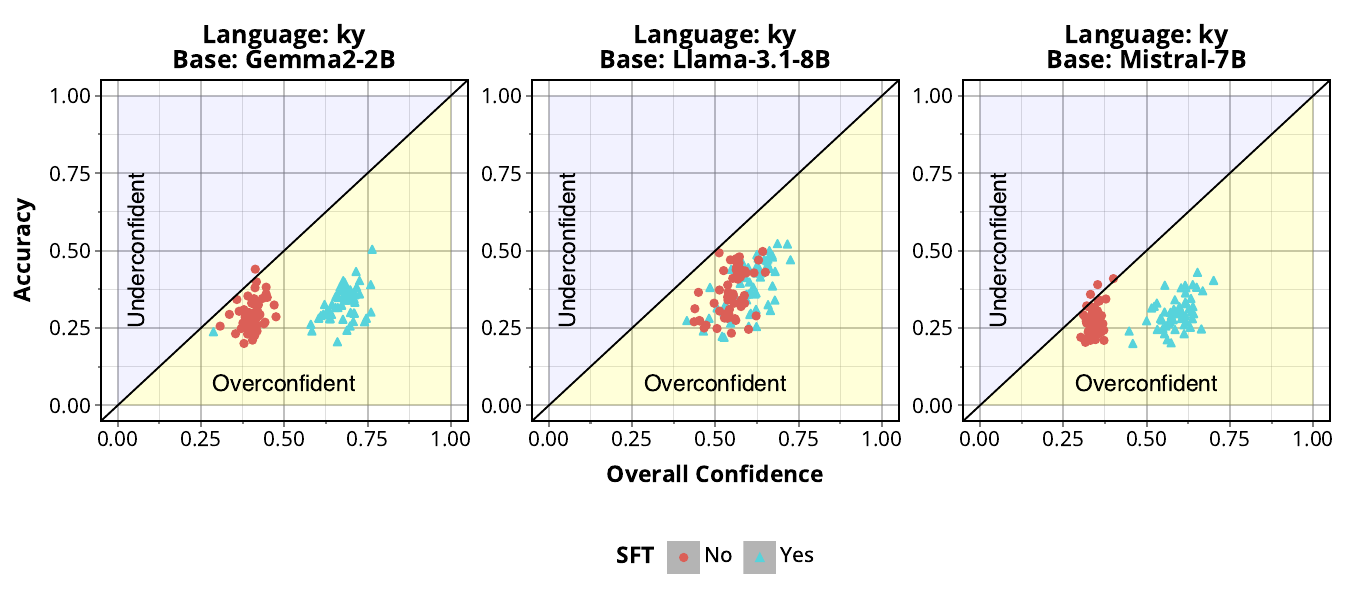}}
\caption{Reliability diagrams for the \textbf{\texttt{GlobalMMLU}} dataset for the \texttt{ky} language.}\label{fig:globalmmlu-base-ky}\end{figure}
\begin{figure}[h!]\centering\resizebox{\linewidth}{!}{\includegraphics[width=\linewidth]{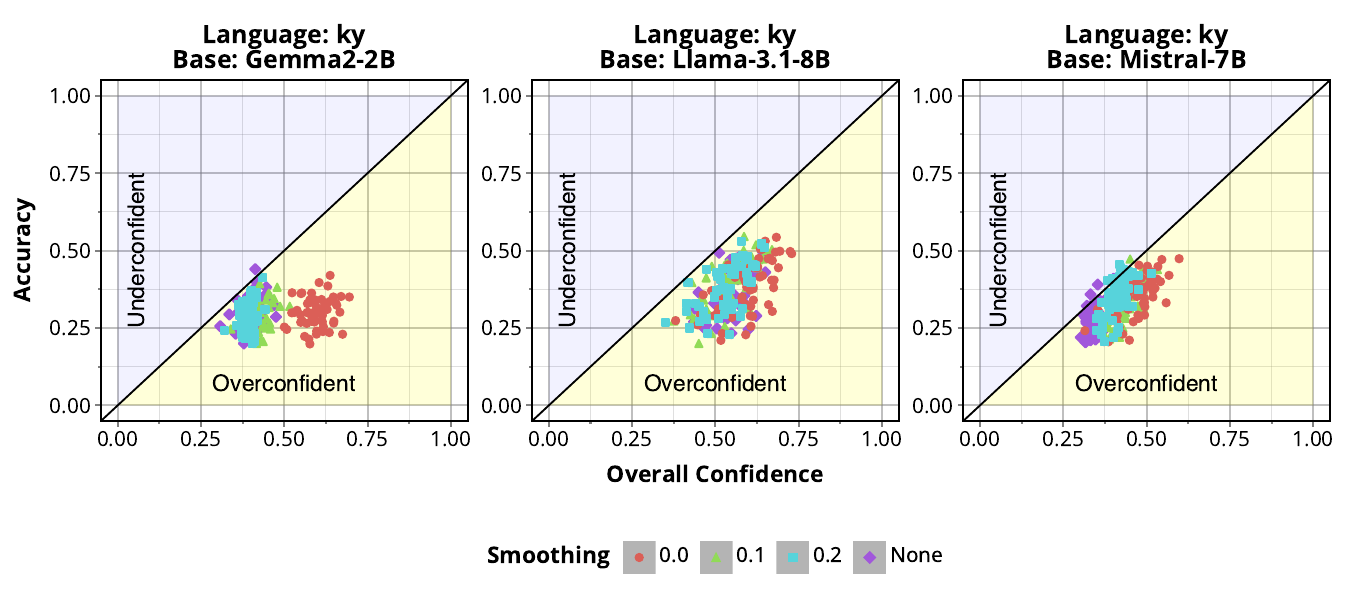}}
\caption{Reliability diagrams for the \textbf{\texttt{GlobalMMLU}} dataset for the \texttt{ky} language after instruction-tuning on the \textbf{\texttt{Tulu3Mixture}} dataset.}\label{fig:globalmmlu-Tulu3Mixture-ky}\end{figure}
\begin{figure}[h!]\centering\resizebox{\linewidth}{!}{\includegraphics[width=\linewidth]{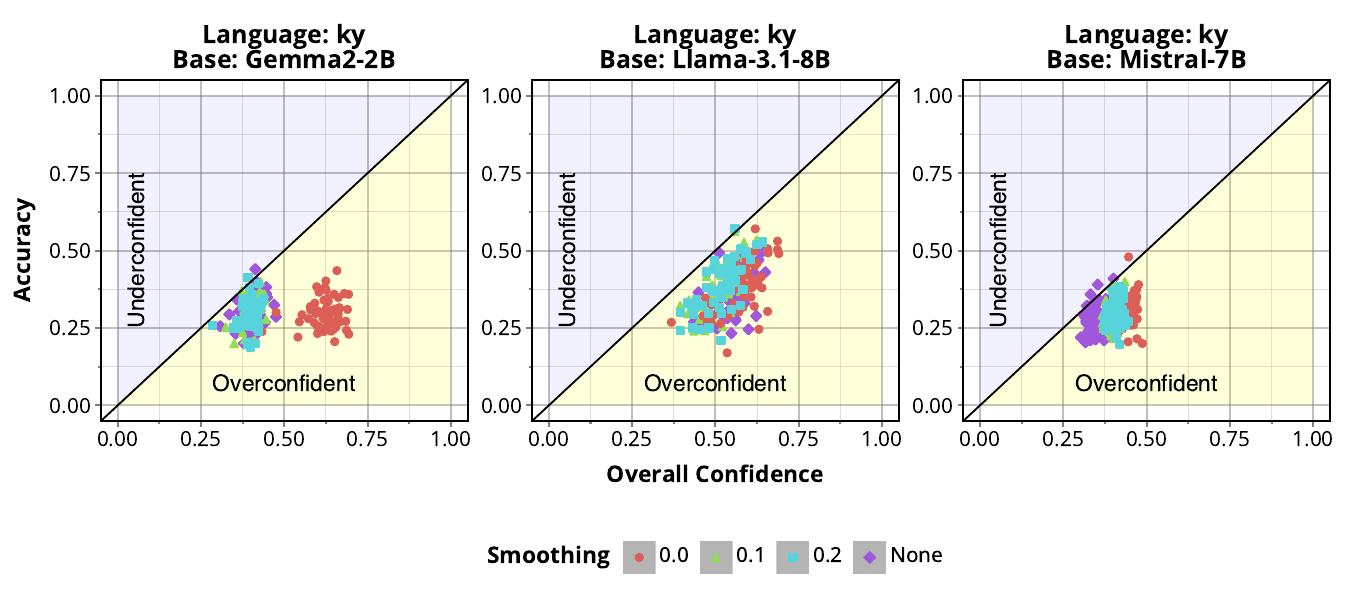}}
\caption{Reliability diagrams for the \textbf{\texttt{GlobalMMLU}} dataset for the \texttt{ky} language after instruction-tuning on the \textbf{\texttt{OpenHermes}} dataset.}\label{fig:globalmmlu-OpenHermes-ky}\end{figure}

\begin{figure}[h!]\centering\resizebox{\linewidth}{!}{\includegraphics[width=\linewidth]{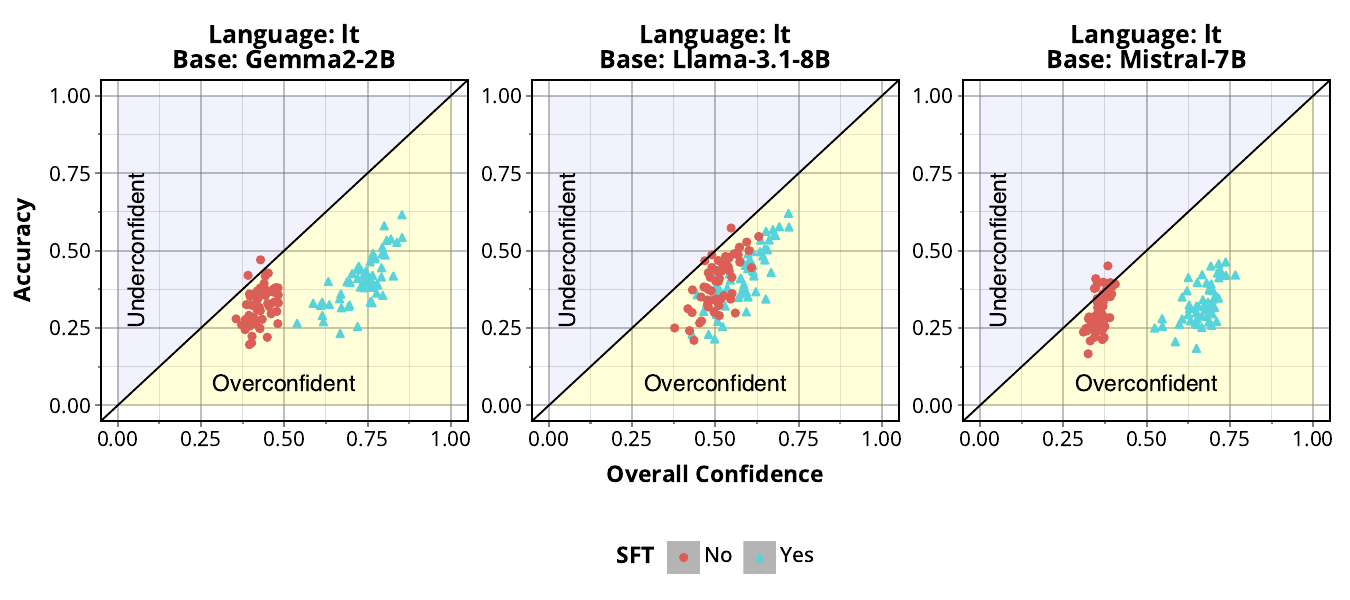}}
\caption{Reliability diagrams for the \textbf{\texttt{GlobalMMLU}} dataset for the \texttt{lt} language.}\label{fig:globalmmlu-base-lt}\end{figure}
\begin{figure}[h!]\centering\resizebox{\linewidth}{!}{\includegraphics[width=\linewidth]{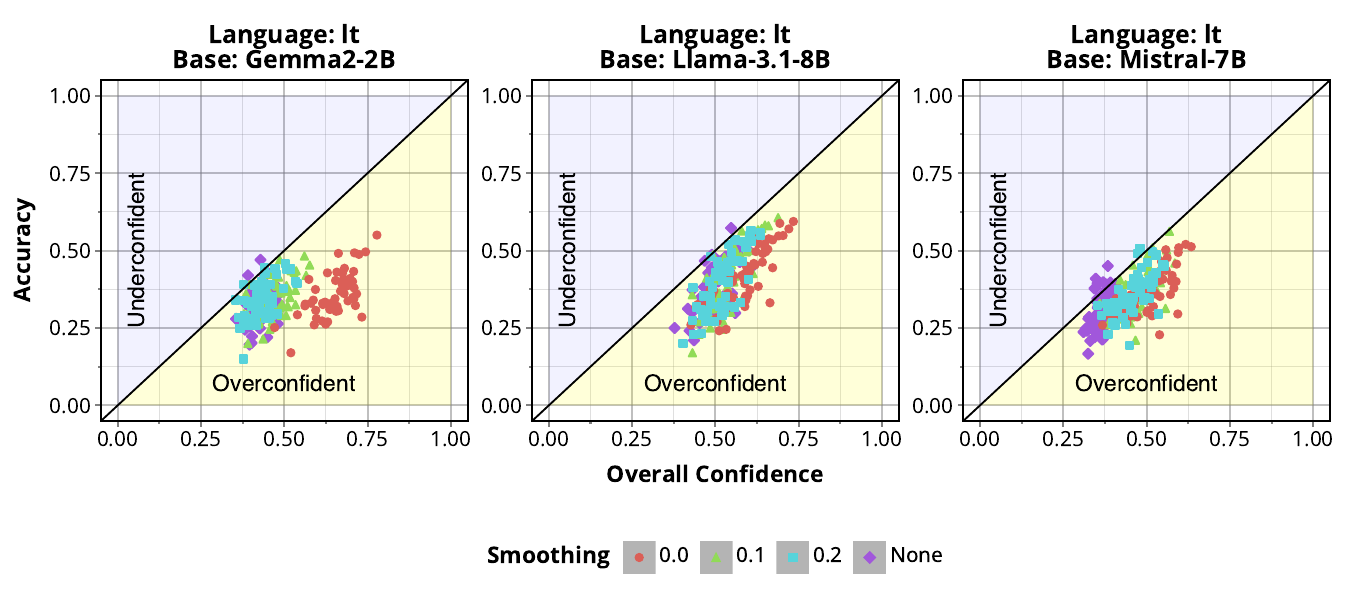}}
\caption{Reliability diagrams for the \textbf{\texttt{GlobalMMLU}} dataset for the \texttt{lt} language after instruction-tuning on the \textbf{\texttt{Tulu3Mixture}} dataset.}\label{fig:globalmmlu-Tulu3Mixture-lt}\end{figure}
\begin{figure}[h!]\centering\resizebox{\linewidth}{!}{\includegraphics[width=\linewidth]{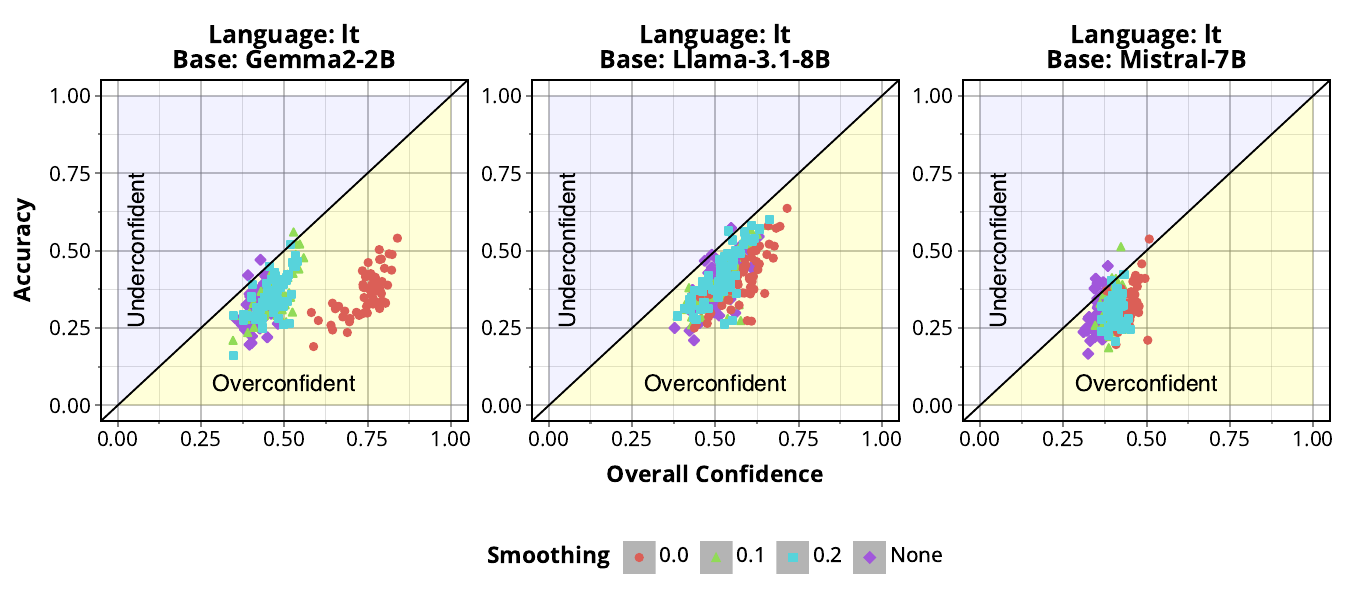}}
\caption{Reliability diagrams for the \textbf{\texttt{GlobalMMLU}} dataset for the \texttt{lt} language after instruction-tuning on the \textbf{\texttt{OpenHermes}} dataset.}\label{fig:globalmmlu-OpenHermes-lt}\end{figure}

\begin{figure}[h!]\centering\resizebox{\linewidth}{!}{\includegraphics[width=\linewidth]{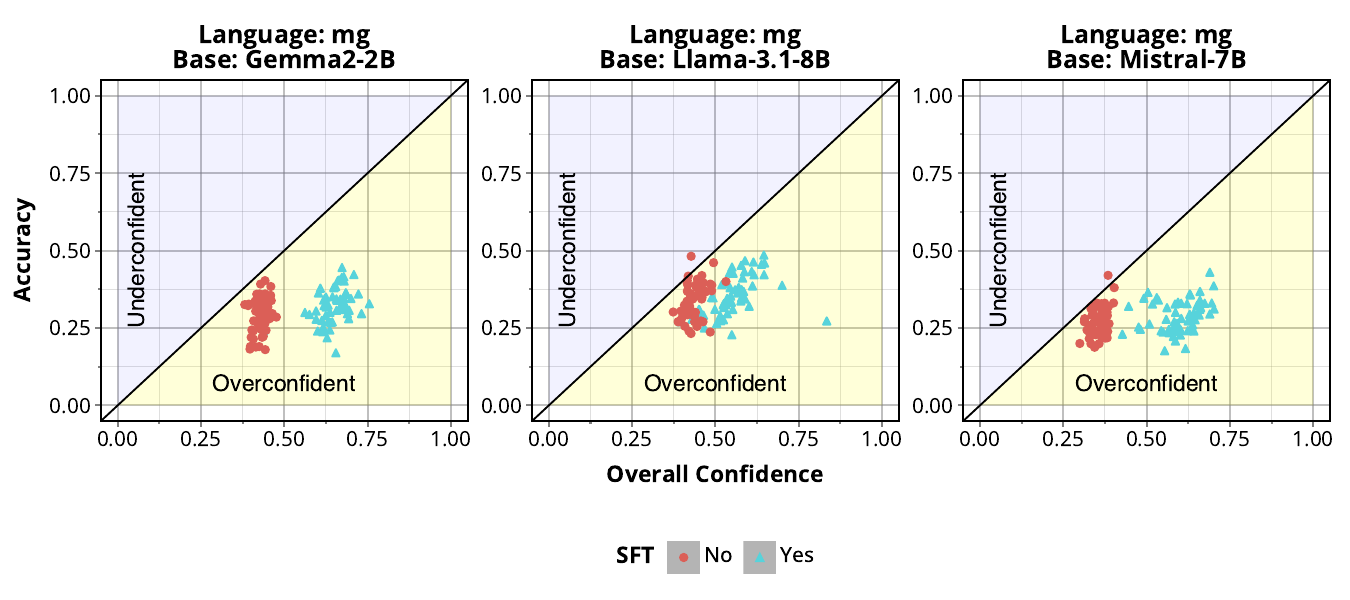}}
\caption{Reliability diagrams for the \textbf{\texttt{GlobalMMLU}} dataset for the \texttt{mg} language.}\label{fig:globalmmlu-base-mg}\end{figure}
\begin{figure}[h!]\centering\resizebox{\linewidth}{!}{\includegraphics[width=\linewidth]{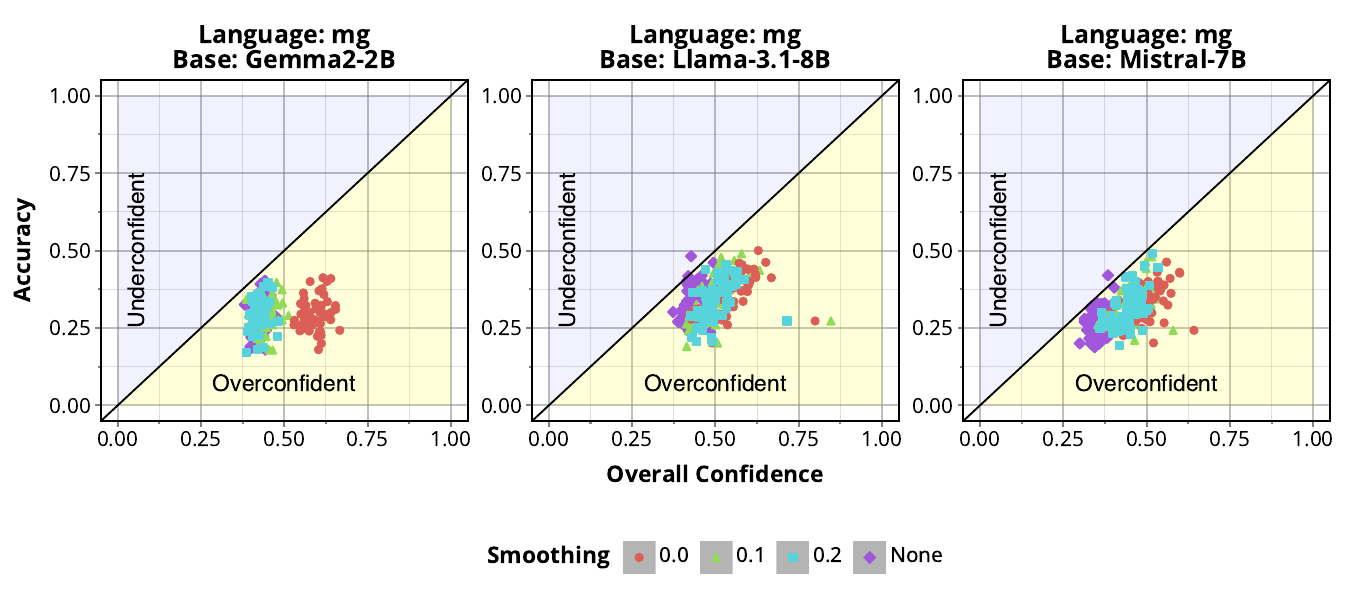}}
\caption{Reliability diagrams for the \textbf{\texttt{GlobalMMLU}} dataset for the \texttt{mg} language after instruction-tuning on the \textbf{\texttt{Tulu3Mixture}} dataset.}\label{fig:globalmmlu-Tulu3Mixture-mg}\end{figure}
\begin{figure}[h!]\centering\resizebox{\linewidth}{!}{\includegraphics[width=\linewidth]{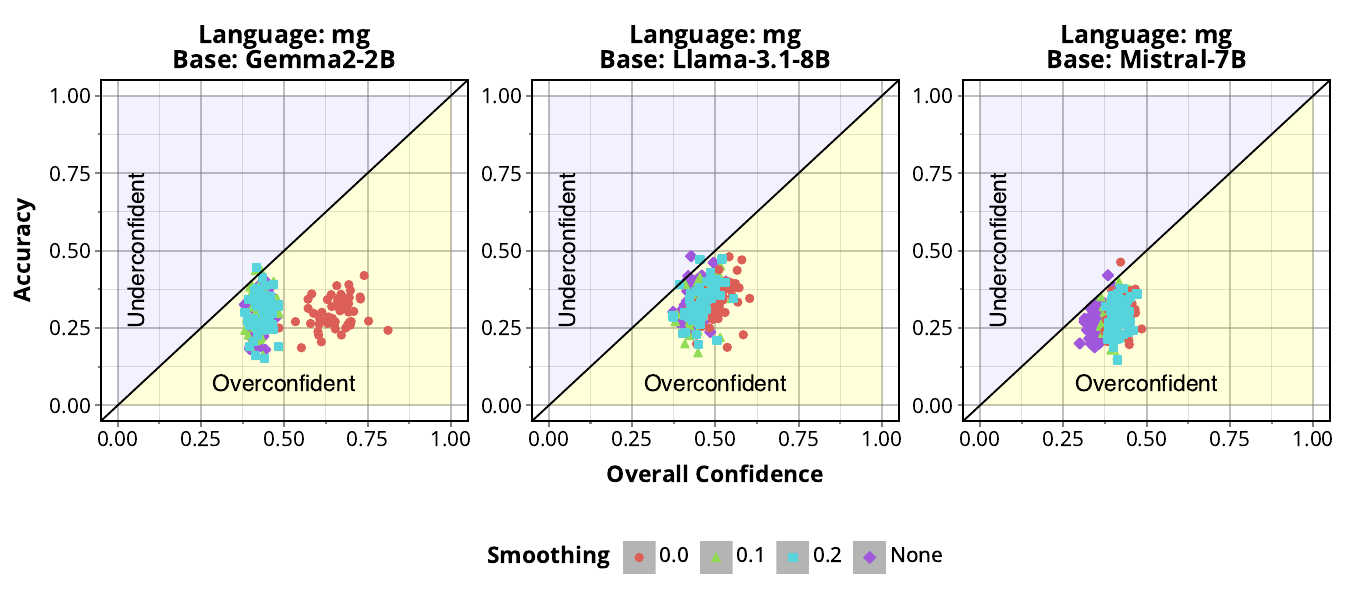}}
\caption{Reliability diagrams for the \textbf{\texttt{GlobalMMLU}} dataset for the \texttt{mg} language after instruction-tuning on the \textbf{\texttt{OpenHermes}} dataset.}\label{fig:globalmmlu-OpenHermes-mg}\end{figure}

\begin{figure}[h!]\centering\resizebox{\linewidth}{!}{\includegraphics[width=\linewidth]{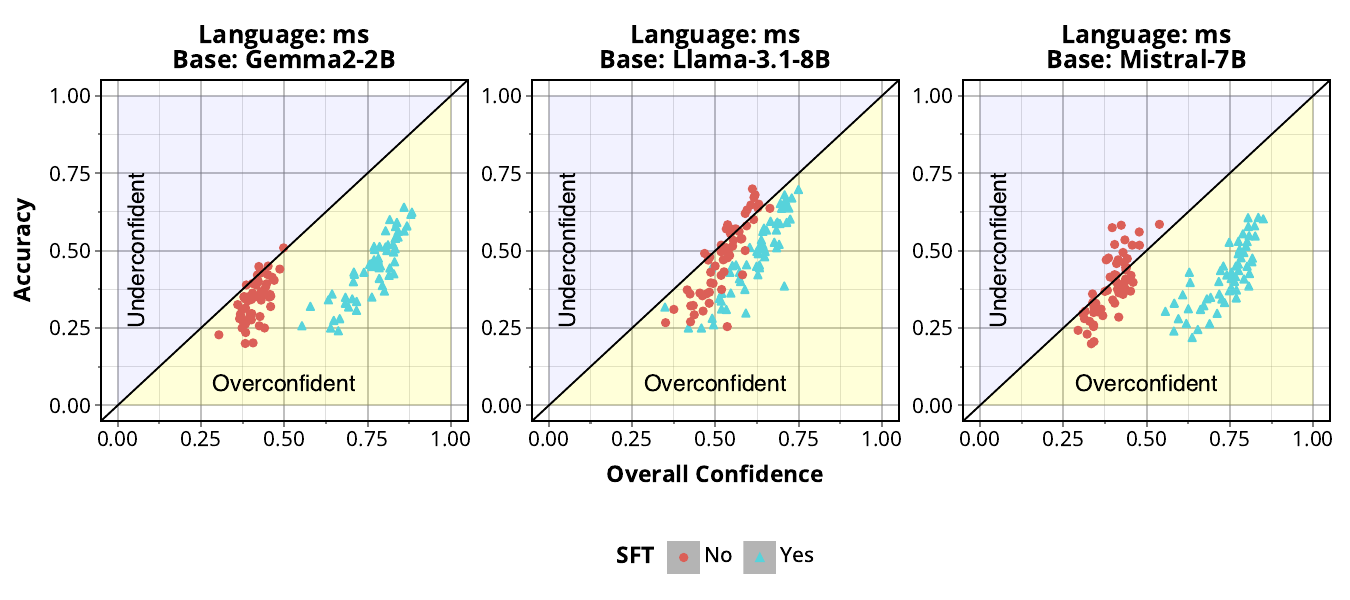}}
\caption{Reliability diagrams for the \textbf{\texttt{GlobalMMLU}} dataset for the \texttt{ms} language.}\label{fig:globalmmlu-base-ms}\end{figure}
\begin{figure}[h!]\centering\resizebox{\linewidth}{!}{\includegraphics[width=\linewidth]{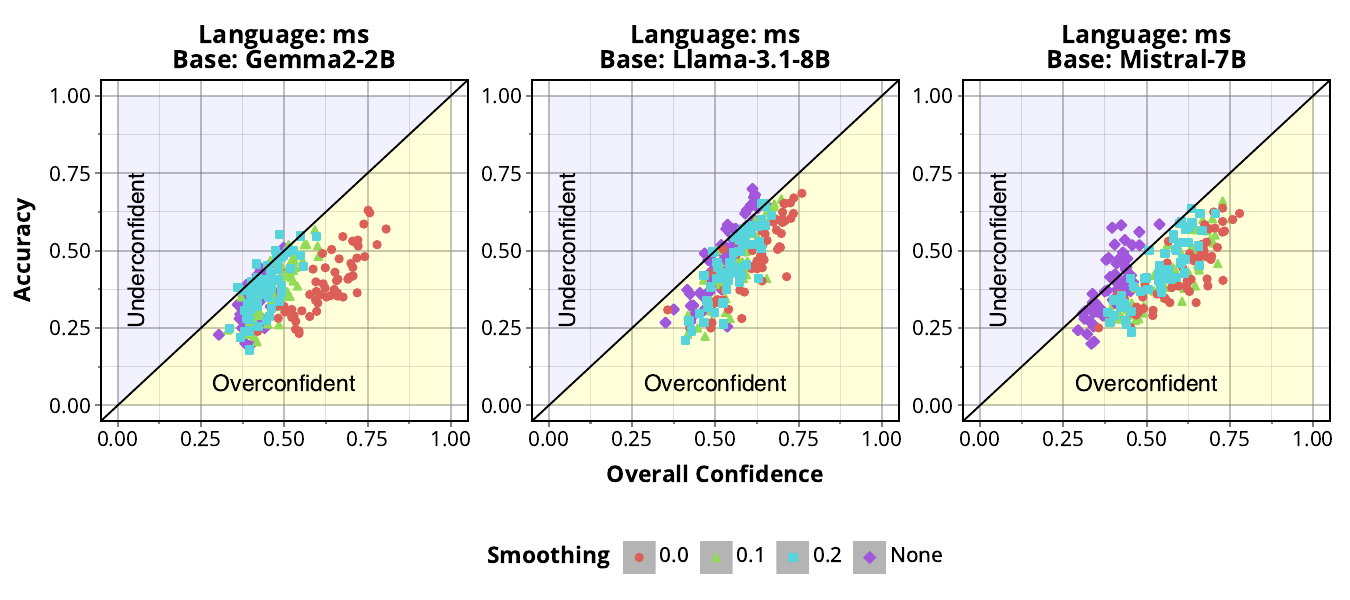}}
\caption{Reliability diagrams for the \textbf{\texttt{GlobalMMLU}} dataset for the \texttt{ms} language after instruction-tuning on the \textbf{\texttt{Tulu3Mixture}} dataset.}\label{fig:globalmmlu-Tulu3Mixture-ms}\end{figure}
\begin{figure}[h!]\centering\resizebox{\linewidth}{!}{\includegraphics[width=\linewidth]{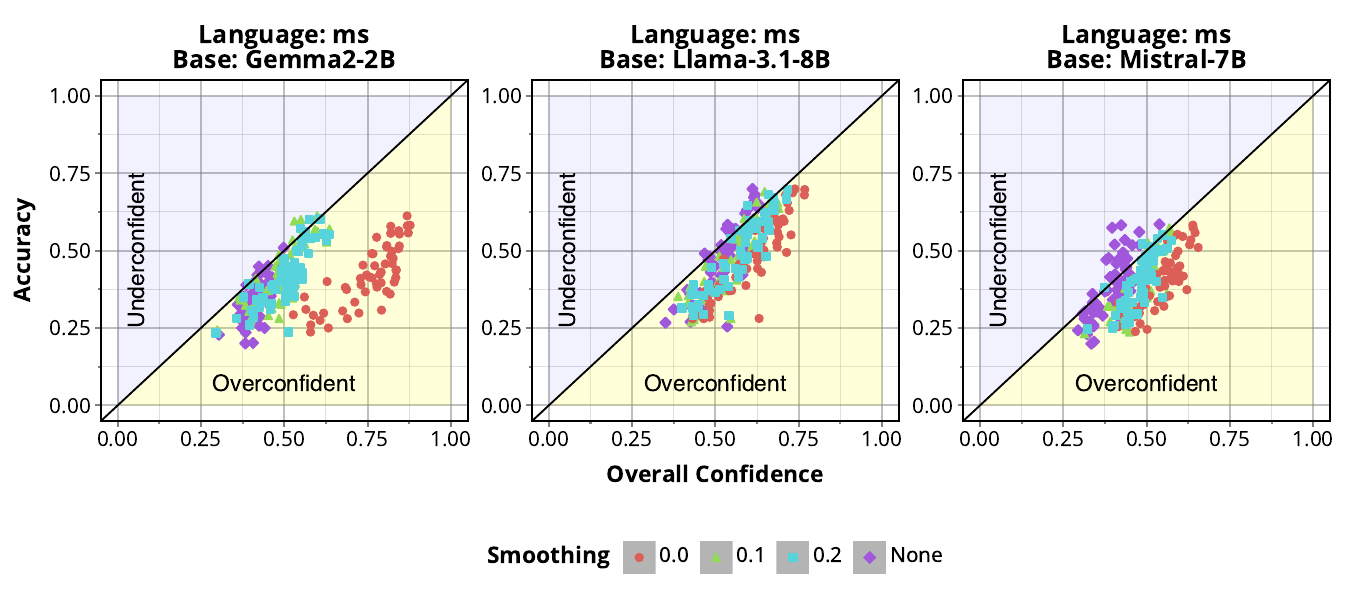}}
\caption{Reliability diagrams for the \textbf{\texttt{GlobalMMLU}} dataset for the \texttt{ms} language after instruction-tuning on the \textbf{\texttt{OpenHermes}} dataset.}\label{fig:globalmmlu-OpenHermes-ms}\end{figure}

\begin{figure}[h!]\centering\resizebox{\linewidth}{!}{\includegraphics[width=\linewidth]{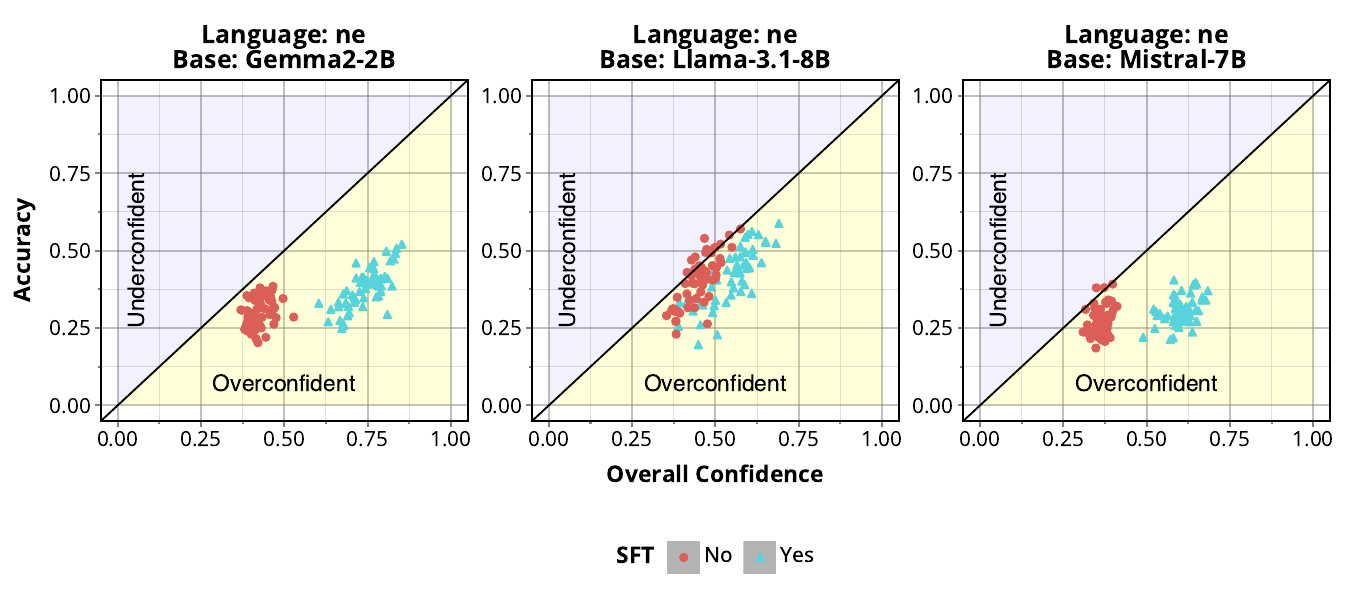}}
\caption{Reliability diagrams for the \textbf{\texttt{GlobalMMLU}} dataset for the \texttt{ne} language.}\label{fig:globalmmlu-base-ne}\end{figure}
\begin{figure}[h!]\centering\resizebox{\linewidth}{!}{\includegraphics[width=\linewidth]{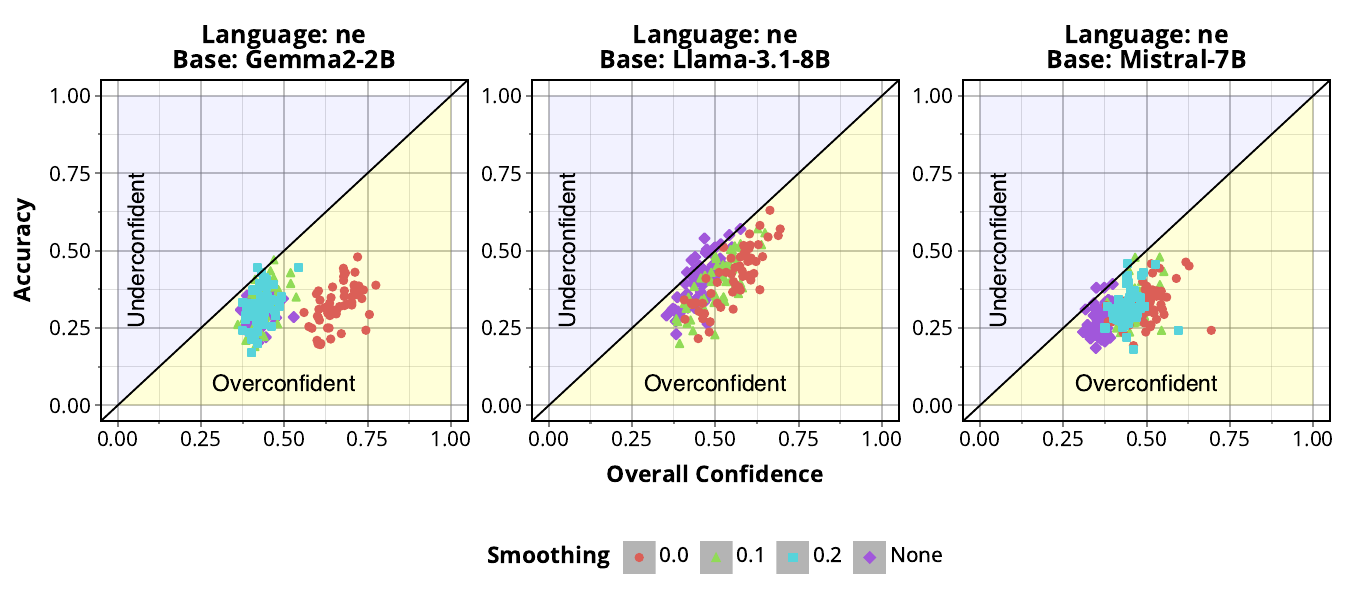}}
\caption{Reliability diagrams for the \textbf{\texttt{GlobalMMLU}} dataset for the \texttt{ne} language after instruction-tuning on the \textbf{\texttt{Tulu3Mixture}} dataset.}\label{fig:globalmmlu-Tulu3Mixture-ne}\end{figure}
\begin{figure}[h!]\centering\resizebox{\linewidth}{!}{\includegraphics[width=\linewidth]{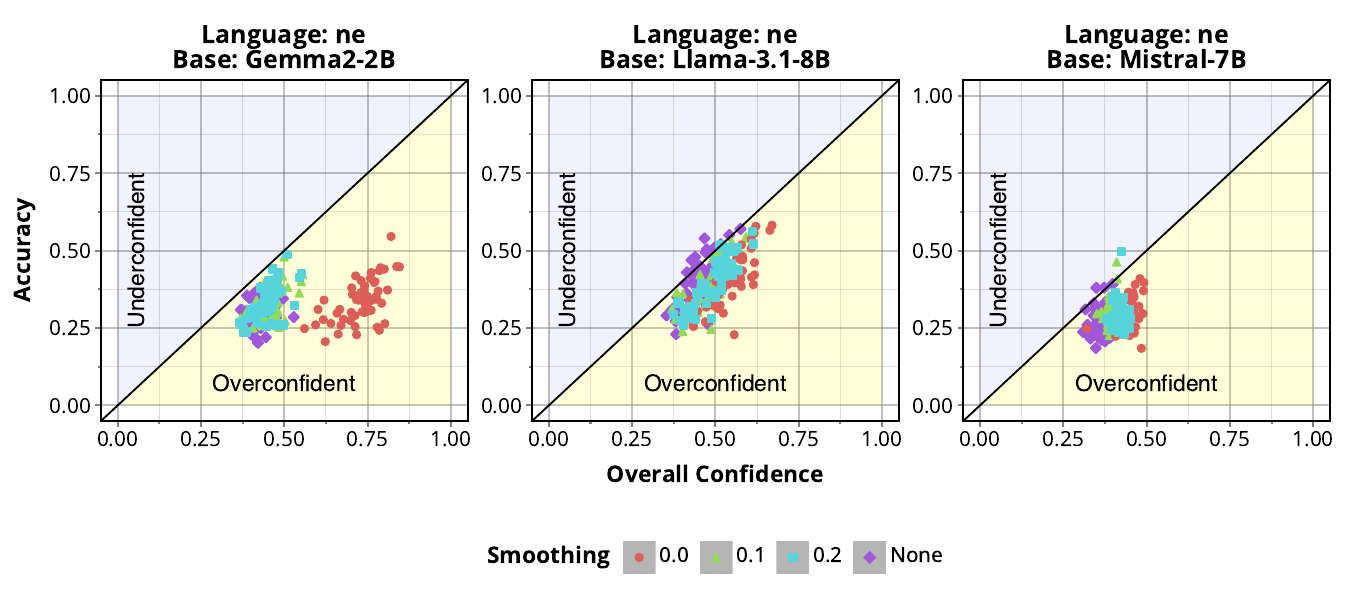}}
\caption{Reliability diagrams for the \textbf{\texttt{GlobalMMLU}} dataset for the \texttt{ne} language after instruction-tuning on the \textbf{\texttt{OpenHermes}} dataset.}\label{fig:globalmmlu-OpenHermes-ne}\end{figure}

\begin{figure}[h!]\centering\resizebox{\linewidth}{!}{\includegraphics[width=\linewidth]{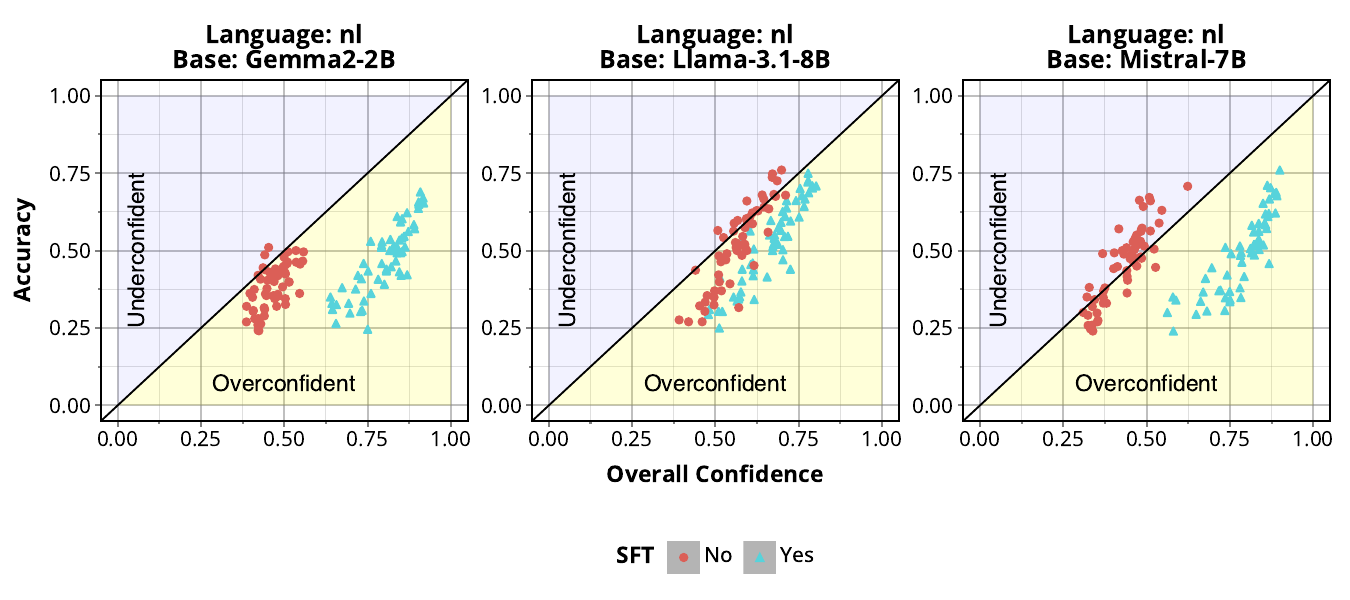}}
\caption{Reliability diagrams for the \textbf{\texttt{GlobalMMLU}} dataset for the \texttt{nl} language.}\label{fig:globalmmlu-base-nl}\end{figure}
\begin{figure}[h!]\centering\resizebox{\linewidth}{!}{\includegraphics[width=\linewidth]{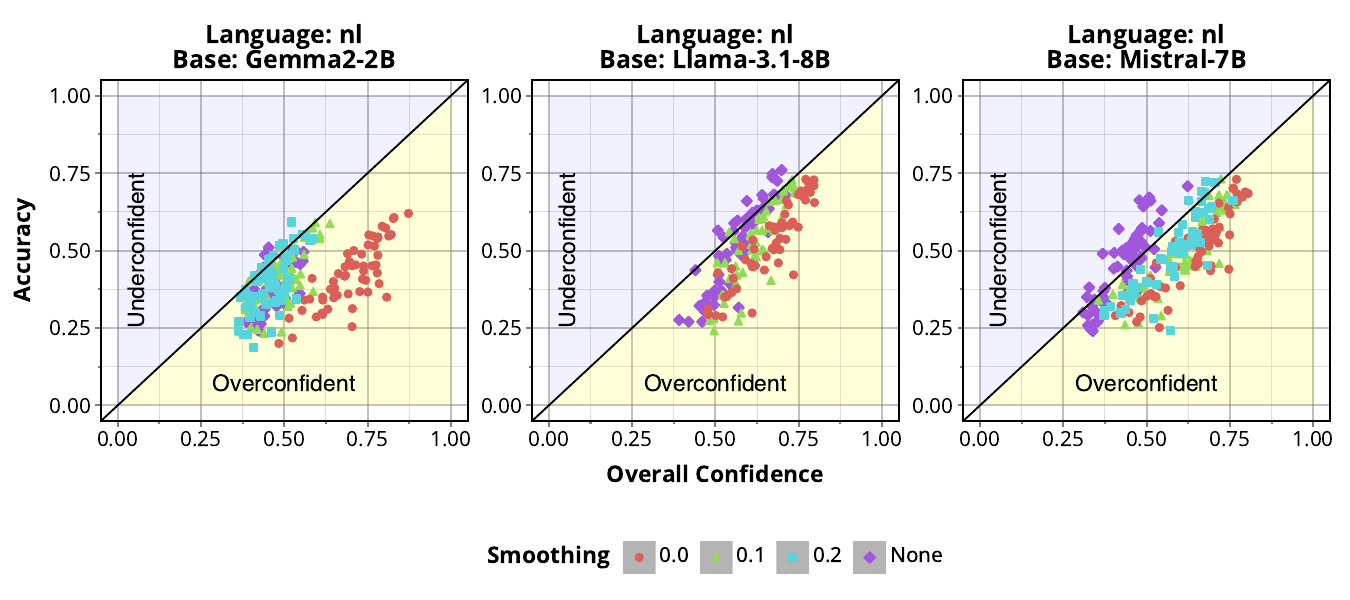}}
\caption{Reliability diagrams for the \textbf{\texttt{GlobalMMLU}} dataset for the \texttt{nl} language after instruction-tuning on the \textbf{\texttt{Tulu3Mixture}} dataset.}\label{fig:globalmmlu-Tulu3Mixture-nl}\end{figure}
\begin{figure}[h!]\centering\resizebox{\linewidth}{!}{\includegraphics[width=\linewidth]{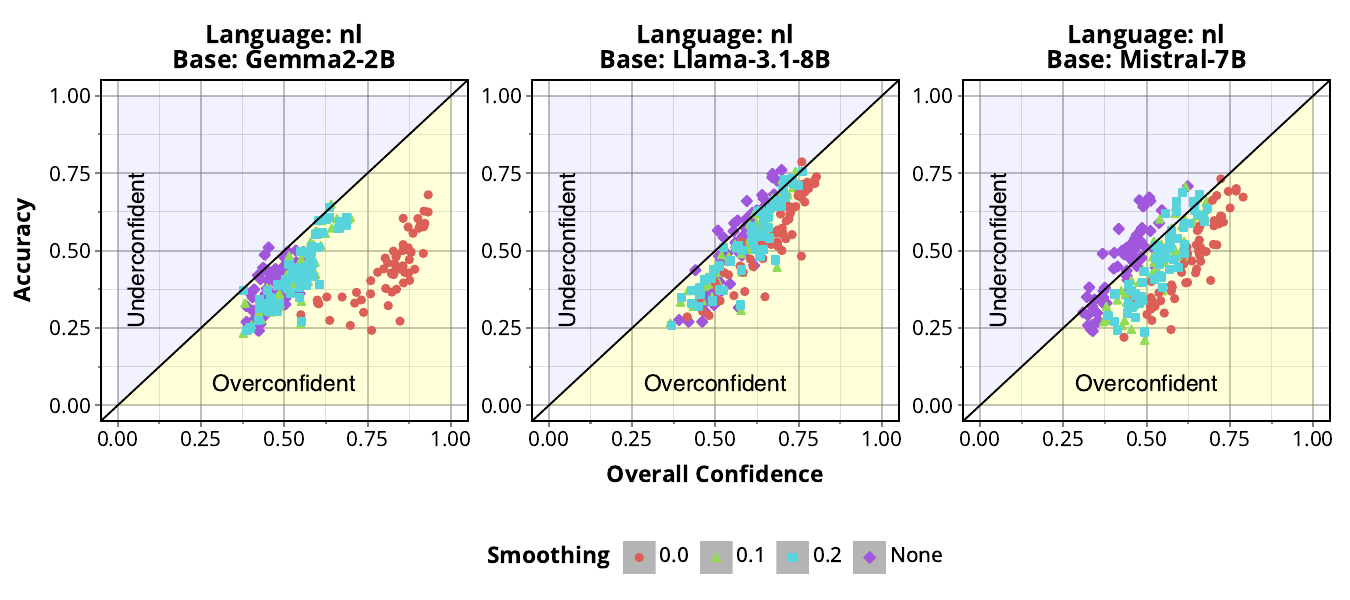}}
\caption{Reliability diagrams for the \textbf{\texttt{GlobalMMLU}} dataset for the \texttt{nl} language after instruction-tuning on the \textbf{\texttt{OpenHermes}} dataset.}\label{fig:globalmmlu-OpenHermes-nl}\end{figure}

\begin{figure}[h!]\centering\resizebox{\linewidth}{!}{\includegraphics[width=\linewidth]{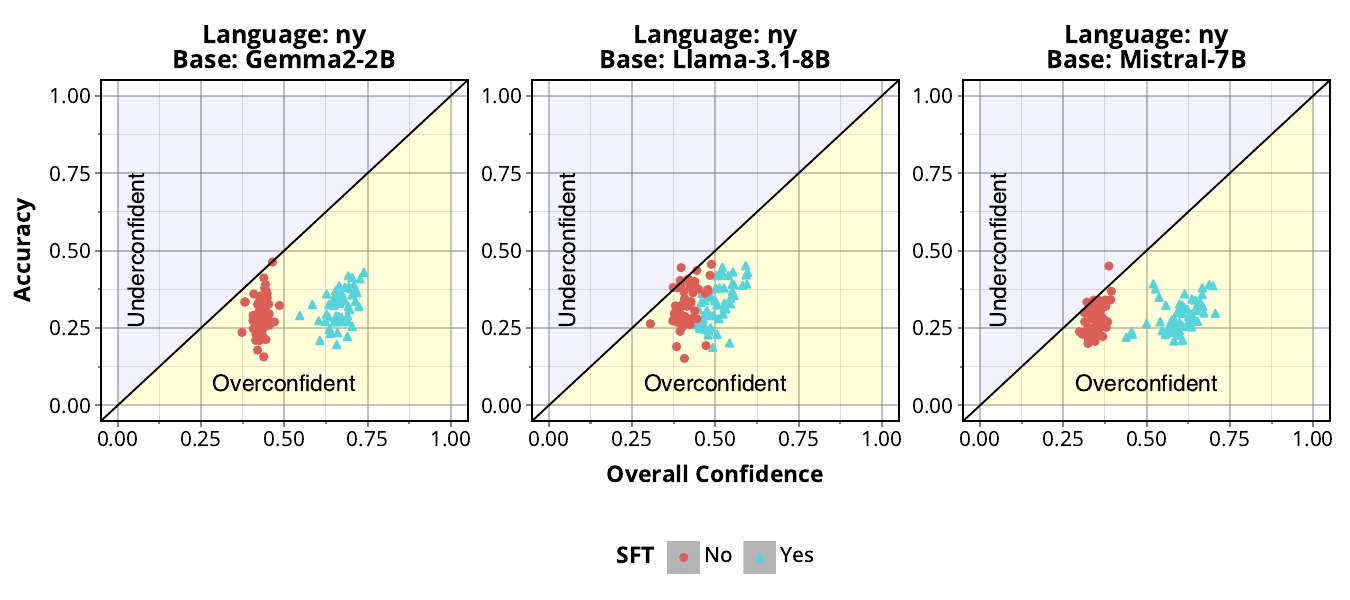}}
\caption{Reliability diagrams for the \textbf{\texttt{GlobalMMLU}} dataset for the \texttt{ny} language.}\label{fig:globalmmlu-base-ny}\end{figure}
\begin{figure}[h!]\centering\resizebox{\linewidth}{!}{\includegraphics[width=\linewidth]{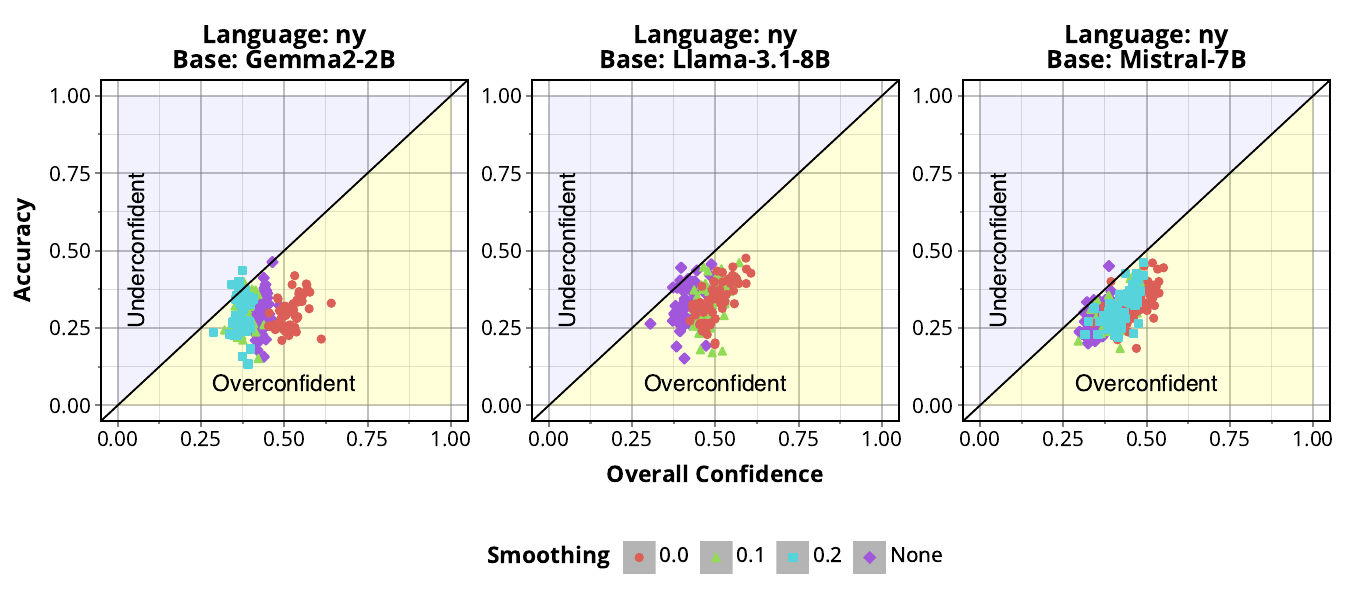}}
\caption{Reliability diagrams for the \textbf{\texttt{GlobalMMLU}} dataset for the \texttt{ny} language after instruction-tuning on the \textbf{\texttt{Tulu3Mixture}} dataset.}\label{fig:globalmmlu-Tulu3Mixture-ny}\end{figure}
\begin{figure}[h!]\centering\resizebox{\linewidth}{!}{\includegraphics[width=\linewidth]{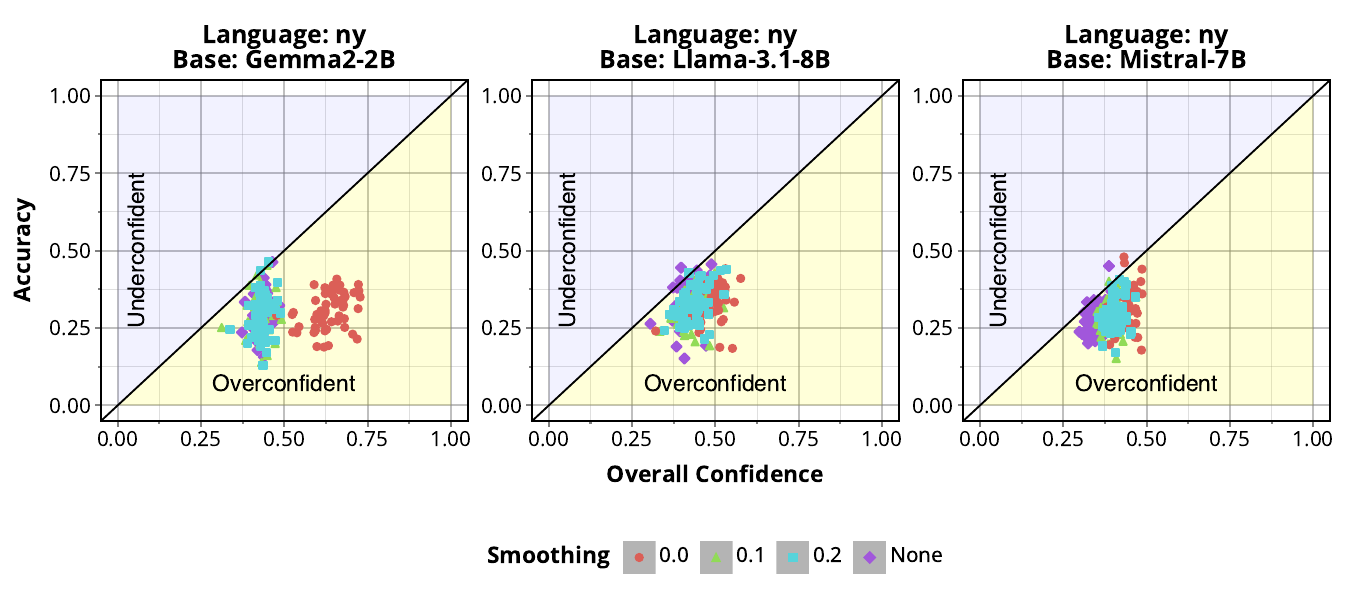}}
\caption{Reliability diagrams for the \textbf{\texttt{GlobalMMLU}} dataset for the \texttt{ny} language after instruction-tuning on the \textbf{\texttt{OpenHermes}} dataset.}\label{fig:globalmmlu-OpenHermes-ny}\end{figure}

\begin{figure}[h!]\centering\resizebox{\linewidth}{!}{\includegraphics[width=\linewidth]{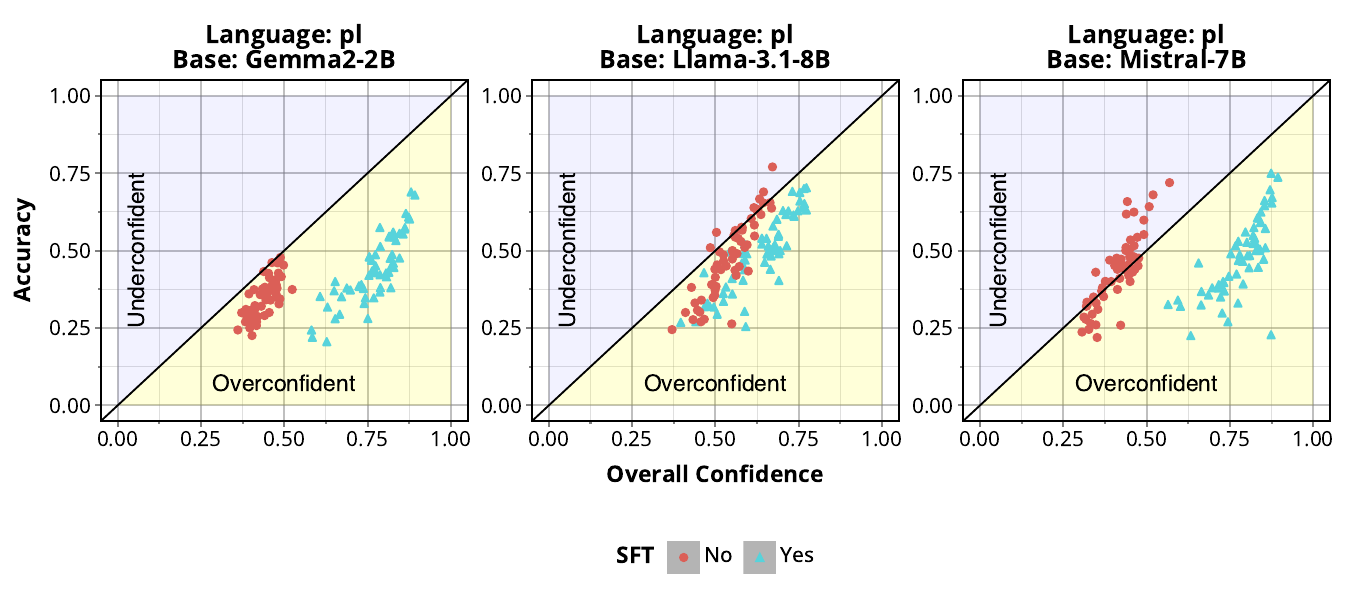}}
\caption{Reliability diagrams for the \textbf{\texttt{GlobalMMLU}} dataset for the \texttt{pl} language.}\label{fig:globalmmlu-base-pl}\end{figure}
\begin{figure}[h!]\centering\resizebox{\linewidth}{!}{\includegraphics[width=\linewidth]{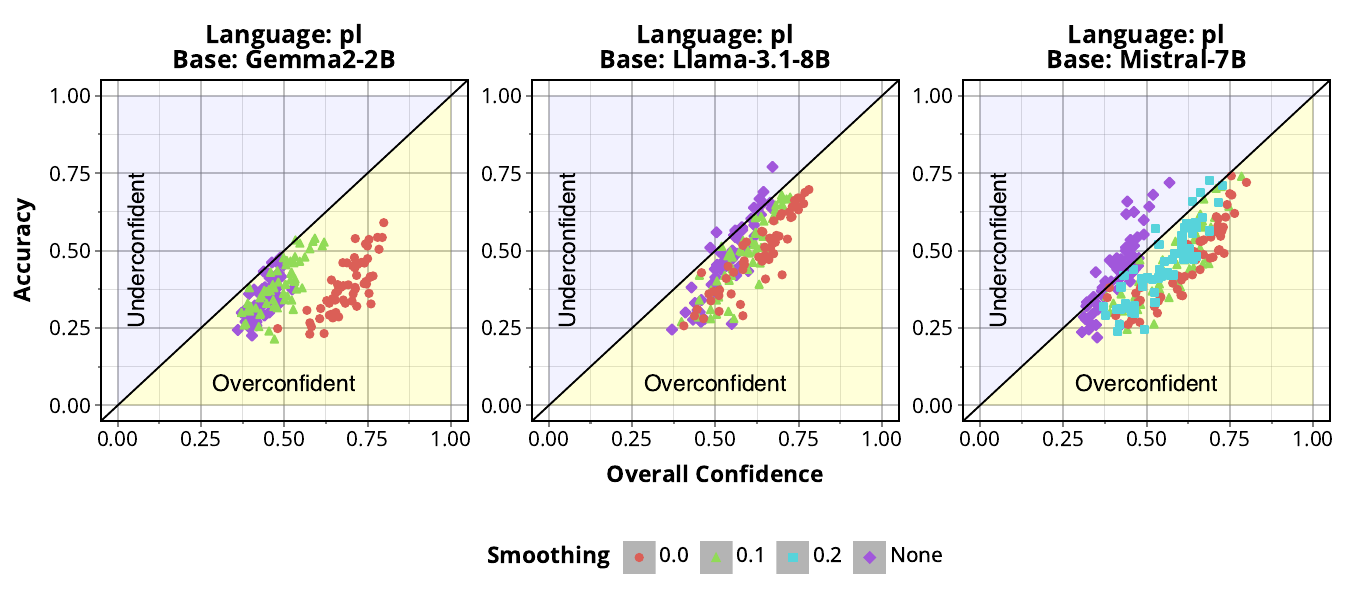}}
\caption{Reliability diagrams for the \textbf{\texttt{GlobalMMLU}} dataset for the \texttt{pl} language after instruction-tuning on the \textbf{\texttt{Tulu3Mixture}} dataset.}\label{fig:globalmmlu-Tulu3Mixture-pl}\end{figure}
\begin{figure}[h!]\centering\resizebox{\linewidth}{!}{\includegraphics[width=\linewidth]{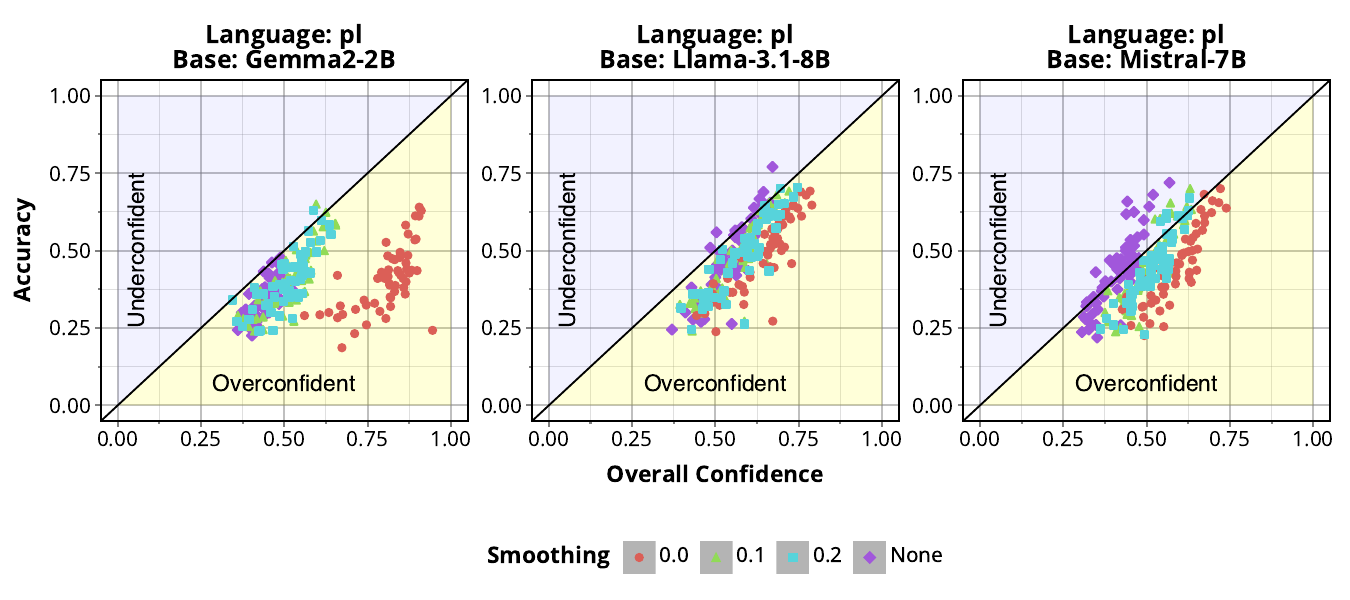}}
\caption{Reliability diagrams for the \textbf{\texttt{GlobalMMLU}} dataset for the \texttt{pl} language after instruction-tuning on the \textbf{\texttt{OpenHermes}} dataset.}\label{fig:globalmmlu-OpenHermes-pl}\end{figure}

\begin{figure}[h!]\centering\resizebox{\linewidth}{!}{\includegraphics[width=\linewidth]{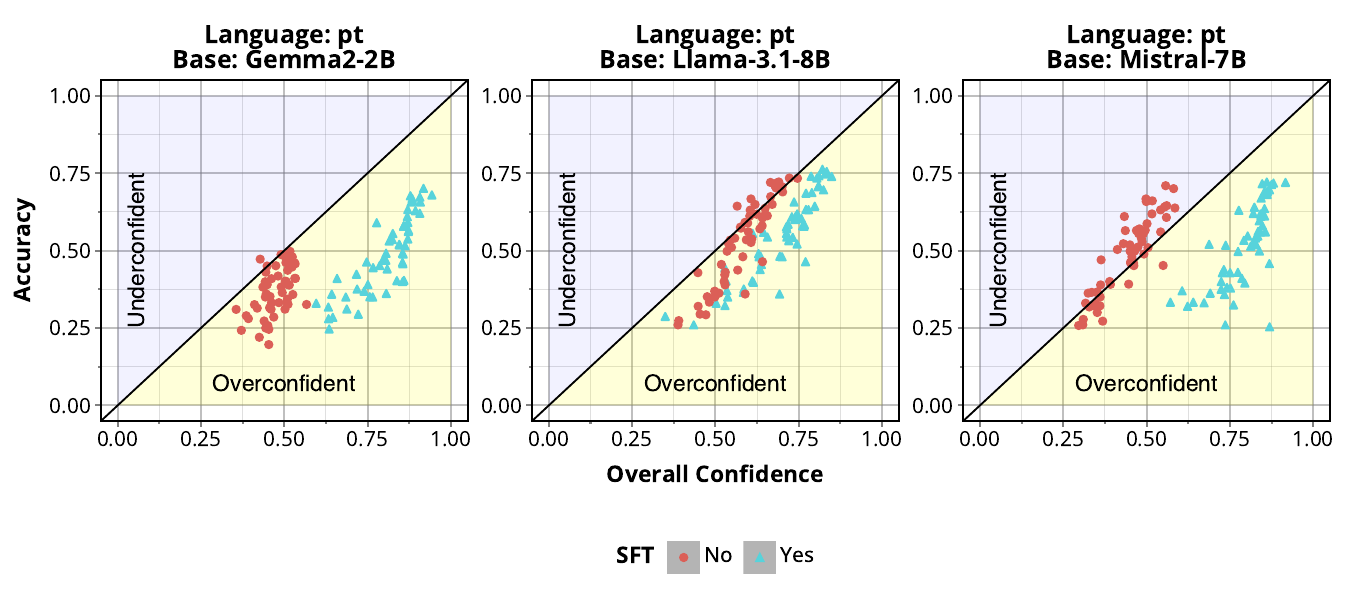}}
\caption{Reliability diagrams for the \textbf{\texttt{GlobalMMLU}} dataset for the \texttt{pt} language.}\label{fig:globalmmlu-base-pt}\end{figure}
\begin{figure}[h!]\centering\resizebox{\linewidth}{!}{\includegraphics[width=\linewidth]{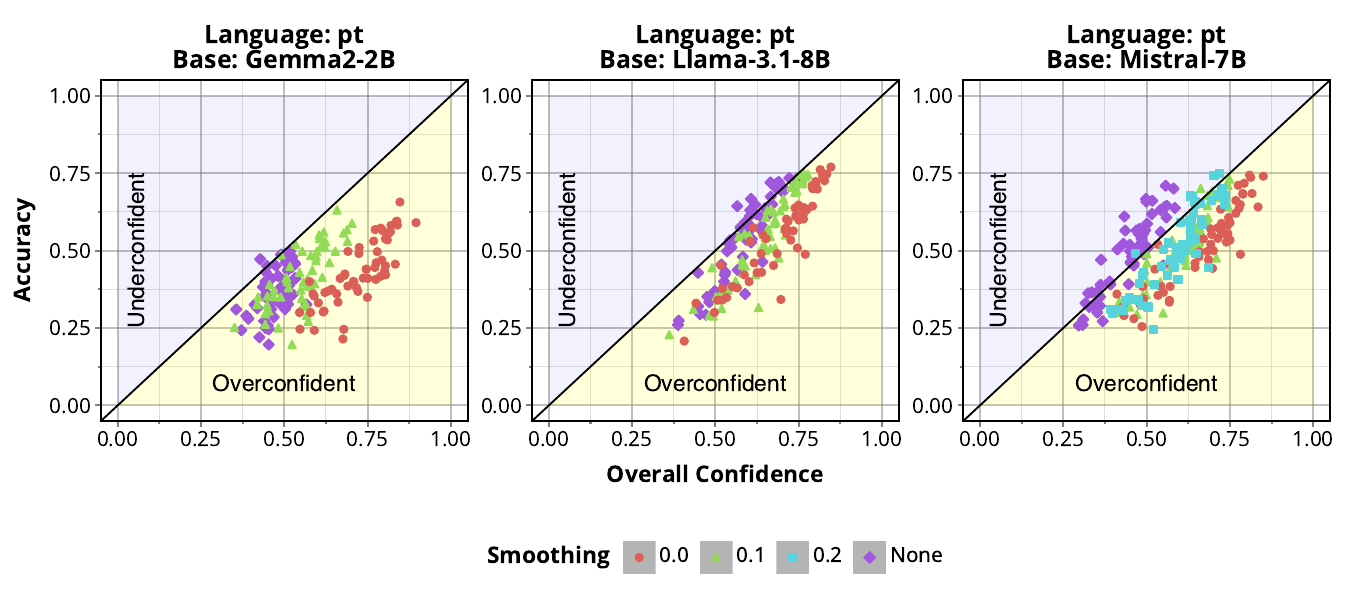}}
\caption{Reliability diagrams for the \textbf{\texttt{GlobalMMLU}} dataset for the \texttt{pt} language after instruction-tuning on the \textbf{\texttt{Tulu3Mixture}} dataset.}\label{fig:globalmmlu-Tulu3Mixture-pt}\end{figure}
\begin{figure}[h!]\centering\resizebox{\linewidth}{!}{\includegraphics[width=\linewidth]{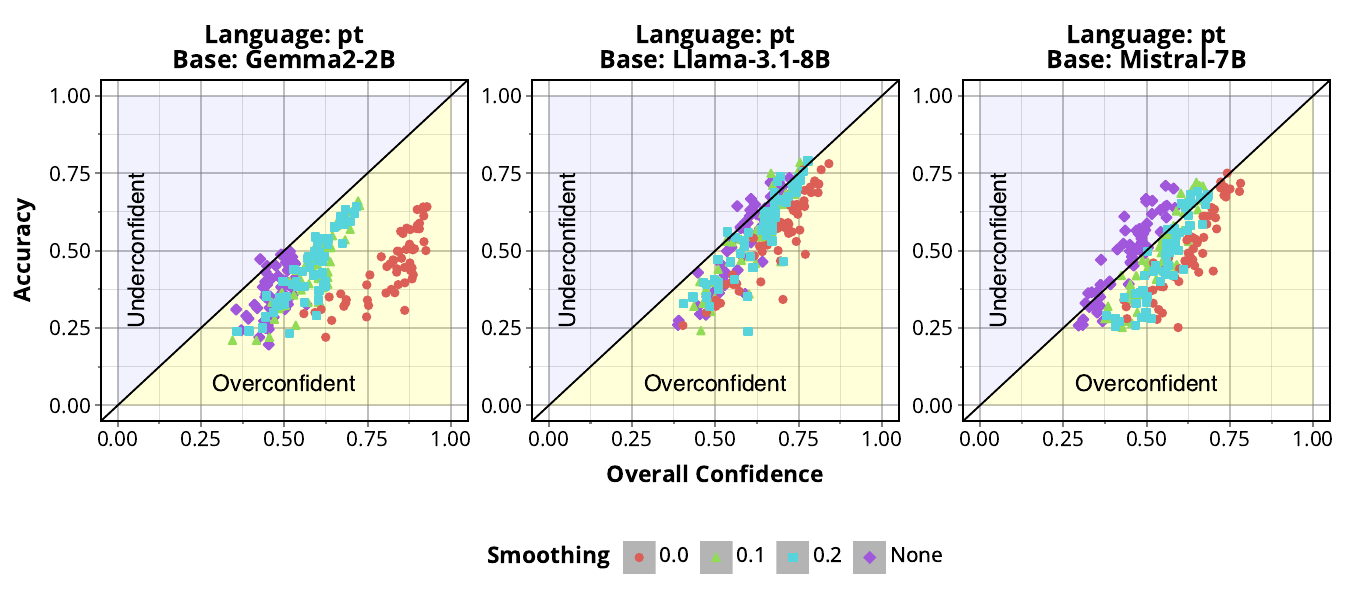}}
\caption{Reliability diagrams for the \textbf{\texttt{GlobalMMLU}} dataset for the \texttt{pt} language after instruction-tuning on the \textbf{\texttt{OpenHermes}} dataset.}\label{fig:globalmmlu-OpenHermes-pt}\end{figure}

\begin{figure}[h!]\centering\resizebox{\linewidth}{!}{\includegraphics[width=\linewidth]{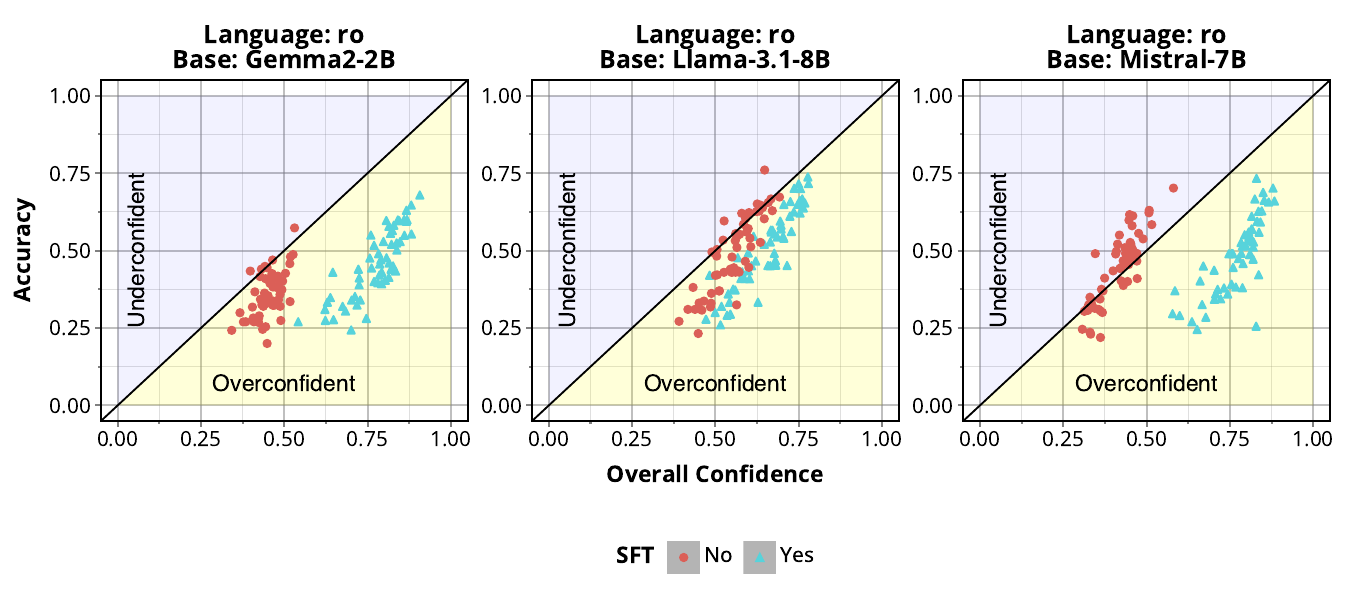}}
\caption{Reliability diagrams for the \textbf{\texttt{GlobalMMLU}} dataset for the \texttt{ro} language.}\label{fig:globalmmlu-base-ro}\end{figure}
\begin{figure}[h!]\centering\resizebox{\linewidth}{!}{\includegraphics[width=\linewidth]{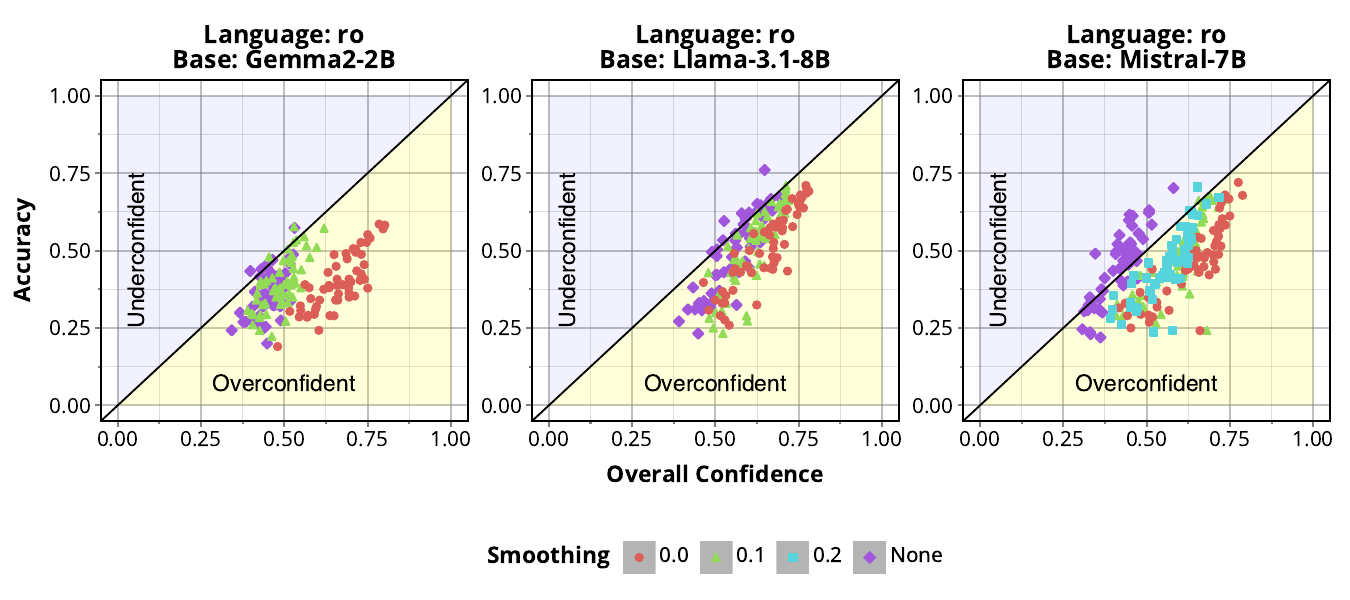}}
\caption{Reliability diagrams for the \textbf{\texttt{GlobalMMLU}} dataset for the \texttt{ro} language after instruction-tuning on the \textbf{\texttt{Tulu3Mixture}} dataset.}\label{fig:globalmmlu-Tulu3Mixture-ro}\end{figure}
\begin{figure}[h!]\centering\resizebox{\linewidth}{!}{\includegraphics[width=\linewidth]{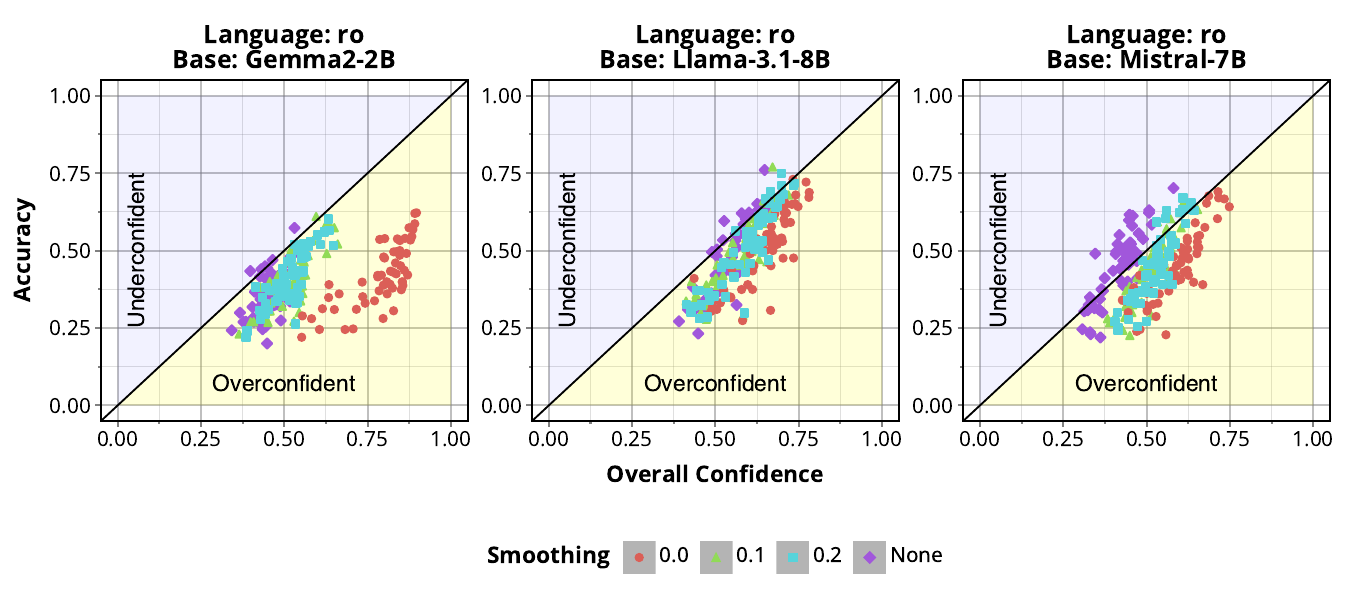}}
\caption{Reliability diagrams for the \textbf{\texttt{GlobalMMLU}} dataset for the \texttt{ro} language after instruction-tuning on the \textbf{\texttt{OpenHermes}} dataset.}\label{fig:globalmmlu-OpenHermes-ro}\end{figure}

\begin{figure}[h!]\centering\resizebox{\linewidth}{!}{\includegraphics[width=\linewidth]{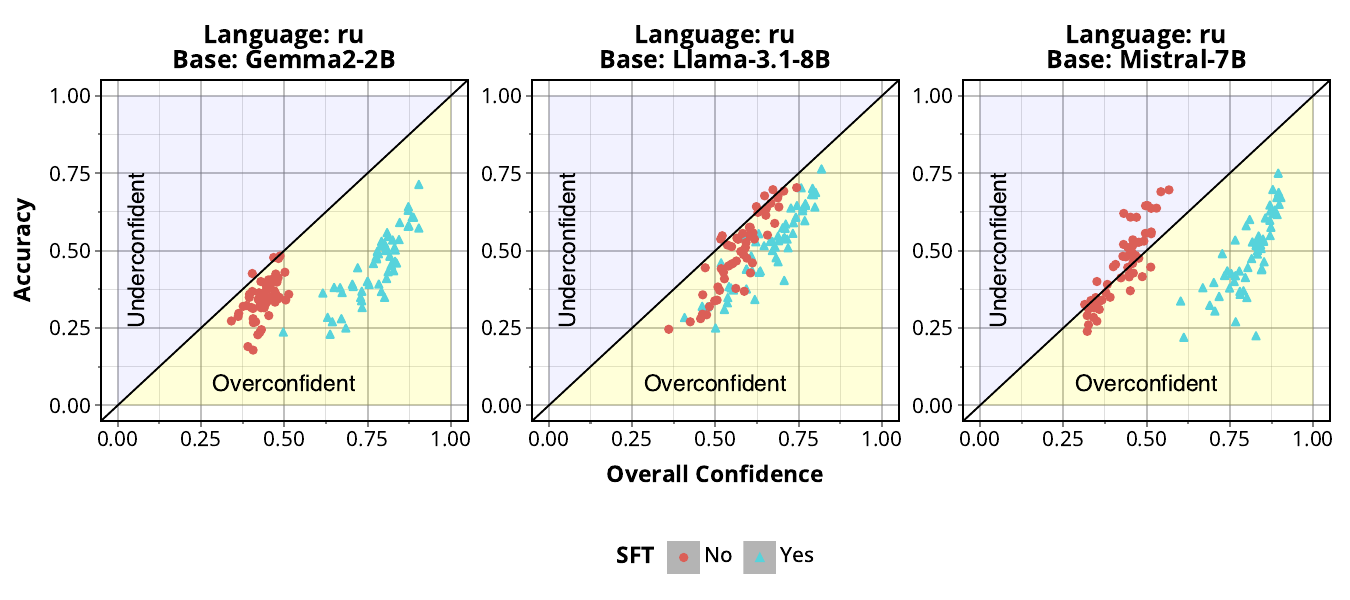}}
\caption{Reliability diagrams for the \textbf{\texttt{GlobalMMLU}} dataset for the \texttt{ru} language.}\label{fig:globalmmlu-base-ru}\end{figure}
\begin{figure}[h!]\centering\resizebox{\linewidth}{!}{\includegraphics[width=\linewidth]{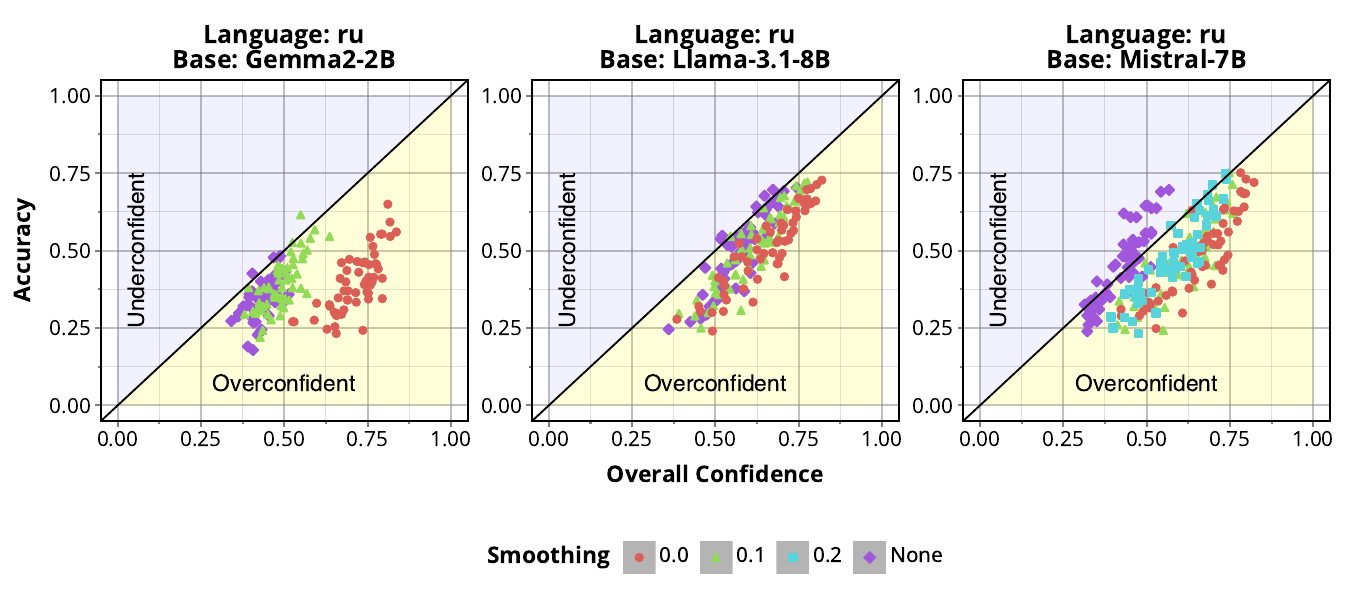}}
\caption{Reliability diagrams for the \textbf{\texttt{GlobalMMLU}} dataset for the \texttt{ru} language after instruction-tuning on the \textbf{\texttt{Tulu3Mixture}} dataset.}\label{fig:globalmmlu-Tulu3Mixture-ru}\end{figure}
\begin{figure}[h!]\centering\resizebox{\linewidth}{!}{\includegraphics[width=\linewidth]{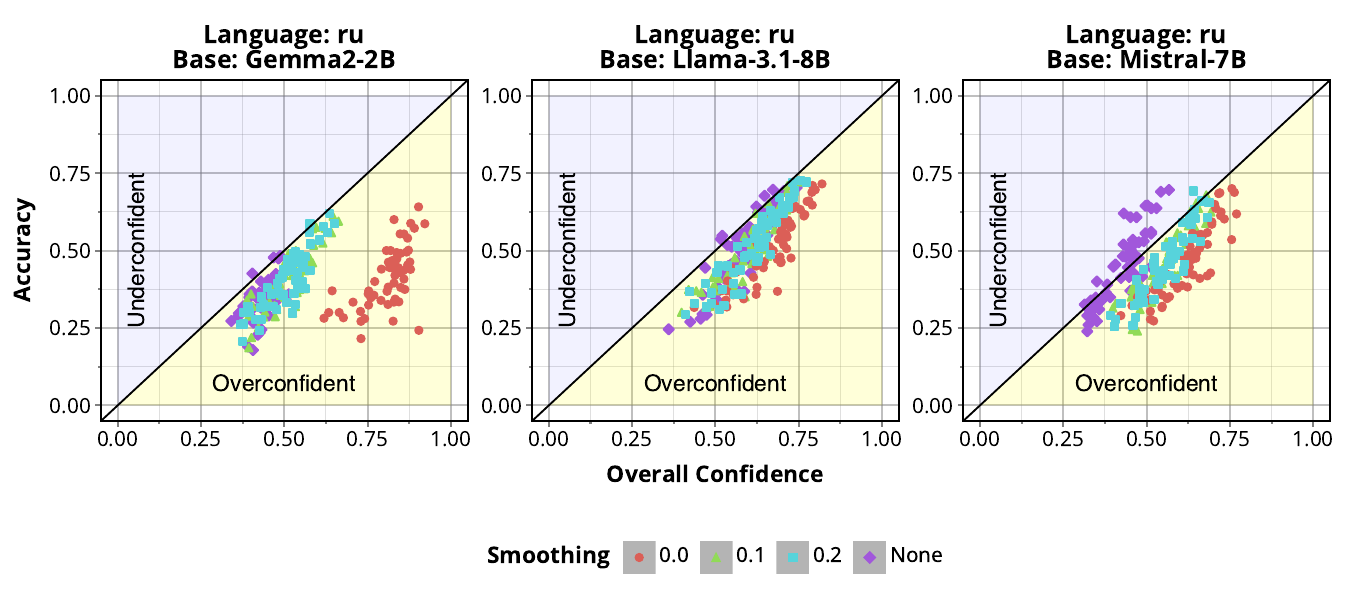}}
\caption{Reliability diagrams for the \textbf{\texttt{GlobalMMLU}} dataset for the \texttt{ru} language after instruction-tuning on the \textbf{\texttt{OpenHermes}} dataset.}\label{fig:globalmmlu-OpenHermes-ru}\end{figure}

\begin{figure}[h!]\centering\resizebox{\linewidth}{!}{\includegraphics[width=\linewidth]{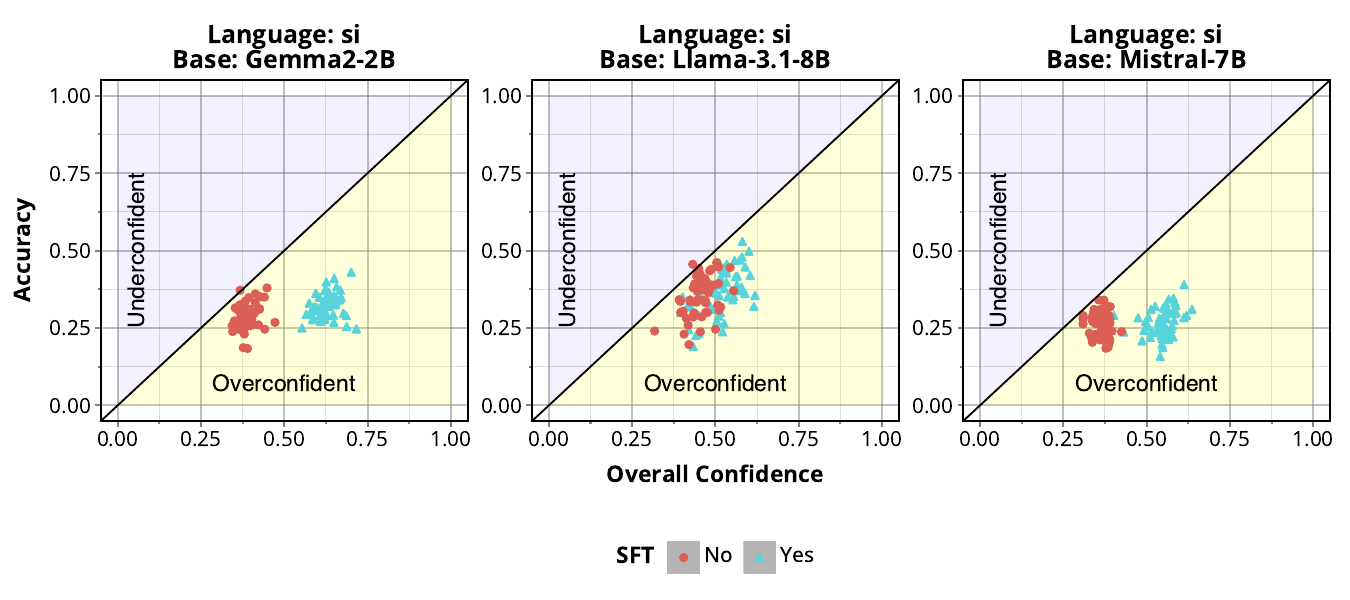}}
\caption{Reliability diagrams for the \textbf{\texttt{GlobalMMLU}} dataset for the \texttt{si} language.}\label{fig:globalmmlu-base-si}\end{figure}
\begin{figure}[h!]\centering\resizebox{\linewidth}{!}{\includegraphics[width=\linewidth]{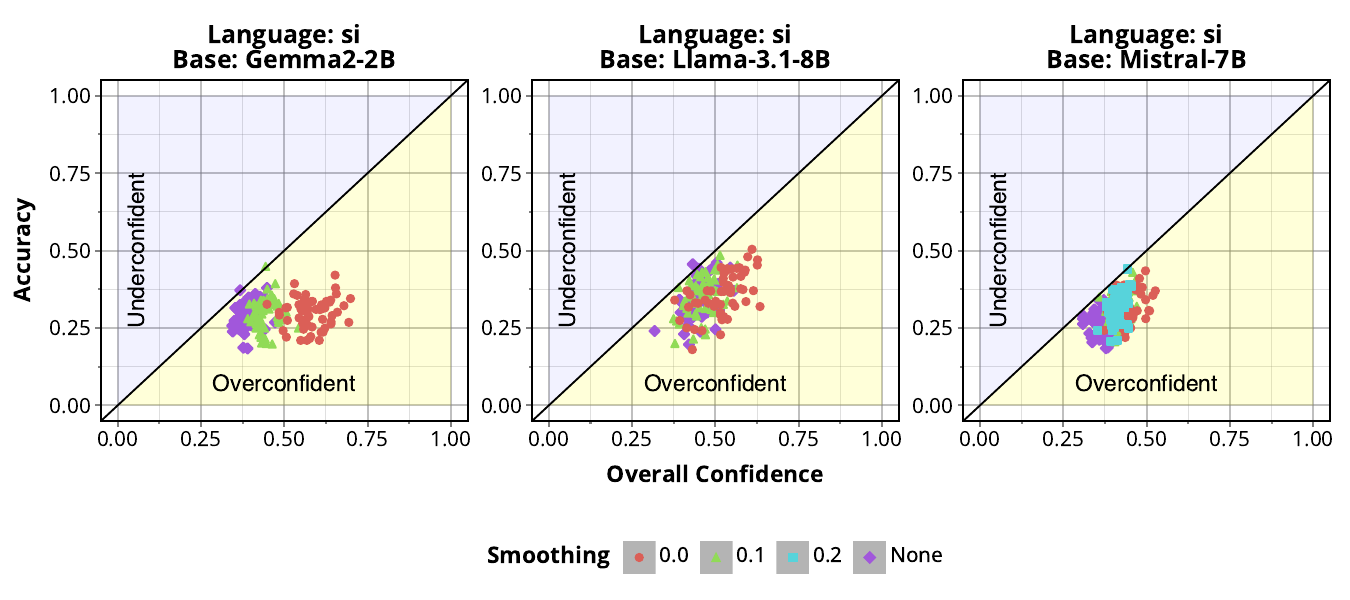}}
\caption{Reliability diagrams for the \textbf{\texttt{GlobalMMLU}} dataset for the \texttt{si} language after instruction-tuning on the \textbf{\texttt{Tulu3Mixture}} dataset.}\label{fig:globalmmlu-Tulu3Mixture-si}\end{figure}
\begin{figure}[h!]\centering\resizebox{\linewidth}{!}{\includegraphics[width=\linewidth]{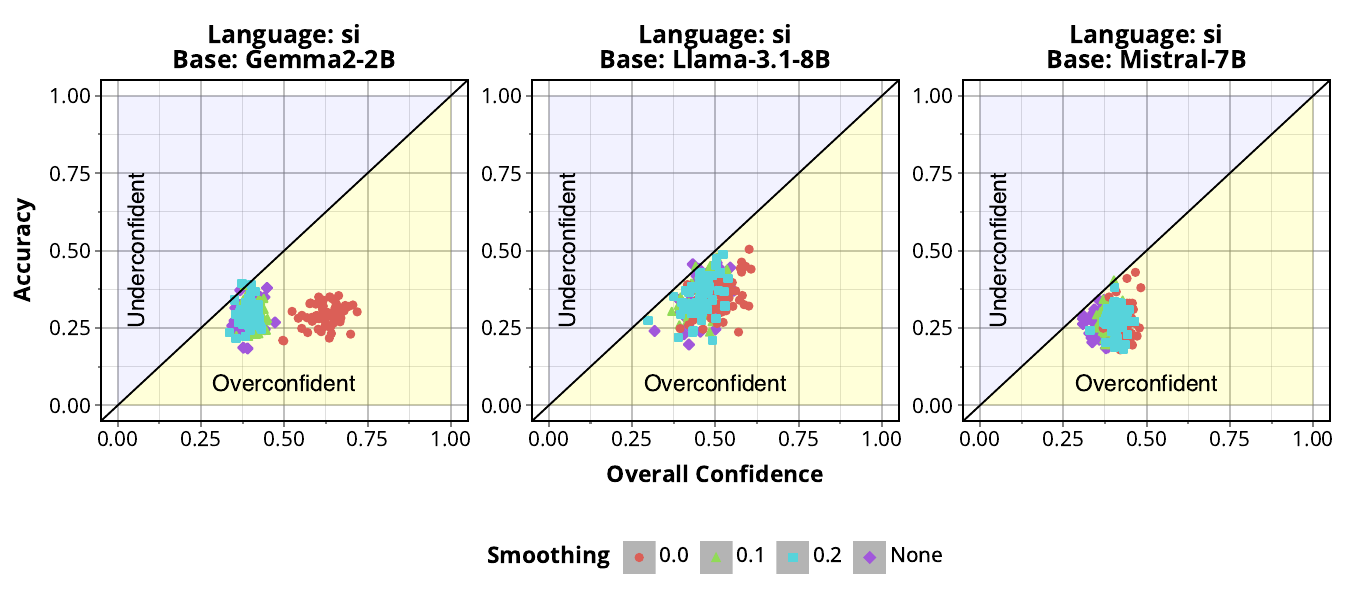}}
\caption{Reliability diagrams for the \textbf{\texttt{GlobalMMLU}} dataset for the \texttt{si} language after instruction-tuning on the \textbf{\texttt{OpenHermes}} dataset.}\label{fig:globalmmlu-OpenHermes-si}\end{figure}

\begin{figure}[h!]\centering\resizebox{\linewidth}{!}{\includegraphics[width=\linewidth]{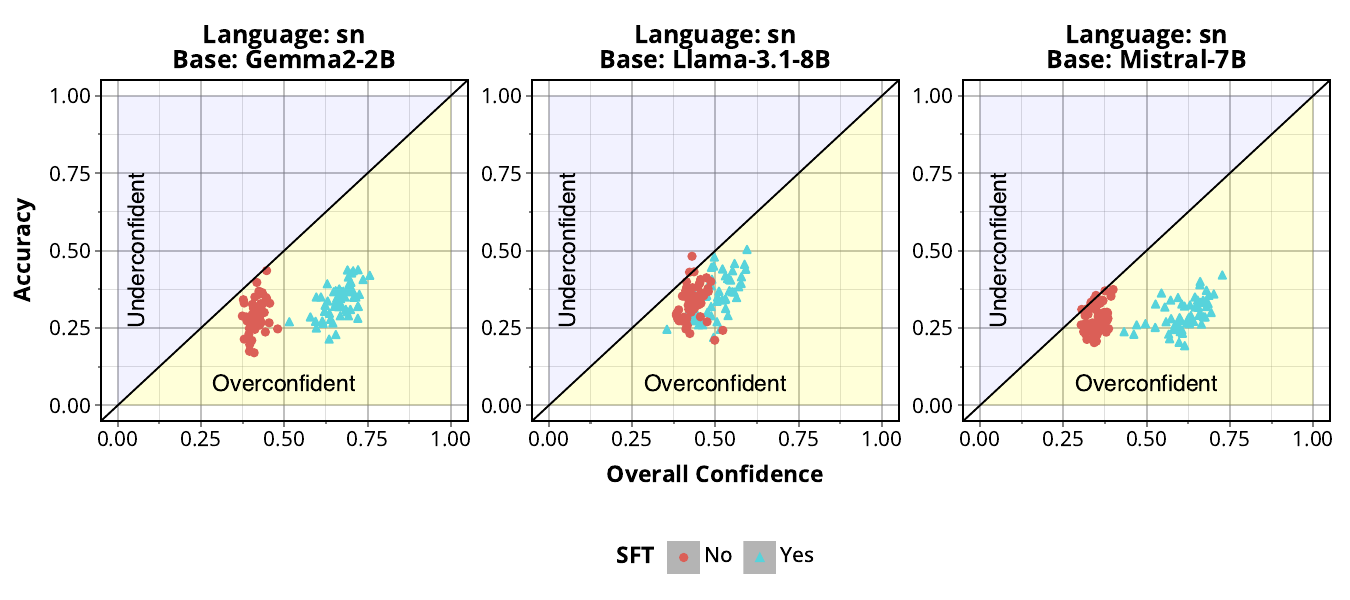}}
\caption{Reliability diagrams for the \textbf{\texttt{GlobalMMLU}} dataset for the \texttt{sn} language.}\label{fig:globalmmlu-base-sn}\end{figure}
\begin{figure}[h!]\centering\resizebox{\linewidth}{!}{\includegraphics[width=\linewidth]{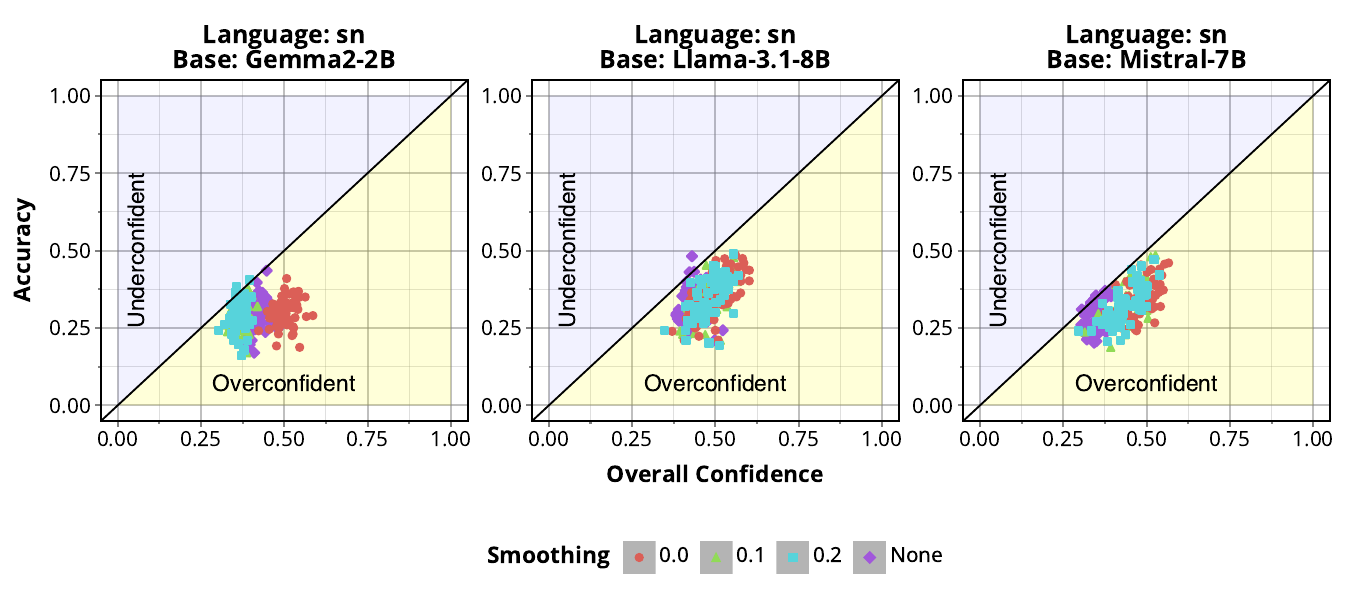}}
\caption{Reliability diagrams for the \textbf{\texttt{GlobalMMLU}} dataset for the \texttt{sn} language after instruction-tuning on the \textbf{\texttt{Tulu3Mixture}} dataset.}\label{fig:globalmmlu-Tulu3Mixture-sn}\end{figure}
\begin{figure}[h!]\centering\resizebox{\linewidth}{!}{\includegraphics[width=\linewidth]{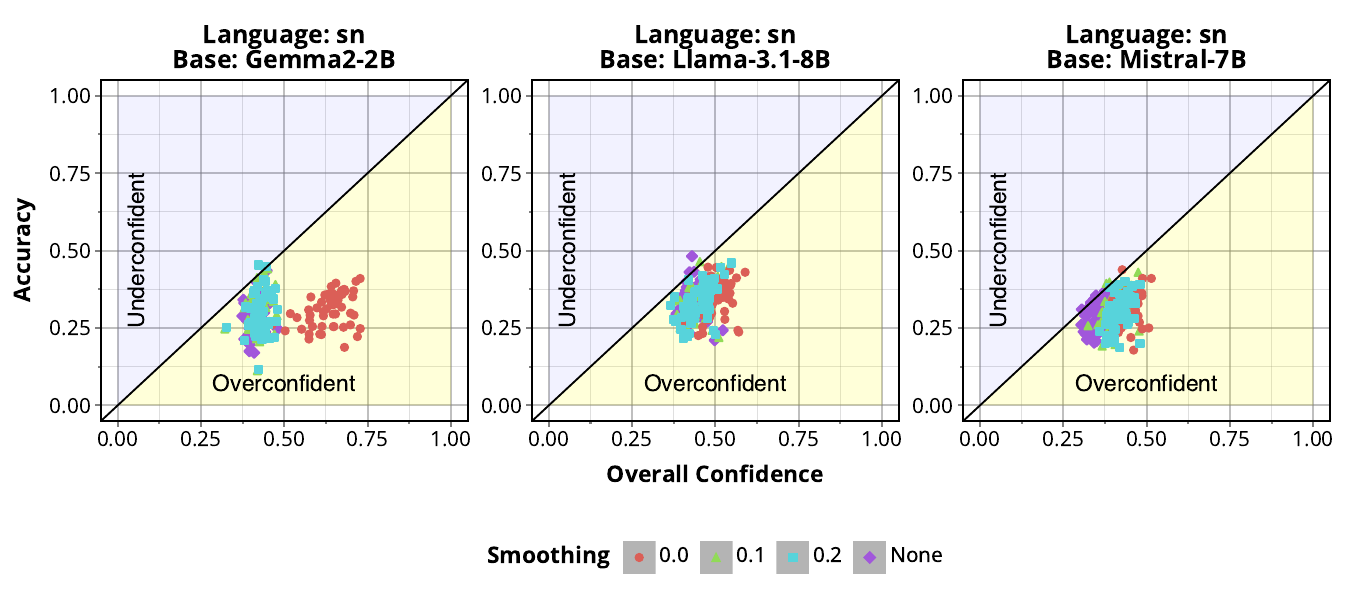}}
\caption{Reliability diagrams for the \textbf{\texttt{GlobalMMLU}} dataset for the \texttt{sn} language after instruction-tuning on the \textbf{\texttt{OpenHermes}} dataset.}\label{fig:globalmmlu-OpenHermes-sn}\end{figure}

\clearpage
\begin{figure}[h!]\centering\resizebox{\linewidth}{!}{\includegraphics[width=\linewidth]{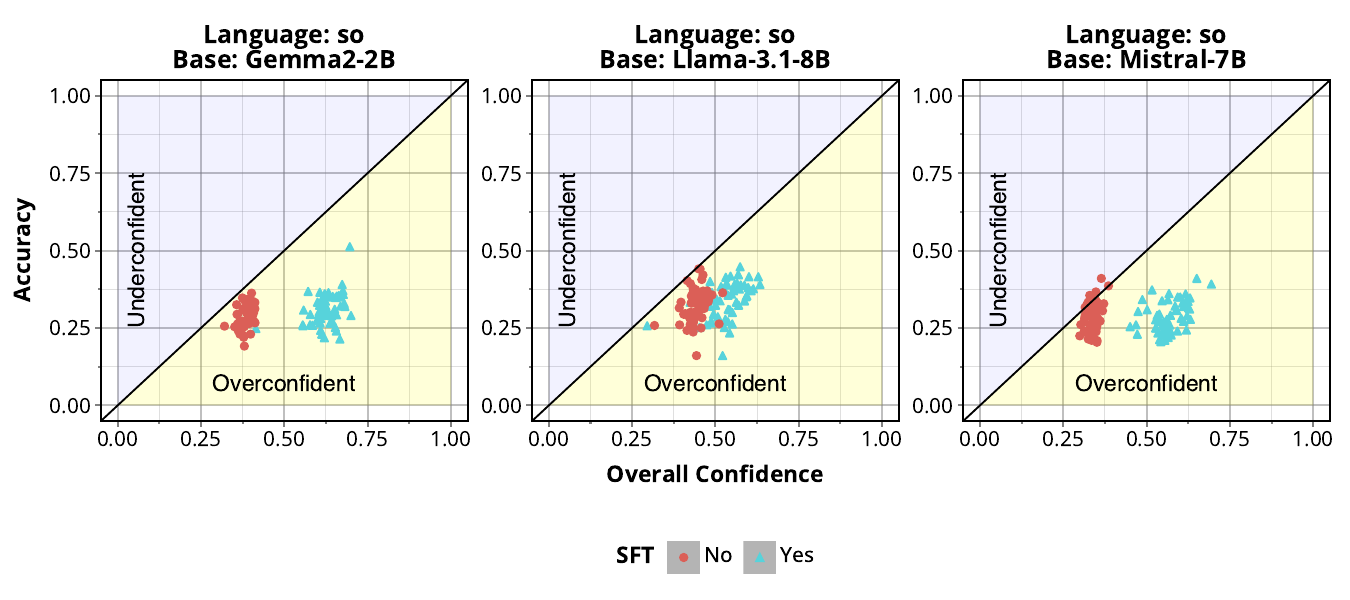}}
\caption{Reliability diagrams for the \textbf{\texttt{GlobalMMLU}} dataset for the \texttt{so} language.}\label{fig:globalmmlu-base-so}\end{figure}
\begin{figure}[h!]\centering\resizebox{\linewidth}{!}{\includegraphics[width=\linewidth]{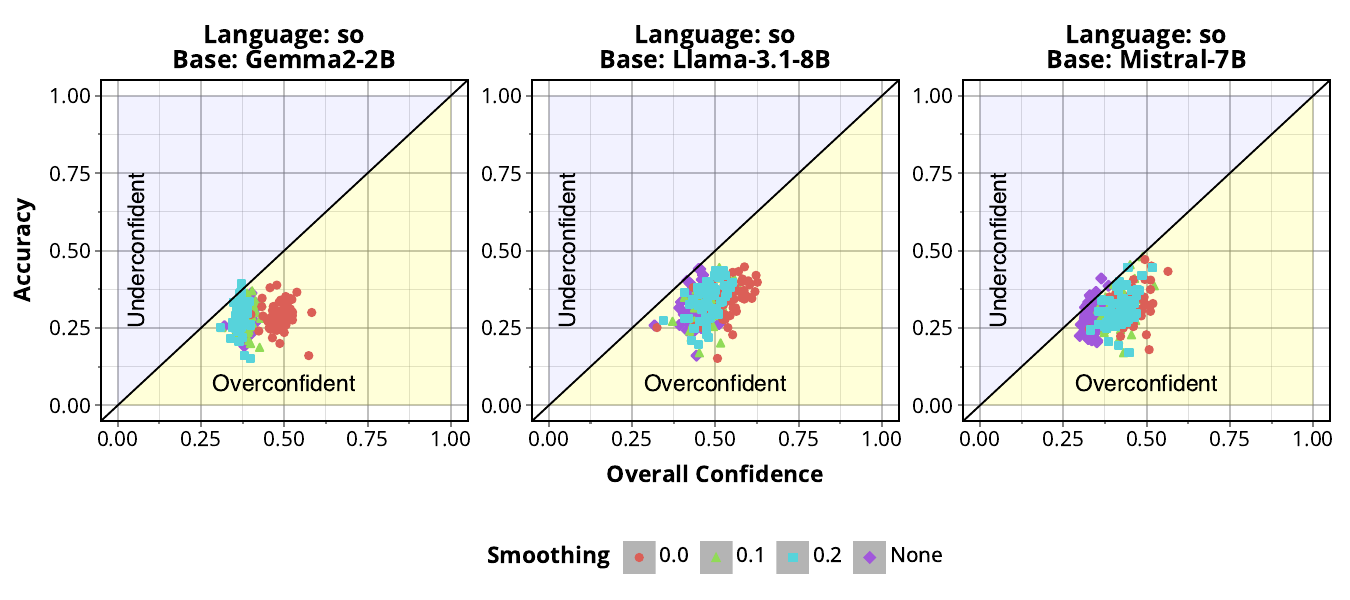}}
\caption{Reliability diagrams for the \textbf{\texttt{GlobalMMLU}} dataset for the \texttt{so} language after instruction-tuning on the \textbf{\texttt{Tulu3Mixture}} dataset.}\label{fig:globalmmlu-Tulu3Mixture-so}\end{figure}
\begin{figure}[h!]\centering\resizebox{\linewidth}{!}{\includegraphics[width=\linewidth]{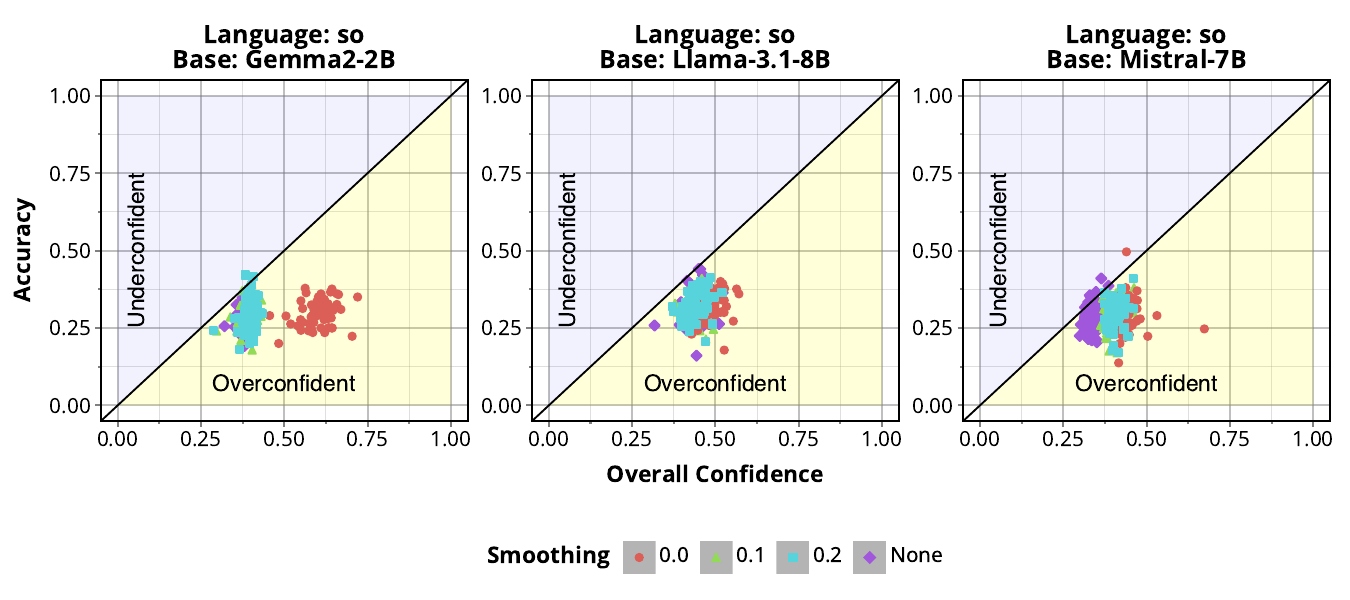}}
\caption{Reliability diagrams for the \textbf{\texttt{GlobalMMLU}} dataset for the \texttt{so} language after instruction-tuning on the \textbf{\texttt{OpenHermes}} dataset.}\label{fig:globalmmlu-OpenHermes-so}\end{figure}

\begin{figure}[h!]\centering\resizebox{\linewidth}{!}{\includegraphics[width=\linewidth]{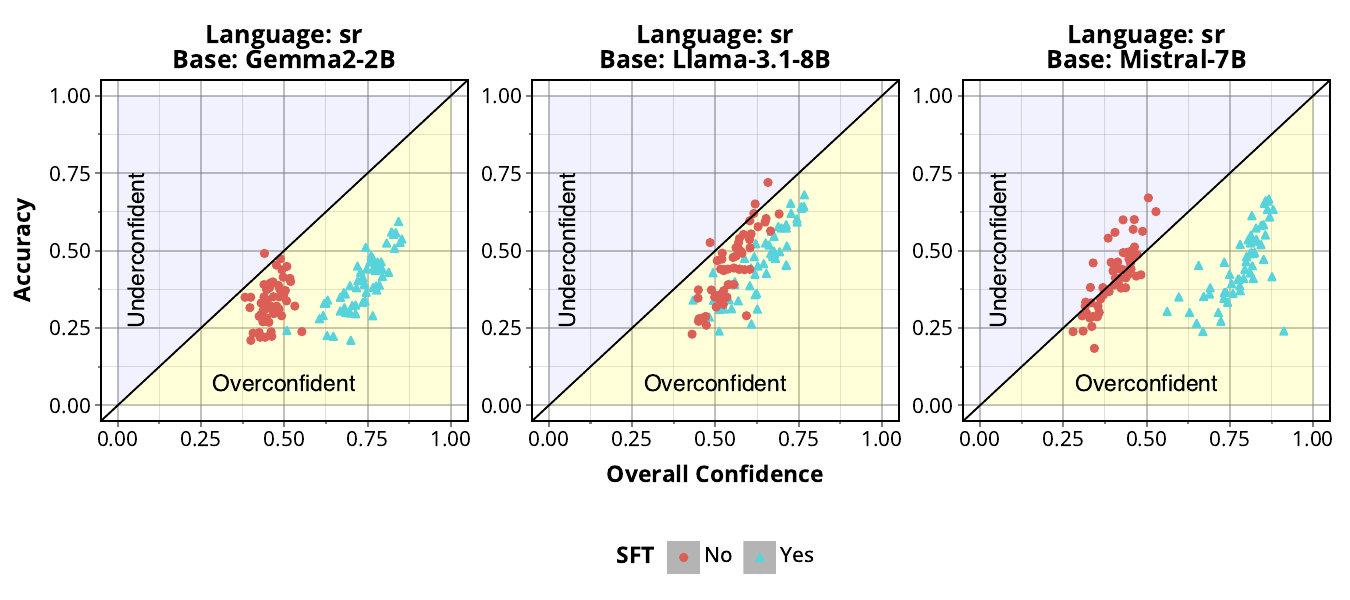}}
\caption{Reliability diagrams for the \textbf{\texttt{GlobalMMLU}} dataset for the \texttt{sr} language.}\label{fig:globalmmlu-base-sr}\end{figure}
\begin{figure}[h!]\centering\resizebox{\linewidth}{!}{\includegraphics[width=\linewidth]{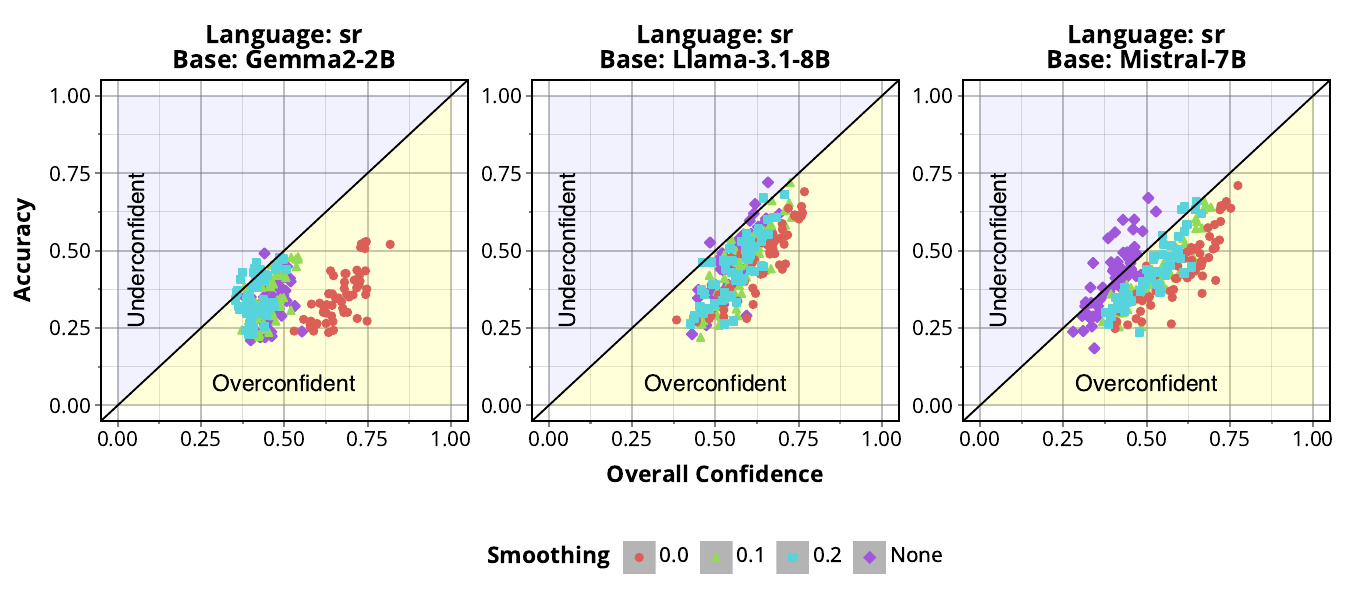}}
\caption{Reliability diagrams for the \textbf{\texttt{GlobalMMLU}} dataset for the \texttt{sr} language after instruction-tuning on the \textbf{\texttt{Tulu3Mixture}} dataset.}\label{fig:globalmmlu-Tulu3Mixture-sr}\end{figure}
\begin{figure}[h!]\centering\resizebox{\linewidth}{!}{\includegraphics[width=\linewidth]{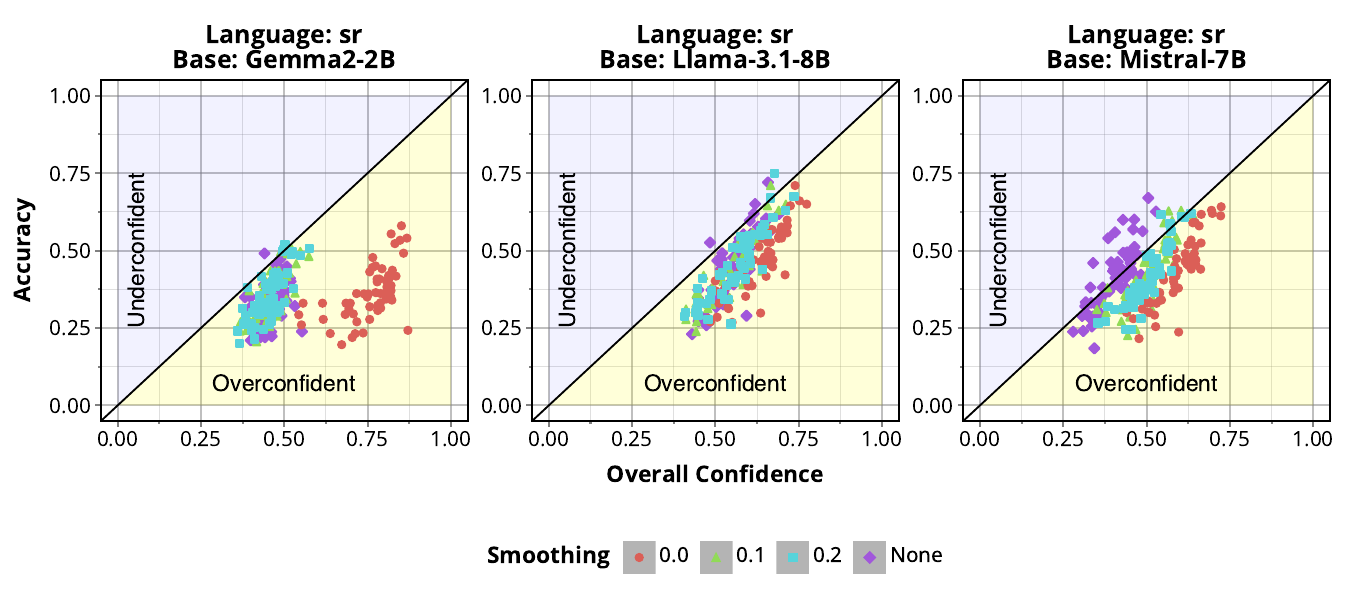}}
\caption{Reliability diagrams for the \textbf{\texttt{GlobalMMLU}} dataset for the \texttt{sr} language after instruction-tuning on the \textbf{\texttt{OpenHermes}} dataset.}\label{fig:globalmmlu-OpenHermes-sr}\end{figure}

\begin{figure}[h!]\centering\resizebox{\linewidth}{!}{\includegraphics[width=\linewidth]{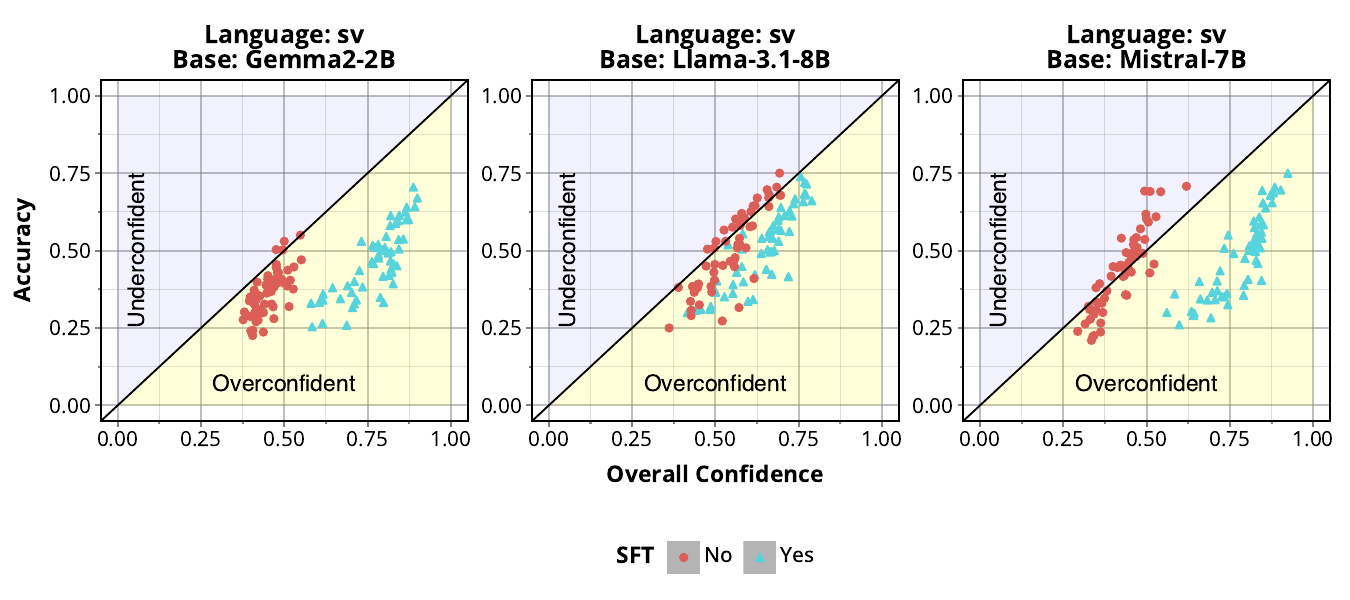}}
\caption{Reliability diagrams for the \textbf{\texttt{GlobalMMLU}} dataset for the \texttt{sv} language.}\label{fig:globalmmlu-base-sv}\end{figure}
\begin{figure}[h!]\centering\resizebox{\linewidth}{!}{\includegraphics[width=\linewidth]{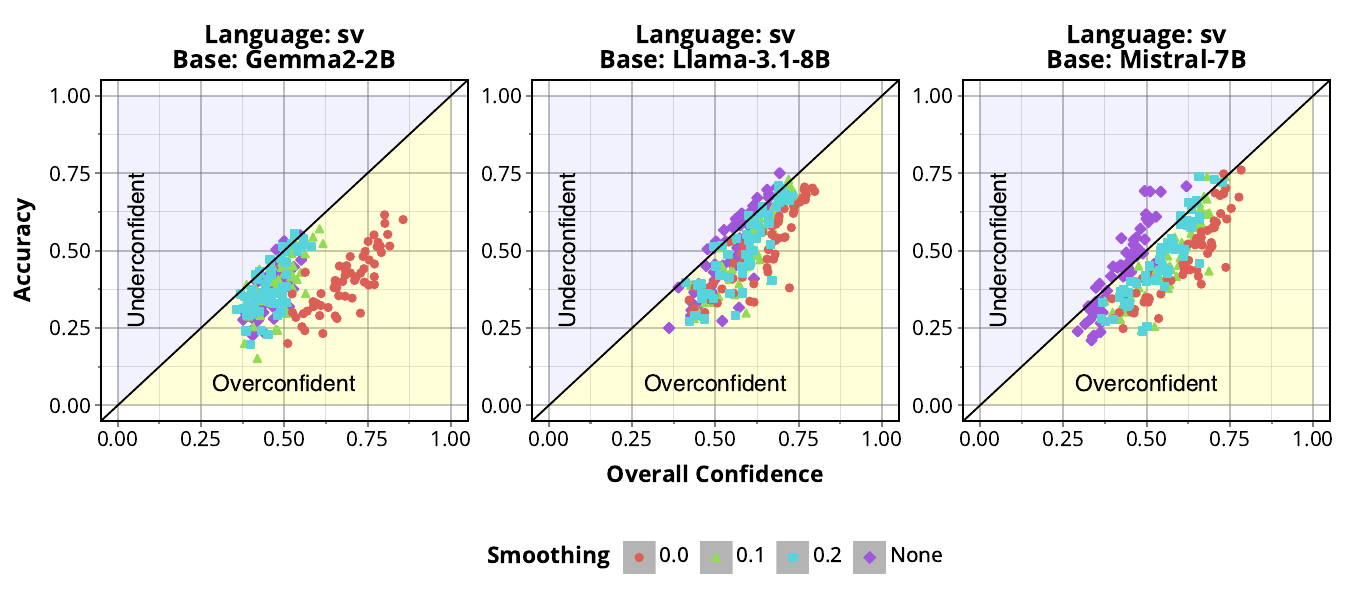}}
\caption{Reliability diagrams for the \textbf{\texttt{GlobalMMLU}} dataset for the \texttt{sv} language after instruction-tuning on the \textbf{\texttt{Tulu3Mixture}} dataset.}\label{fig:globalmmlu-Tulu3Mixture-sv}\end{figure}
\begin{figure}[h!]\centering\resizebox{\linewidth}{!}{\includegraphics[width=\linewidth]{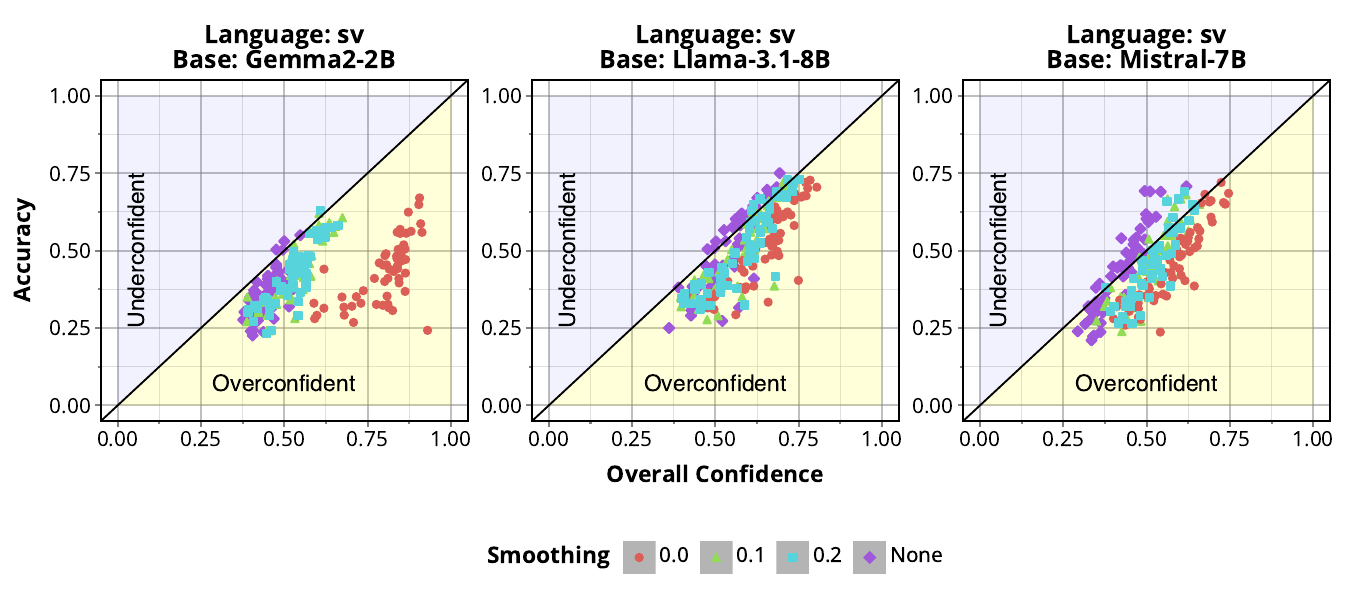}}
\caption{Reliability diagrams for the \textbf{\texttt{GlobalMMLU}} dataset for the \texttt{sv} language after instruction-tuning on the \textbf{\texttt{OpenHermes}} dataset.}\label{fig:globalmmlu-OpenHermes-sv}\end{figure}

\begin{figure}[h!]\centering\resizebox{\linewidth}{!}{\includegraphics[width=\linewidth]{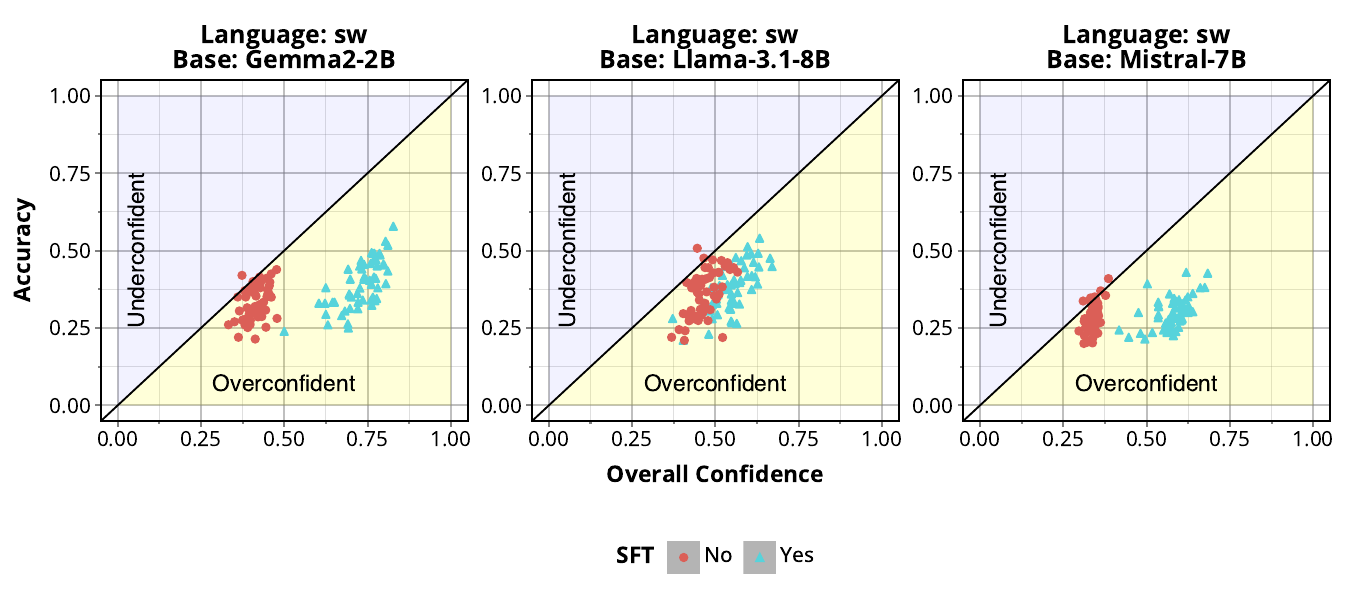}}
\caption{Reliability diagrams for the \textbf{\texttt{GlobalMMLU}} dataset for the \texttt{sw} language.}\label{fig:globalmmlu-base-sw}\end{figure}
\begin{figure}[h!]\centering\resizebox{\linewidth}{!}{\includegraphics[width=\linewidth]{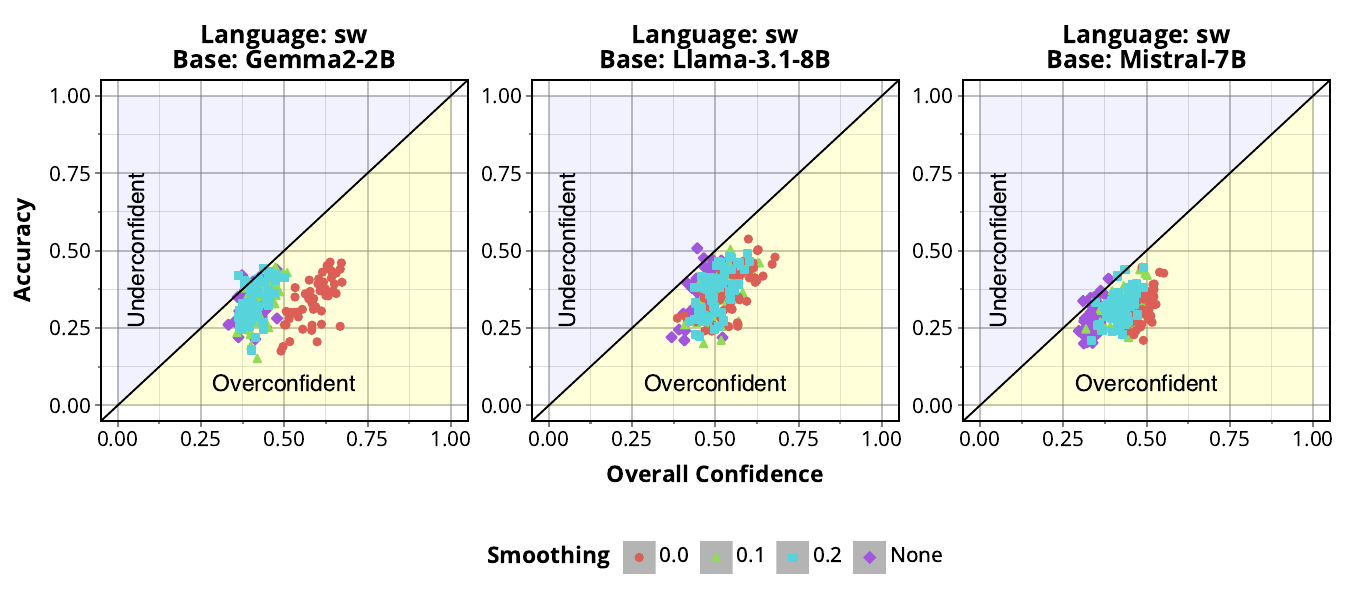}}
\caption{Reliability diagrams for the \textbf{\texttt{GlobalMMLU}} dataset for the \texttt{sw} language after instruction-tuning on the \textbf{\texttt{Tulu3Mixture}} dataset.}\label{fig:globalmmlu-Tulu3Mixture-sw}\end{figure}
\begin{figure}[h!]\centering\resizebox{\linewidth}{!}{\includegraphics[width=\linewidth]{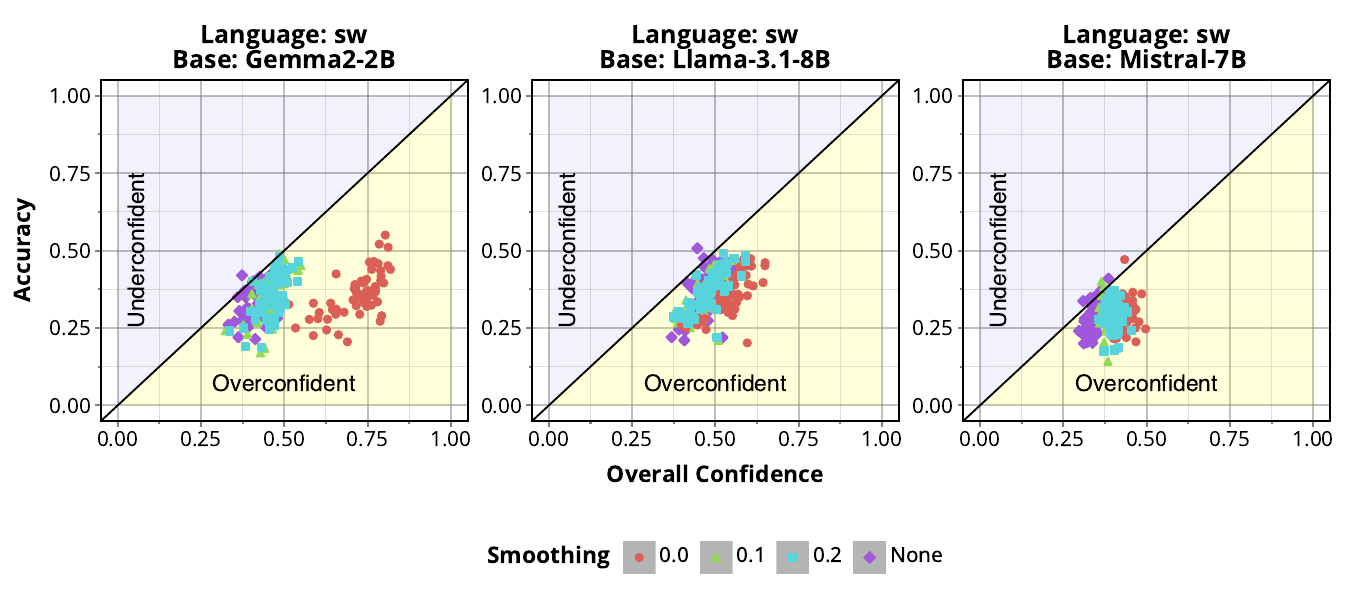}}
\caption{Reliability diagrams for the \textbf{\texttt{GlobalMMLU}} dataset for the \texttt{sw} language after instruction-tuning on the \textbf{\texttt{OpenHermes}} dataset.}\label{fig:globalmmlu-OpenHermes-sw}\end{figure}

\begin{figure}[h!]\centering\resizebox{\linewidth}{!}{\includegraphics[width=\linewidth]{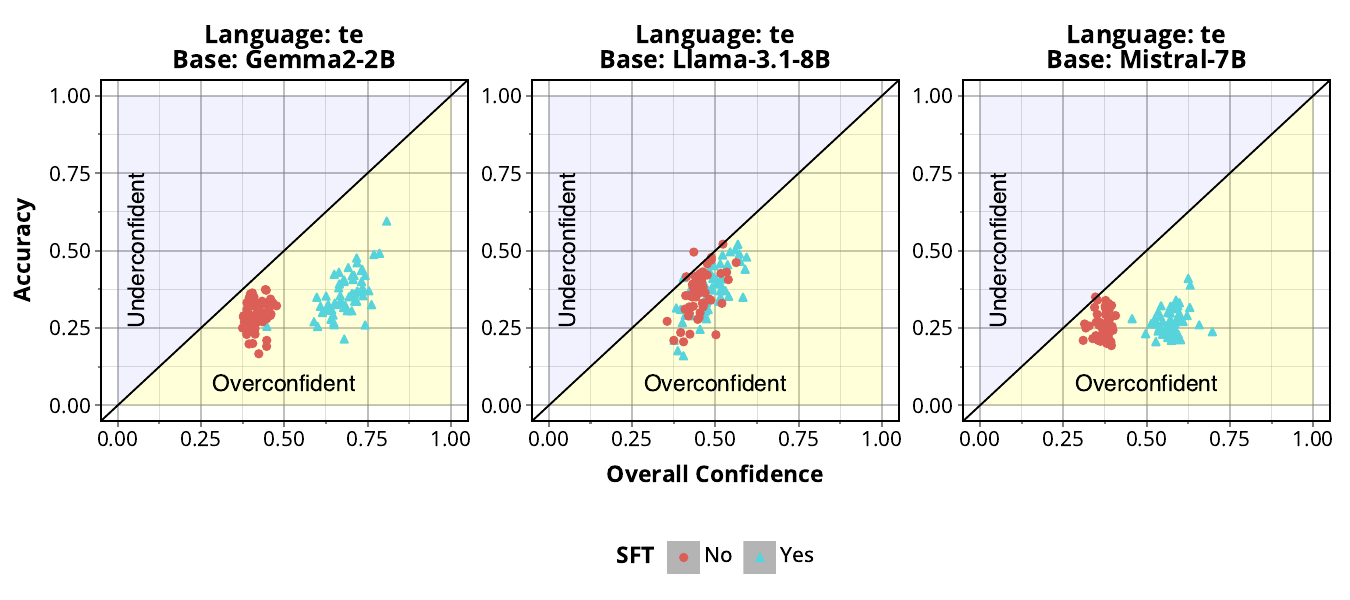}}
\caption{Reliability diagrams for the \textbf{\texttt{GlobalMMLU}} dataset for the \texttt{te} language.}\label{fig:globalmmlu-base-te}\end{figure}
\begin{figure}[h!]\centering\resizebox{\linewidth}{!}{\includegraphics[width=\linewidth]{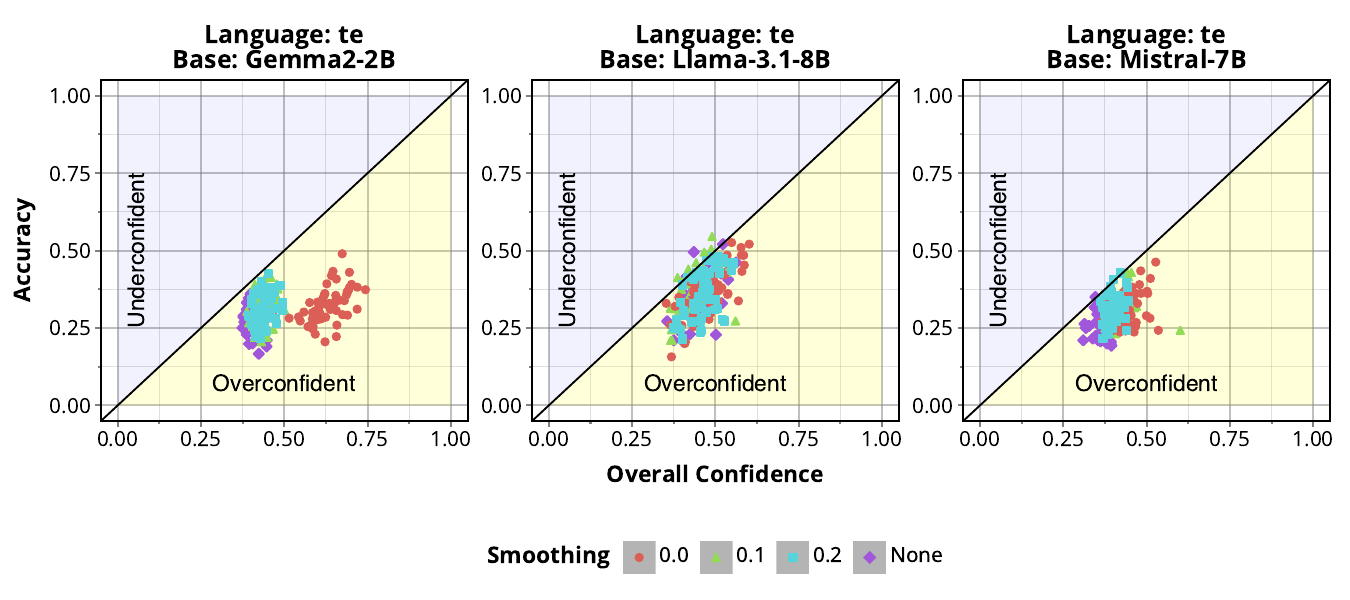}}
\caption{Reliability diagrams for the \textbf{\texttt{GlobalMMLU}} dataset for the \texttt{te} language after instruction-tuning on the \textbf{\texttt{Tulu3Mixture}} dataset.}\label{fig:globalmmlu-Tulu3Mixture-te}\end{figure}
\begin{figure}[h!]\centering\resizebox{\linewidth}{!}{\includegraphics[width=\linewidth]{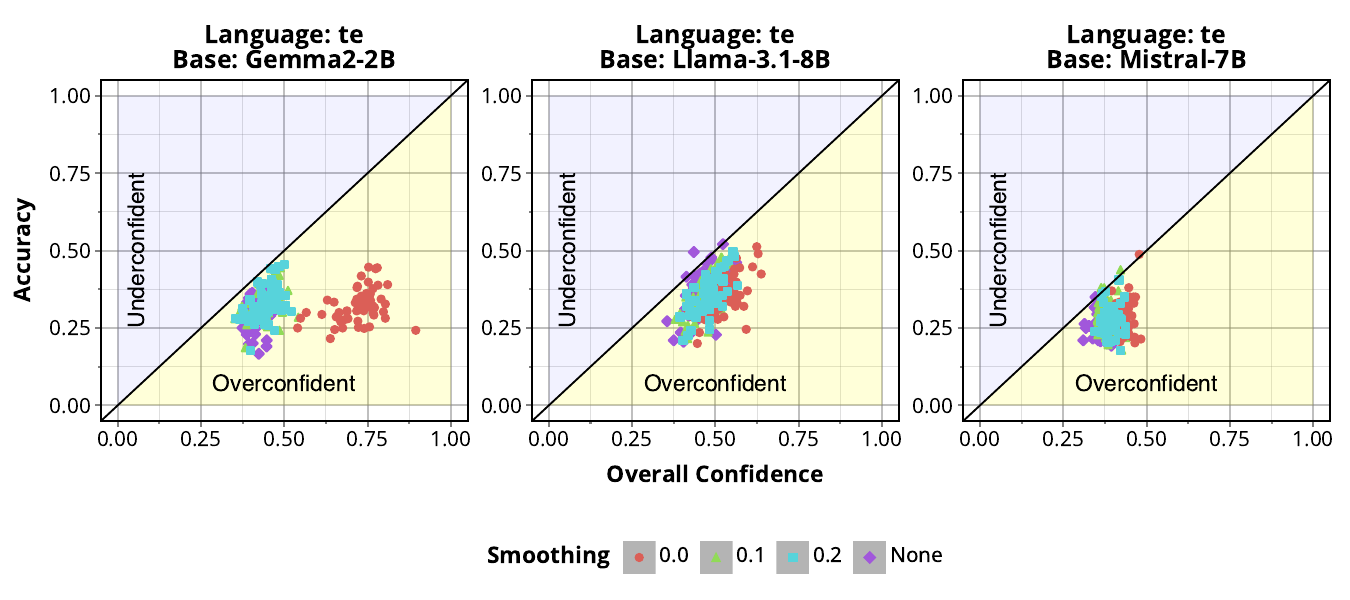}}
\caption{Reliability diagrams for the \textbf{\texttt{GlobalMMLU}} dataset for the \texttt{te} language after instruction-tuning on the \textbf{\texttt{OpenHermes}} dataset.}\label{fig:globalmmlu-OpenHermes-te}\end{figure}

\begin{figure}[h!]\centering\resizebox{\linewidth}{!}{\includegraphics[width=\linewidth]{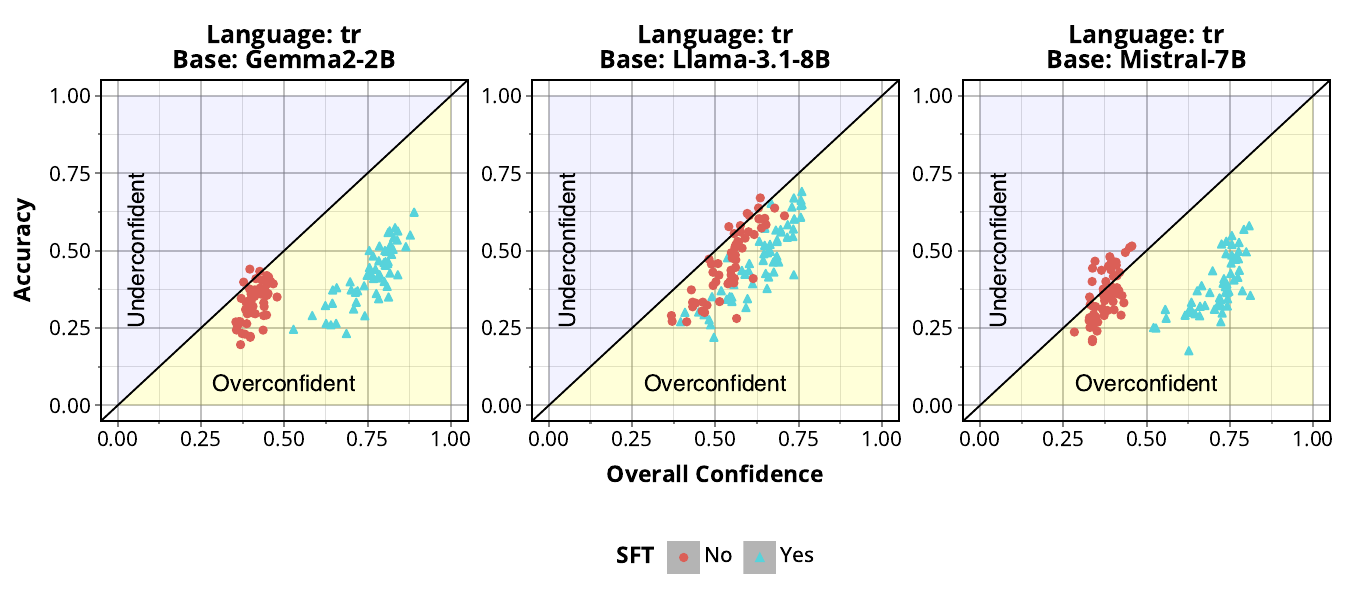}}
\caption{Reliability diagrams for the \textbf{\texttt{GlobalMMLU}} dataset for the \texttt{tr} language.}\label{fig:globalmmlu-base-tr}\end{figure}
\begin{figure}[h!]\centering\resizebox{\linewidth}{!}{\includegraphics[width=\linewidth]{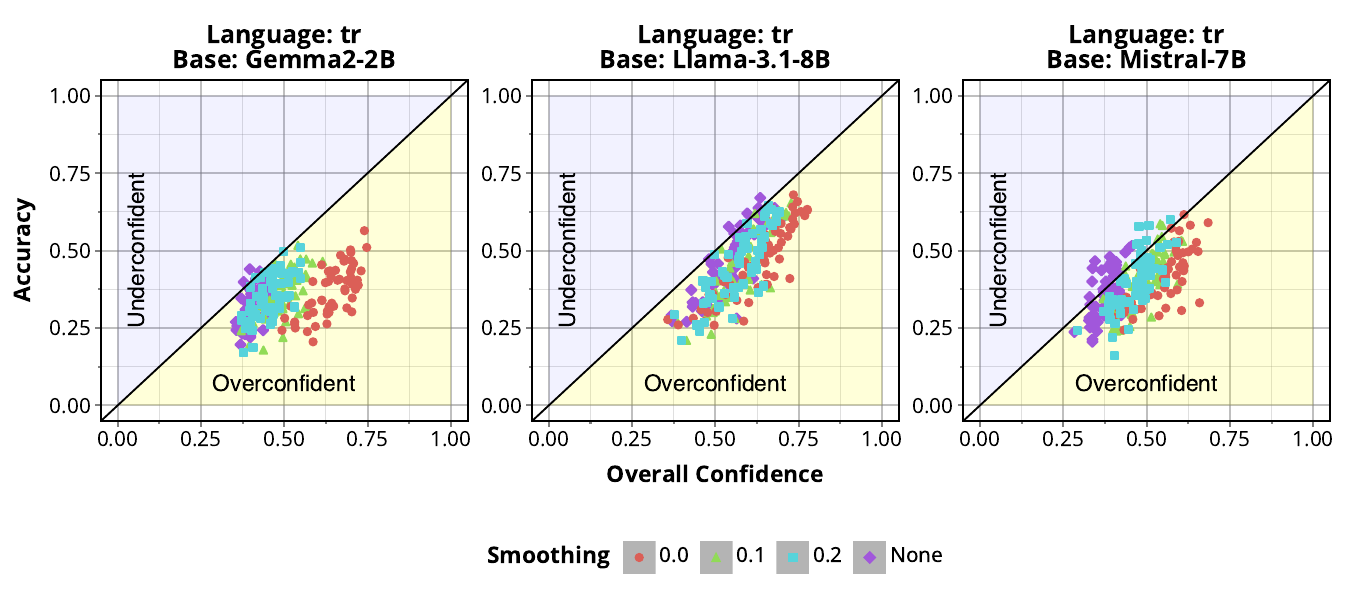}}
\caption{Reliability diagrams for the \textbf{\texttt{GlobalMMLU}} dataset for the \texttt{tr} language after instruction-tuning on the \textbf{\texttt{Tulu3Mixture}} dataset.}\label{fig:globalmmlu-Tulu3Mixture-tr}\end{figure}
\begin{figure}[h!]\centering\resizebox{\linewidth}{!}{\includegraphics[width=\linewidth]{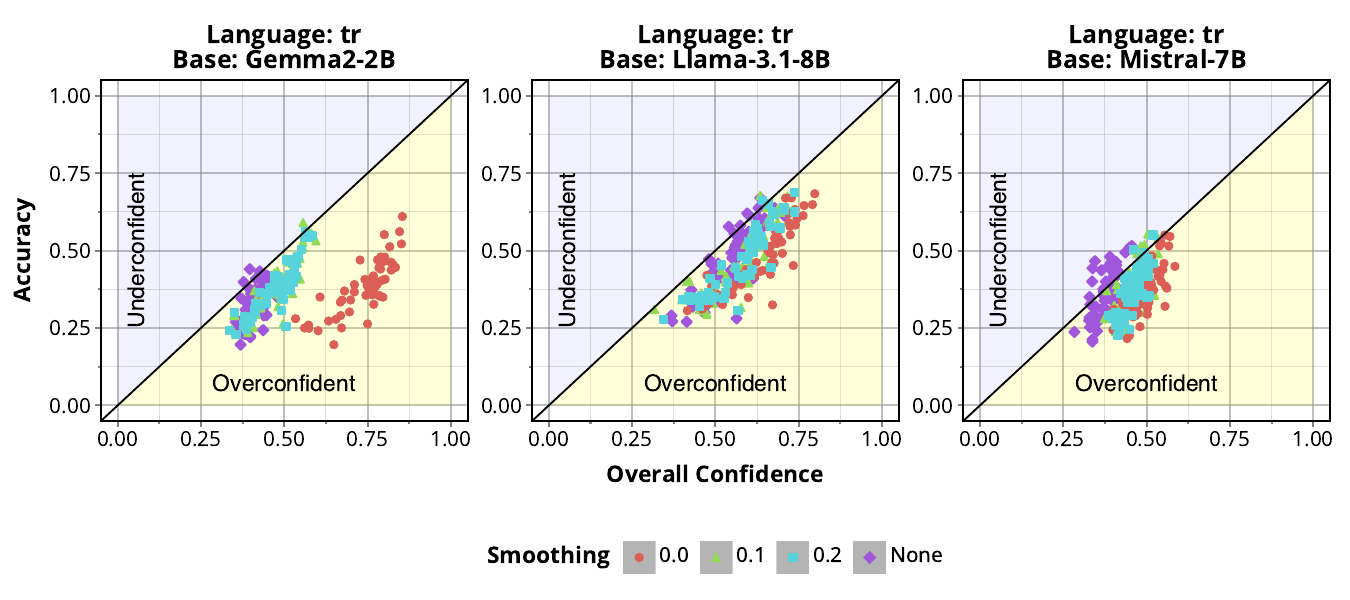}}
\caption{Reliability diagrams for the \textbf{\texttt{GlobalMMLU}} dataset for the \texttt{tr} language after instruction-tuning on the \textbf{\texttt{OpenHermes}} dataset.}\label{fig:globalmmlu-OpenHermes-tr}\end{figure}

\begin{figure}[h!]\centering\resizebox{\linewidth}{!}{\includegraphics[width=\linewidth]{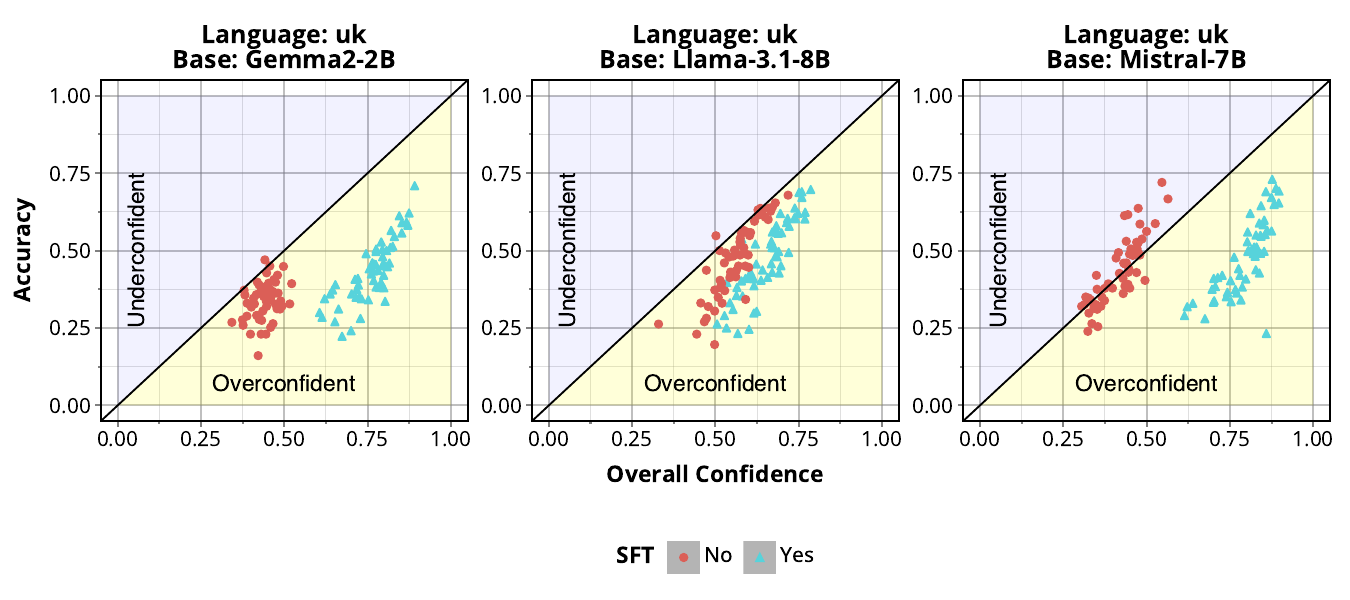}}
\caption{Reliability diagrams for the \textbf{\texttt{GlobalMMLU}} dataset for the \texttt{uk} language.}\label{fig:globalmmlu-base-uk}\end{figure}
\begin{figure}[h!]\centering\resizebox{\linewidth}{!}{\includegraphics[width=\linewidth]{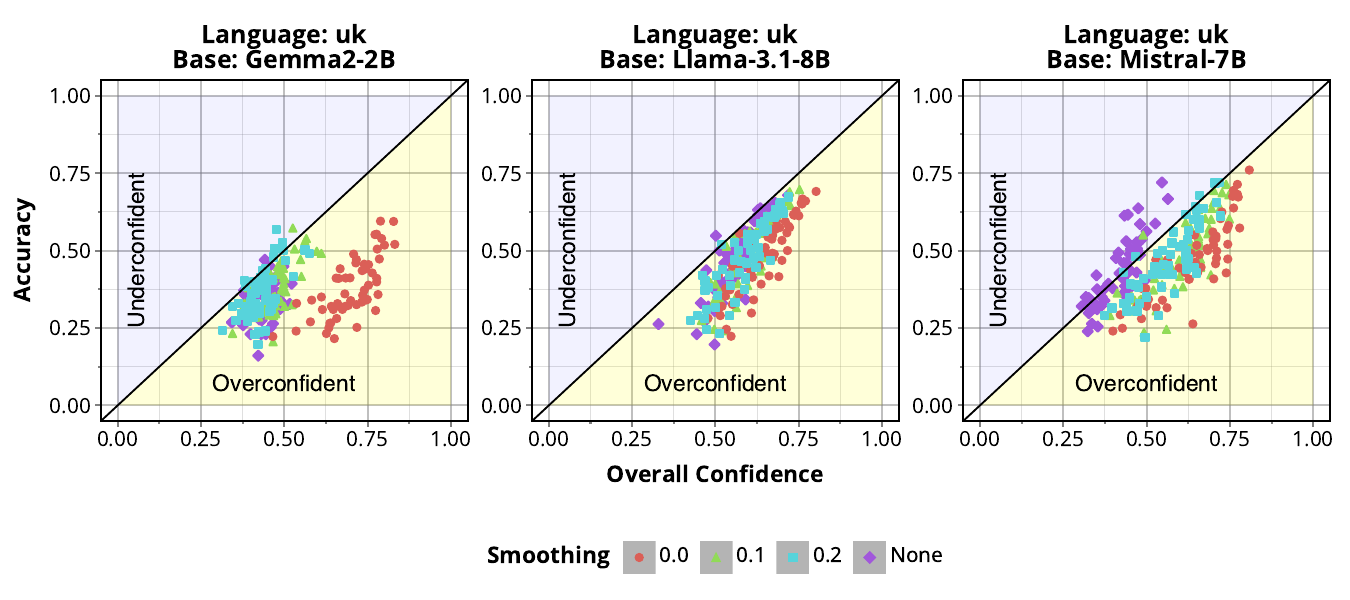}}
\caption{Reliability diagrams for the \textbf{\texttt{GlobalMMLU}} dataset for the \texttt{uk} language after instruction-tuning on the \textbf{\texttt{Tulu3Mixture}} dataset.}\label{fig:globalmmlu-Tulu3Mixture-uk}\end{figure}
\begin{figure}[h!]\centering\resizebox{\linewidth}{!}{\includegraphics[width=\linewidth]{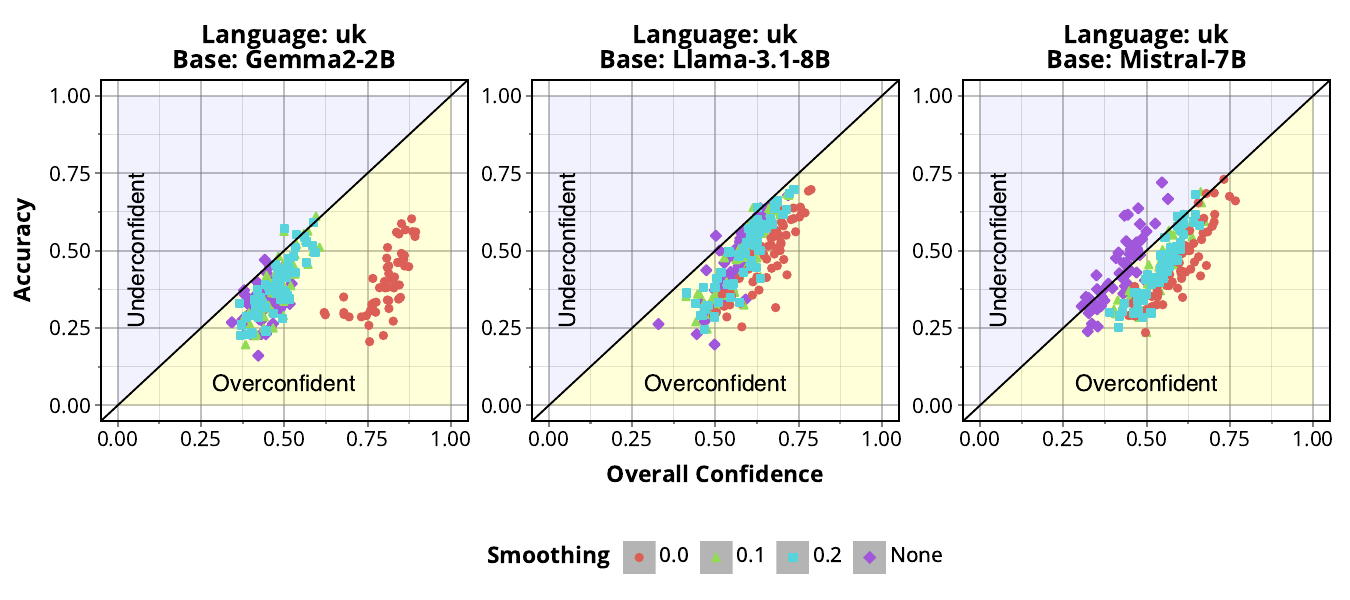}}
\caption{Reliability diagrams for the \textbf{\texttt{GlobalMMLU}} dataset for the \texttt{uk} language after instruction-tuning on the \textbf{\texttt{OpenHermes}} dataset.}\label{fig:globalmmlu-OpenHermes-uk}\end{figure}

\begin{figure}[h!]\centering\resizebox{\linewidth}{!}{\includegraphics[width=\linewidth]{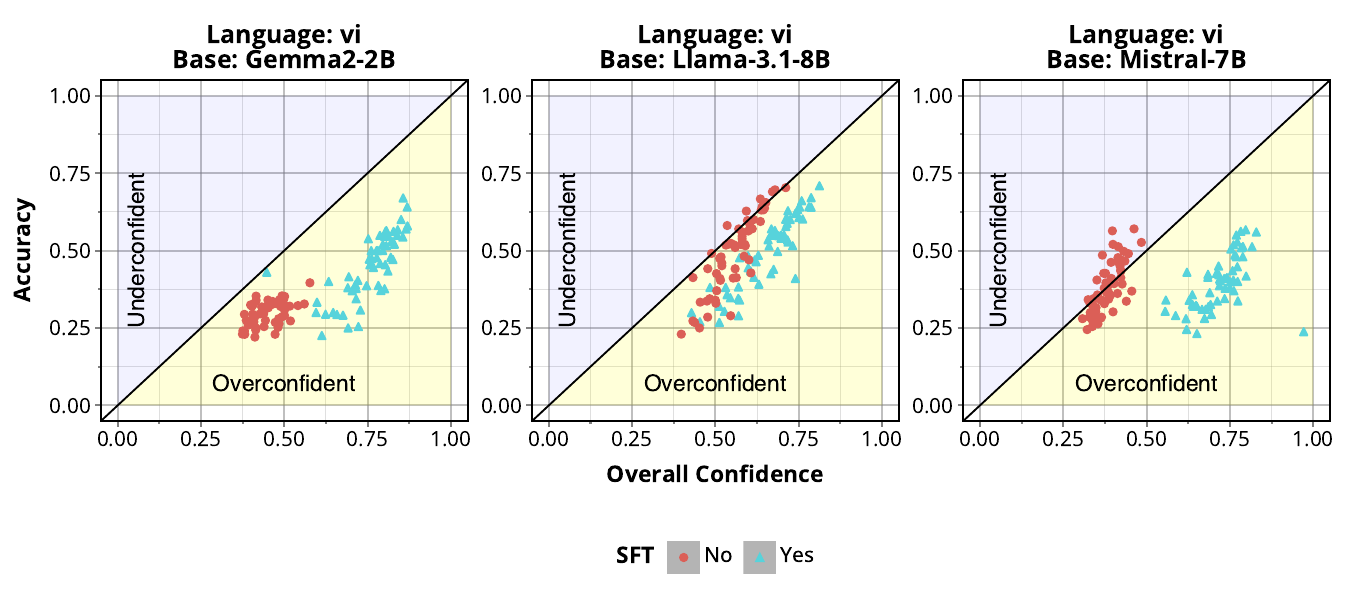}}
\caption{Reliability diagrams for the \textbf{\texttt{GlobalMMLU}} dataset for the \texttt{vi} language.}\label{fig:globalmmlu-base-vi}\end{figure}
\begin{figure}[h!]\centering\resizebox{\linewidth}{!}{\includegraphics[width=\linewidth]{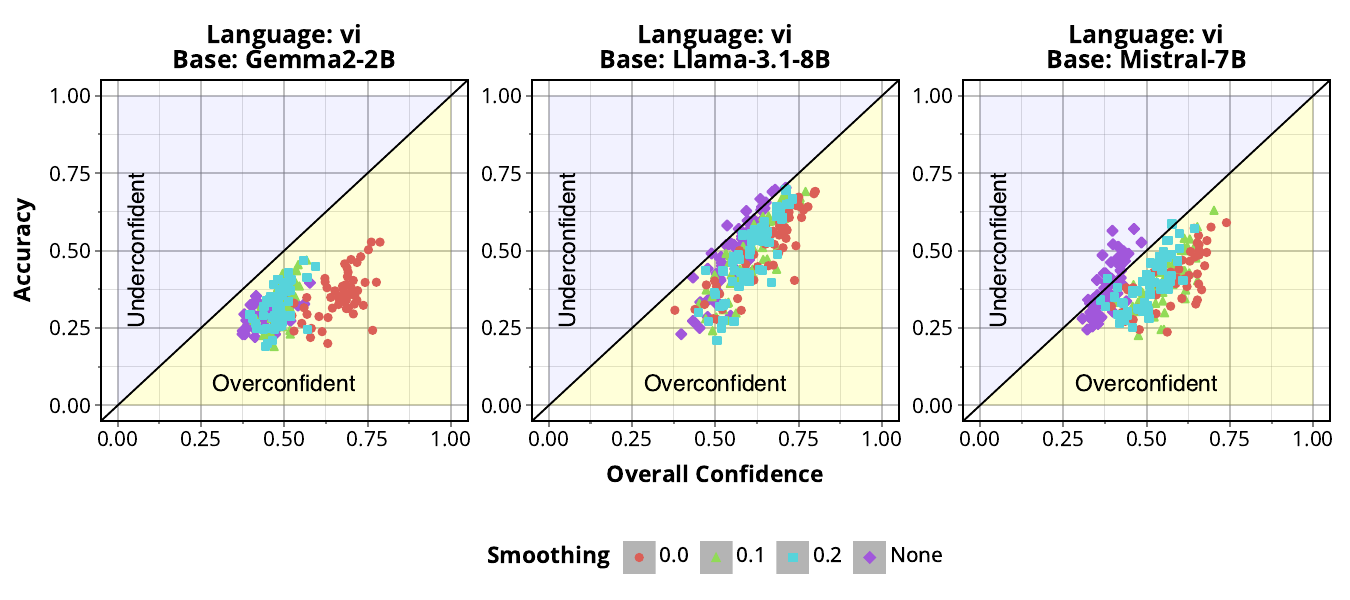}}
\caption{Reliability diagrams for the \textbf{\texttt{GlobalMMLU}} dataset for the \texttt{vi} language after instruction-tuning on the \textbf{\texttt{Tulu3Mixture}} dataset.}\label{fig:globalmmlu-Tulu3Mixture-vi}\end{figure}
\begin{figure}[h!]\centering\resizebox{\linewidth}{!}{\includegraphics[width=\linewidth]{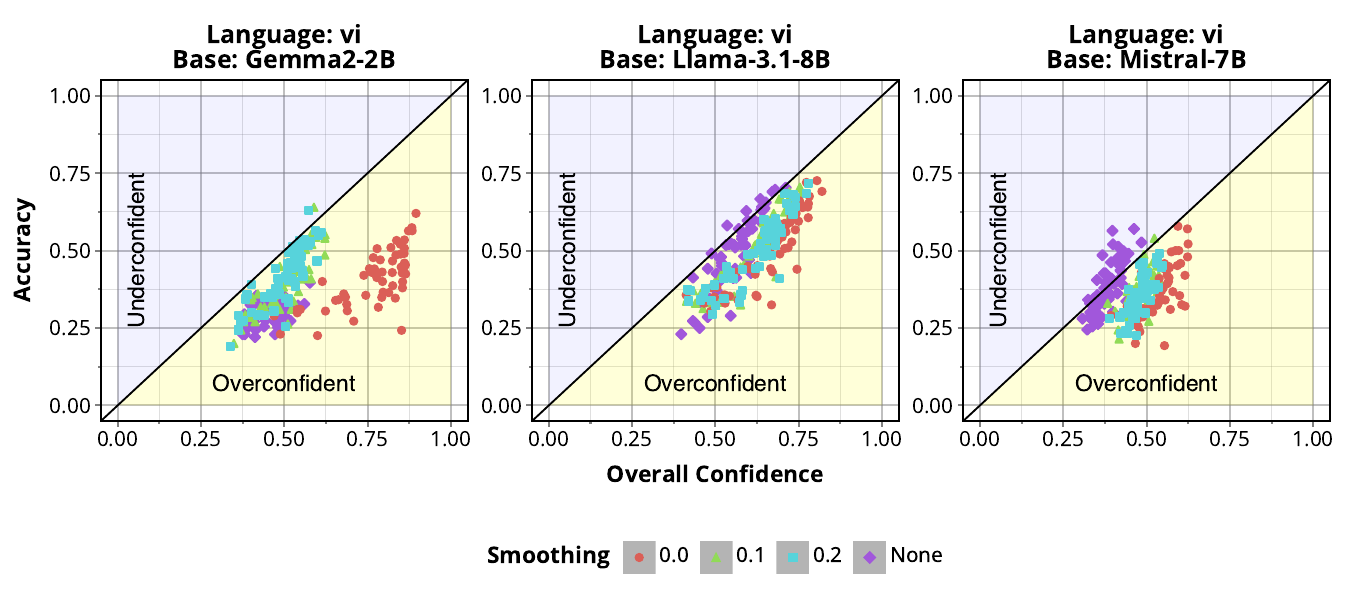}}
\caption{Reliability diagrams for the \textbf{\texttt{GlobalMMLU}} dataset for the \texttt{vi} language after instruction-tuning on the \textbf{\texttt{OpenHermes}} dataset.}\label{fig:globalmmlu-OpenHermes-vi}\end{figure}

\begin{figure}[h!]\centering\resizebox{\linewidth}{!}{\includegraphics[width=\linewidth]{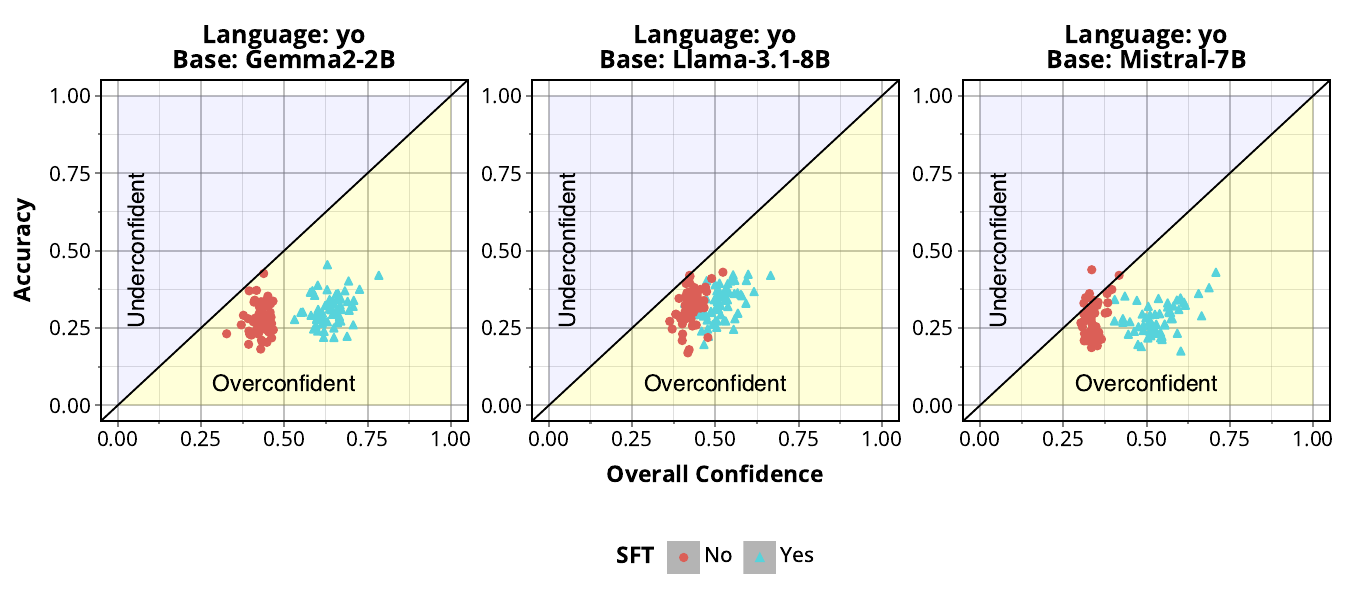}}
\caption{Reliability diagrams for the \textbf{\texttt{GlobalMMLU}} dataset for the \texttt{yo} language.}\label{fig:globalmmlu-base-yo}\end{figure}
\begin{figure}[h!]\centering\resizebox{\linewidth}{!}{\includegraphics[width=\linewidth]{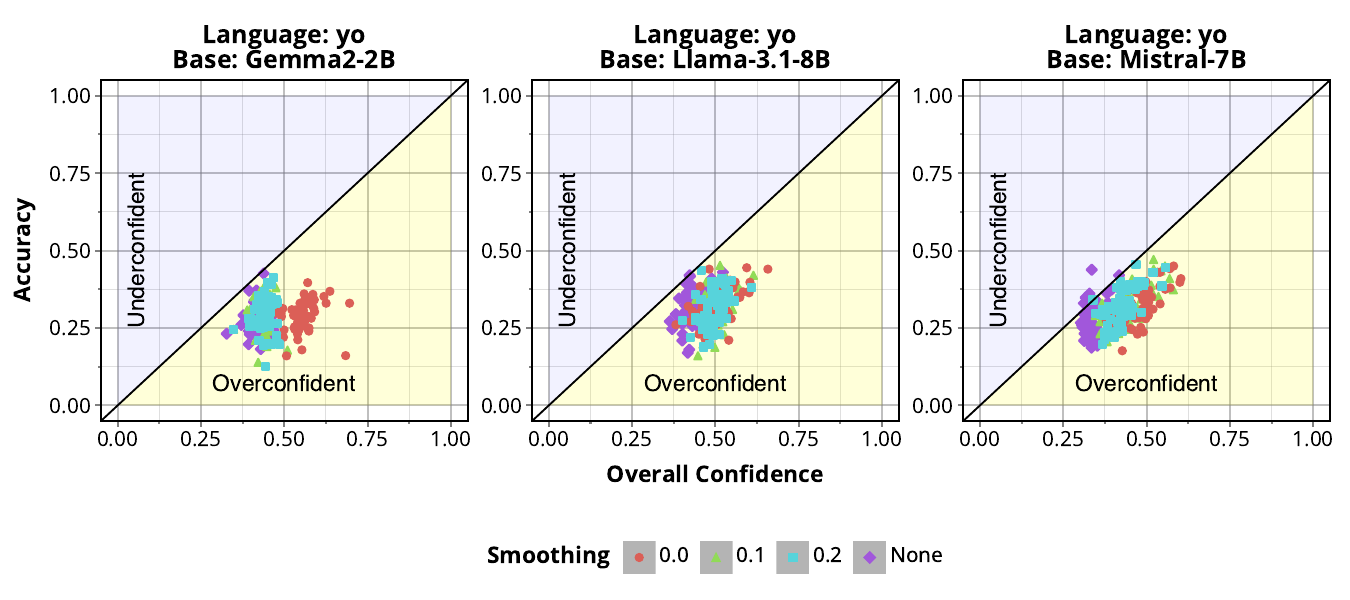}}
\caption{Reliability diagrams for the \textbf{\texttt{GlobalMMLU}} dataset for the \texttt{yo} language after instruction-tuning on the \textbf{\texttt{Tulu3Mixture}} dataset.}\label{fig:globalmmlu-Tulu3Mixture-yo}\end{figure}
\begin{figure}[h!]\centering\resizebox{\linewidth}{!}{\includegraphics[width=\linewidth]{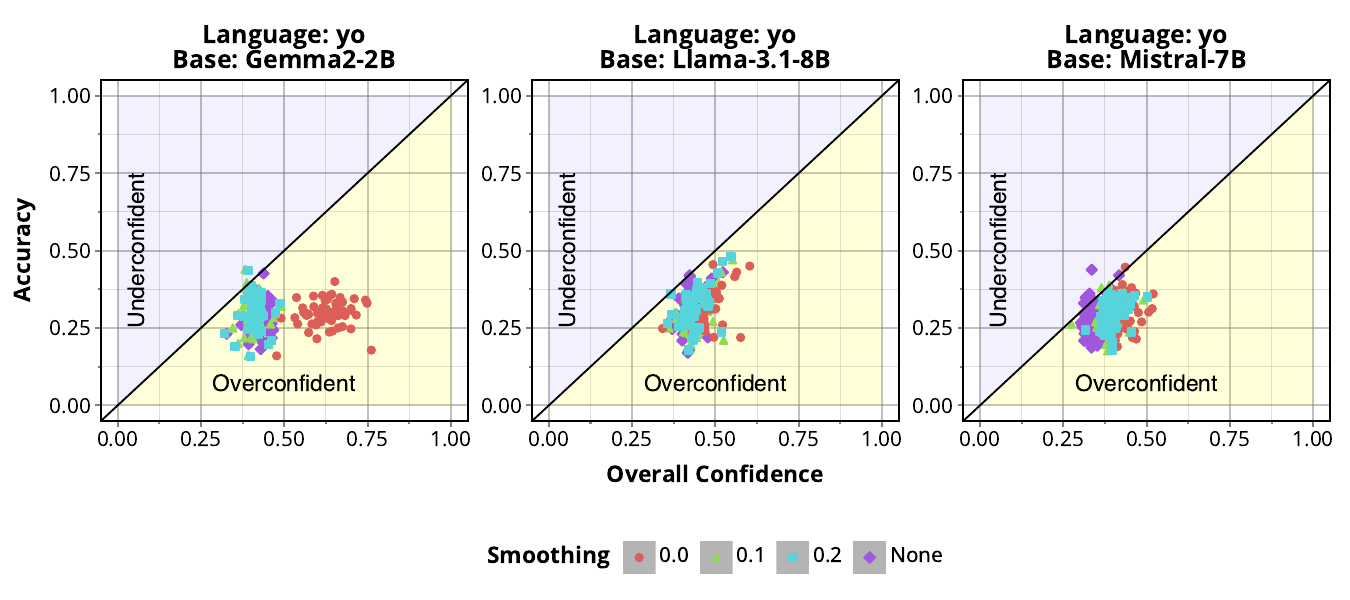}}
\caption{Reliability diagrams for the \textbf{\texttt{GlobalMMLU}} dataset for the \texttt{yo} language after instruction-tuning on the \textbf{\texttt{OpenHermes}} dataset.}\label{fig:globalmmlu-OpenHermes-yo}\end{figure}

\begin{figure}[h!]\centering\resizebox{\linewidth}{!}{\includegraphics[width=\linewidth]{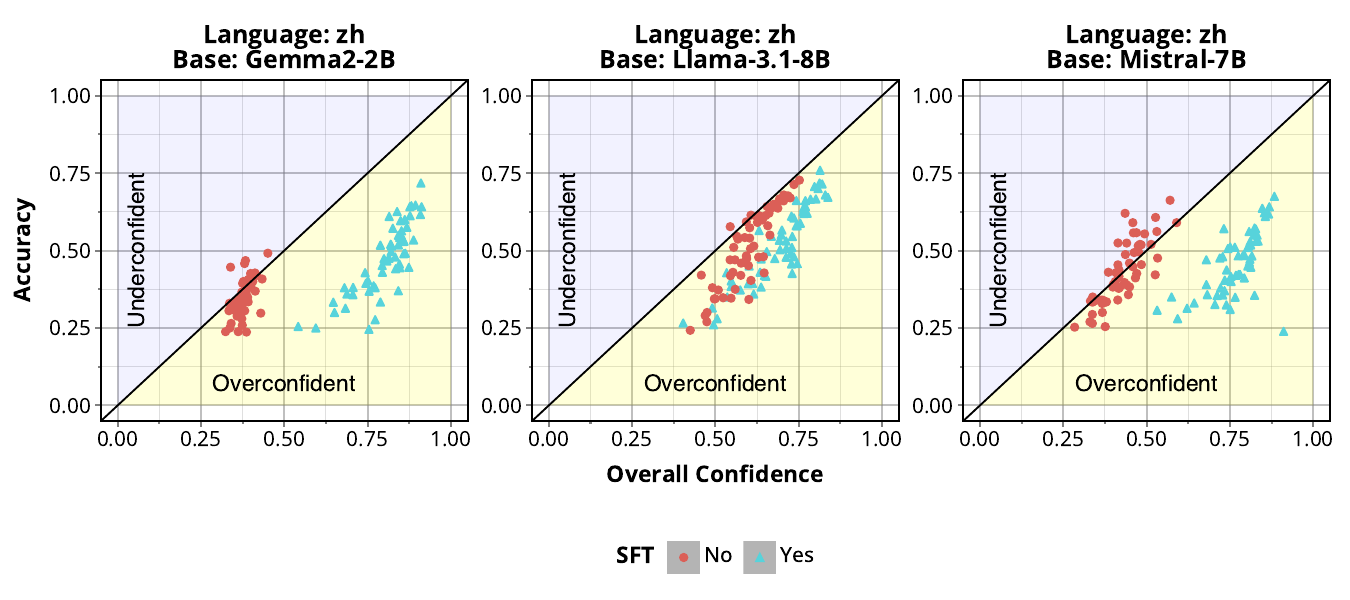}}
\caption{Reliability diagrams for the \textbf{\texttt{GlobalMMLU}} dataset for the \texttt{zh} language.}\label{fig:globalmmlu-base-zh}\end{figure}
\begin{figure}[h!]\centering\resizebox{\linewidth}{!}{\includegraphics[width=\linewidth]{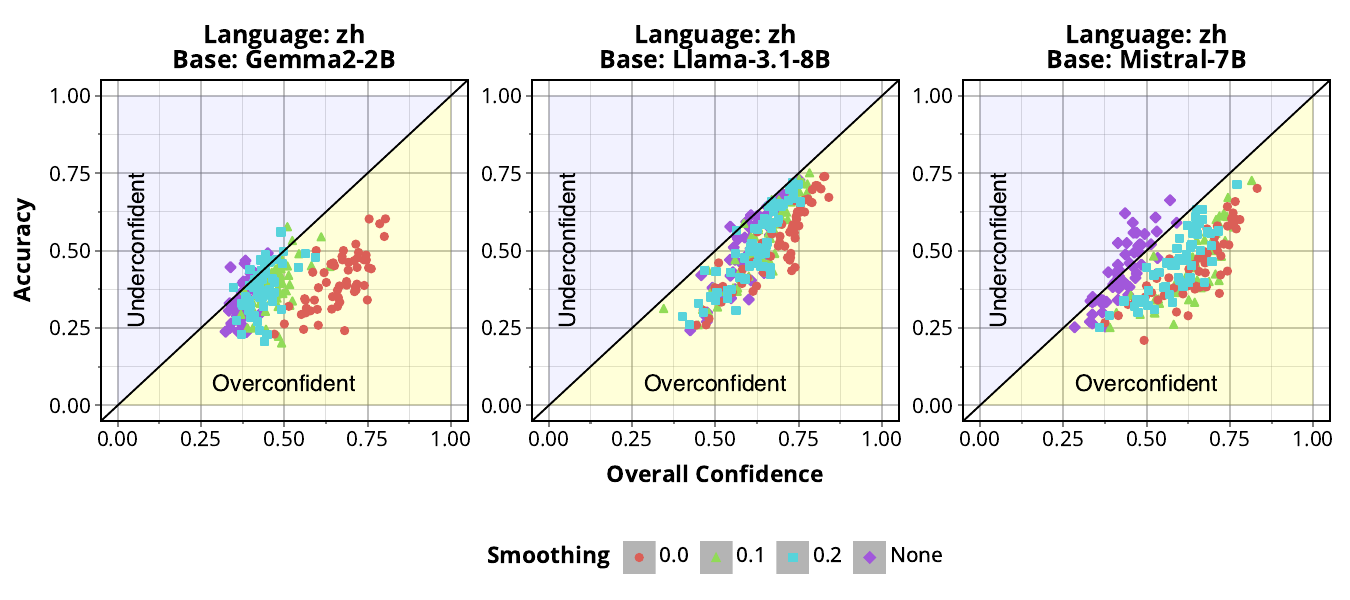}}
\caption{Reliability diagrams for the \textbf{\texttt{GlobalMMLU}} dataset for the \texttt{zh} language after instruction-tuning on the \textbf{\texttt{Tulu3Mixture}} dataset.}\label{fig:globalmmlu-Tulu3Mixture-zh}\end{figure}
\begin{figure}[h!]\centering\resizebox{\linewidth}{!}{\includegraphics[width=\linewidth]{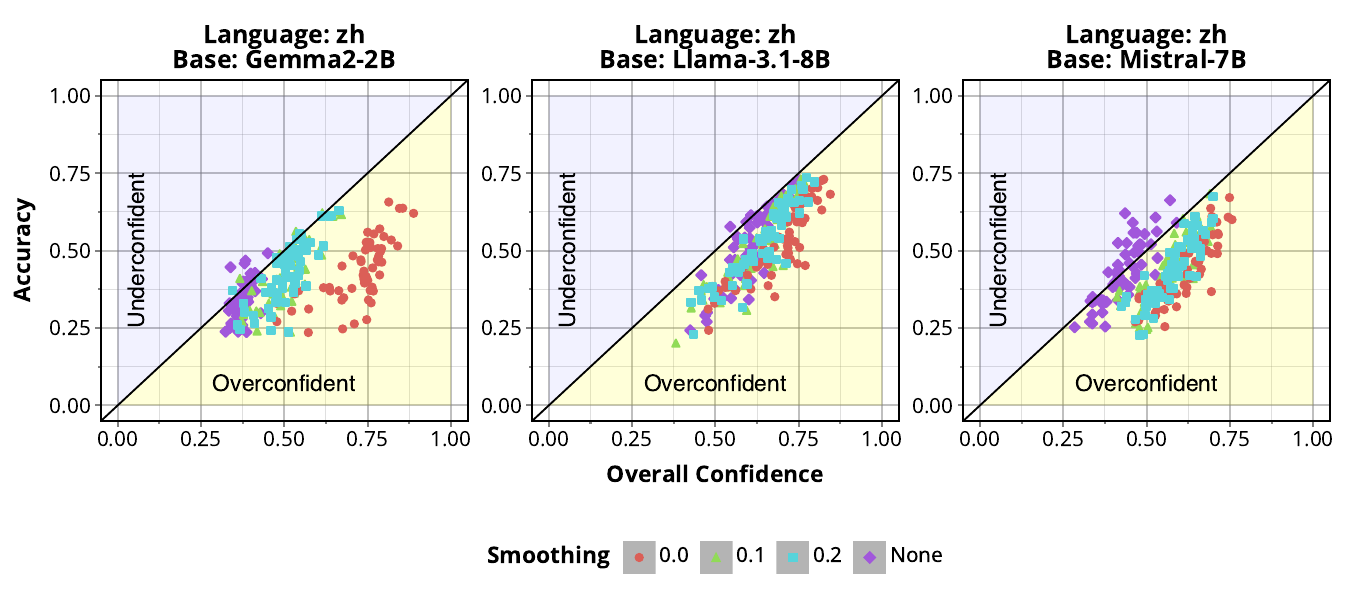}}
\caption{Reliability diagrams for the \textbf{\texttt{GlobalMMLU}} dataset for the \texttt{zh} language after instruction-tuning on the \textbf{\texttt{OpenHermes}} dataset.}\label{fig:globalmmlu-OpenHermes-zh}\end{figure}

\subsection{MMLU-ProX}

\begin{figure}[h!]\centering\resizebox{\linewidth}{!}{\includegraphics[width=\linewidth]{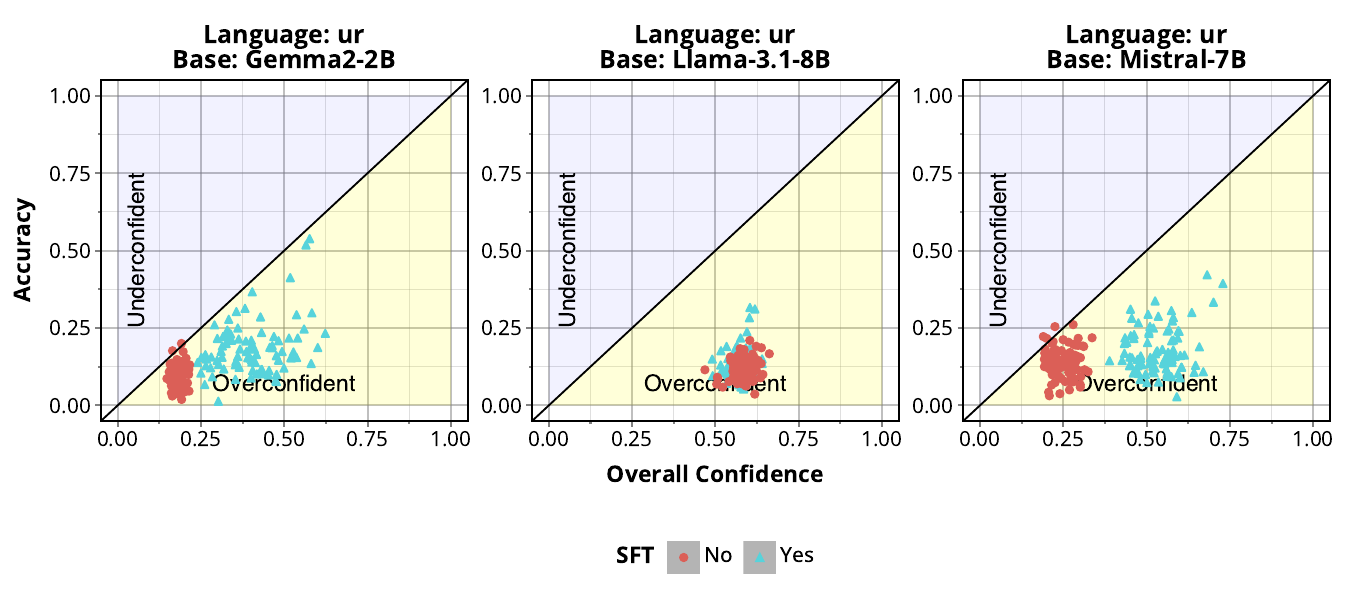}}
\caption{Reliability diagrams for the \textbf{\texttt{MMLU-ProX}} dataset for the \texttt{ur} language.}\label{fig:mmluprox-base-ur}\end{figure}
\begin{figure}[h!]\centering\resizebox{\linewidth}{!}{\includegraphics[width=\linewidth]{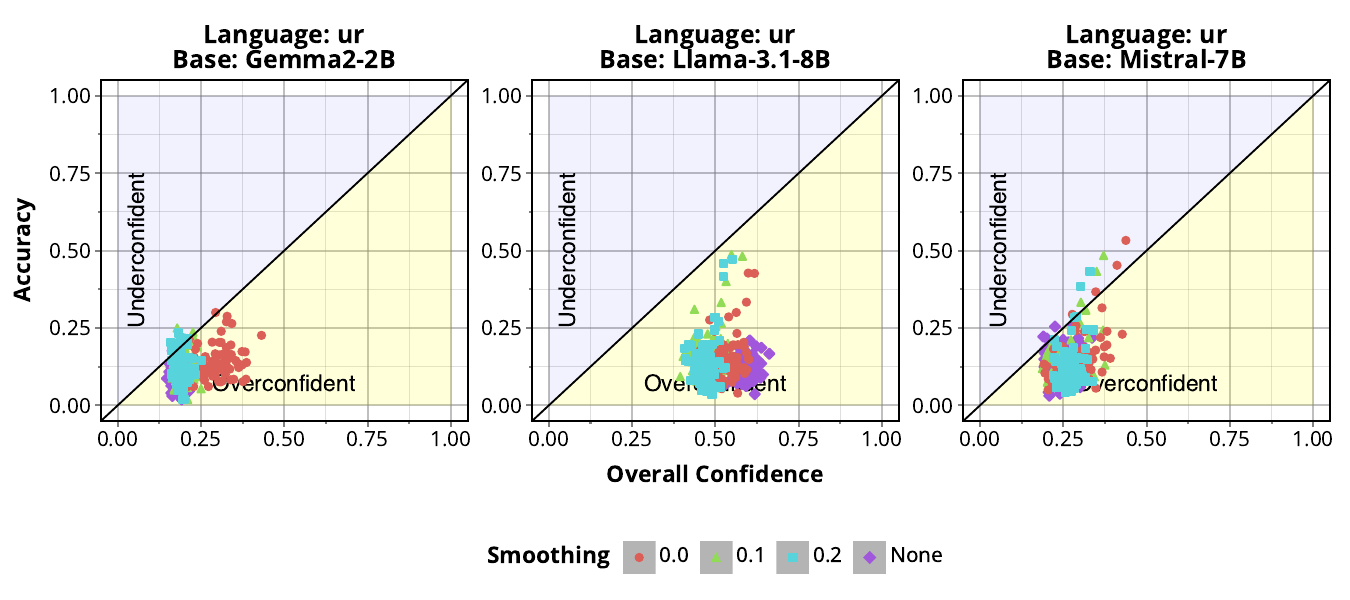}}
\caption{Reliability diagrams for the \textbf{\texttt{MMLU-ProX}} dataset for the \texttt{ur} language after instruction-tuning on the \textbf{\texttt{Tulu3Mixture}} dataset.}\label{fig:mmluprox-Tulu3Mixture-ur}\end{figure}
\begin{figure}[h!]\centering\resizebox{\linewidth}{!}{\includegraphics[width=\linewidth]{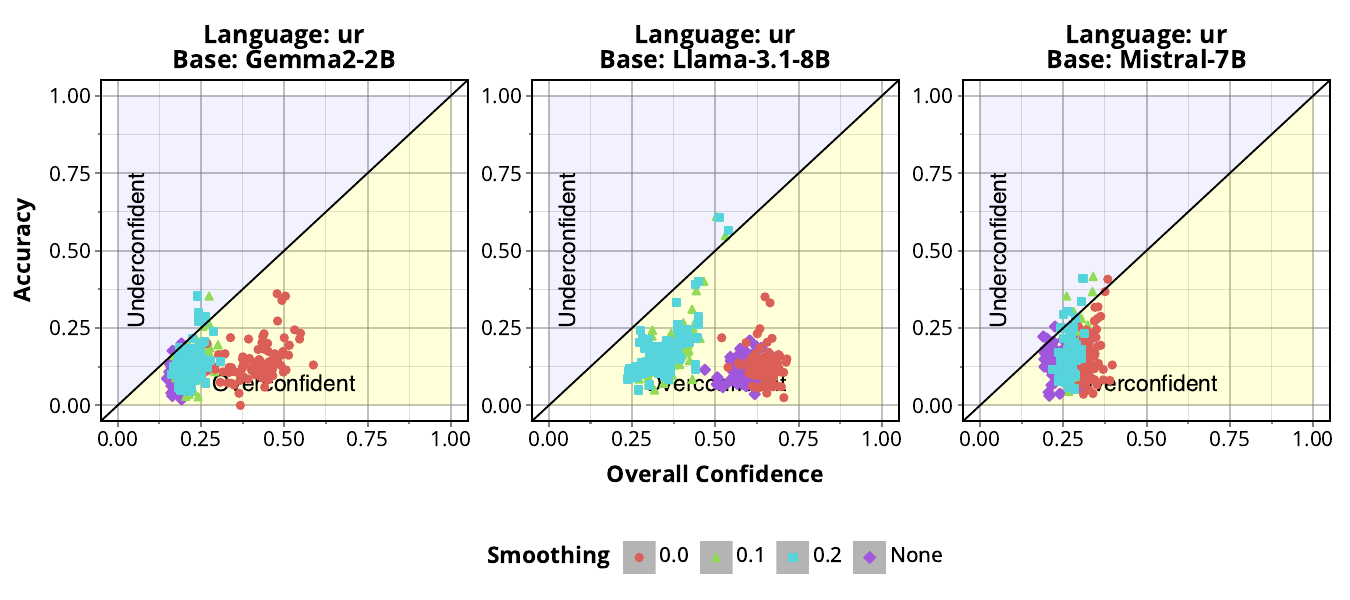}}
\caption{Reliability diagrams for the \textbf{\texttt{MMLU-ProX}} dataset for the \texttt{ur} language after instruction-tuning on the \textbf{\texttt{OpenHermes}} dataset.}\label{fig:mmluprox-OpenHermes-ur}\end{figure}

\begin{figure}[h!]\centering\resizebox{\linewidth}{!}{\includegraphics[width=\linewidth]{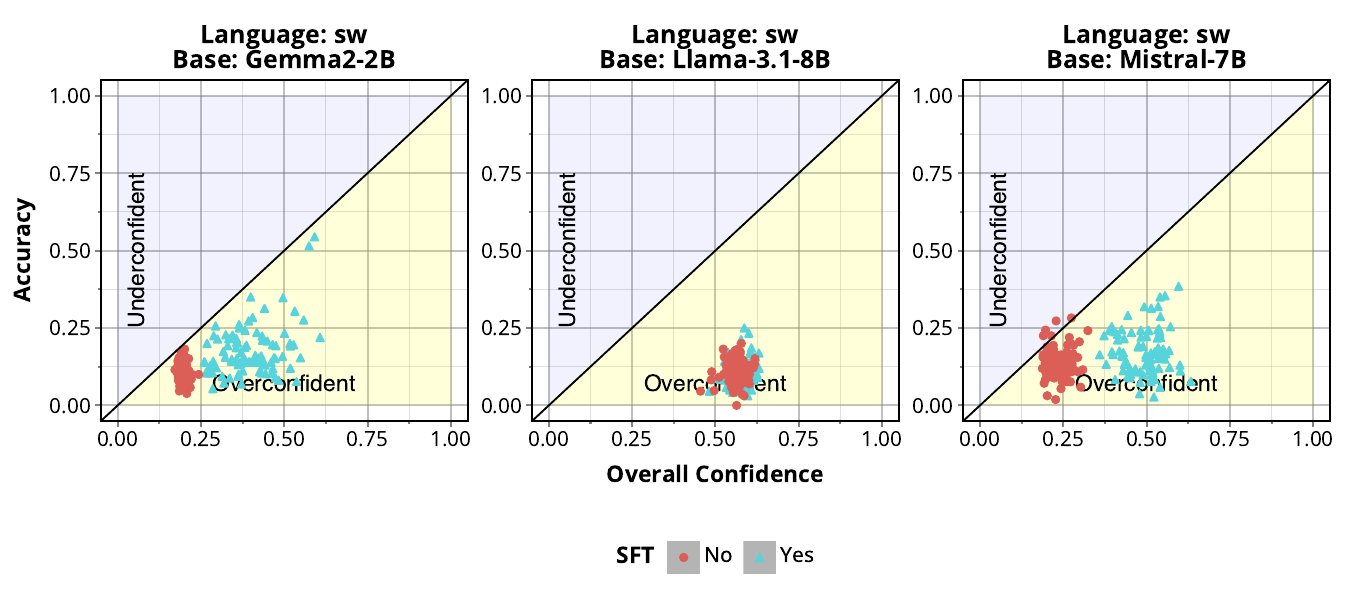}}
\caption{Reliability diagrams for the \textbf{\texttt{MMLU-ProX}} dataset for the \texttt{sw} language.}\label{fig:mmluprox-base-sw}\end{figure}
\begin{figure}[h!]\centering\resizebox{\linewidth}{!}{\includegraphics[width=\linewidth]{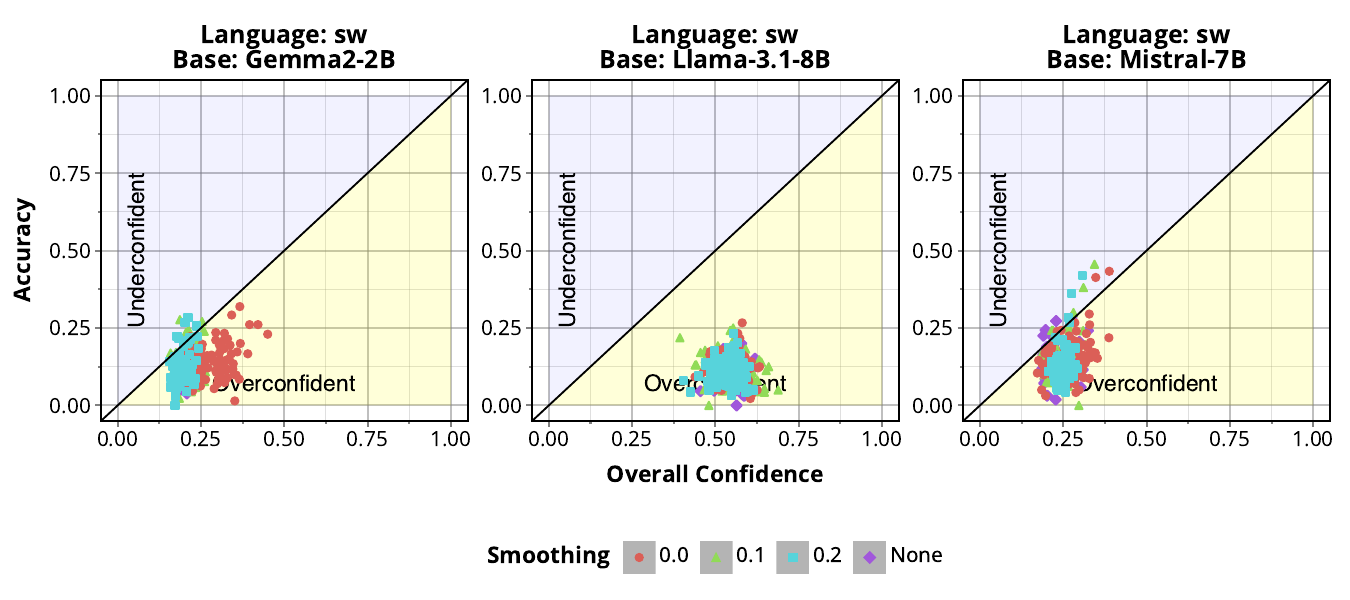}}
\caption{Reliability diagrams for the \textbf{\texttt{MMLU-ProX}} dataset for the \texttt{sw} language after instruction-tuning on the \textbf{\texttt{Tulu3Mixture}} dataset.}\label{fig:mmluprox-Tulu3Mixture-sw}\end{figure}
\begin{figure}[h!]\centering\resizebox{\linewidth}{!}{\includegraphics[width=\linewidth]{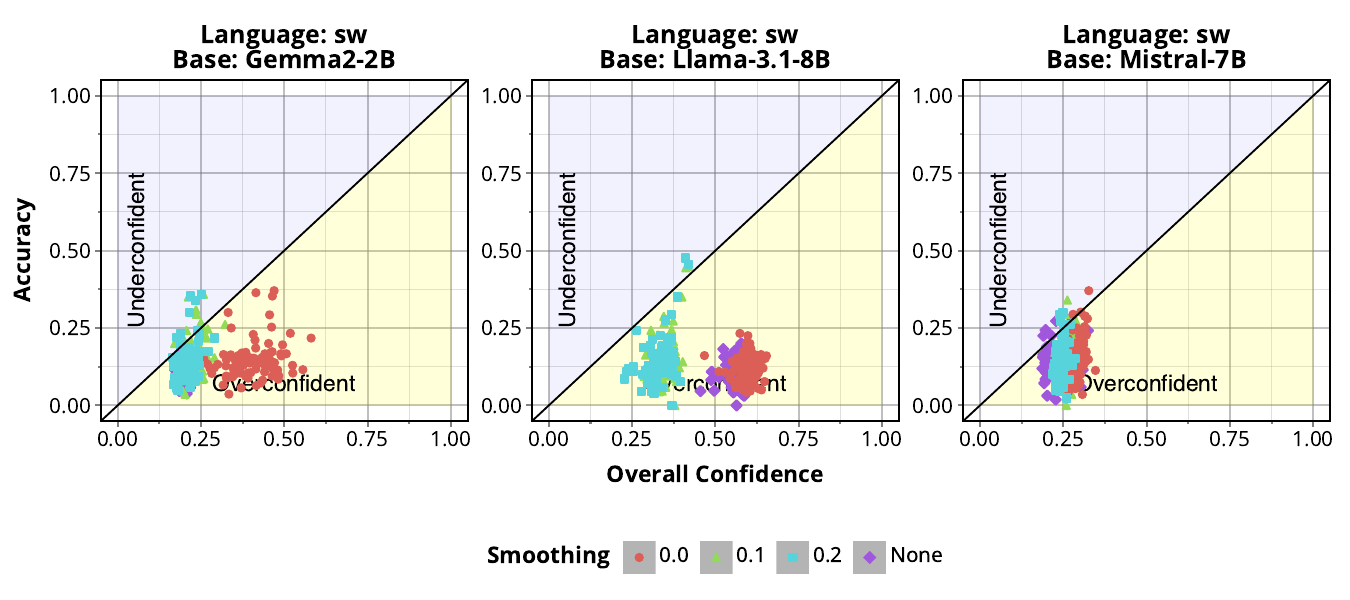}}
\caption{Reliability diagrams for the \textbf{\texttt{MMLU-ProX}} dataset for the \texttt{sw} language after instruction-tuning on the \textbf{\texttt{OpenHermes}} dataset.}\label{fig:mmluprox-OpenHermes-sw}\end{figure}

\begin{figure}[h!]\centering\resizebox{\linewidth}{!}{\includegraphics[width=\linewidth]{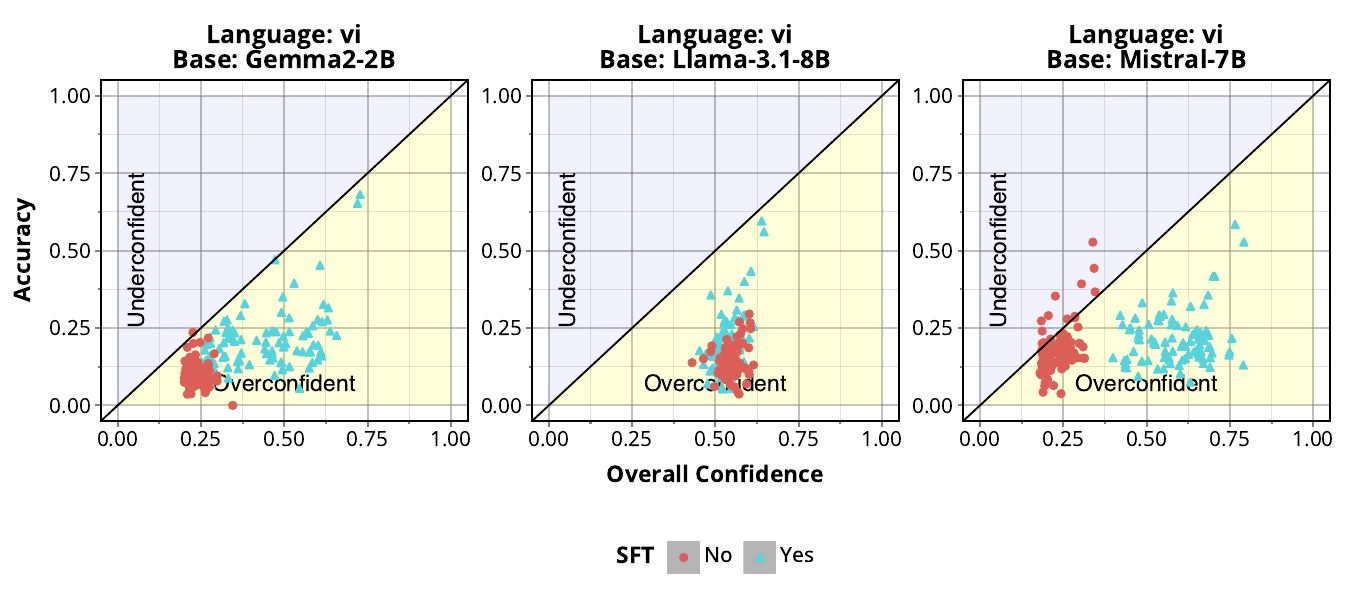}}
\caption{Reliability diagrams for the \textbf{\texttt{MMLU-ProX}} dataset for the \texttt{vi} language.}\label{fig:mmluprox-base-vi}\end{figure}
\begin{figure}[h!]\centering\resizebox{\linewidth}{!}{\includegraphics[width=\linewidth]{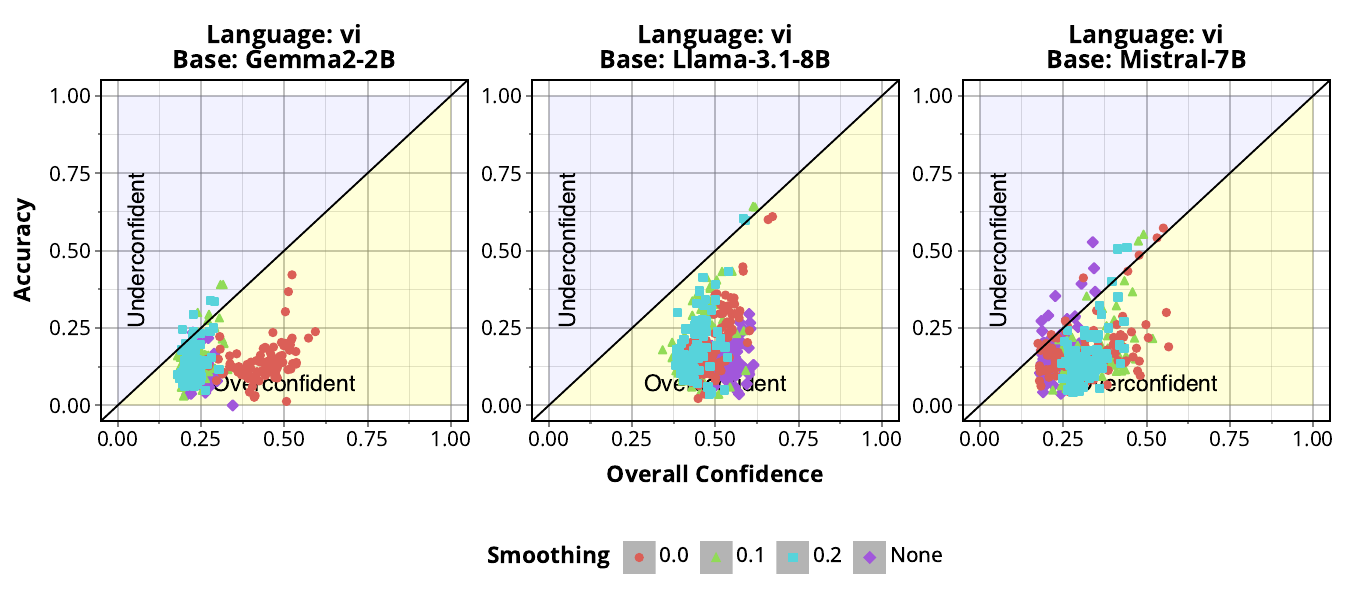}}
\caption{Reliability diagrams for the \textbf{\texttt{MMLU-ProX}} dataset for the \texttt{vi} language after instruction-tuning on the \textbf{\texttt{Tulu3Mixture}} dataset.}\label{fig:mmluprox-Tulu3Mixture-vi}\end{figure}
\begin{figure}[h!]\centering\resizebox{\linewidth}{!}{\includegraphics[width=\linewidth]{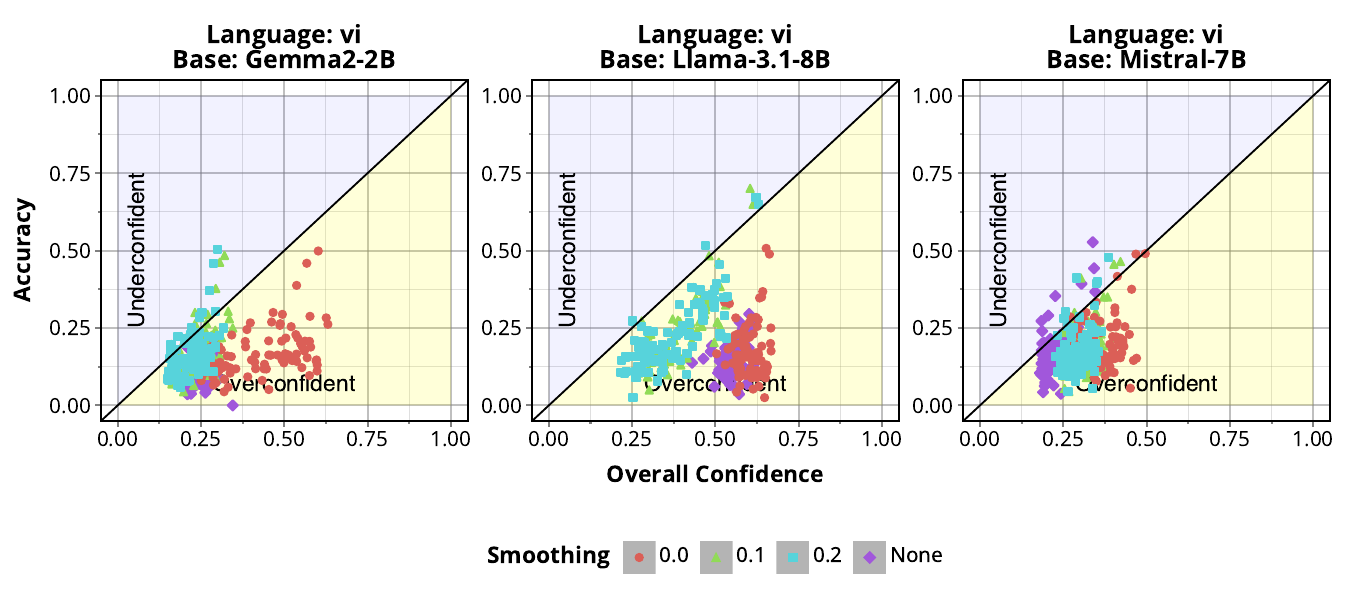}}
\caption{Reliability diagrams for the \textbf{\texttt{MMLU-ProX}} dataset for the \texttt{vi} language after instruction-tuning on the \textbf{\texttt{OpenHermes}} dataset.}\label{fig:mmluprox-OpenHermes-vi}\end{figure}

\begin{figure}[h!]\centering\resizebox{\linewidth}{!}{\includegraphics[width=\linewidth]{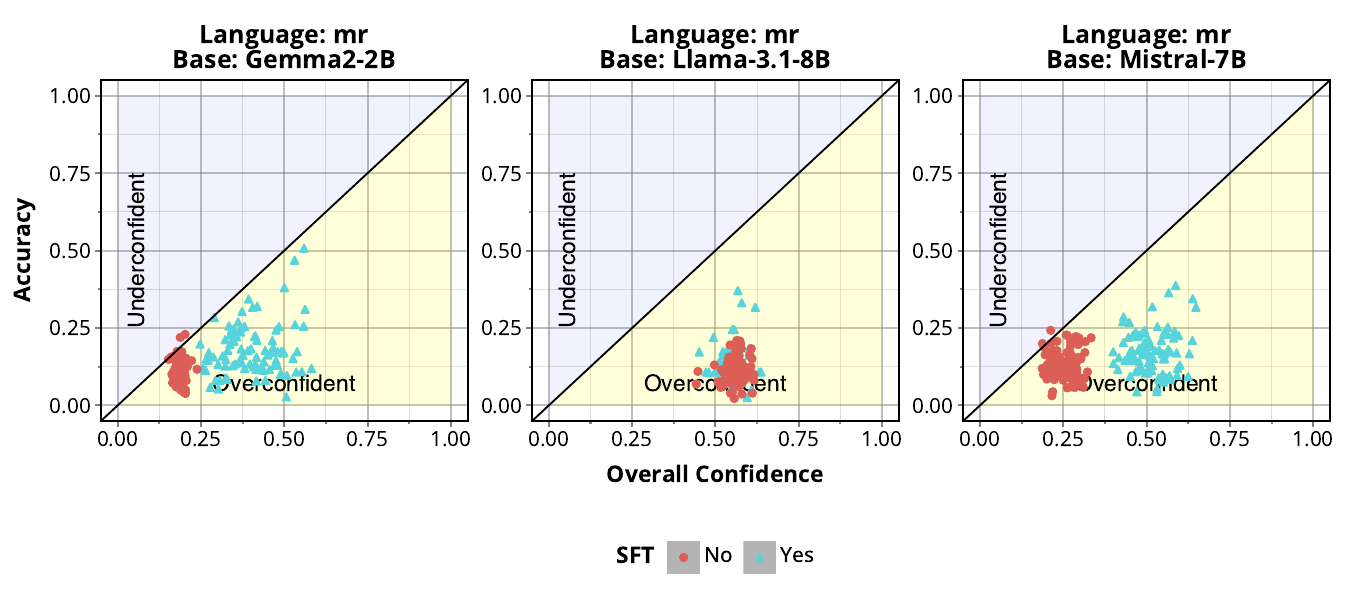}}
\caption{Reliability diagrams for the \textbf{\texttt{MMLU-ProX}} dataset for the \texttt{mr} language.}\label{fig:mmluprox-base-mr}\end{figure}
\begin{figure}[h!]\centering\resizebox{\linewidth}{!}{\includegraphics[width=\linewidth]{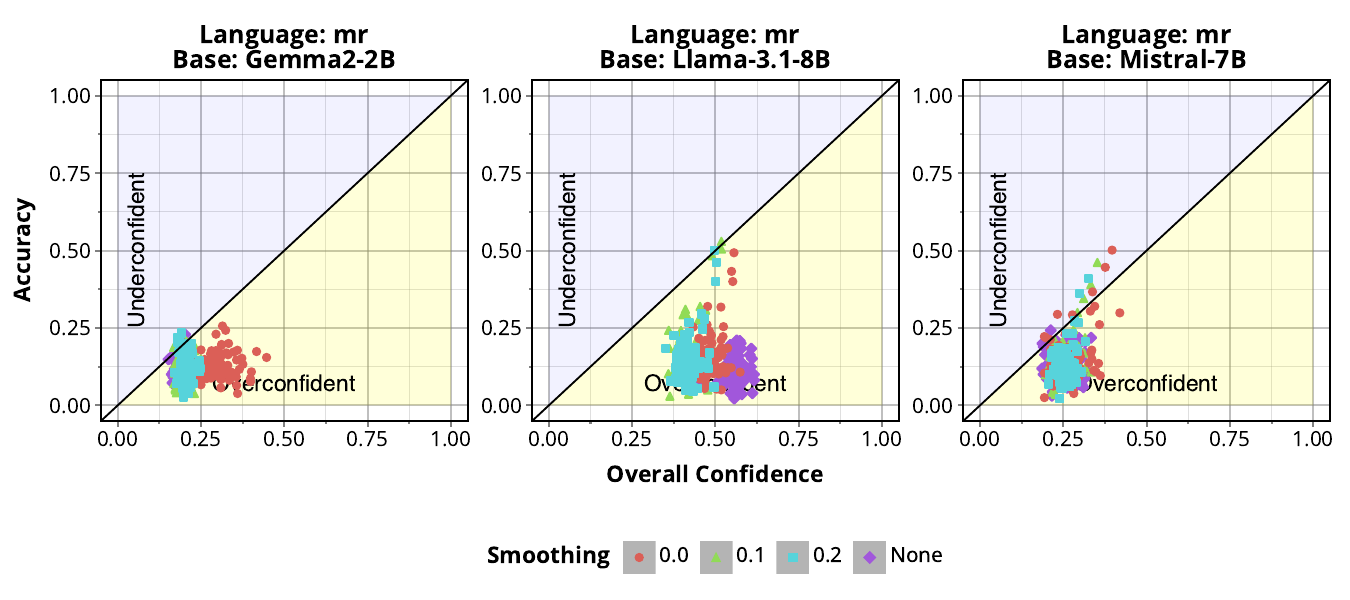}}
\caption{Reliability diagrams for the \textbf{\texttt{MMLU-ProX}} dataset for the \texttt{mr} language after instruction-tuning on the \textbf{\texttt{Tulu3Mixture}} dataset.}\label{fig:mmluprox-Tulu3Mixture-mr}\end{figure}
\begin{figure}[h!]\centering\resizebox{\linewidth}{!}{\includegraphics[width=\linewidth]{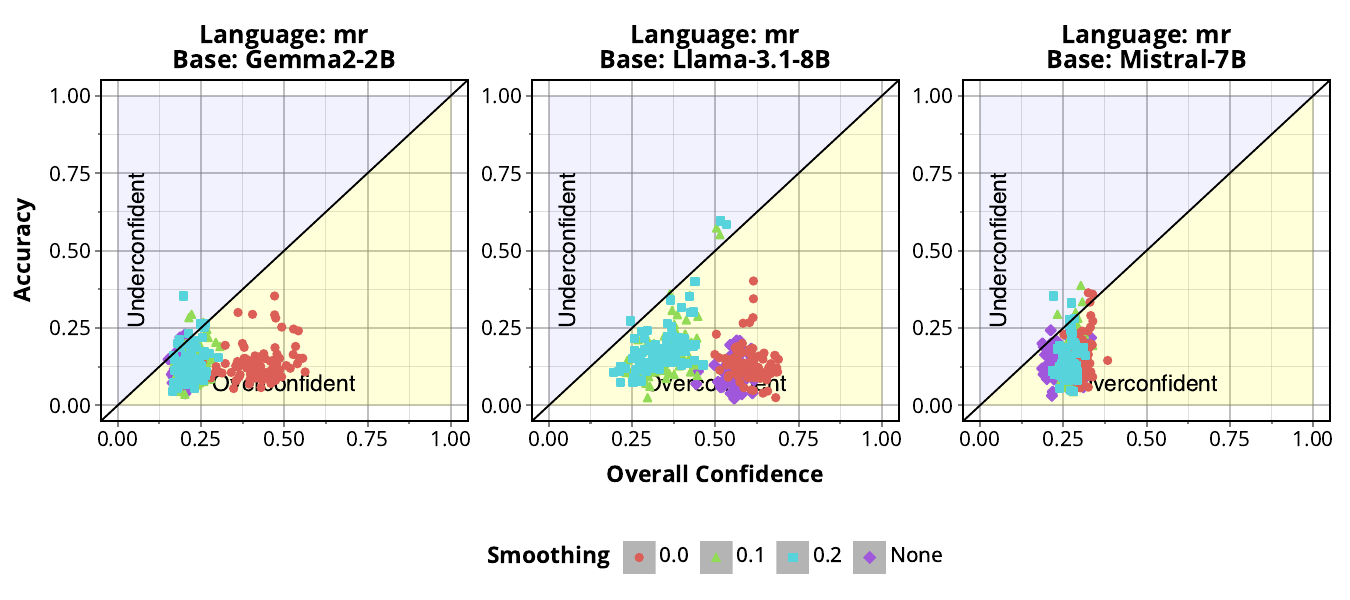}}
\caption{Reliability diagrams for the \textbf{\texttt{MMLU-ProX}} dataset for the \texttt{mr} language after instruction-tuning on the \textbf{\texttt{OpenHermes}} dataset.}\label{fig:mmluprox-OpenHermes-mr}\end{figure}

\begin{figure}[h!]\centering\resizebox{\linewidth}{!}{\includegraphics[width=\linewidth]{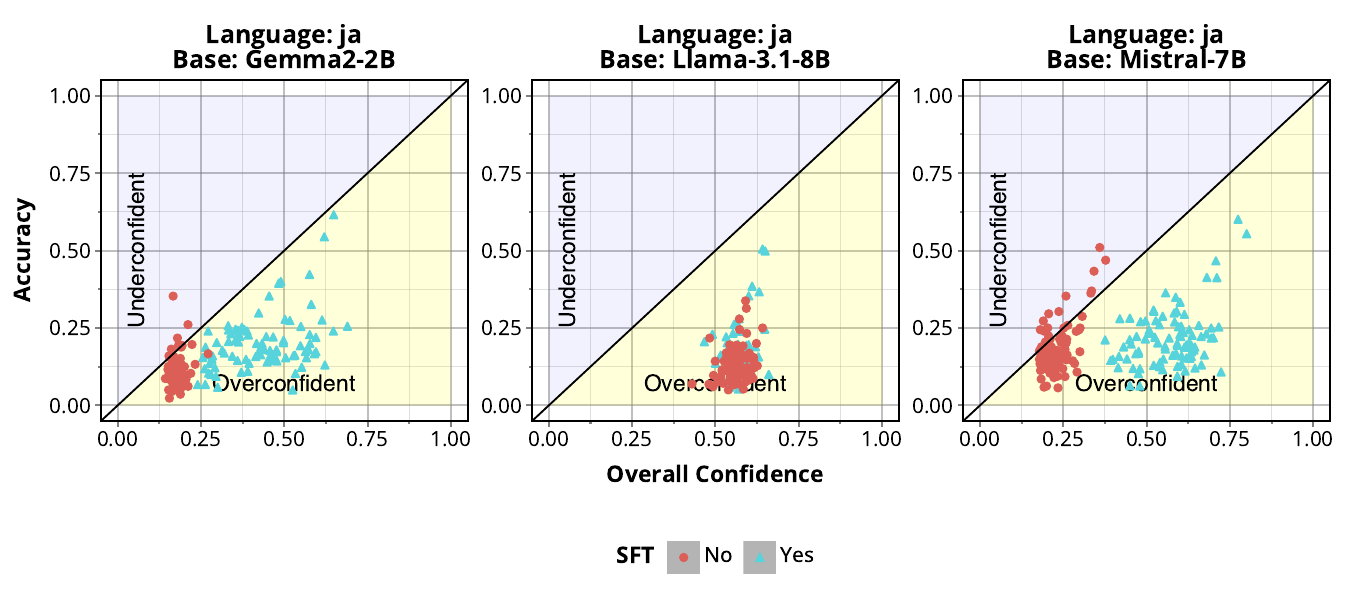}}
\caption{Reliability diagrams for the \textbf{\texttt{MMLU-ProX}} dataset for the \texttt{ja} language.}\label{fig:mmluprox-base-ja}\end{figure}
\begin{figure}[h!]\centering\resizebox{\linewidth}{!}{\includegraphics[width=\linewidth]{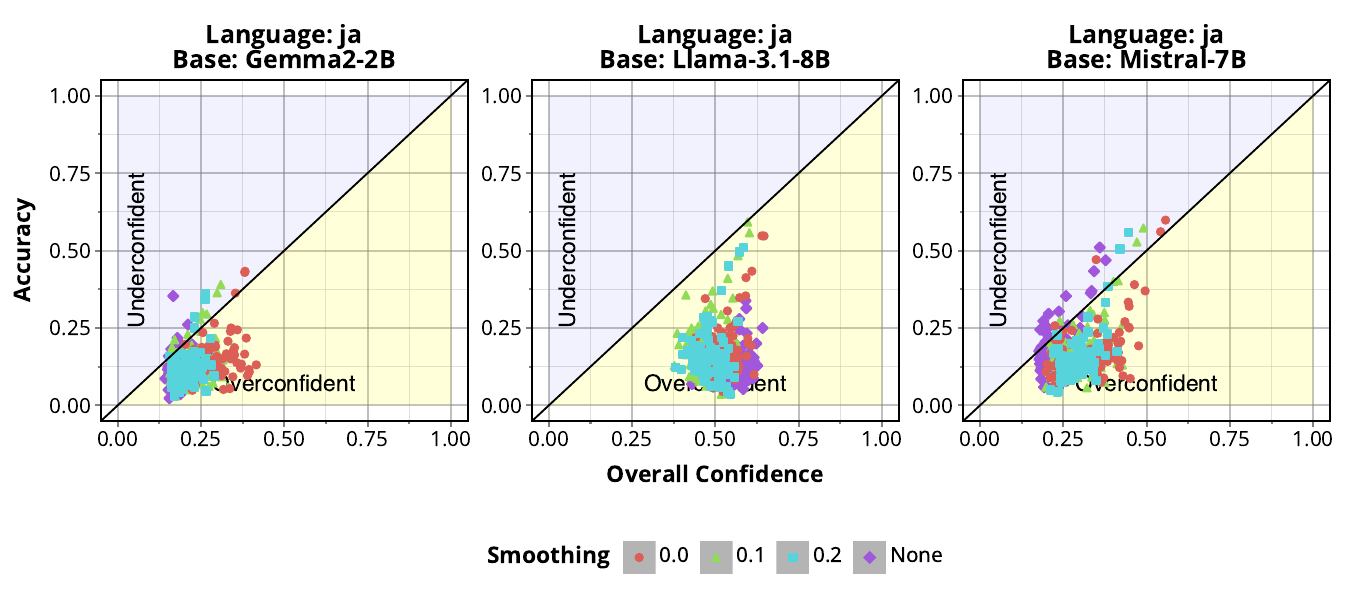}}
\caption{Reliability diagrams for the \textbf{\texttt{MMLU-ProX}} dataset for the \texttt{ja} language after instruction-tuning on the \textbf{\texttt{Tulu3Mixture}} dataset.}\label{fig:mmluprox-Tulu3Mixture-ja}\end{figure}
\begin{figure}[h!]\centering\resizebox{\linewidth}{!}{\includegraphics[width=\linewidth]{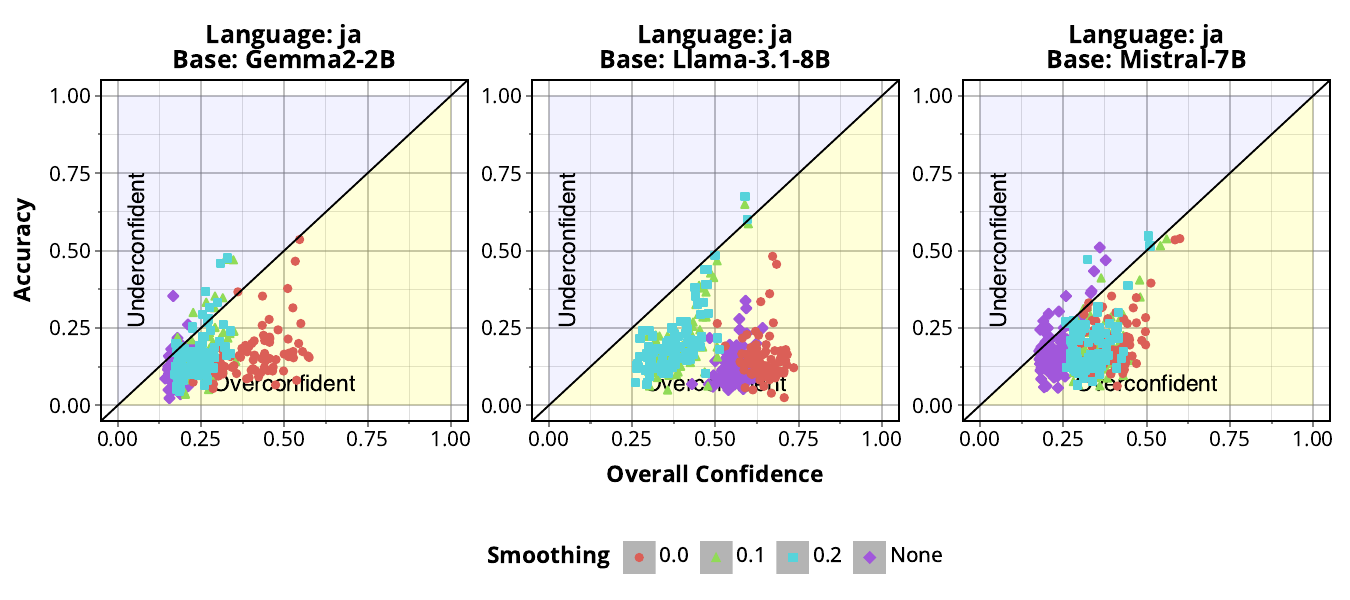}}
\caption{Reliability diagrams for the \textbf{\texttt{MMLU-ProX}} dataset for the \texttt{ja} language after instruction-tuning on the \textbf{\texttt{OpenHermes}} dataset.}\label{fig:mmluprox-OpenHermes-ja}\end{figure}

\begin{figure}[h!]\centering\resizebox{\linewidth}{!}{\includegraphics[width=\linewidth]{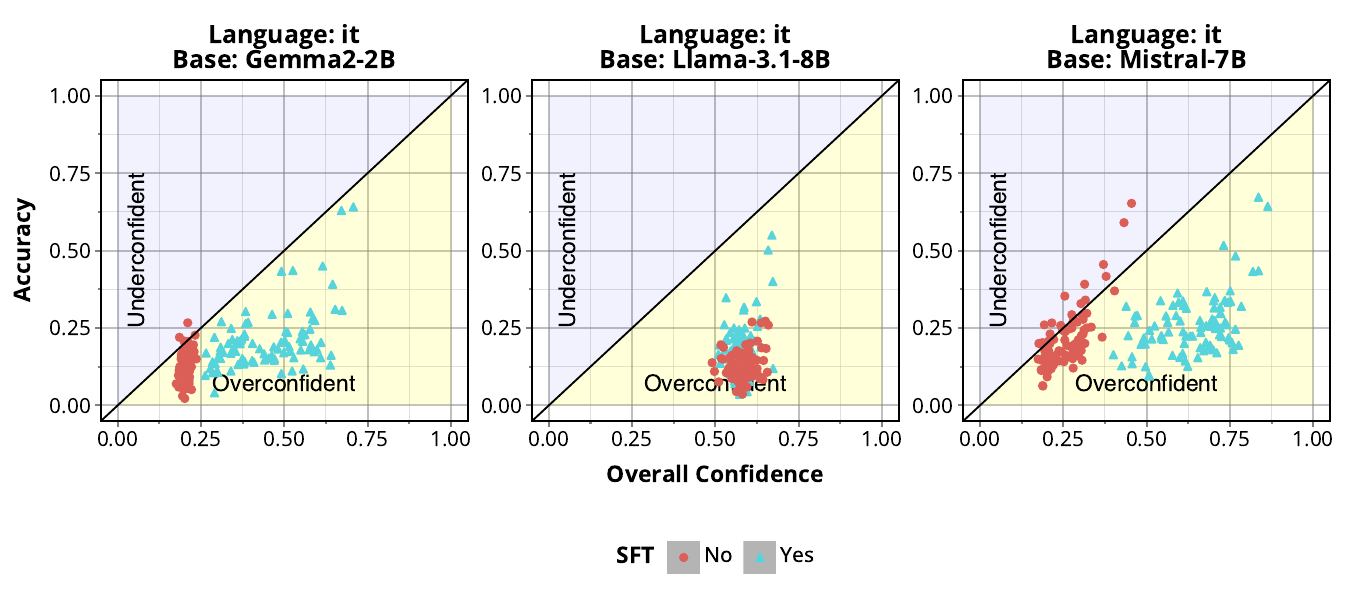}}
\caption{Reliability diagrams for the \textbf{\texttt{MMLU-ProX}} dataset for the \texttt{it} language.}\label{fig:mmluprox-base-it}\end{figure}
\begin{figure}[h!]\centering\resizebox{\linewidth}{!}{\includegraphics[width=\linewidth]{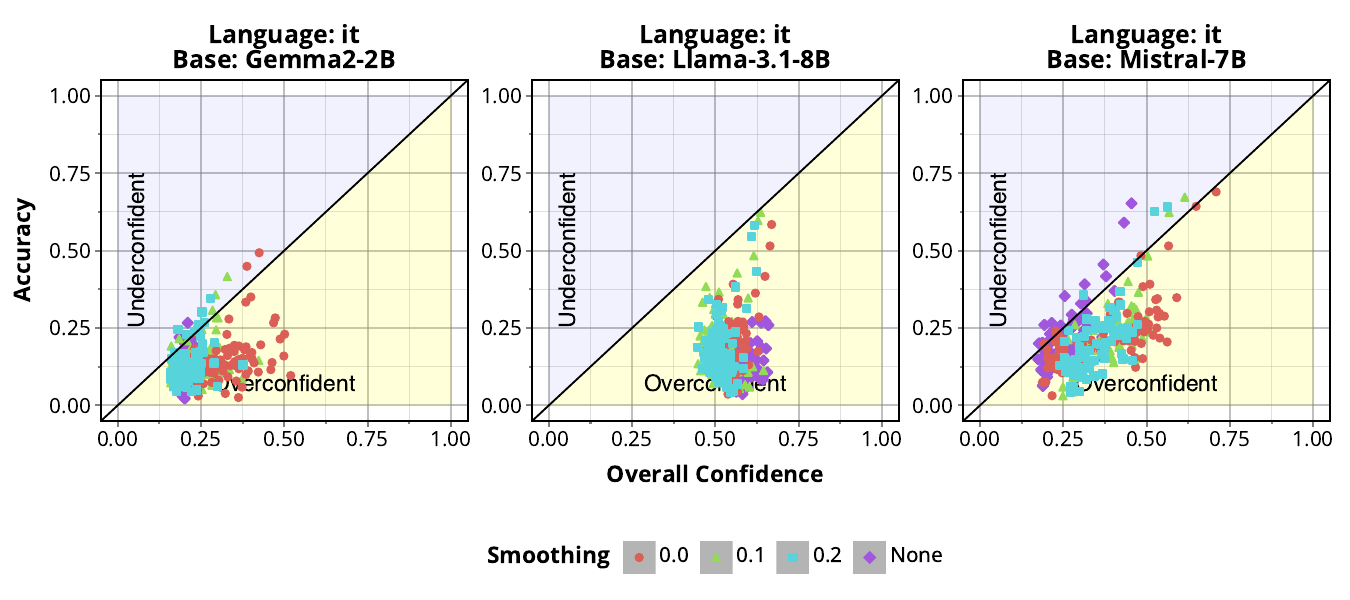}}
\caption{Reliability diagrams for the \textbf{\texttt{MMLU-ProX}} dataset for the \texttt{it} language after instruction-tuning on the \textbf{\texttt{Tulu3Mixture}} dataset.}\label{fig:mmluprox-Tulu3Mixture-it}\end{figure}
\begin{figure}[h!]\centering\resizebox{\linewidth}{!}{\includegraphics[width=\linewidth]{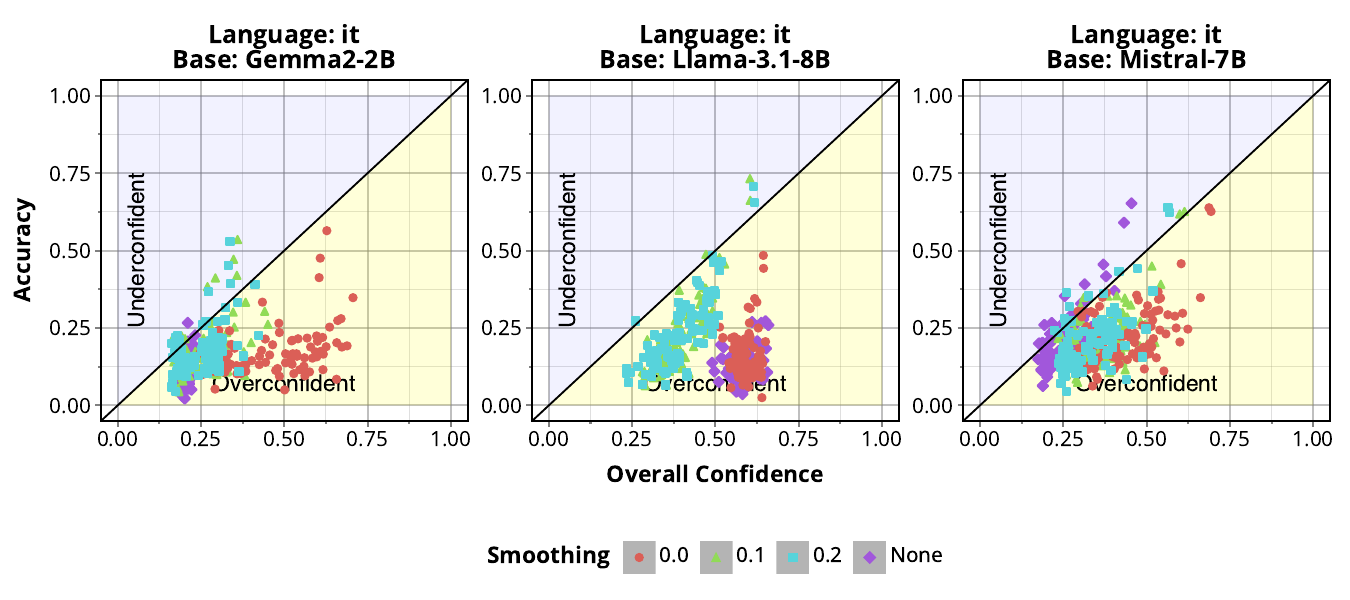}}
\caption{Reliability diagrams for the \textbf{\texttt{MMLU-ProX}} dataset for the \texttt{it} language after instruction-tuning on the \textbf{\texttt{OpenHermes}} dataset.}\label{fig:mmluprox-OpenHermes-it}\end{figure}

\clearpage
\begin{figure}[h!]\centering\resizebox{\linewidth}{!}{\includegraphics[width=\linewidth]{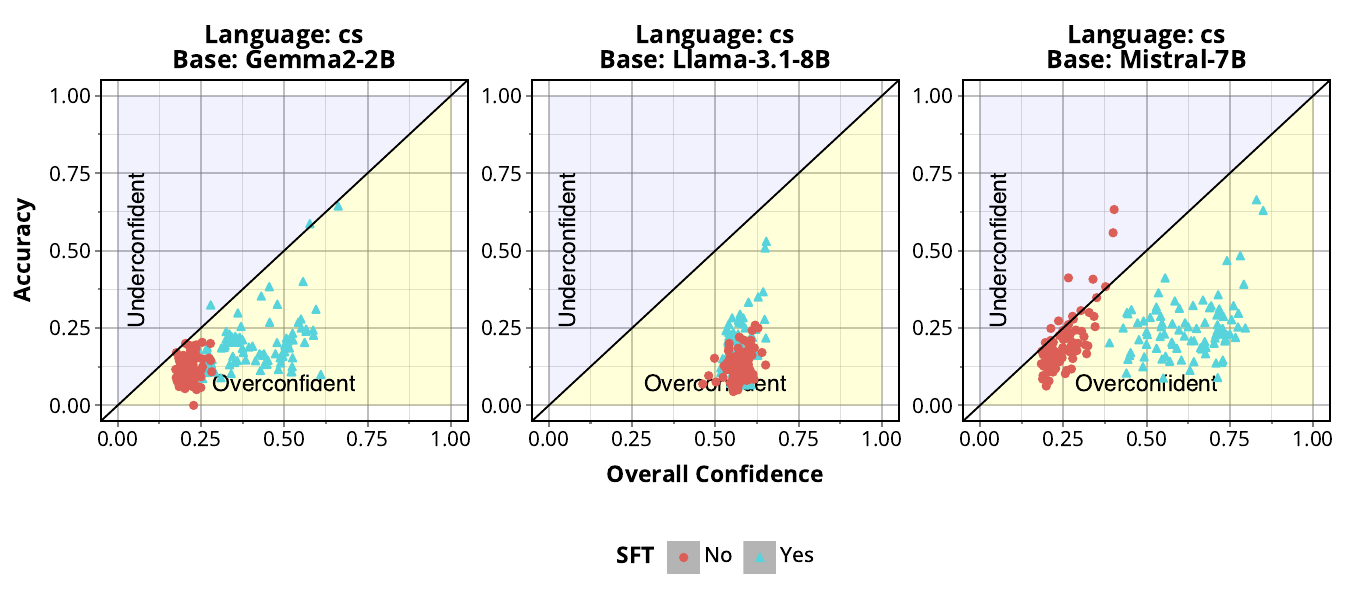}}
\caption{Reliability diagrams for the \textbf{\texttt{MMLU-ProX}} dataset for the \texttt{cs} language.}\label{fig:mmluprox-base-cs}\end{figure}
\begin{figure}[h!]\centering\resizebox{\linewidth}{!}{\includegraphics[width=\linewidth]{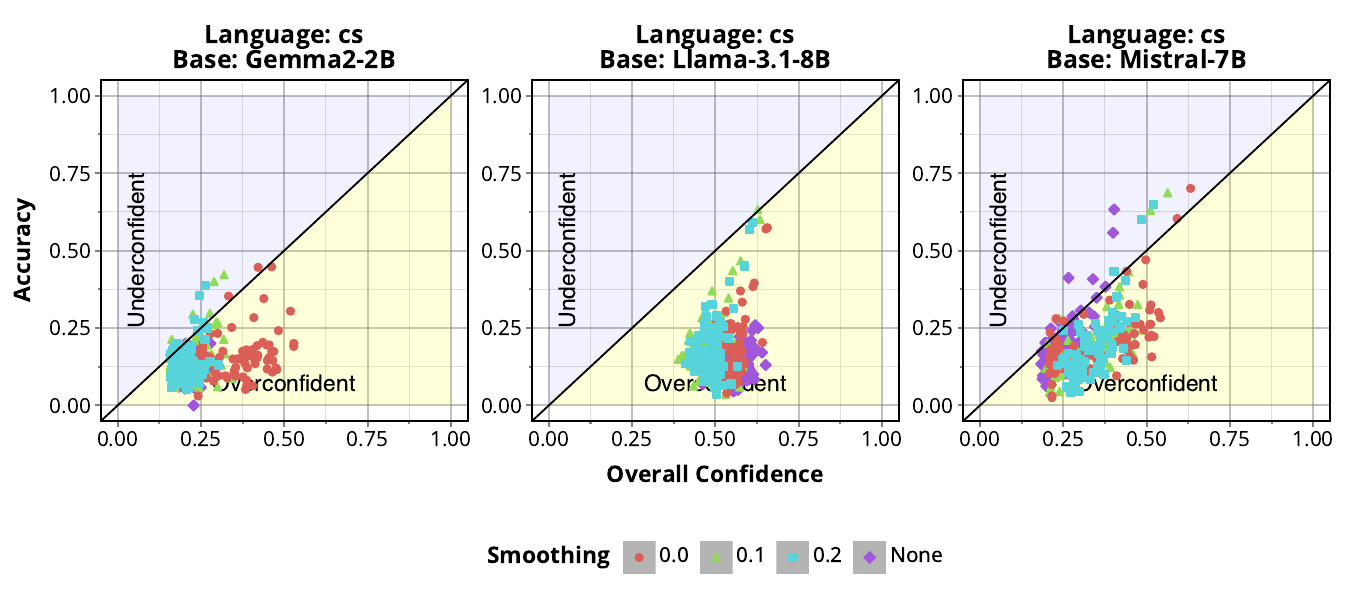}}
\caption{Reliability diagrams for the \textbf{\texttt{MMLU-ProX}} dataset for the \texttt{cs} language after instruction-tuning on the \textbf{\texttt{Tulu3Mixture}} dataset.}\label{fig:mmluprox-Tulu3Mixture-cs}\end{figure}
\begin{figure}[h!]\centering\resizebox{\linewidth}{!}{\includegraphics[width=\linewidth]{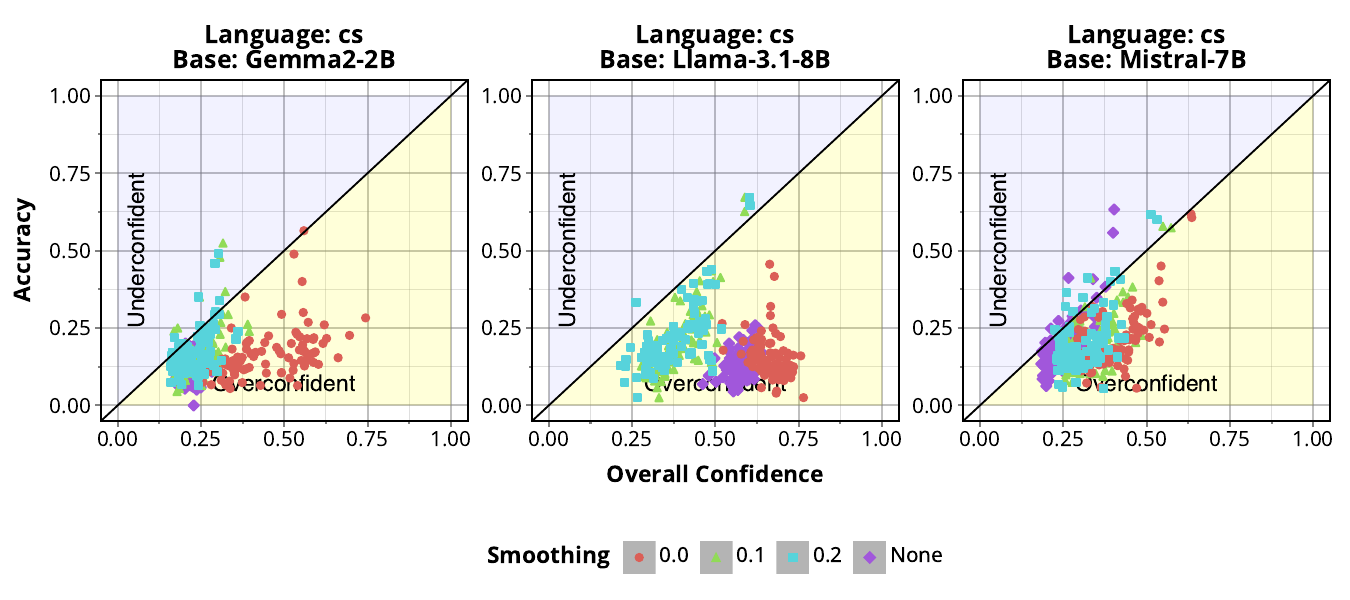}}
\caption{Reliability diagrams for the \textbf{\texttt{MMLU-ProX}} dataset for the \texttt{cs} language after instruction-tuning on the \textbf{\texttt{OpenHermes}} dataset.}\label{fig:mmluprox-OpenHermes-cs}\end{figure}

\begin{figure}[h!]\centering\resizebox{\linewidth}{!}{\includegraphics[width=\linewidth]{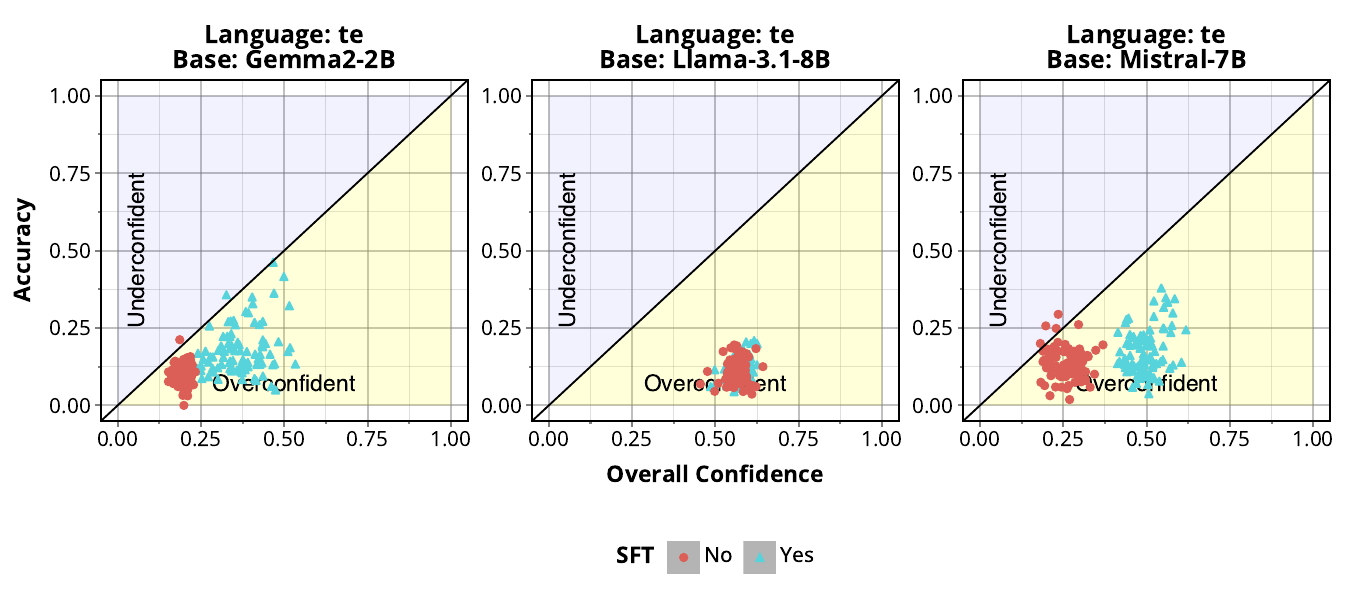}}
\caption{Reliability diagrams for the \textbf{\texttt{MMLU-ProX}} dataset for the \texttt{te} language.}\label{fig:mmluprox-base-te}\end{figure}
\begin{figure}[h!]\centering\resizebox{\linewidth}{!}{\includegraphics[width=\linewidth]{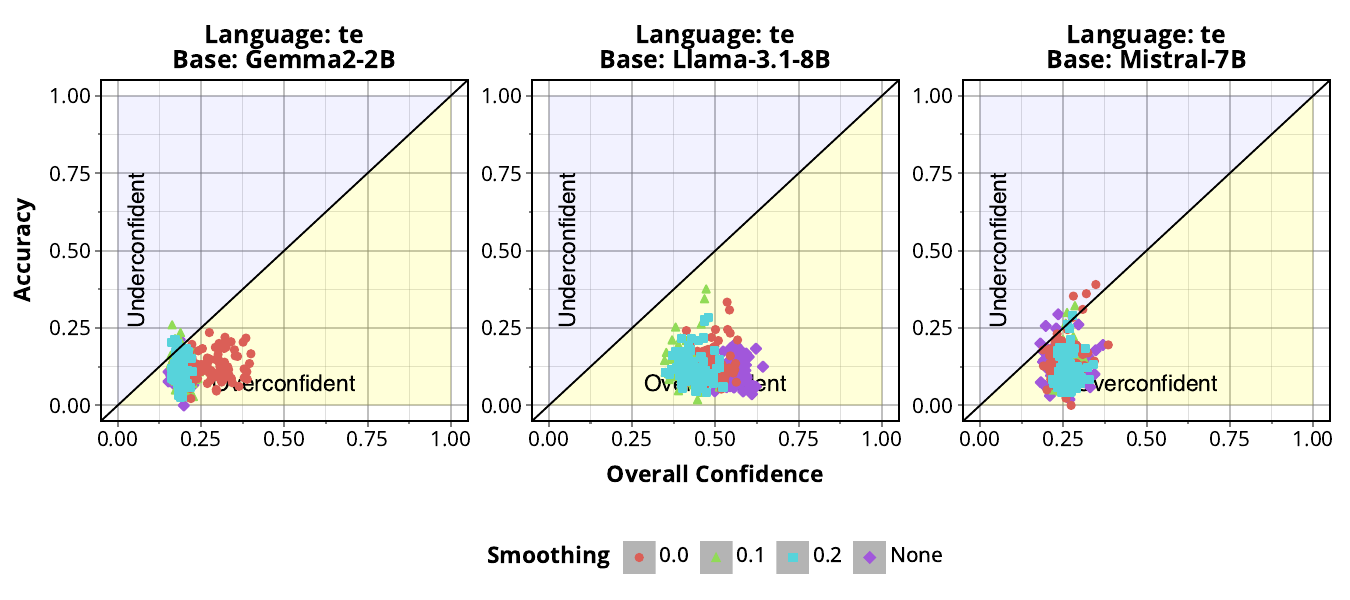}}
\caption{Reliability diagrams for the \textbf{\texttt{MMLU-ProX}} dataset for the \texttt{te} language after instruction-tuning on the \textbf{\texttt{Tulu3Mixture}} dataset.}\label{fig:mmluprox-Tulu3Mixture-te}\end{figure}
\begin{figure}[h!]\centering\resizebox{\linewidth}{!}{\includegraphics[width=\linewidth]{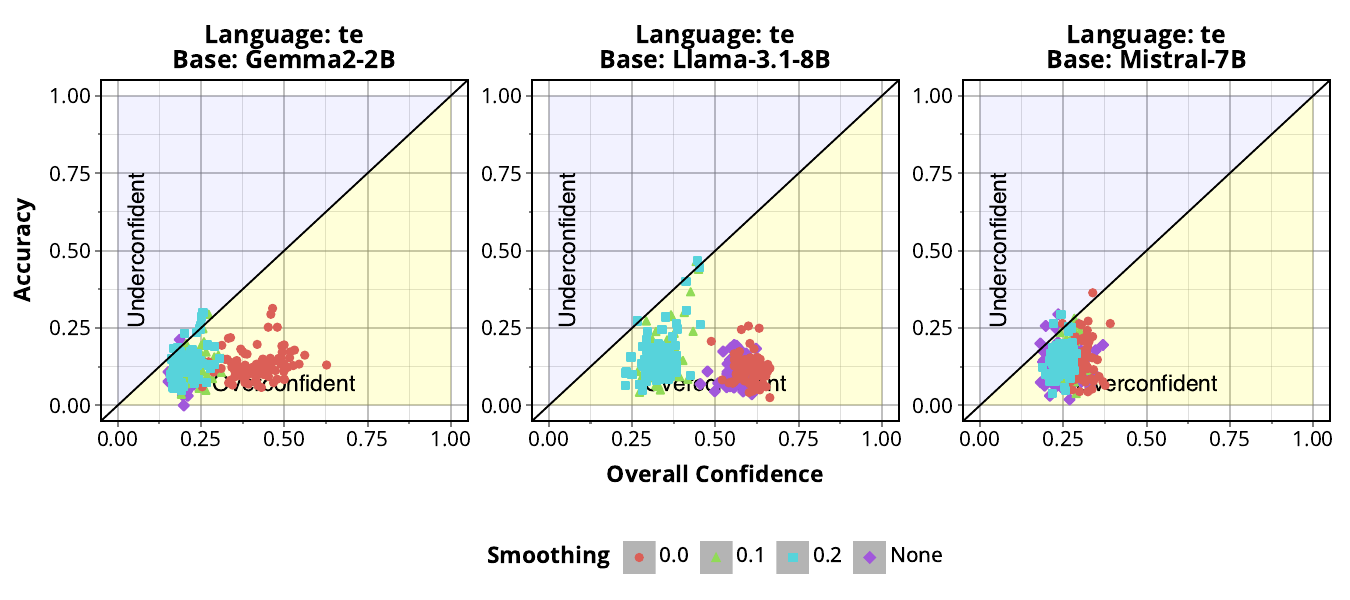}}
\caption{Reliability diagrams for the \textbf{\texttt{MMLU-ProX}} dataset for the \texttt{te} language after instruction-tuning on the \textbf{\texttt{OpenHermes}} dataset.}\label{fig:mmluprox-OpenHermes-te}\end{figure}

\begin{figure}[h!]\centering\resizebox{\linewidth}{!}{\includegraphics[width=\linewidth]{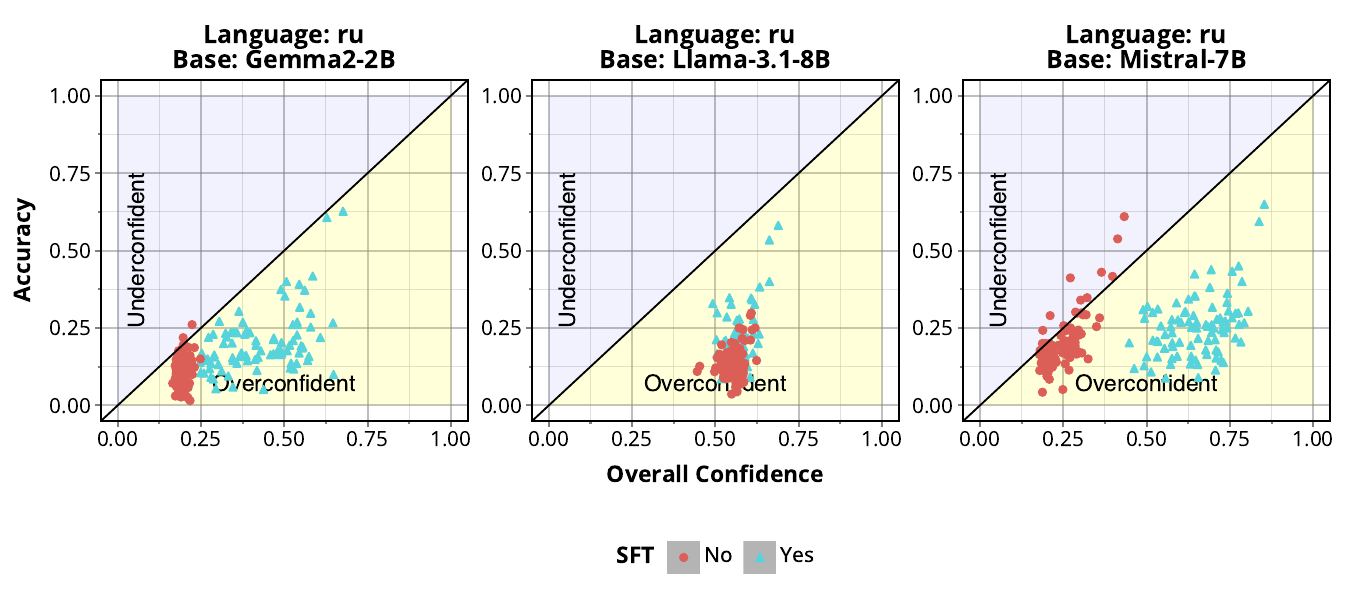}}
\caption{Reliability diagrams for the \textbf{\texttt{MMLU-ProX}} dataset for the \texttt{ru} language.}\label{fig:mmluprox-base-ru}\end{figure}
\begin{figure}[h!]\centering\resizebox{\linewidth}{!}{\includegraphics[width=\linewidth]{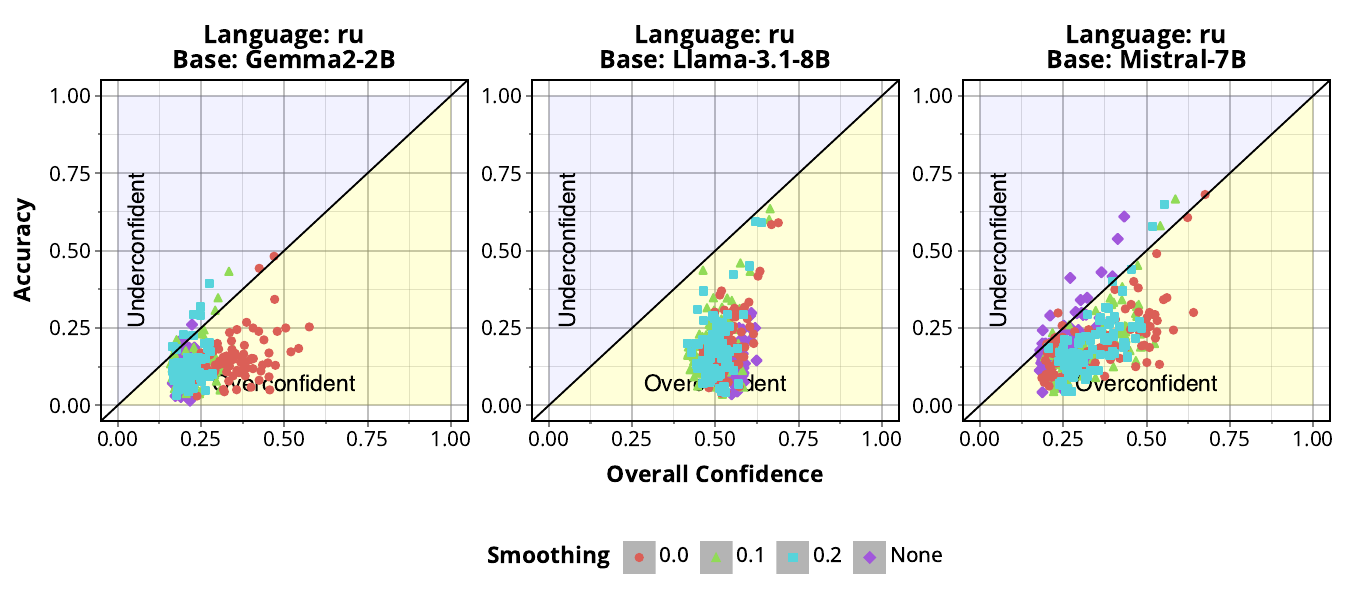}}
\caption{Reliability diagrams for the \textbf{\texttt{MMLU-ProX}} dataset for the \texttt{ru} language after instruction-tuning on the \textbf{\texttt{Tulu3Mixture}} dataset.}\label{fig:mmluprox-Tulu3Mixture-ru}\end{figure}
\begin{figure}[h!]\centering\resizebox{\linewidth}{!}{\includegraphics[width=\linewidth]{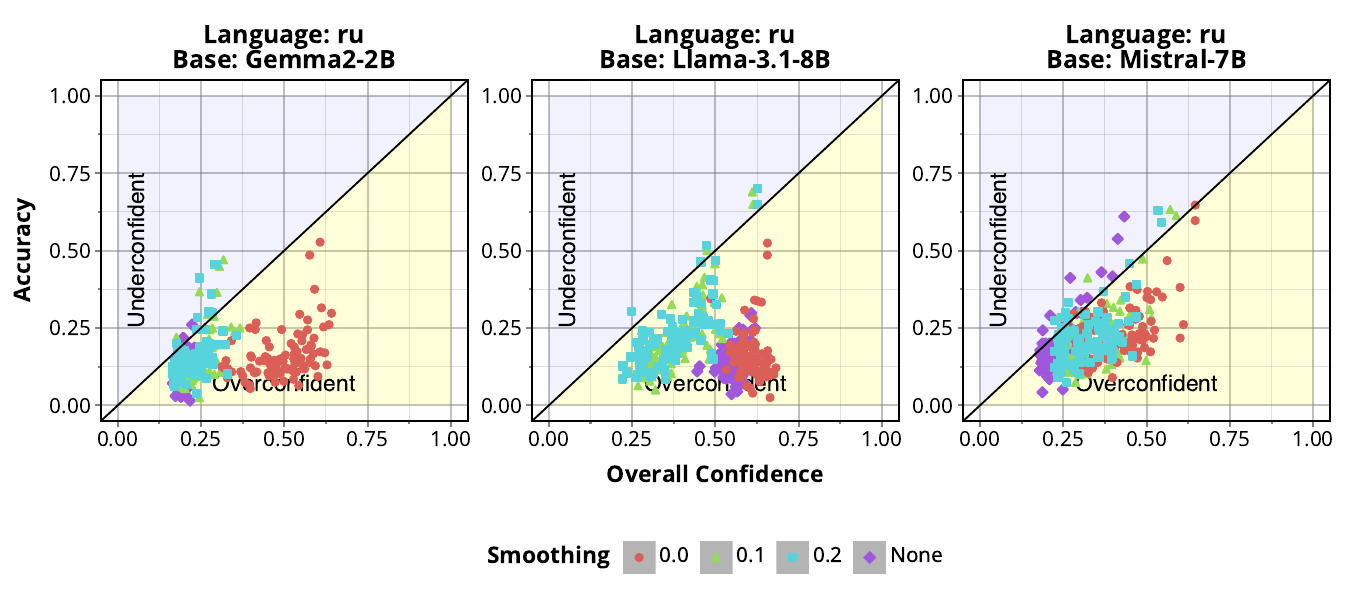}}
\caption{Reliability diagrams for the \textbf{\texttt{MMLU-ProX}} dataset for the \texttt{ru} language after instruction-tuning on the \textbf{\texttt{OpenHermes}} dataset.}\label{fig:mmluprox-OpenHermes-ru}\end{figure}

\begin{figure}[h!]\centering\resizebox{\linewidth}{!}{\includegraphics[width=\linewidth]{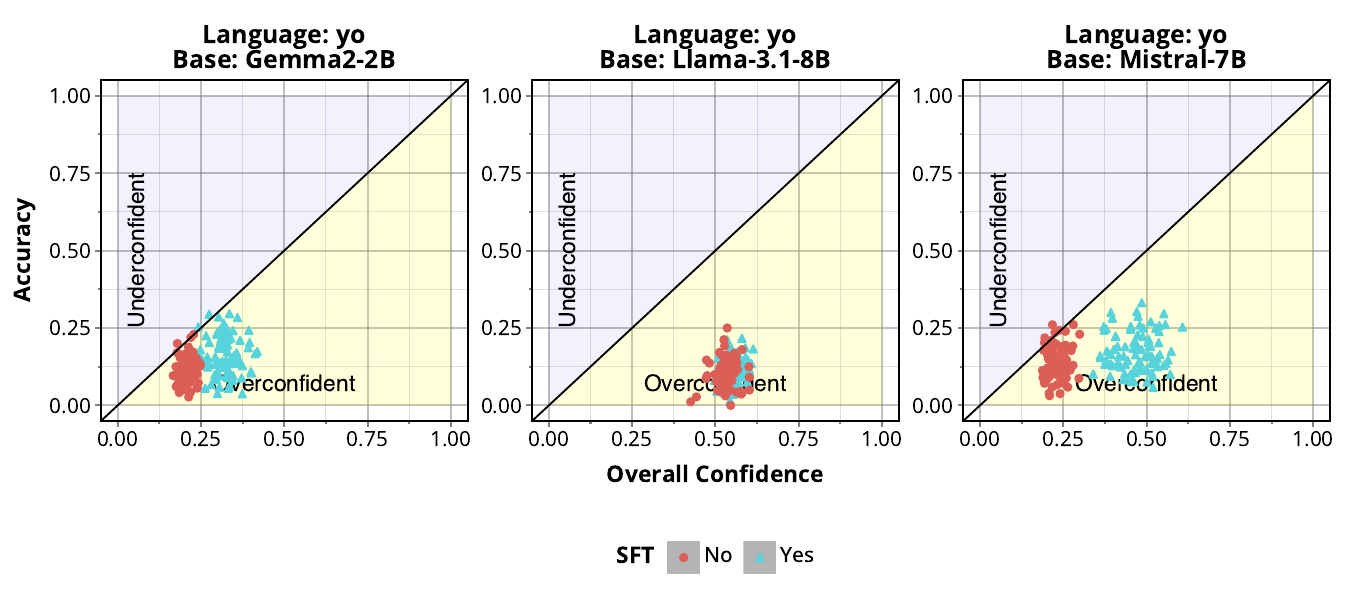}}
\caption{Reliability diagrams for the \textbf{\texttt{MMLU-ProX}} dataset for the \texttt{yo} language.}\label{fig:mmluprox-base-yo}\end{figure}
\begin{figure}[h!]\centering\resizebox{\linewidth}{!}{\includegraphics[width=\linewidth]{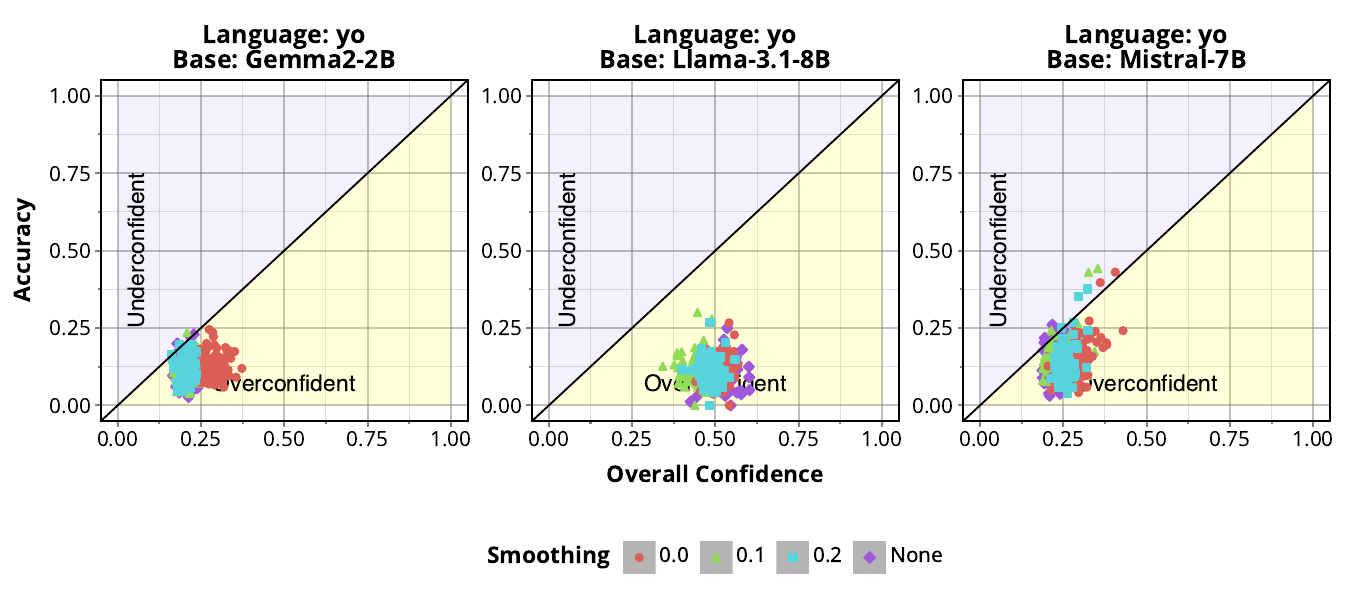}}
\caption{Reliability diagrams for the \textbf{\texttt{MMLU-ProX}} dataset for the \texttt{yo} language after instruction-tuning on the \textbf{\texttt{Tulu3Mixture}} dataset.}\label{fig:mmluprox-Tulu3Mixture-yo}\end{figure}
\begin{figure}[h!]\centering\resizebox{\linewidth}{!}{\includegraphics[width=\linewidth]{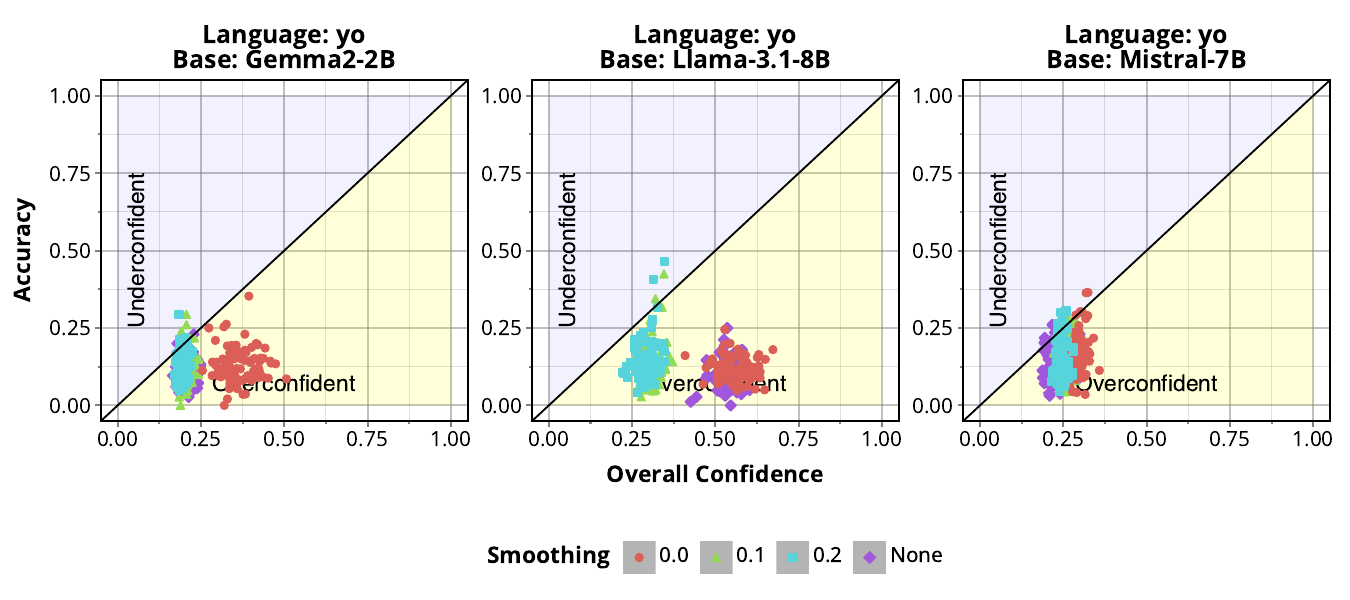}}
\caption{Reliability diagrams for the \textbf{\texttt{MMLU-ProX}} dataset for the \texttt{yo} language after instruction-tuning on the \textbf{\texttt{OpenHermes}} dataset.}\label{fig:mmluprox-OpenHermes-yo}\end{figure}

\begin{figure}[h!]\centering\resizebox{\linewidth}{!}{\includegraphics[width=\linewidth]{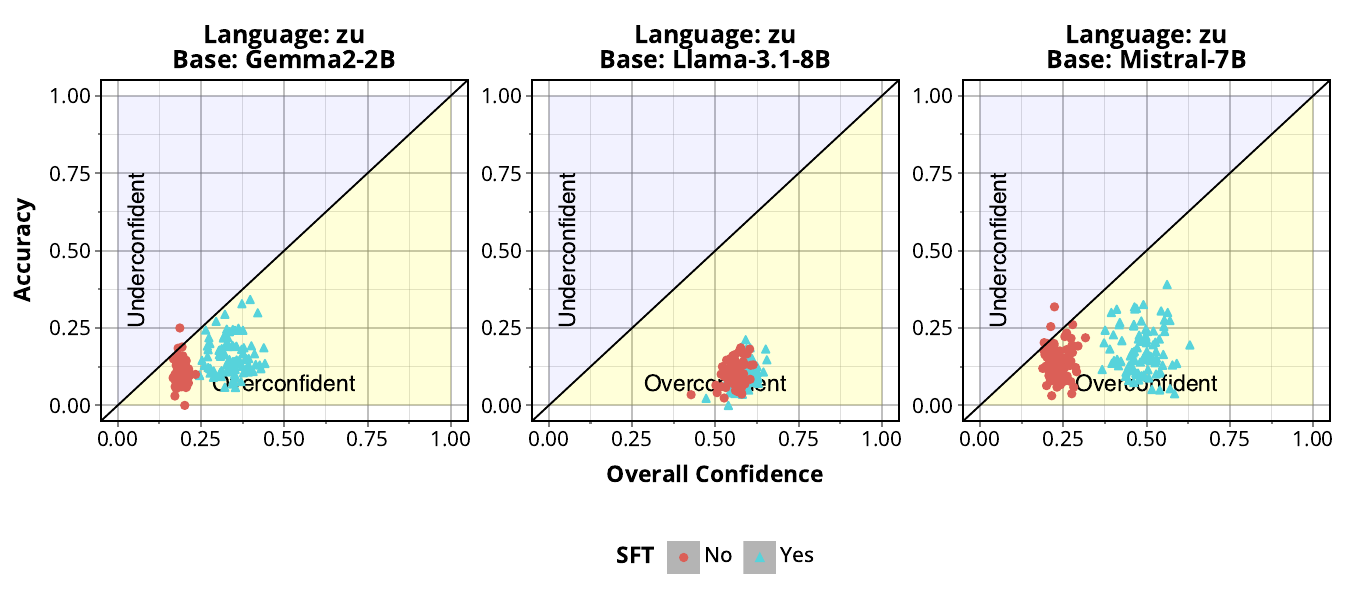}}
\caption{Reliability diagrams for the \textbf{\texttt{MMLU-ProX}} dataset for the \texttt{zu} language.}\label{fig:mmluprox-base-zu}\end{figure}
\begin{figure}[h!]\centering\resizebox{\linewidth}{!}{\includegraphics[width=\linewidth]{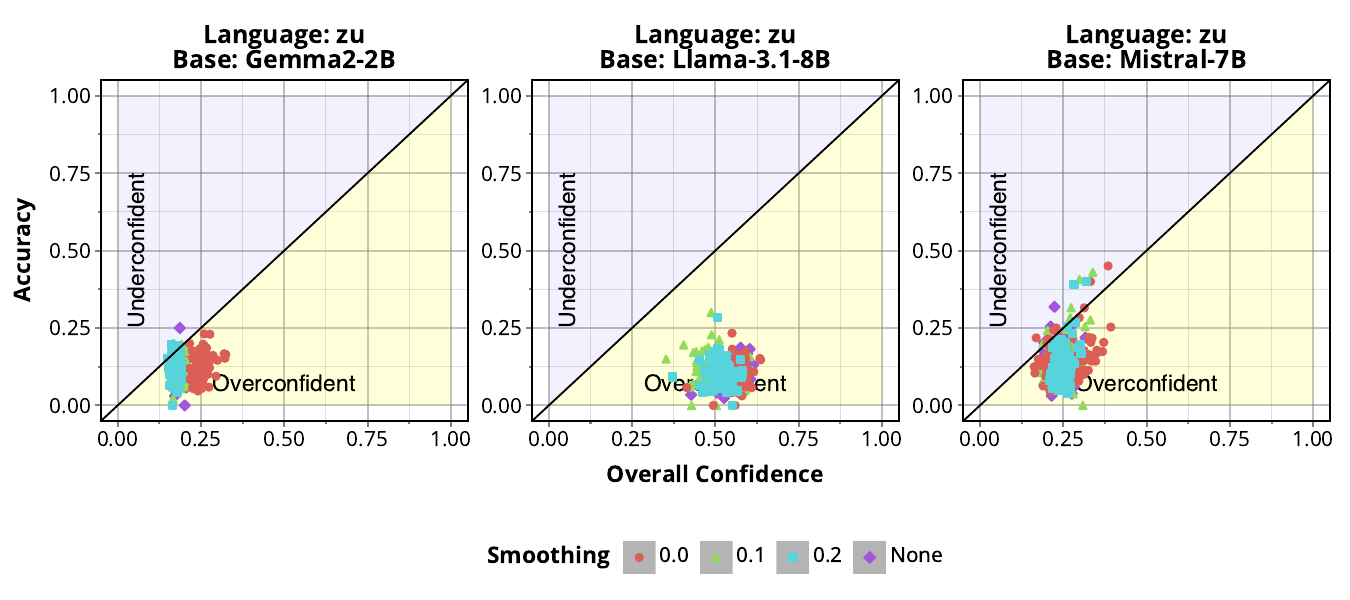}}
\caption{Reliability diagrams for the \textbf{\texttt{MMLU-ProX}} dataset for the \texttt{zu} language after instruction-tuning on the \textbf{\texttt{Tulu3Mixture}} dataset.}\label{fig:mmluprox-Tulu3Mixture-zu}\end{figure}
\begin{figure}[h!]\centering\resizebox{\linewidth}{!}{\includegraphics[width=\linewidth]{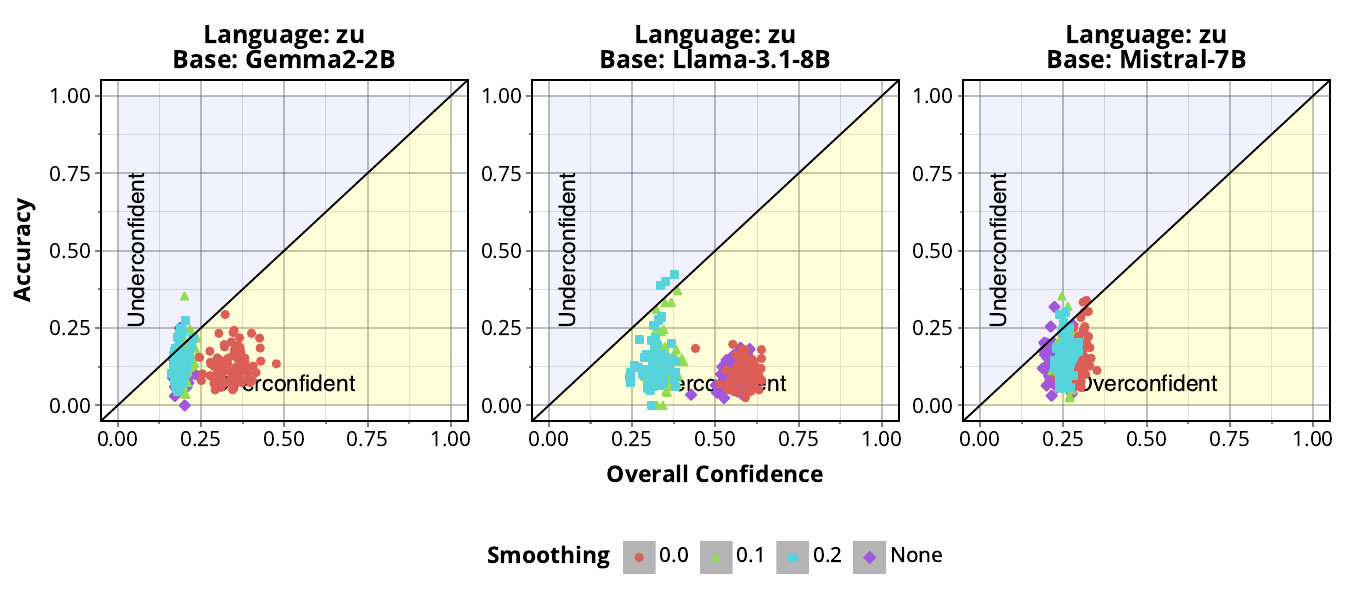}}
\caption{Reliability diagrams for the \textbf{\texttt{MMLU-ProX}} dataset for the \texttt{zu} language after instruction-tuning on the \textbf{\texttt{OpenHermes}} dataset.}\label{fig:mmluprox-OpenHermes-zu}\end{figure}

\begin{figure}[h!]\centering\resizebox{\linewidth}{!}{\includegraphics[width=\linewidth]{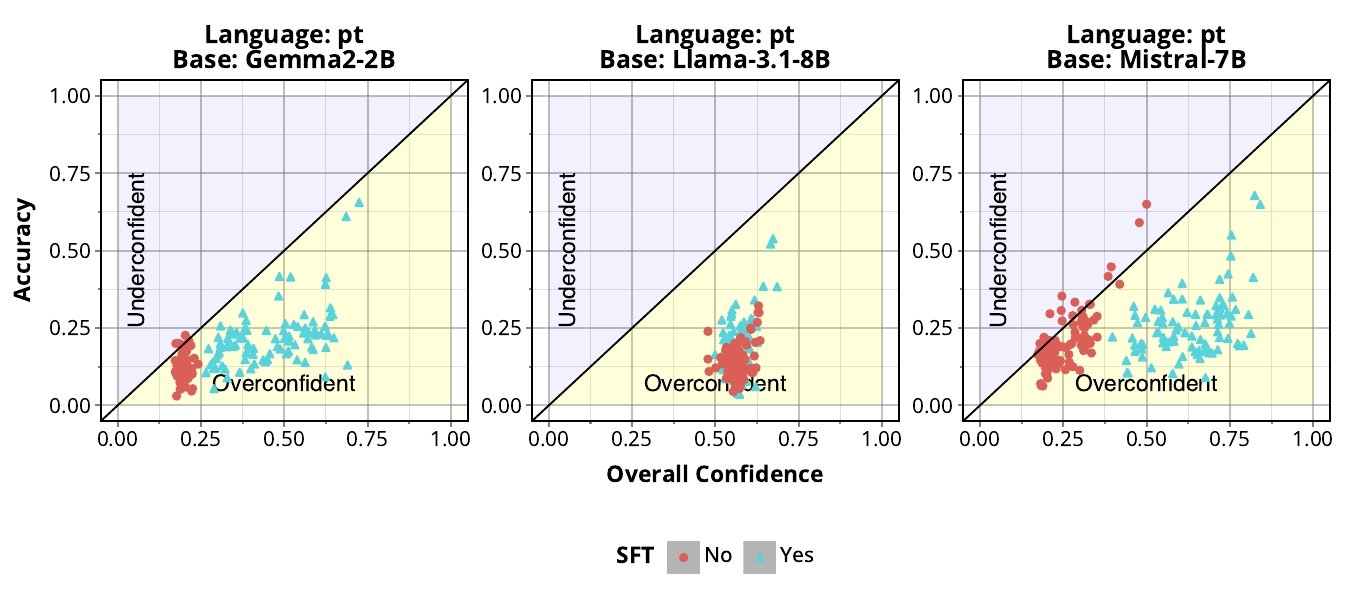}}
\caption{Reliability diagrams for the \textbf{\texttt{MMLU-ProX}} dataset for the \texttt{pt} language.}\label{fig:mmluprox-base-pt}\end{figure}
\begin{figure}[h!]\centering\resizebox{\linewidth}{!}{\includegraphics[width=\linewidth]{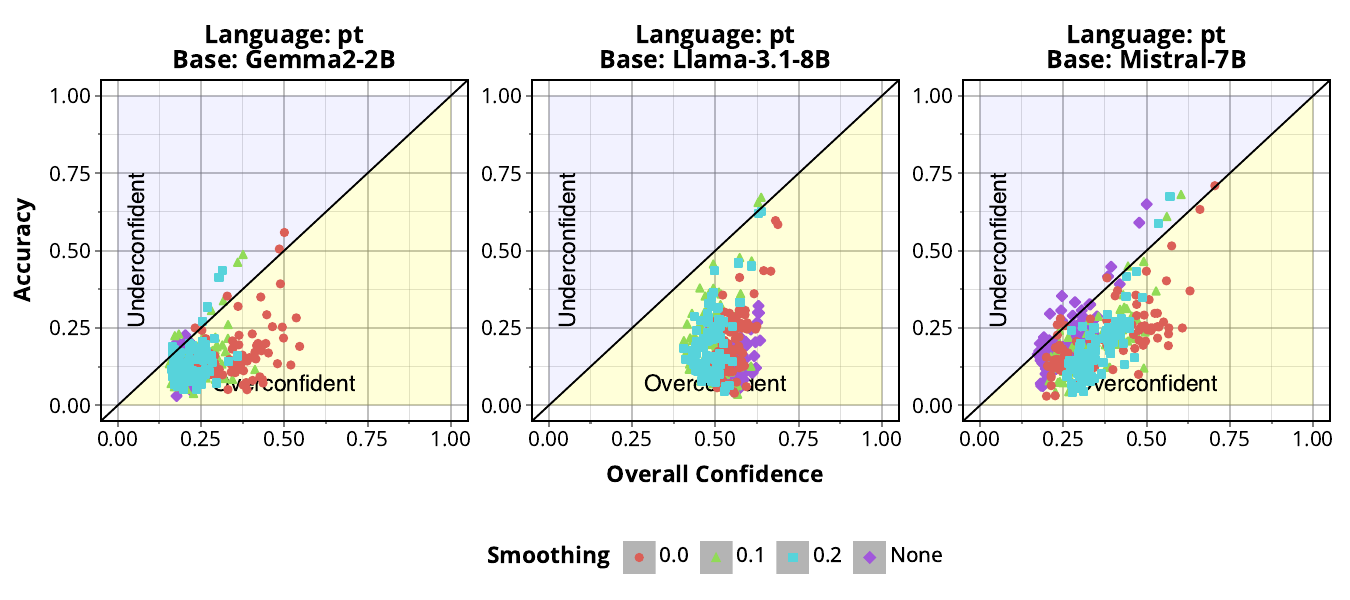}}
\caption{Reliability diagrams for the \textbf{\texttt{MMLU-ProX}} dataset for the \texttt{pt} language after instruction-tuning on the \textbf{\texttt{Tulu3Mixture}} dataset.}\label{fig:mmluprox-Tulu3Mixture-pt}\end{figure}
\begin{figure}[h!]\centering\resizebox{\linewidth}{!}{\includegraphics[width=\linewidth]{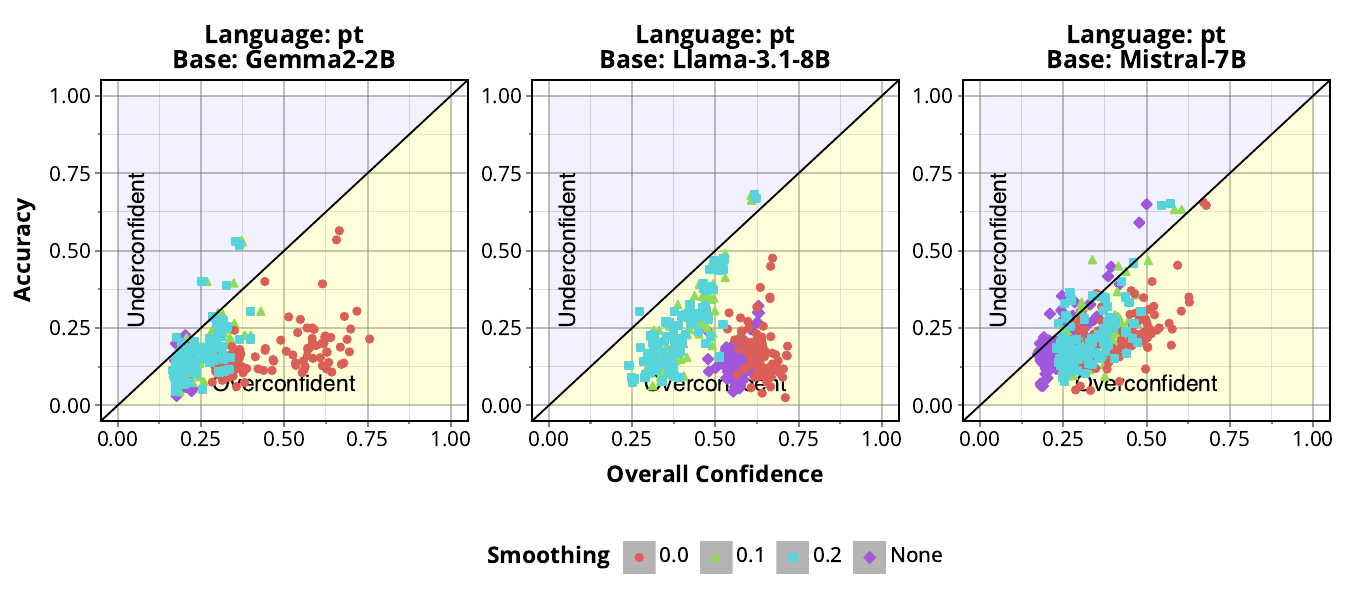}}
\caption{Reliability diagrams for the \textbf{\texttt{MMLU-ProX}} dataset for the \texttt{pt} language after instruction-tuning on the \textbf{\texttt{OpenHermes}} dataset.}\label{fig:mmluprox-OpenHermes-pt}\end{figure}

\begin{figure}[h!]\centering\resizebox{\linewidth}{!}{\includegraphics[width=\linewidth]{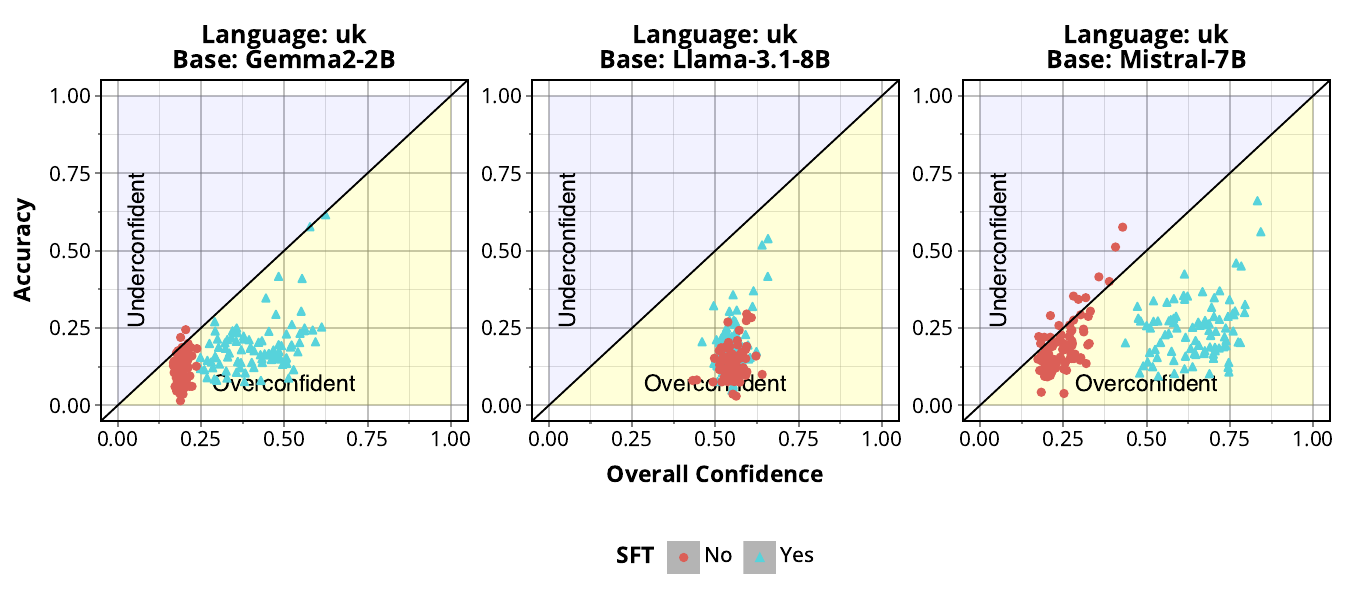}}
\caption{Reliability diagrams for the \textbf{\texttt{MMLU-ProX}} dataset for the \texttt{uk} language.}\label{fig:mmluprox-base-uk}\end{figure}
\begin{figure}[h!]\centering\resizebox{\linewidth}{!}{\includegraphics[width=\linewidth]{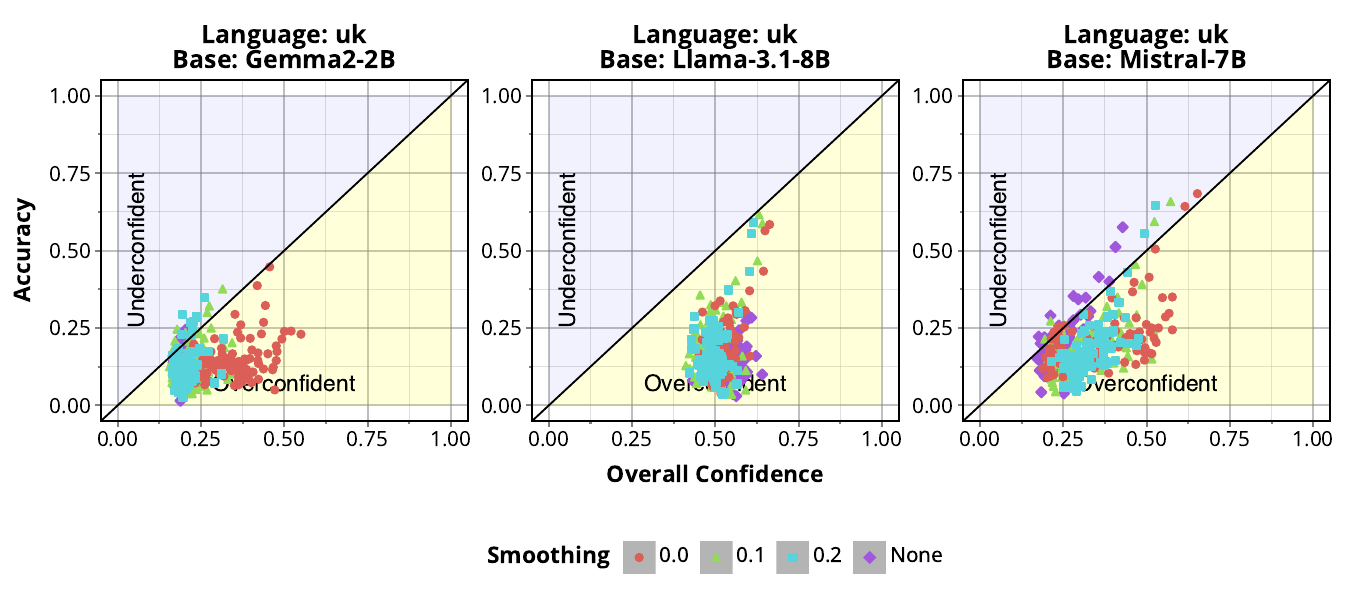}}
\caption{Reliability diagrams for the \textbf{\texttt{MMLU-ProX}} dataset for the \texttt{uk} language after instruction-tuning on the \textbf{\texttt{Tulu3Mixture}} dataset.}\label{fig:mmluprox-Tulu3Mixture-uk}\end{figure}
\begin{figure}[h!]\centering\resizebox{\linewidth}{!}{\includegraphics[width=\linewidth]{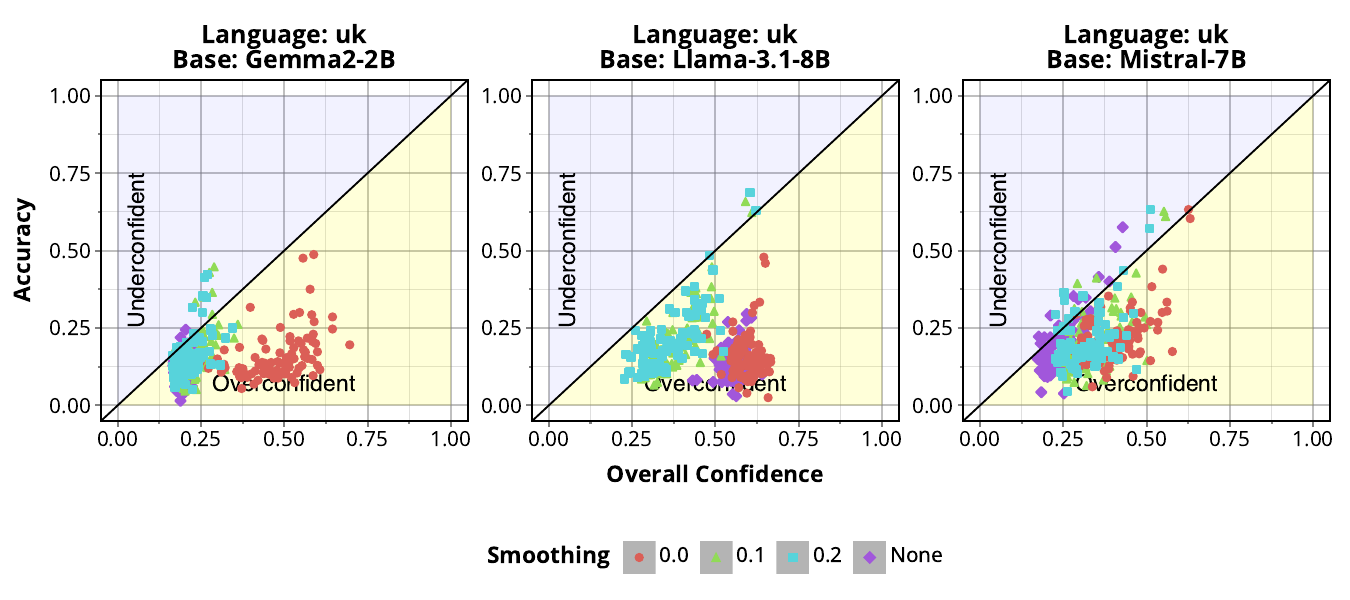}}
\caption{Reliability diagrams for the \textbf{\texttt{MMLU-ProX}} dataset for the \texttt{uk} language after instruction-tuning on the \textbf{\texttt{OpenHermes}} dataset.}\label{fig:mmluprox-OpenHermes-uk}\end{figure}

\begin{figure}[h!]\centering\resizebox{\linewidth}{!}{\includegraphics[width=\linewidth]{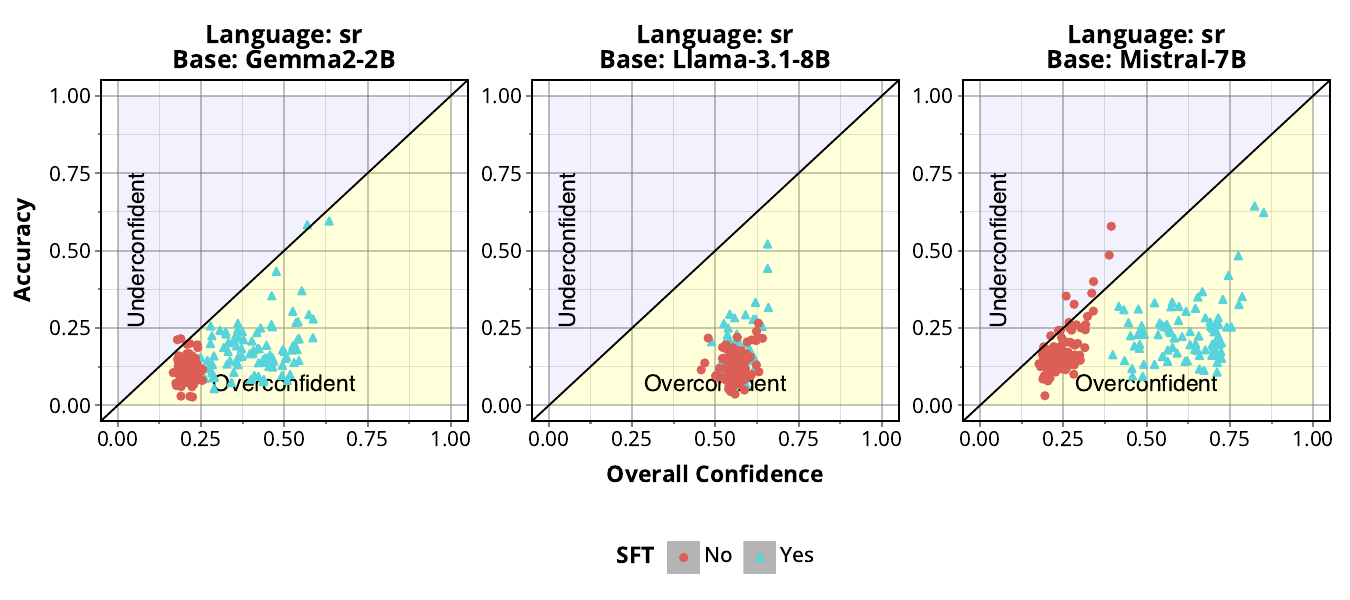}}
\caption{Reliability diagrams for the \textbf{\texttt{MMLU-ProX}} dataset for the \texttt{sr} language.}\label{fig:mmluprox-base-sr}\end{figure}
\begin{figure}[h!]\centering\resizebox{\linewidth}{!}{\includegraphics[width=\linewidth]{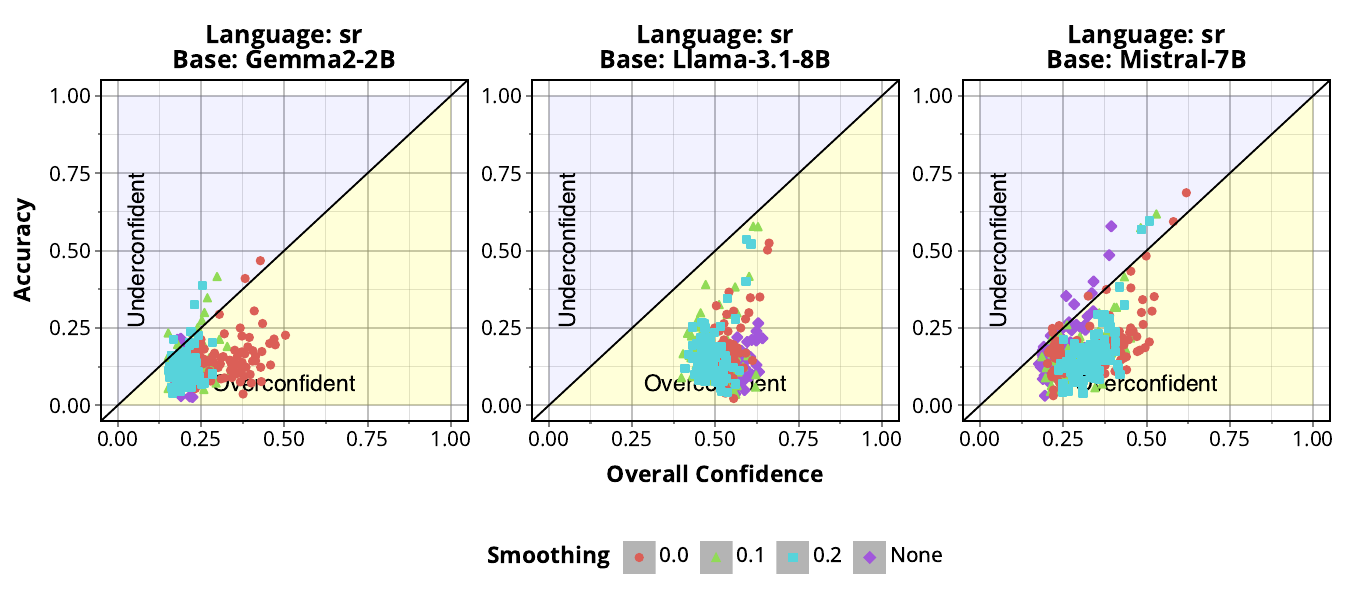}}
\caption{Reliability diagrams for the \textbf{\texttt{MMLU-ProX}} dataset for the \texttt{sr} language after instruction-tuning on the \textbf{\texttt{Tulu3Mixture}} dataset.}\label{fig:mmluprox-Tulu3Mixture-sr}\end{figure}
\begin{figure}[h!]\centering\resizebox{\linewidth}{!}{\includegraphics[width=\linewidth]{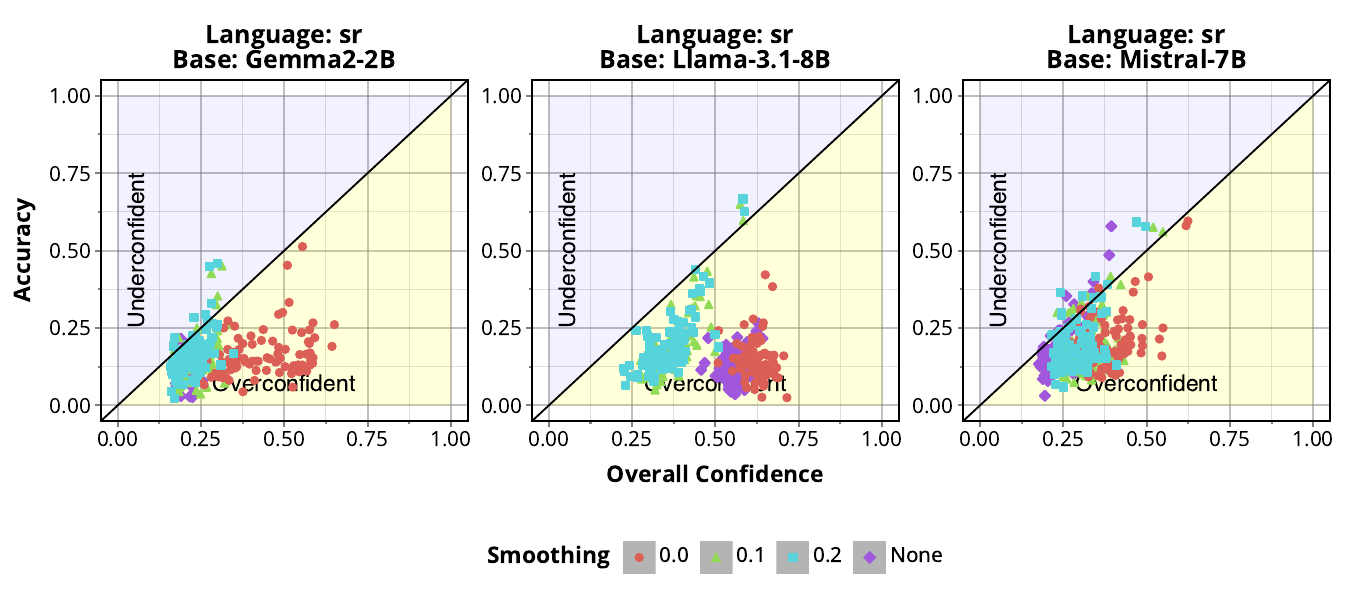}}
\caption{Reliability diagrams for the \textbf{\texttt{MMLU-ProX}} dataset for the \texttt{sr} language after instruction-tuning on the \textbf{\texttt{OpenHermes}} dataset.}\label{fig:mmluprox-OpenHermes-sr}\end{figure}

\clearpage
\begin{figure}[h!]\centering\resizebox{\linewidth}{!}{\includegraphics[width=\linewidth]{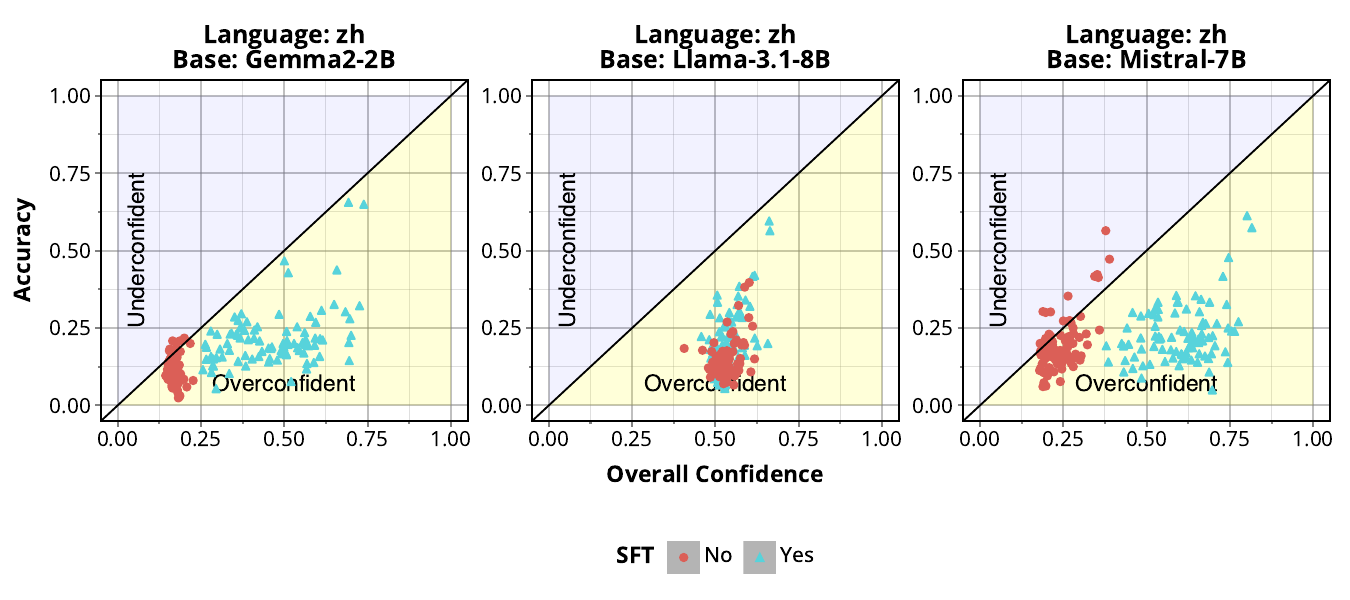}}
\caption{Reliability diagrams for the \textbf{\texttt{MMLU-ProX}} dataset for the \texttt{zh} language.}\label{fig:mmluprox-base-zh}\end{figure}
\begin{figure}[h!]\centering\resizebox{\linewidth}{!}{\includegraphics[width=\linewidth]{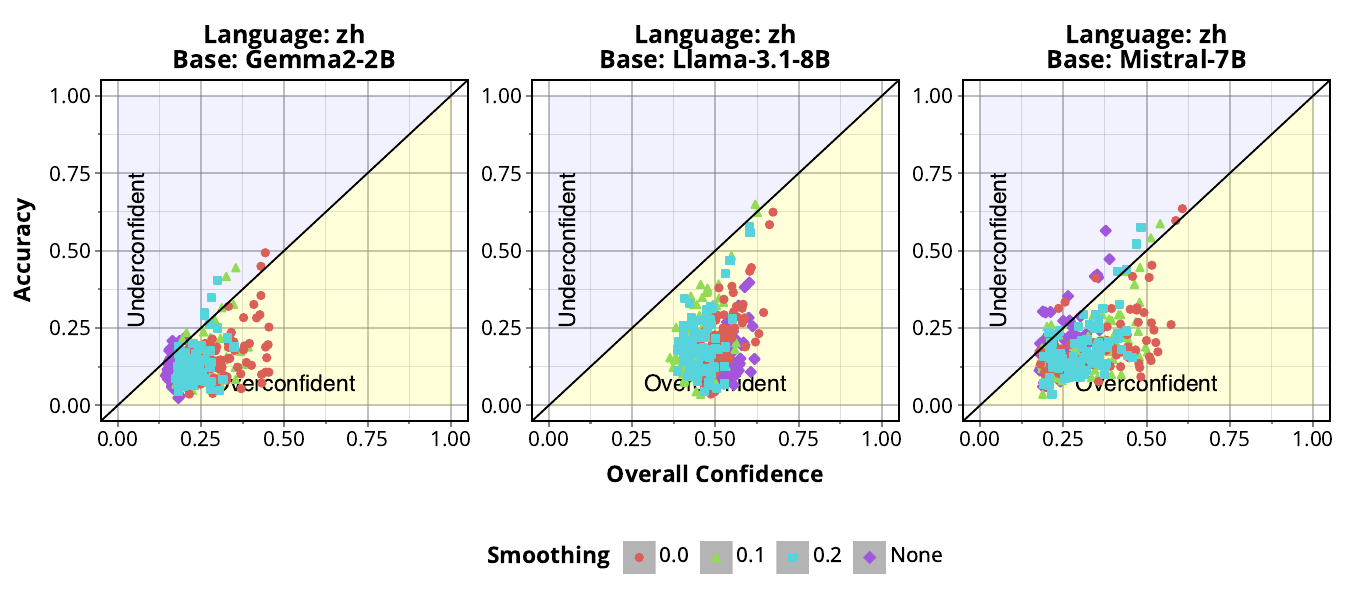}}
\caption{Reliability diagrams for the \textbf{\texttt{MMLU-ProX}} dataset for the \texttt{zh} language after instruction-tuning on the \textbf{\texttt{Tulu3Mixture}} dataset.}\label{fig:mmluprox-Tulu3Mixture-zh}\end{figure}
\begin{figure}[h!]\centering\resizebox{\linewidth}{!}{\includegraphics[width=\linewidth]{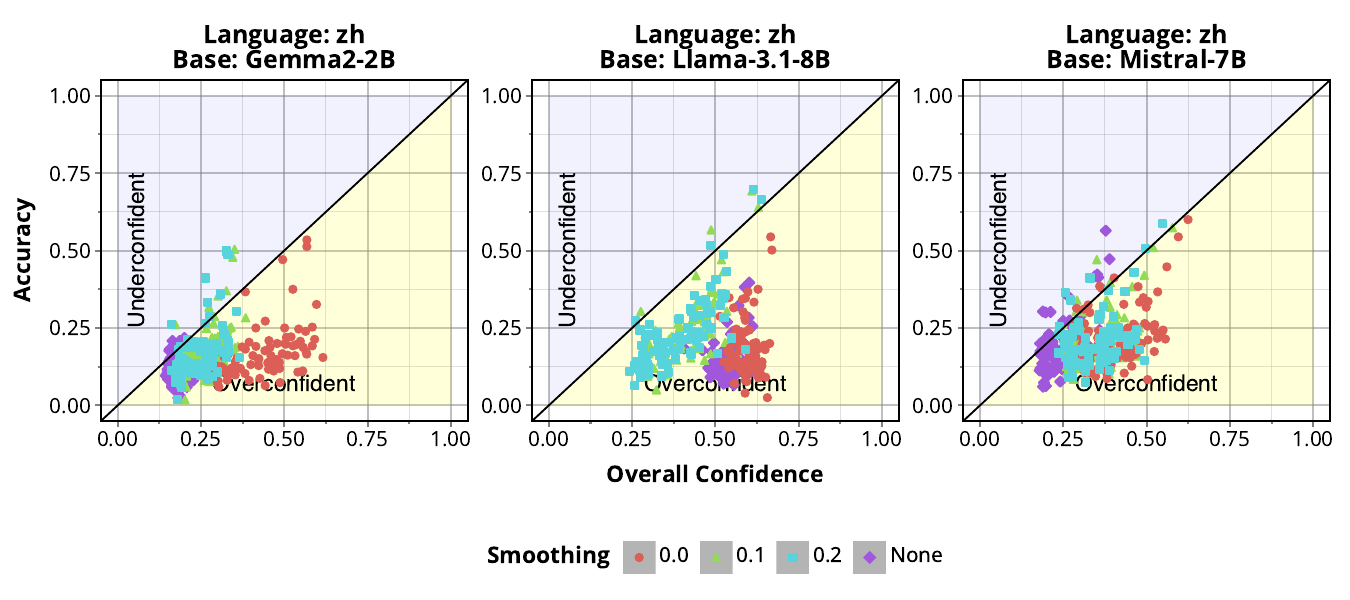}}
\caption{Reliability diagrams for the \textbf{\texttt{MMLU-ProX}} dataset for the \texttt{zh} language after instruction-tuning on the \textbf{\texttt{OpenHermes}} dataset.}\label{fig:mmluprox-OpenHermes-zh}\end{figure}

\begin{figure}[h!]\centering\resizebox{\linewidth}{!}{\includegraphics[width=\linewidth]{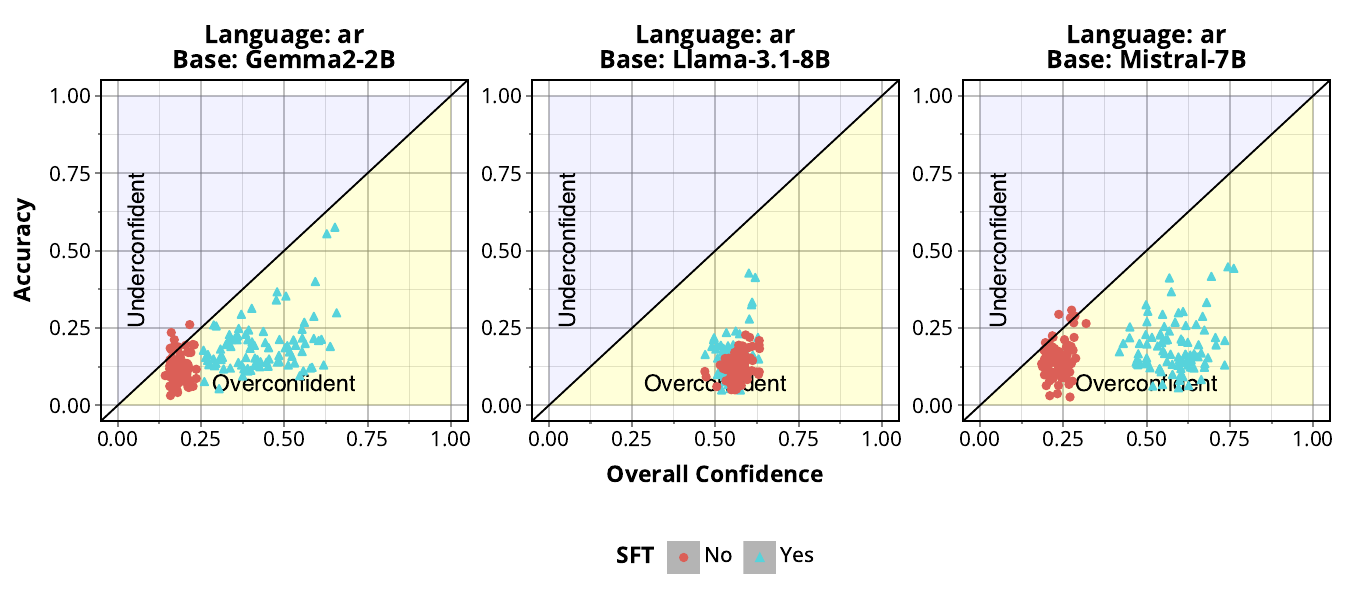}}
\caption{Reliability diagrams for the \textbf{\texttt{MMLU-ProX}} dataset for the \texttt{ar} language.}\label{fig:mmluprox-base-ar}\end{figure}
\begin{figure}[h!]\centering\resizebox{\linewidth}{!}{\includegraphics[width=\linewidth]{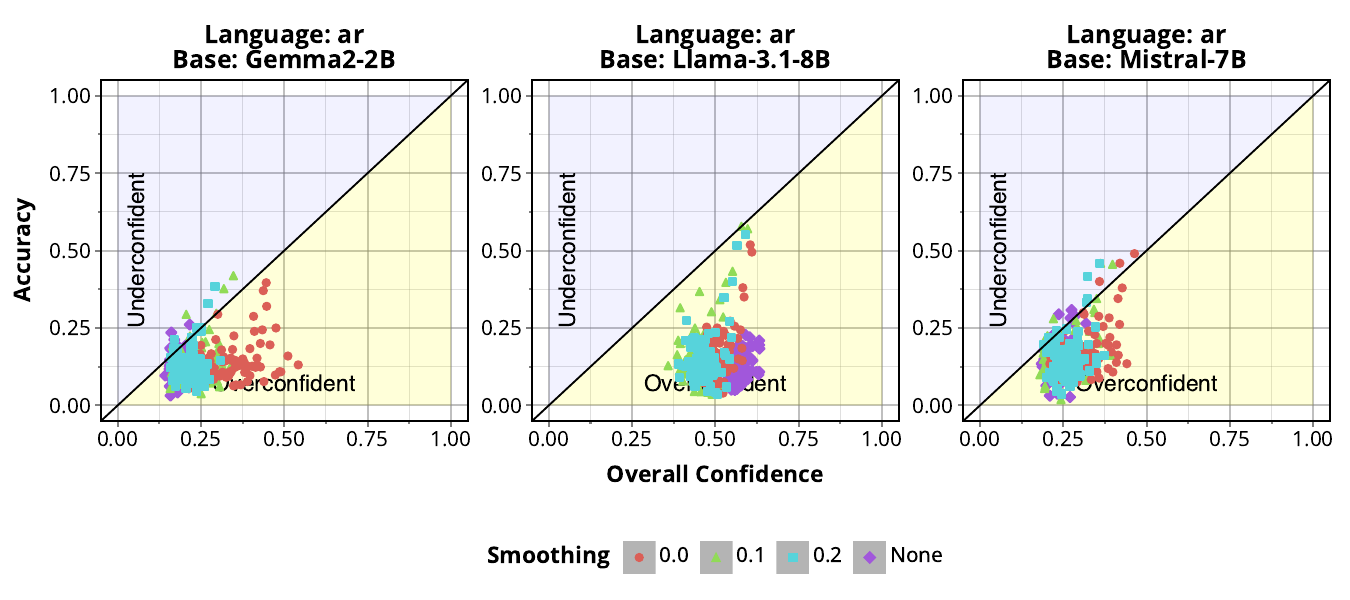}}
\caption{Reliability diagrams for the \textbf{\texttt{MMLU-ProX}} dataset for the \texttt{ar} language after instruction-tuning on the \textbf{\texttt{Tulu3Mixture}} dataset.}\label{fig:mmluprox-Tulu3Mixture-ar}\end{figure}
\begin{figure}[h!]\centering\resizebox{\linewidth}{!}{\includegraphics[width=\linewidth]{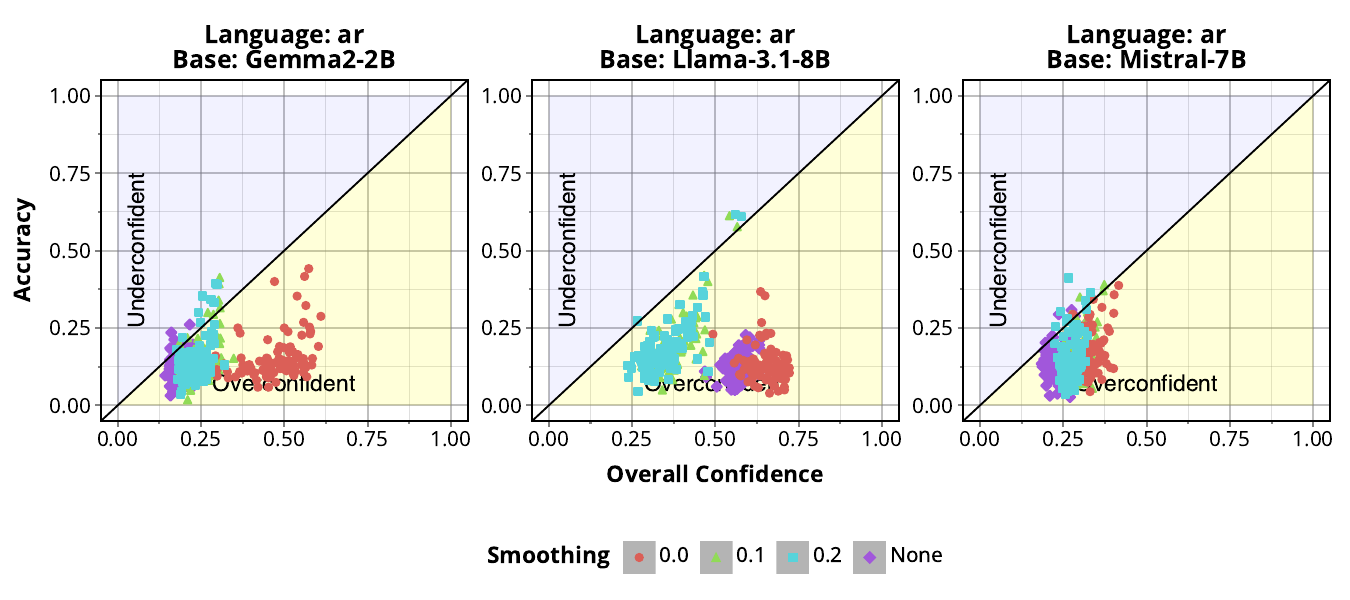}}
\caption{Reliability diagrams for the \textbf{\texttt{MMLU-ProX}} dataset for the \texttt{ar} language after instruction-tuning on the \textbf{\texttt{OpenHermes}} dataset.}\label{fig:mmluprox-OpenHermes-ar}\end{figure}

\begin{figure}[h!]\centering\resizebox{\linewidth}{!}{\includegraphics[width=\linewidth]{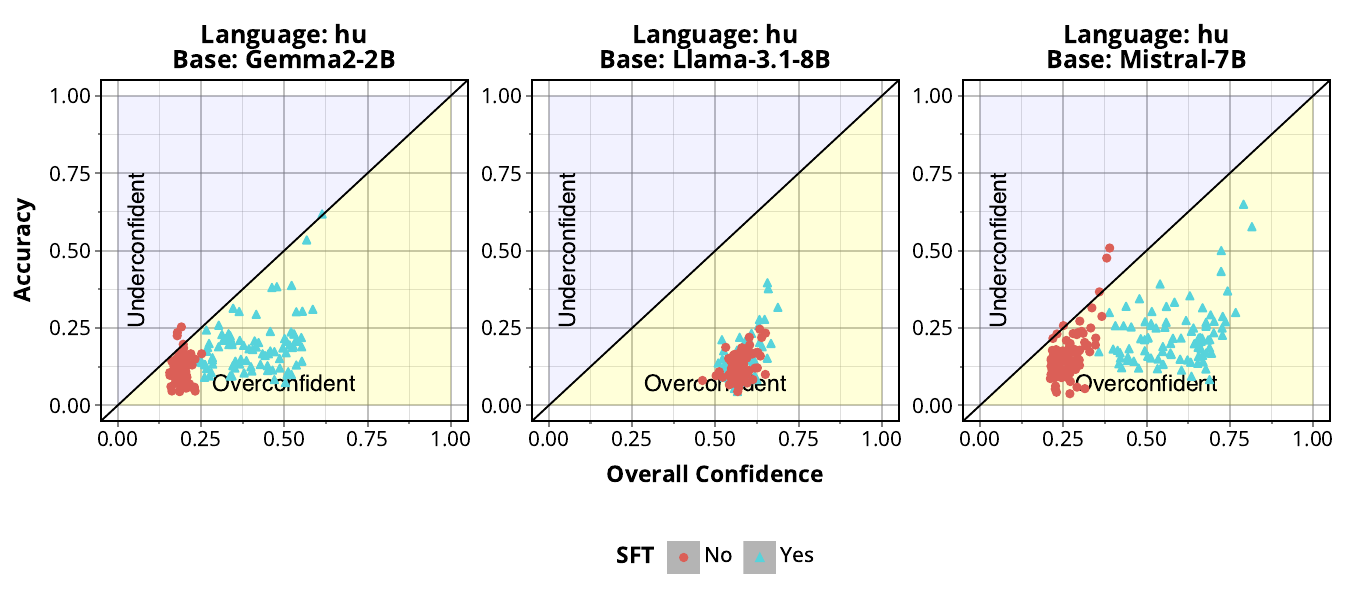}}
\caption{Reliability diagrams for the \textbf{\texttt{MMLU-ProX}} dataset for the \texttt{hu} language.}\label{fig:mmluprox-base-hu}\end{figure}
\begin{figure}[h!]\centering\resizebox{\linewidth}{!}{\includegraphics[width=\linewidth]{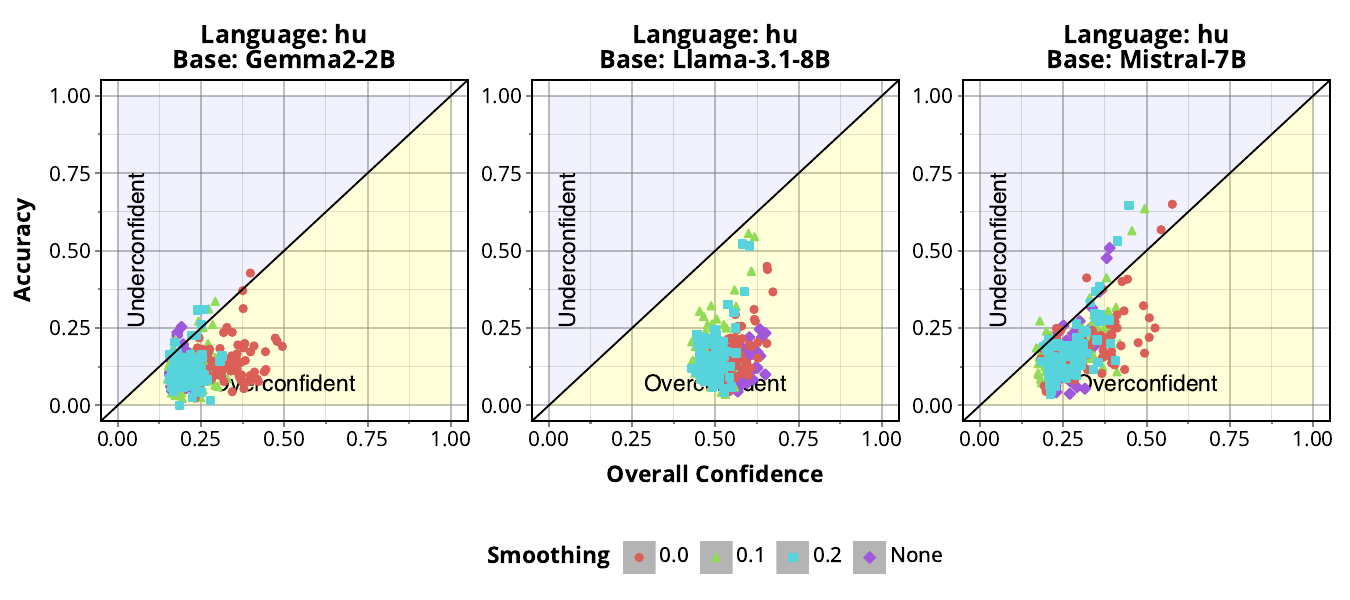}}
\caption{Reliability diagrams for the \textbf{\texttt{MMLU-ProX}} dataset for the \texttt{hu} language after instruction-tuning on the \textbf{\texttt{Tulu3Mixture}} dataset.}\label{fig:mmluprox-Tulu3Mixture-hu}\end{figure}
\begin{figure}[h!]\centering\resizebox{\linewidth}{!}{\includegraphics[width=\linewidth]{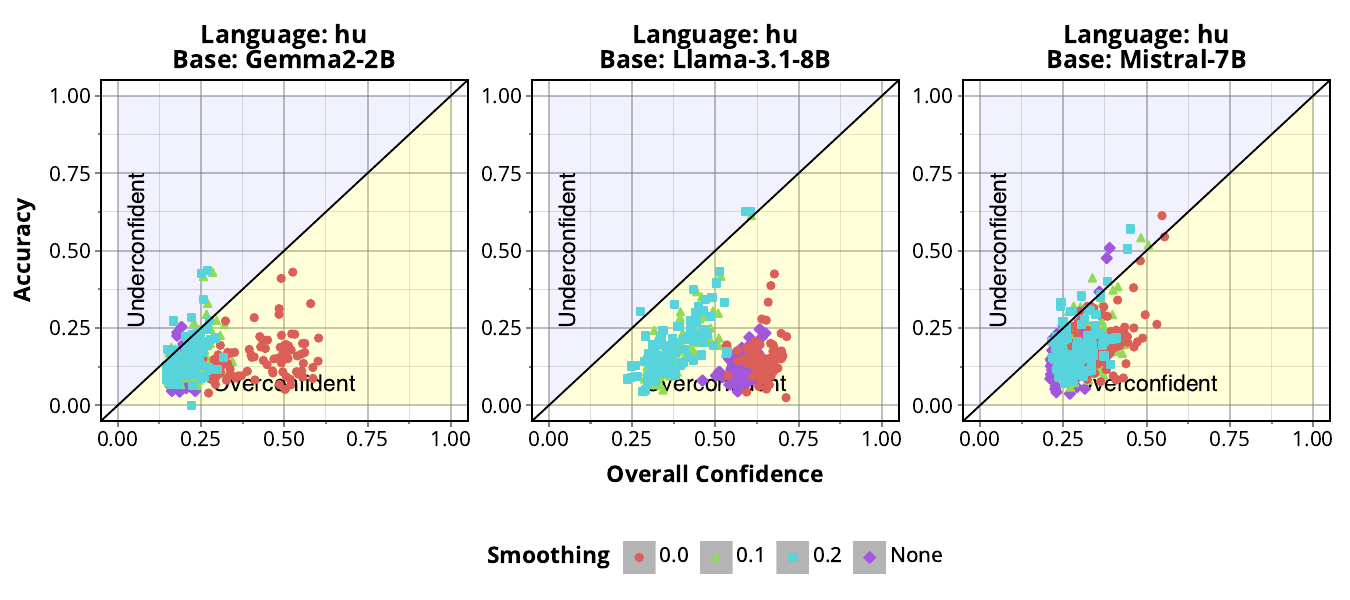}}
\caption{Reliability diagrams for the \textbf{\texttt{MMLU-ProX}} dataset for the \texttt{hu} language after instruction-tuning on the \textbf{\texttt{OpenHermes}} dataset.}\label{fig:mmluprox-OpenHermes-hu}\end{figure}

\begin{figure}[h!]\centering\resizebox{\linewidth}{!}{\includegraphics[width=\linewidth]{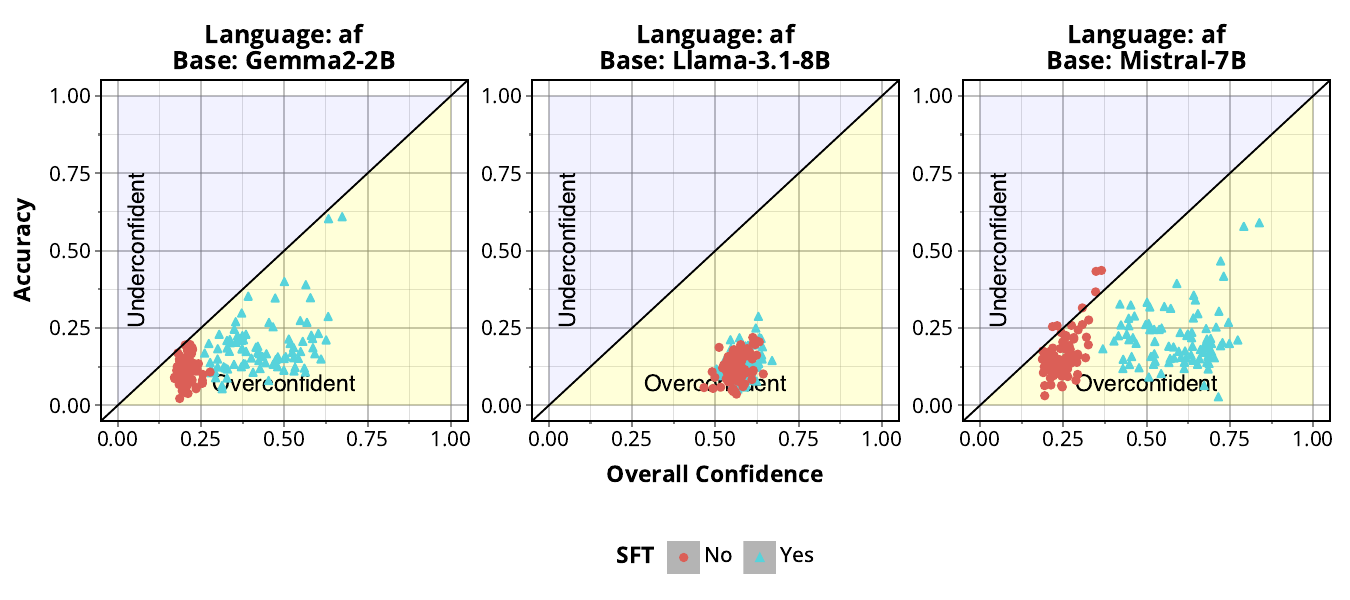}}
\caption{Reliability diagrams for the \textbf{\texttt{MMLU-ProX}} dataset for the \texttt{af} language.}\label{fig:mmluprox-base-af}\end{figure}
\begin{figure}[h!]\centering\resizebox{\linewidth}{!}{\includegraphics[width=\linewidth]{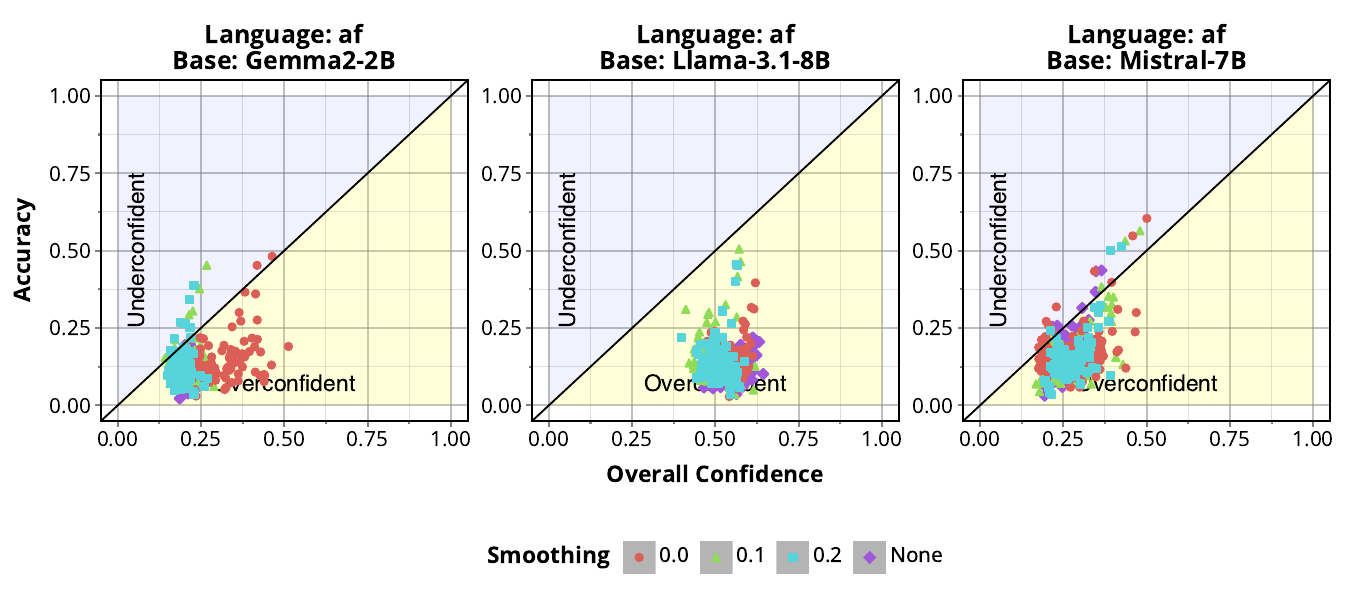}}
\caption{Reliability diagrams for the \textbf{\texttt{MMLU-ProX}} dataset for the \texttt{af} language after instruction-tuning on the \textbf{\texttt{Tulu3Mixture}} dataset.}\label{fig:mmluprox-Tulu3Mixture-af}\end{figure}
\begin{figure}[h!]\centering\resizebox{\linewidth}{!}{\includegraphics[width=\linewidth]{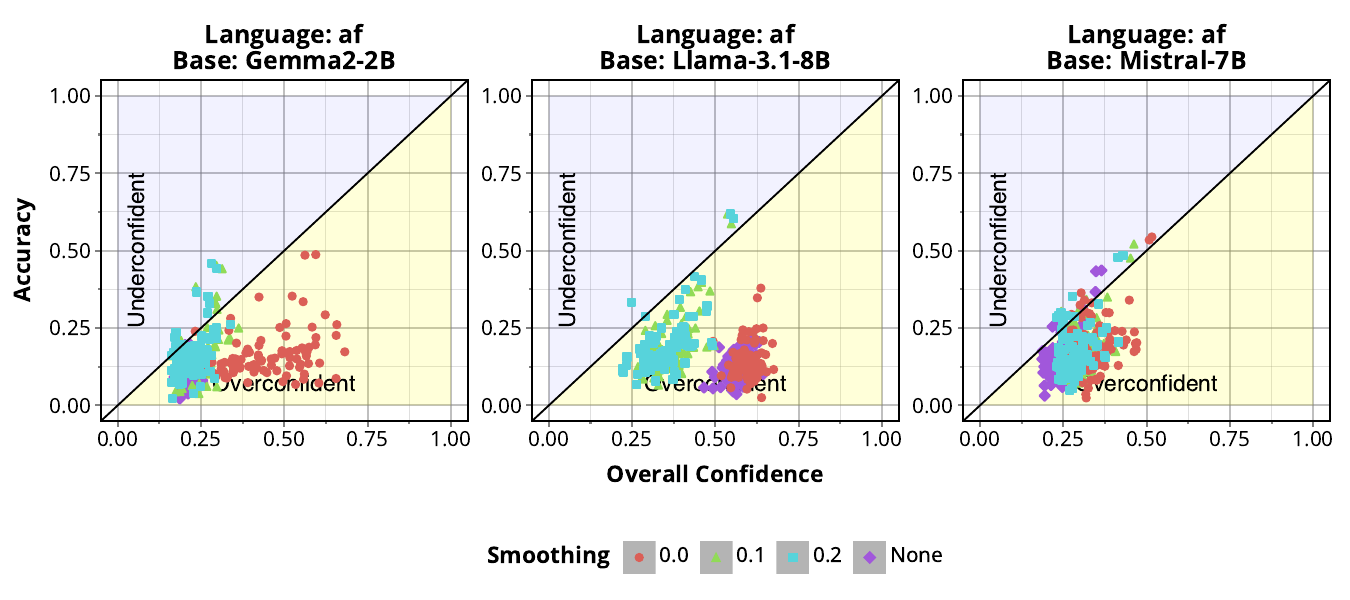}}
\caption{Reliability diagrams for the \textbf{\texttt{MMLU-ProX}} dataset for the \texttt{af} language after instruction-tuning on the \textbf{\texttt{OpenHermes}} dataset.}\label{fig:mmluprox-OpenHermes-af}\end{figure}

\begin{figure}[h!]\centering\resizebox{\linewidth}{!}{\includegraphics[width=\linewidth]{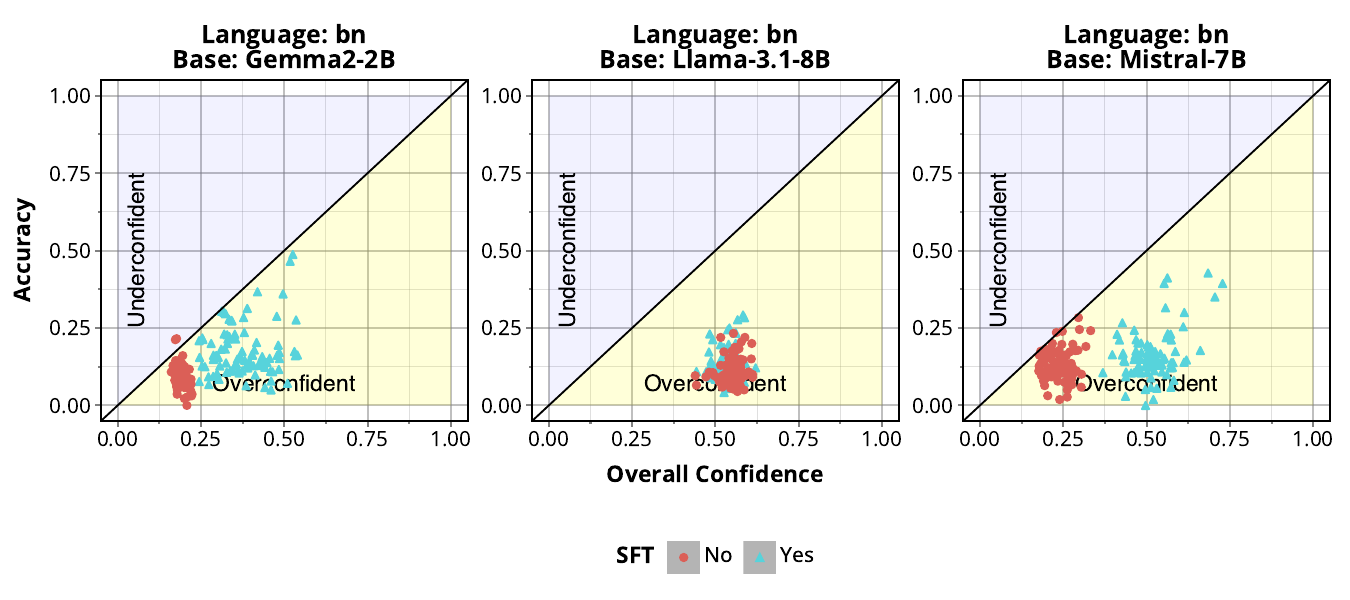}}
\caption{Reliability diagrams for the \textbf{\texttt{MMLU-ProX}} dataset for the \texttt{bn} language.}\label{fig:mmluprox-base-bn}\end{figure}
\begin{figure}[h!]\centering\resizebox{\linewidth}{!}{\includegraphics[width=\linewidth]{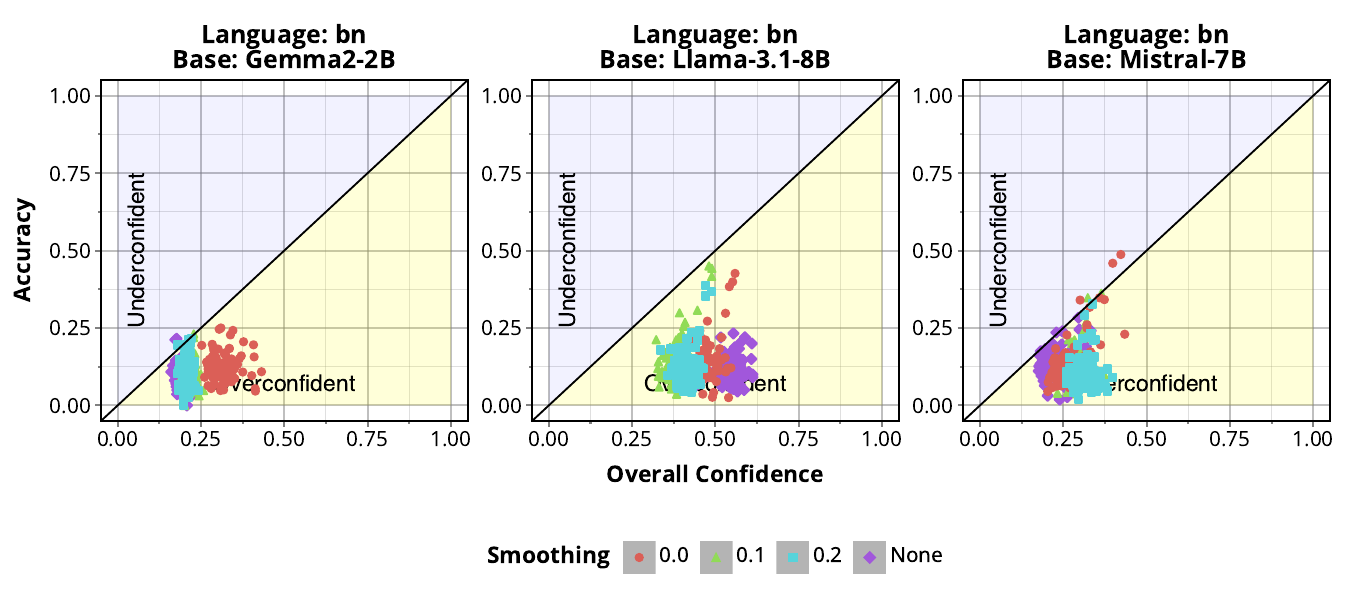}}
\caption{Reliability diagrams for the \textbf{\texttt{MMLU-ProX}} dataset for the \texttt{bn} language after instruction-tuning on the \textbf{\texttt{Tulu3Mixture}} dataset.}\label{fig:mmluprox-Tulu3Mixture-bn}\end{figure}
\begin{figure}[h!]\centering\resizebox{\linewidth}{!}{\includegraphics[width=\linewidth]{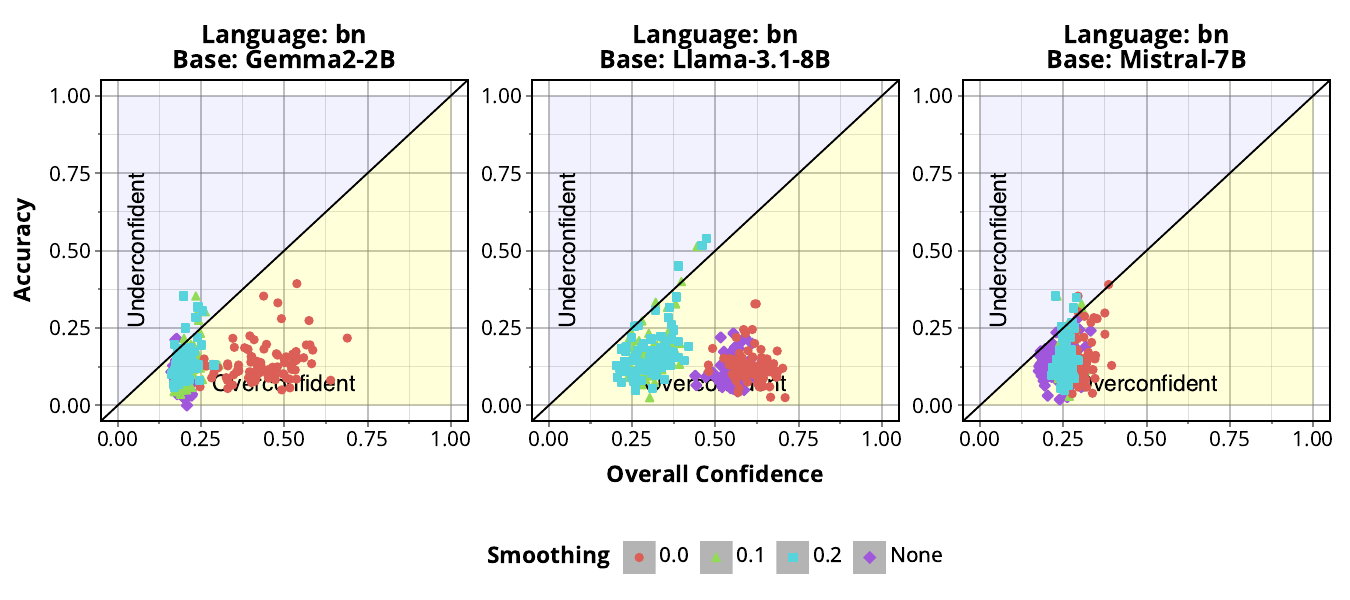}}
\caption{Reliability diagrams for the \textbf{\texttt{MMLU-ProX}} dataset for the \texttt{bn} language after instruction-tuning on the \textbf{\texttt{OpenHermes}} dataset.}\label{fig:mmluprox-OpenHermes-bn}\end{figure}

\begin{figure}[h!]\centering\resizebox{\linewidth}{!}{\includegraphics[width=\linewidth]{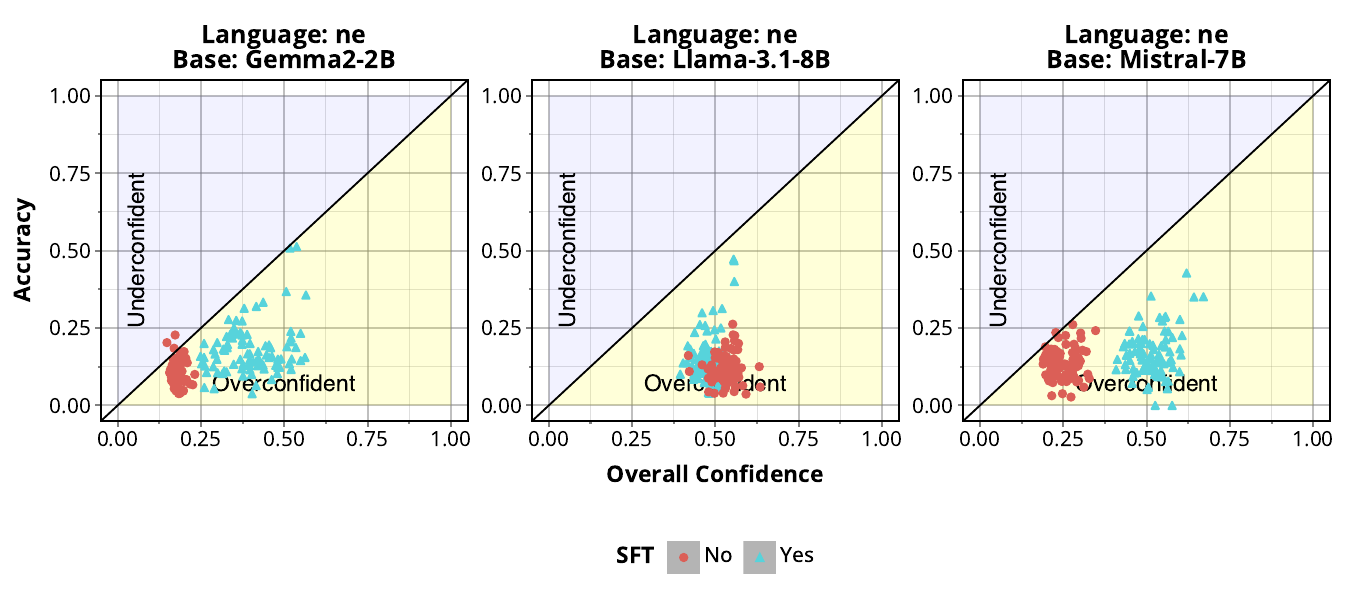}}
\caption{Reliability diagrams for the \textbf{\texttt{MMLU-ProX}} dataset for the \texttt{ne} language.}\label{fig:mmluprox-base-ne}\end{figure}
\begin{figure}[h!]\centering\resizebox{\linewidth}{!}{\includegraphics[width=\linewidth]{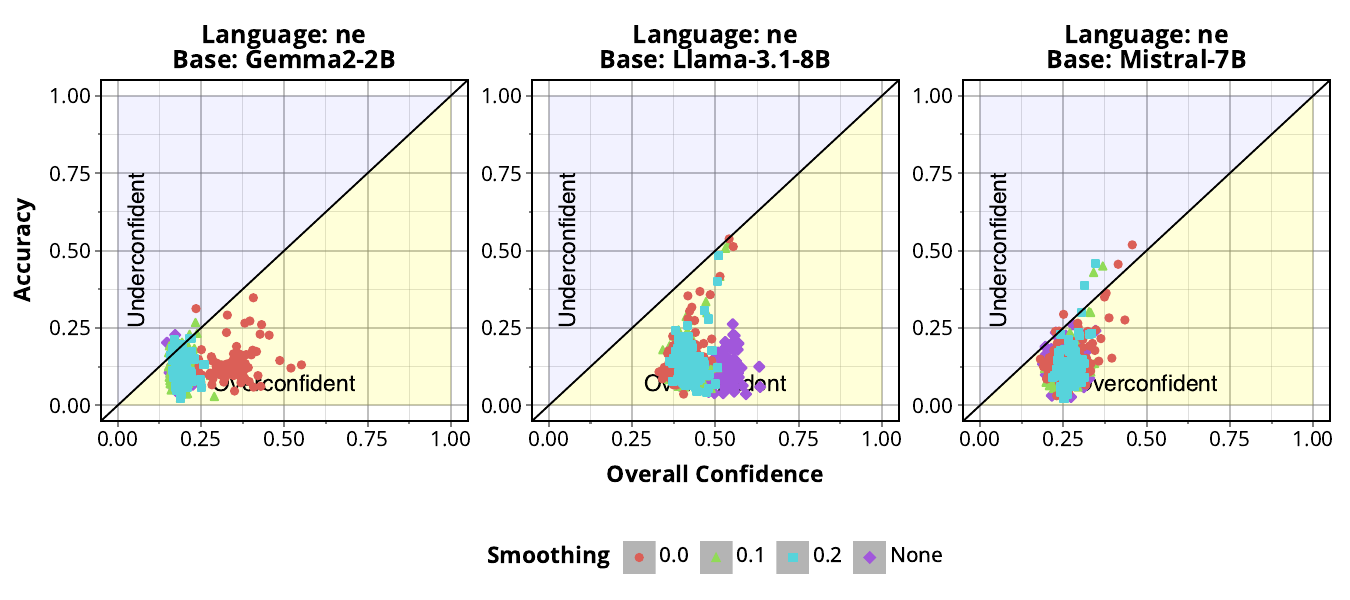}}
\caption{Reliability diagrams for the \textbf{\texttt{MMLU-ProX}} dataset for the \texttt{ne} language after instruction-tuning on the \textbf{\texttt{Tulu3Mixture}} dataset.}\label{fig:mmluprox-Tulu3Mixture-ne}\end{figure}
\begin{figure}[h!]\centering\resizebox{\linewidth}{!}{\includegraphics[width=\linewidth]{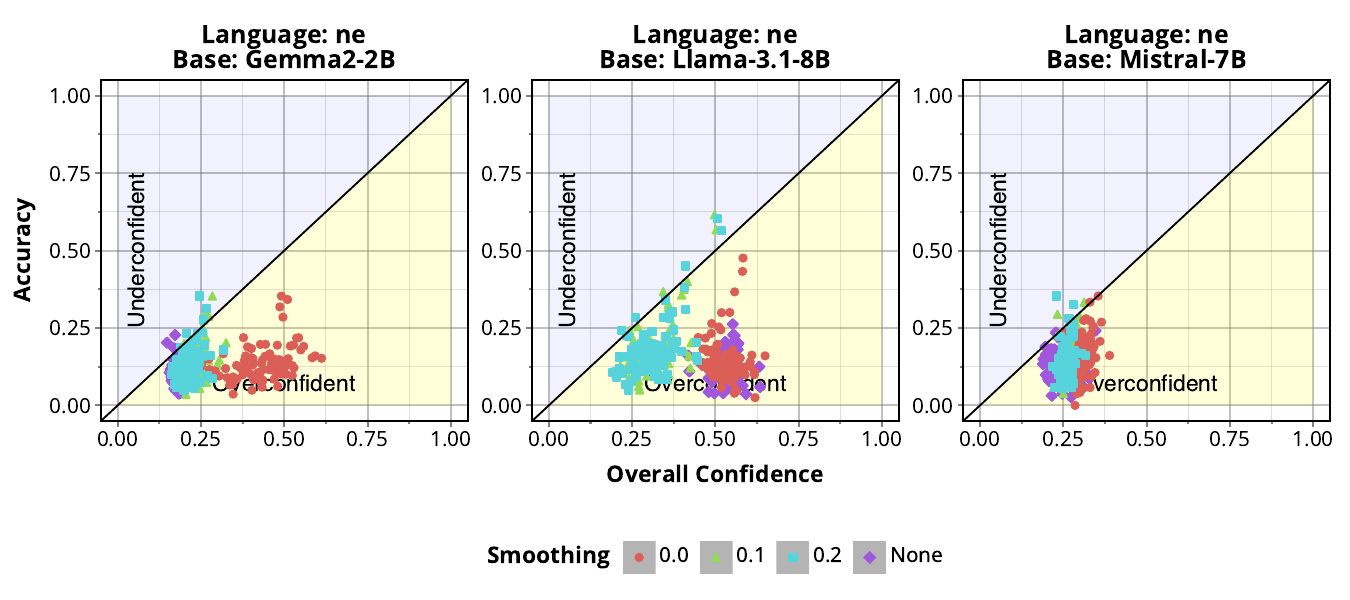}}
\caption{Reliability diagrams for the \textbf{\texttt{MMLU-ProX}} dataset for the \texttt{ne} language after instruction-tuning on the \textbf{\texttt{OpenHermes}} dataset.}\label{fig:mmluprox-OpenHermes-ne}\end{figure}

\begin{figure}[h!]\centering\resizebox{\linewidth}{!}{\includegraphics[width=\linewidth]{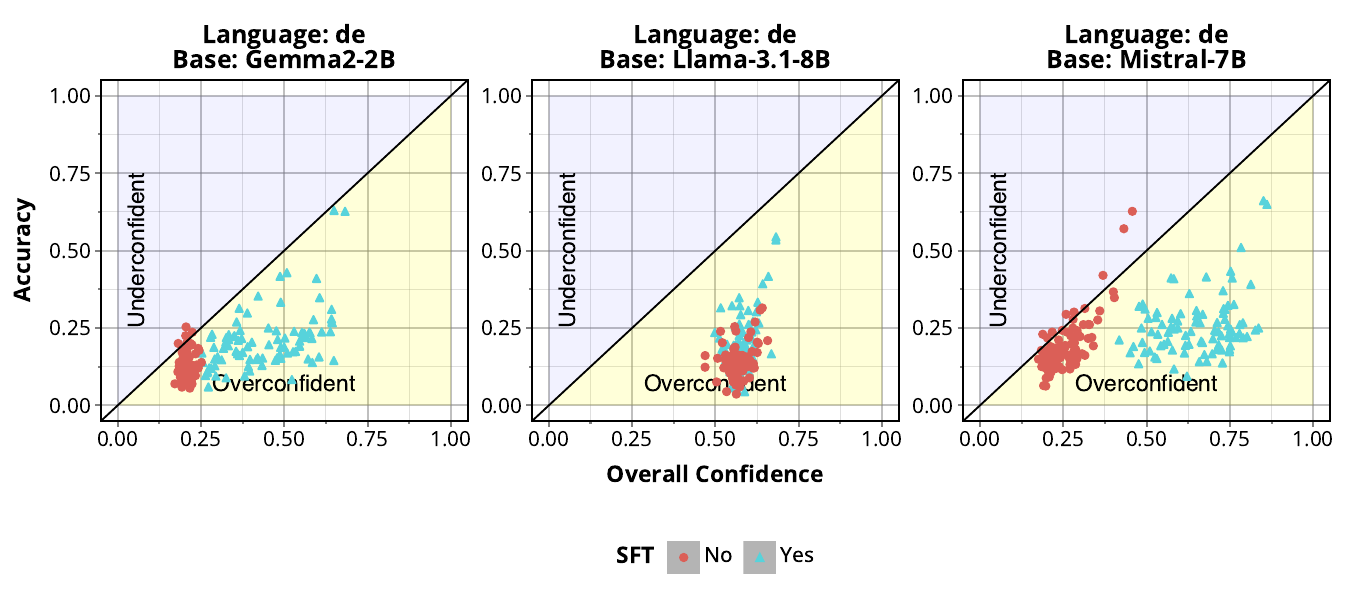}}
\caption{Reliability diagrams for the \textbf{\texttt{MMLU-ProX}} dataset for the \texttt{de} language.}\label{fig:mmluprox-base-de}\end{figure}
\begin{figure}[h!]\centering\resizebox{\linewidth}{!}{\includegraphics[width=\linewidth]{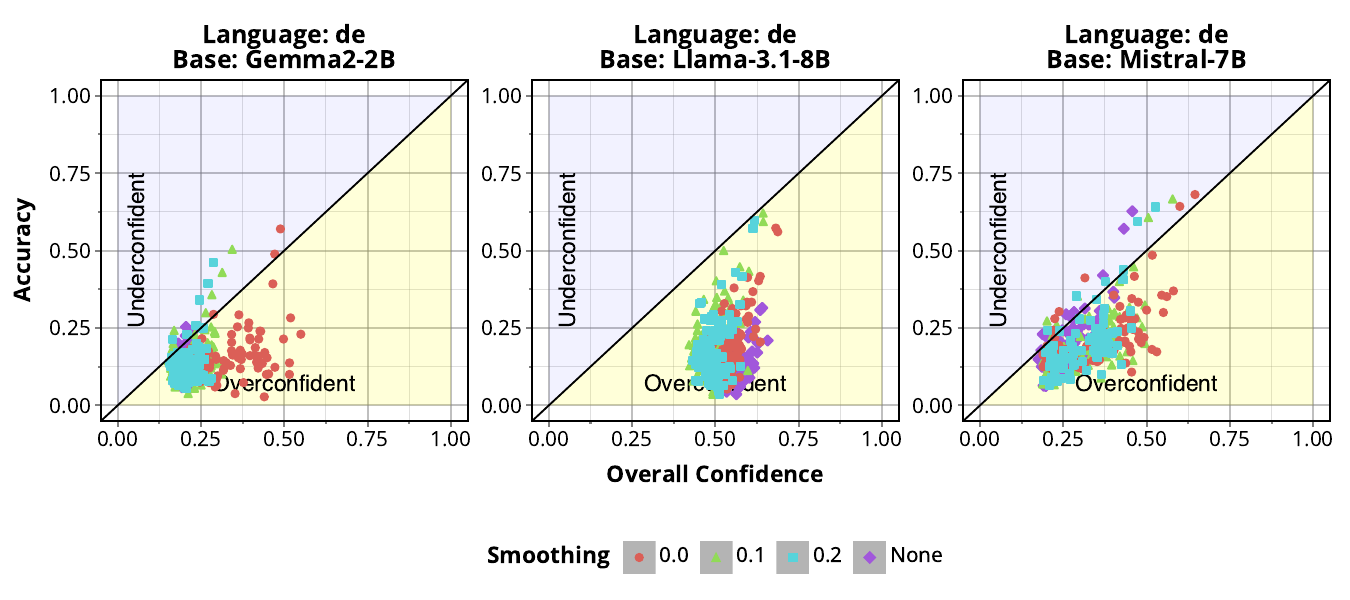}}
\caption{Reliability diagrams for the \textbf{\texttt{MMLU-ProX}} dataset for the \texttt{de} language after instruction-tuning on the \textbf{\texttt{Tulu3Mixture}} dataset.}\label{fig:mmluprox-Tulu3Mixture-de}\end{figure}
\begin{figure}[h!]\centering\resizebox{\linewidth}{!}{\includegraphics[width=\linewidth]{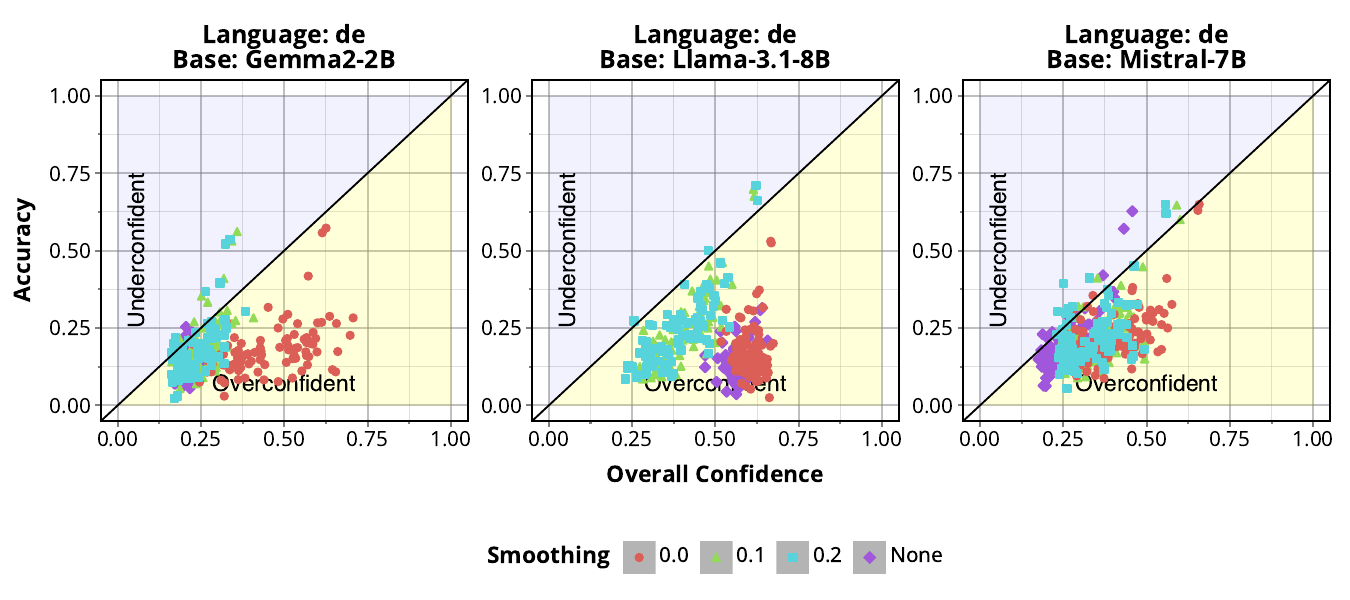}}
\caption{Reliability diagrams for the \textbf{\texttt{MMLU-ProX}} dataset for the \texttt{de} language after instruction-tuning on the \textbf{\texttt{OpenHermes}} dataset.}\label{fig:mmluprox-OpenHermes-de}\end{figure}

\begin{figure}[h!]\centering\resizebox{\linewidth}{!}{\includegraphics[width=\linewidth]{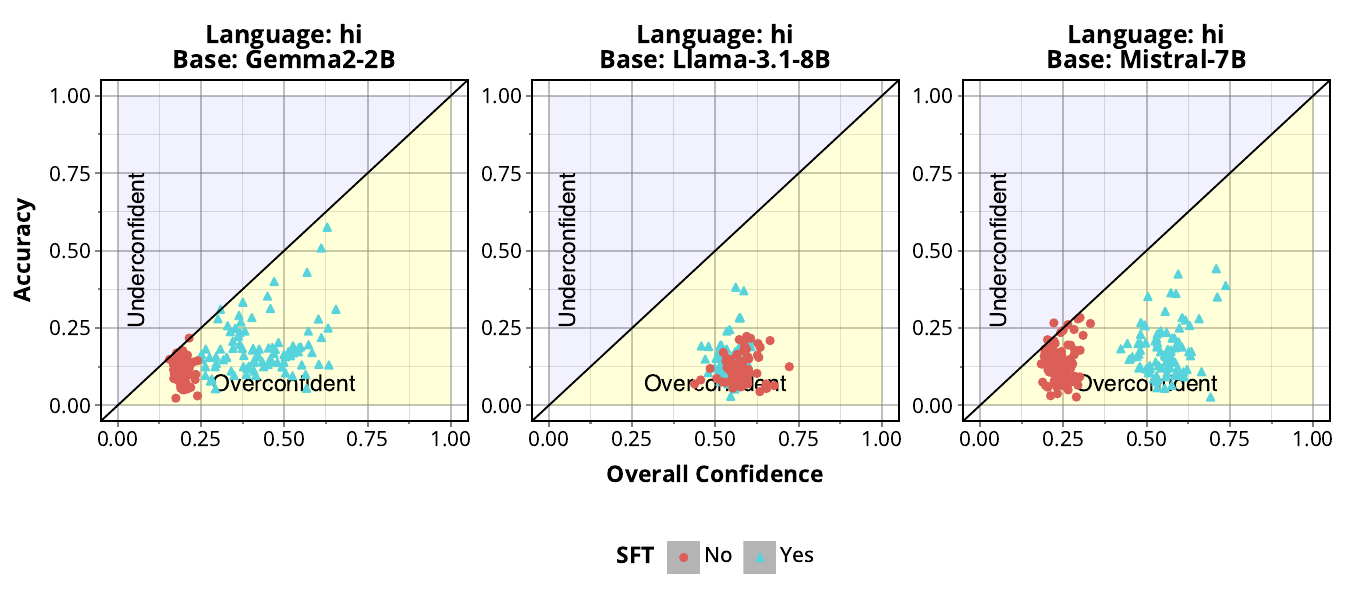}}
\caption{Reliability diagrams for the \textbf{\texttt{MMLU-ProX}} dataset for the \texttt{hi} language.}\label{fig:mmluprox-base-hi}\end{figure}
\begin{figure}[h!]\centering\resizebox{\linewidth}{!}{\includegraphics[width=\linewidth]{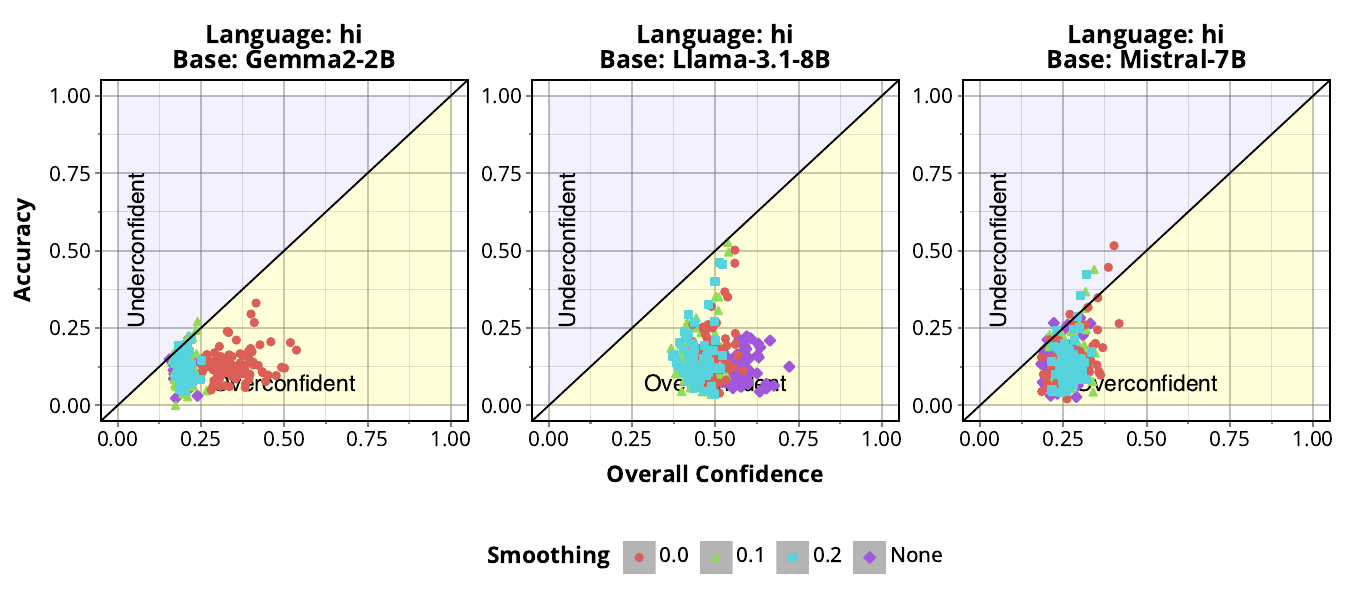}}
\caption{Reliability diagrams for the \textbf{\texttt{MMLU-ProX}} dataset for the \texttt{hi} language after instruction-tuning on the \textbf{\texttt{Tulu3Mixture}} dataset.}\label{fig:mmluprox-Tulu3Mixture-hi}\end{figure}
\begin{figure}[h!]\centering\resizebox{\linewidth}{!}{\includegraphics[width=\linewidth]{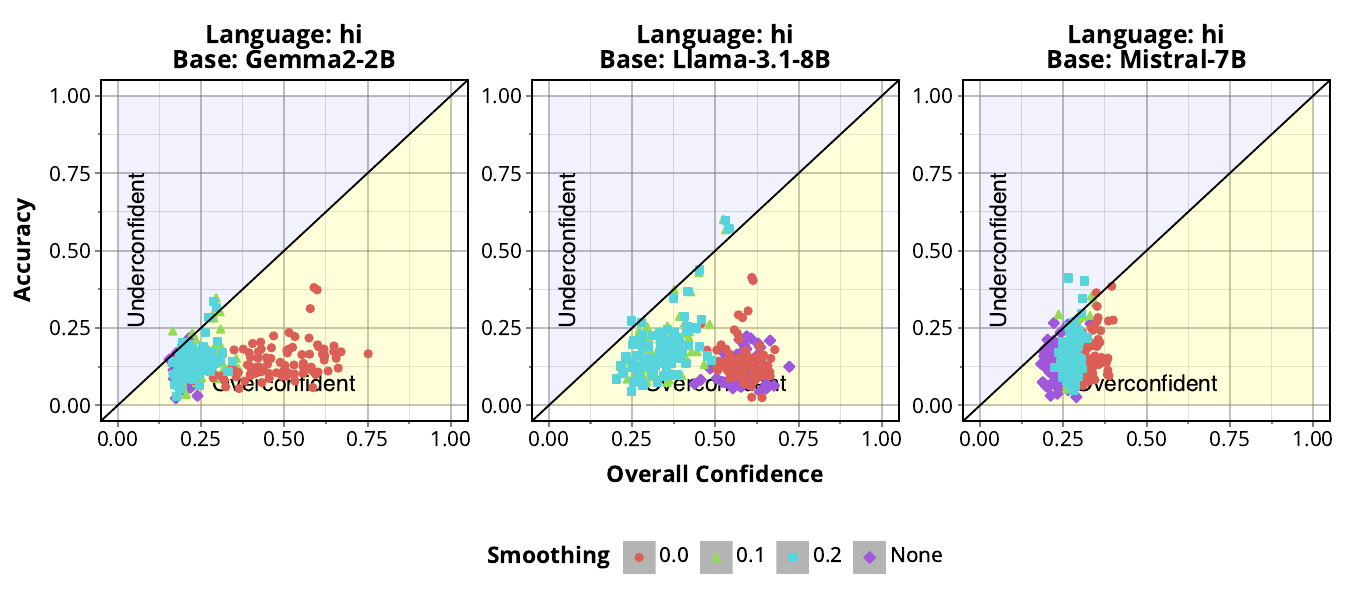}}
\caption{Reliability diagrams for the \textbf{\texttt{MMLU-ProX}} dataset for the \texttt{hi} language after instruction-tuning on the \textbf{\texttt{OpenHermes}} dataset.}\label{fig:mmluprox-OpenHermes-hi}\end{figure}

\begin{figure}[h!]\centering\resizebox{\linewidth}{!}{\includegraphics[width=\linewidth]{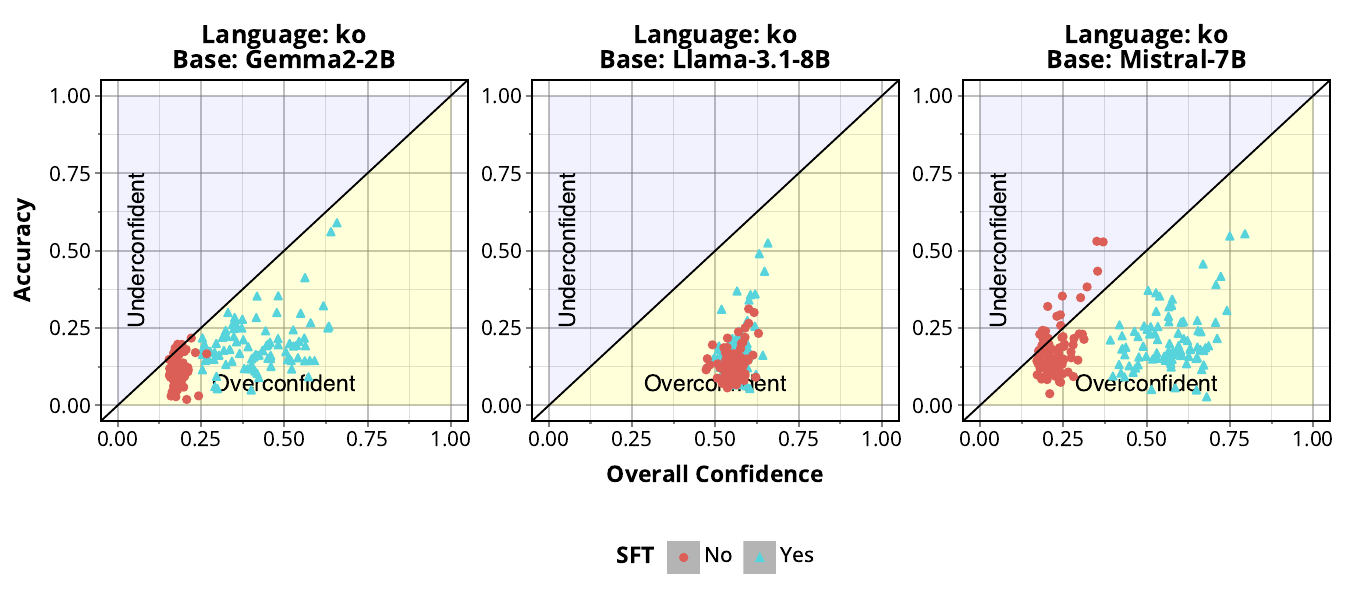}}
\caption{Reliability diagrams for the \textbf{\texttt{MMLU-ProX}} dataset for the \texttt{ko} language.}\label{fig:mmluprox-base-ko}\end{figure}
\begin{figure}[h!]\centering\resizebox{\linewidth}{!}{\includegraphics[width=\linewidth]{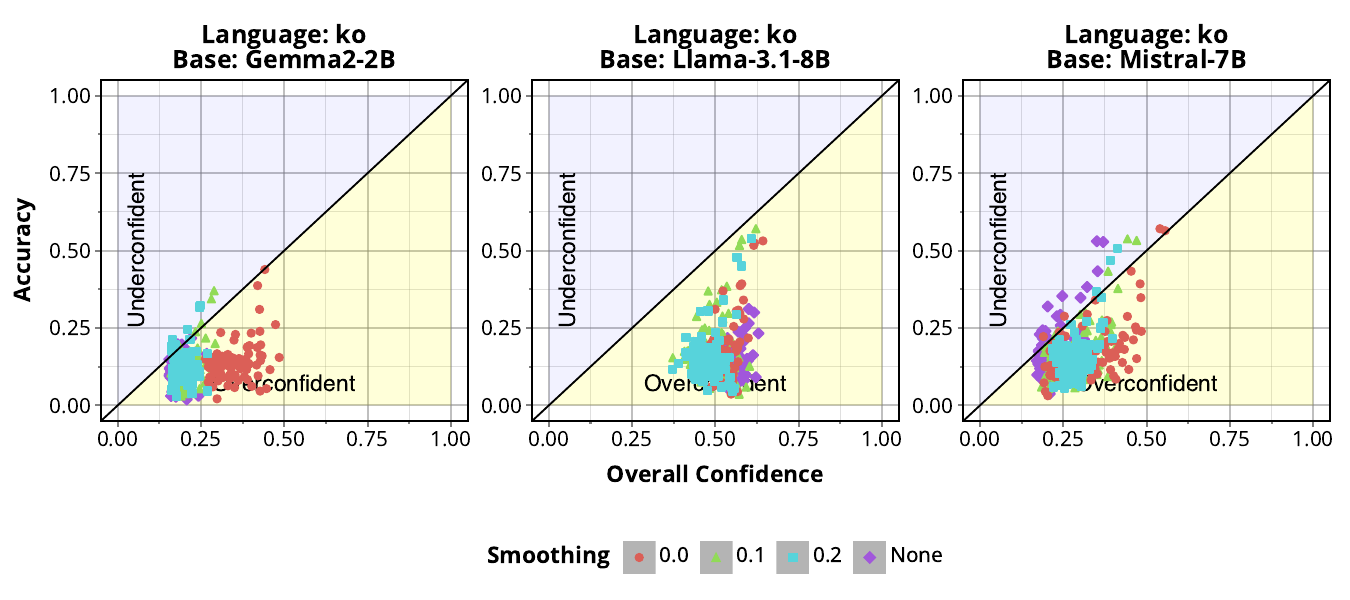}}
\caption{Reliability diagrams for the \textbf{\texttt{MMLU-ProX}} dataset for the \texttt{ko} language after instruction-tuning on the \textbf{\texttt{Tulu3Mixture}} dataset.}\label{fig:mmluprox-Tulu3Mixture-ko}\end{figure}
\begin{figure}[h!]\centering\resizebox{\linewidth}{!}{\includegraphics[width=\linewidth]{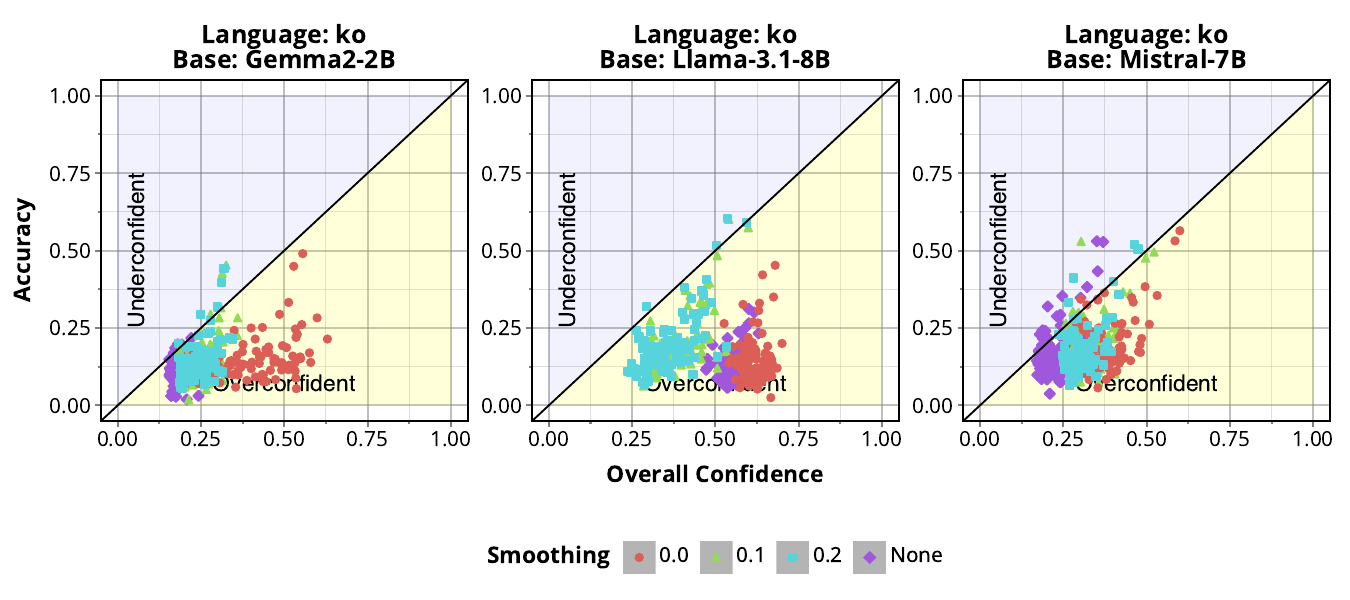}}
\caption{Reliability diagrams for the \textbf{\texttt{MMLU-ProX}} dataset for the \texttt{ko} language after instruction-tuning on the \textbf{\texttt{OpenHermes}} dataset.}\label{fig:mmluprox-OpenHermes-ko}\end{figure}

\begin{figure}[h!]\centering\resizebox{\linewidth}{!}{\includegraphics[width=\linewidth]{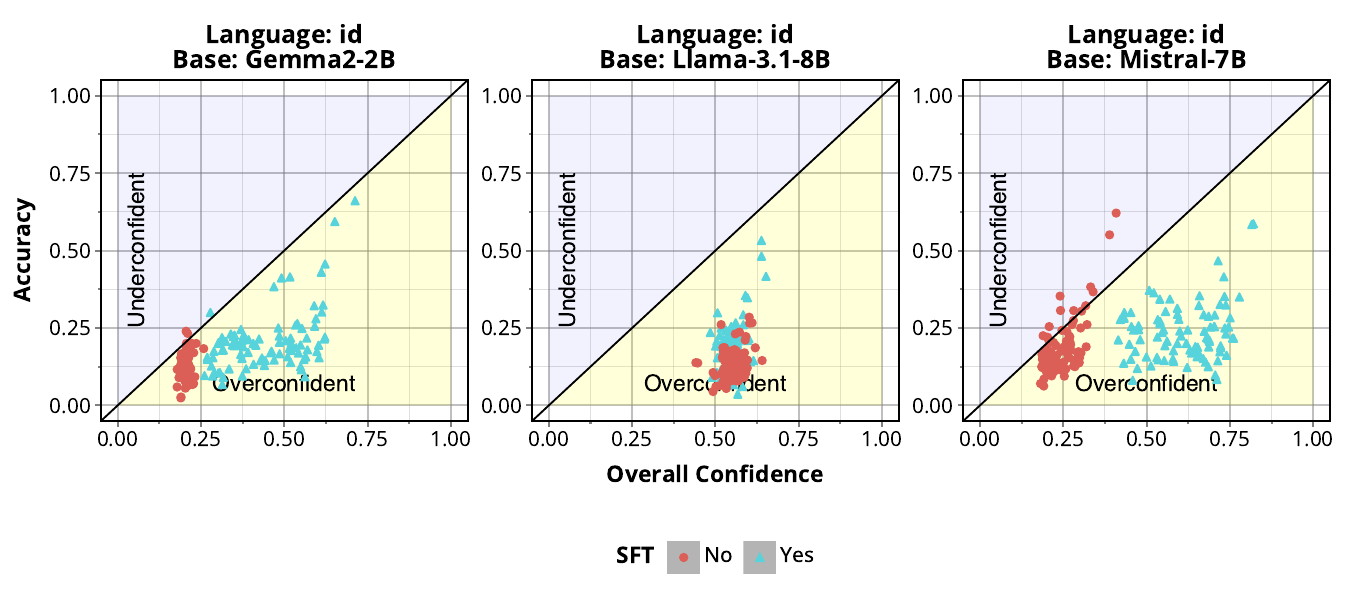}}
\caption{Reliability diagrams for the \textbf{\texttt{MMLU-ProX}} dataset for the \texttt{id} language.}\label{fig:mmluprox-base-id}\end{figure}
\begin{figure}[h!]\centering\resizebox{\linewidth}{!}{\includegraphics[width=\linewidth]{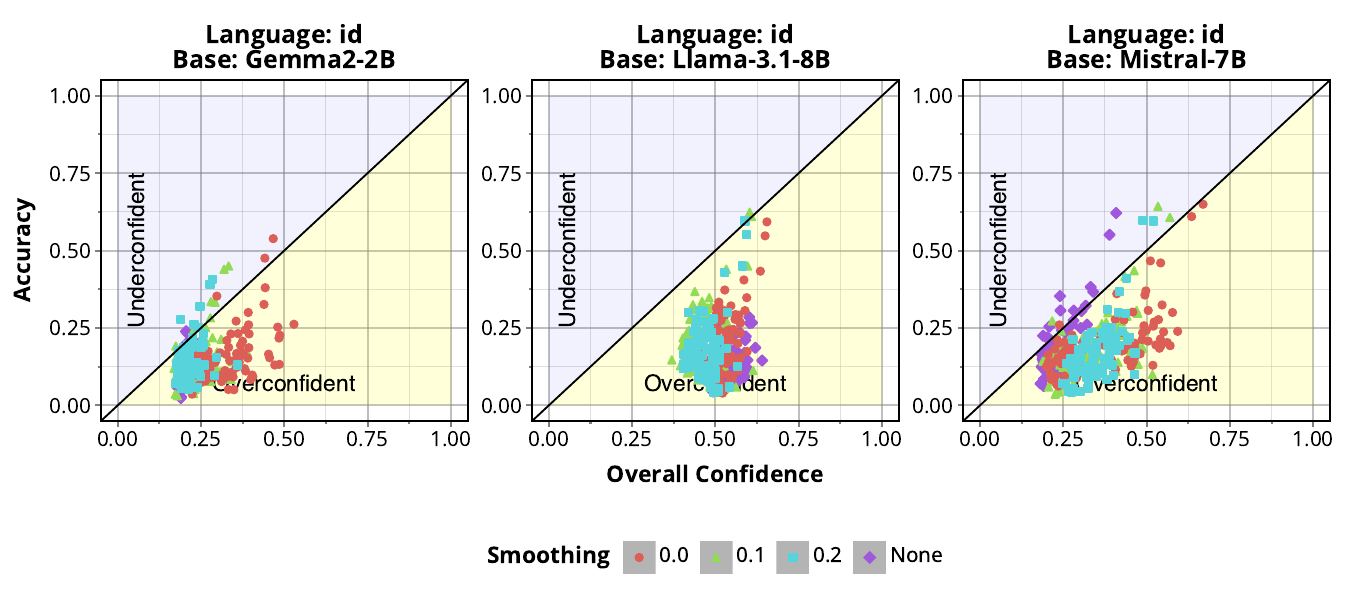}}
\caption{Reliability diagrams for the \textbf{\texttt{MMLU-ProX}} dataset for the \texttt{id} language after instruction-tuning on the \textbf{\texttt{Tulu3Mixture}} dataset.}\label{fig:mmluprox-Tulu3Mixture-id}\end{figure}
\begin{figure}[h!]\centering\resizebox{\linewidth}{!}{\includegraphics[width=\linewidth]{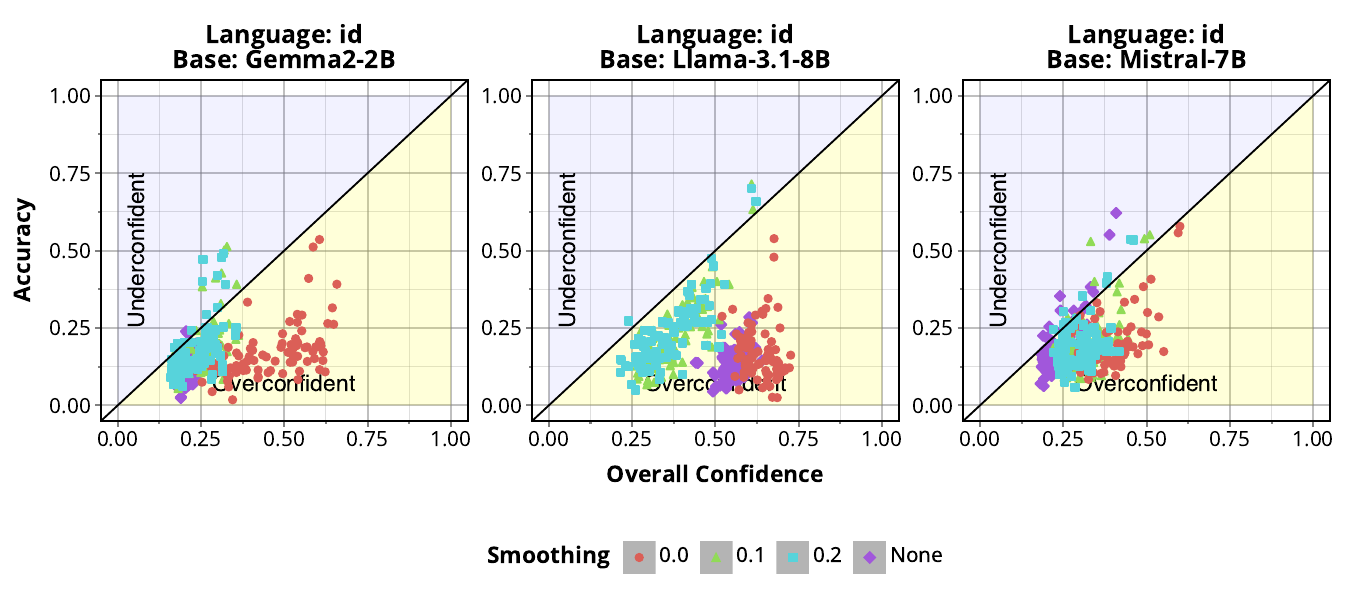}}
\caption{Reliability diagrams for the \textbf{\texttt{MMLU-ProX}} dataset for the \texttt{id} language after instruction-tuning on the \textbf{\texttt{OpenHermes}} dataset.}\label{fig:mmluprox-OpenHermes-id}\end{figure}

\begin{figure}[h!]\centering\resizebox{\linewidth}{!}{\includegraphics[width=\linewidth]{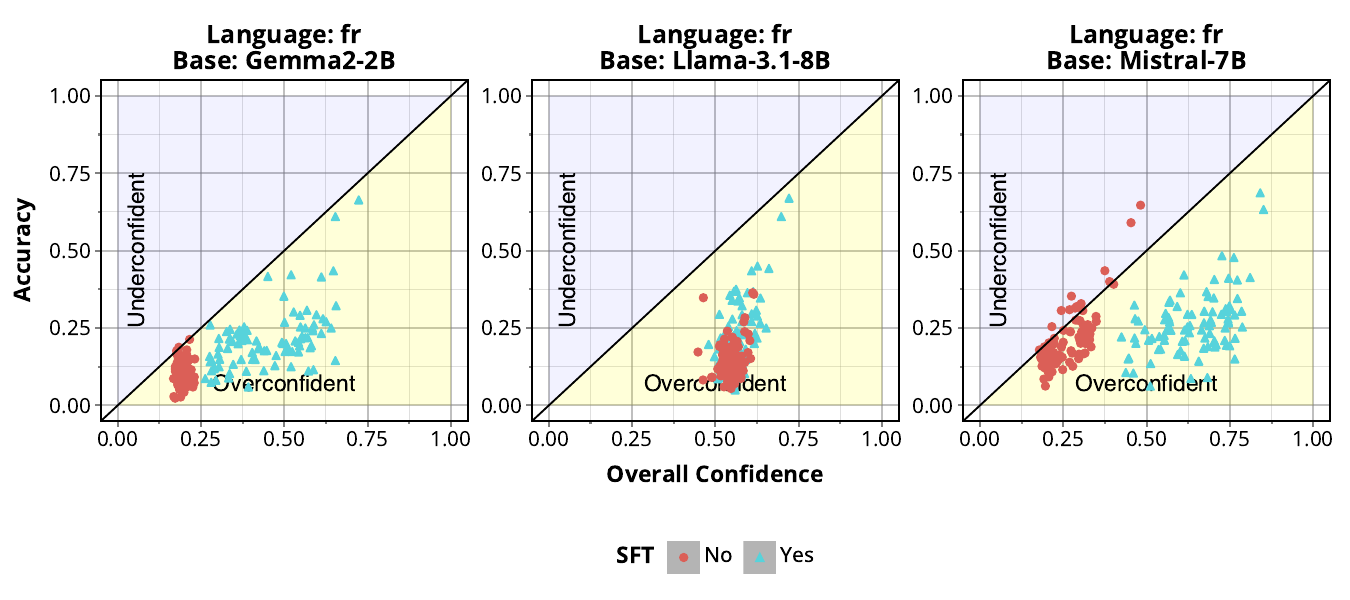}}
\caption{Reliability diagrams for the \textbf{\texttt{MMLU-ProX}} dataset for the \texttt{fr} language.}\label{fig:mmluprox-base-fr}\end{figure}
\begin{figure}[h!]\centering\resizebox{\linewidth}{!}{\includegraphics[width=\linewidth]{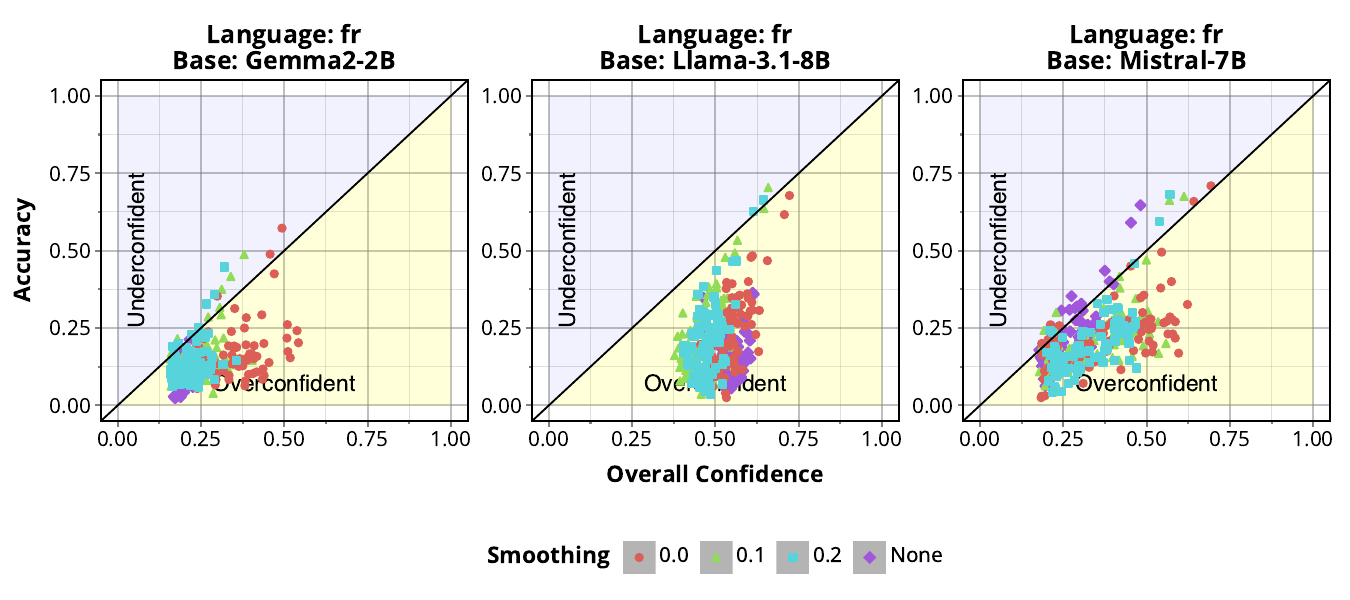}}
\caption{Reliability diagrams for the \textbf{\texttt{MMLU-ProX}} dataset for the \texttt{fr} language after instruction-tuning on the \textbf{\texttt{Tulu3Mixture}} dataset.}\label{fig:mmluprox-Tulu3Mixture-fr}\end{figure}
\begin{figure}[h!]\centering\resizebox{\linewidth}{!}{\includegraphics[width=\linewidth]{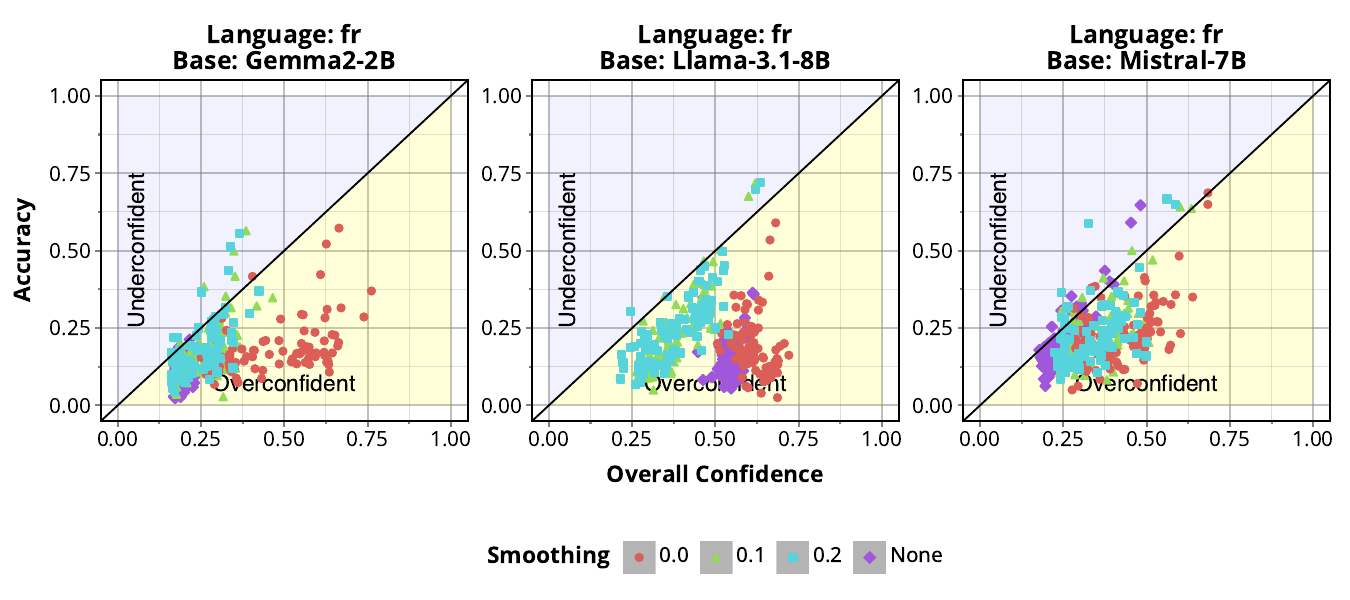}}
\caption{Reliability diagrams for the \textbf{\texttt{MMLU-ProX}} dataset for the \texttt{fr} language after instruction-tuning on the \textbf{\texttt{OpenHermes}} dataset.}\label{fig:mmluprox-OpenHermes-fr}\end{figure}

\begin{figure}[h!]\centering\resizebox{\linewidth}{!}{\includegraphics[width=\linewidth]{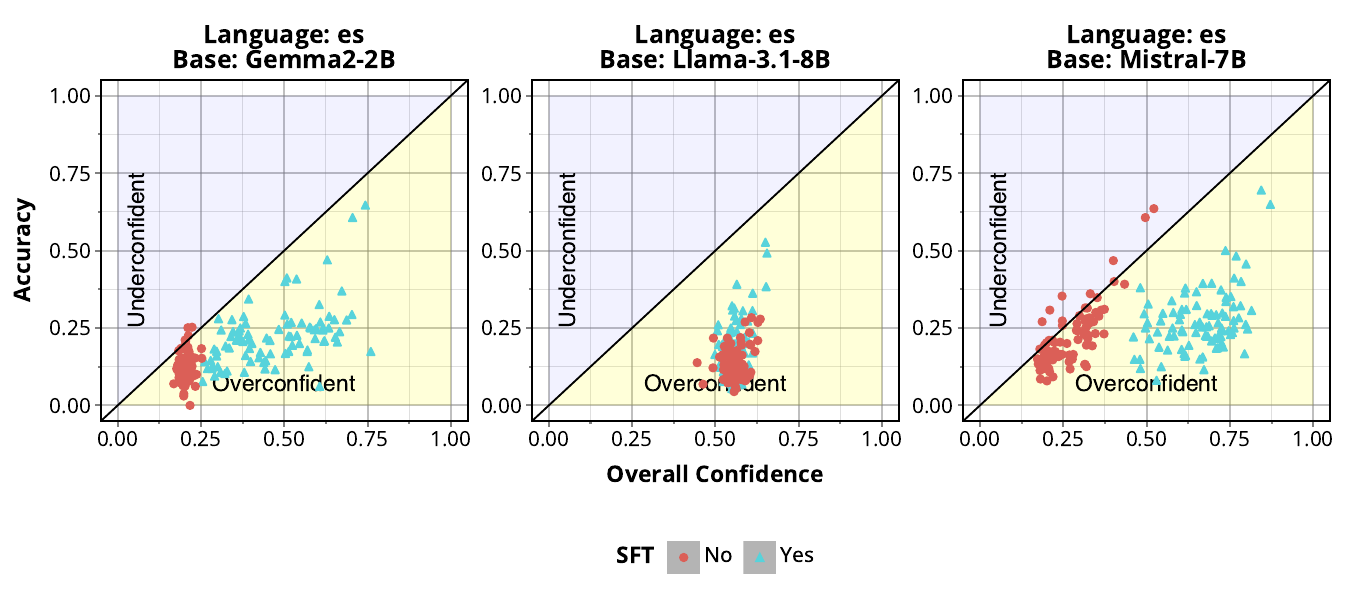}}
\caption{Reliability diagrams for the \textbf{\texttt{MMLU-ProX}} dataset for the \texttt{es} language.}\label{fig:mmluprox-base-es}\end{figure}
\begin{figure}[h!]\centering\resizebox{\linewidth}{!}{\includegraphics[width=\linewidth]{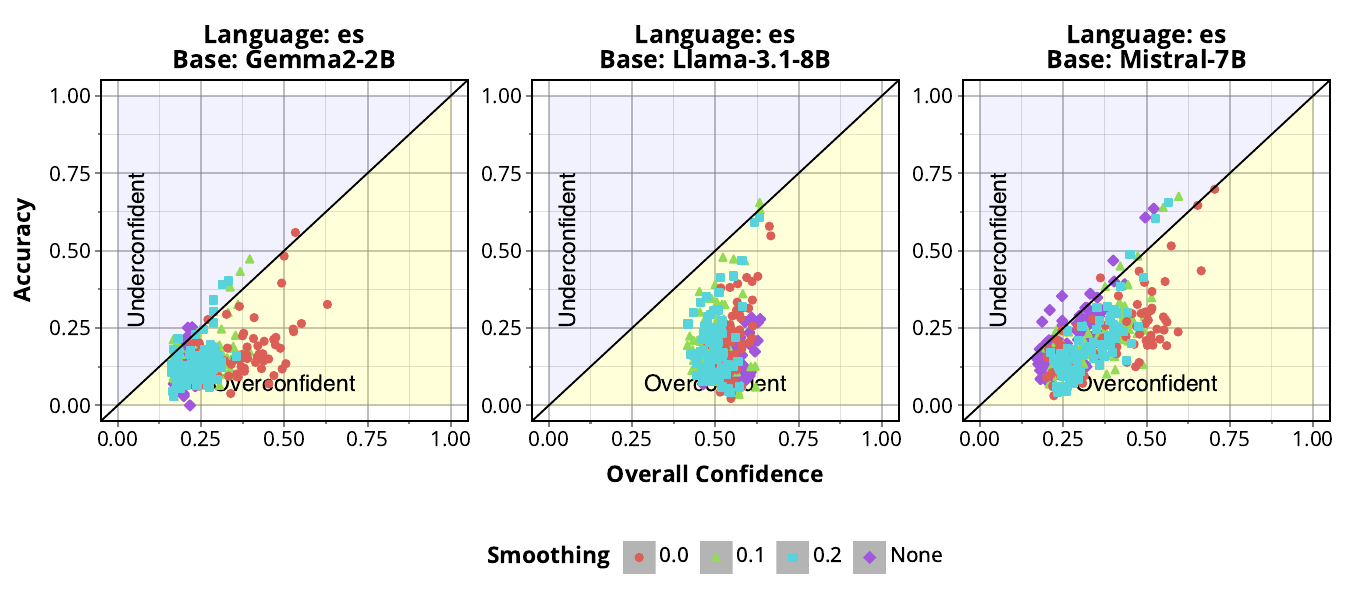}}
\caption{Reliability diagrams for the \textbf{\texttt{MMLU-ProX}} dataset for the \texttt{es} language after instruction-tuning on the \textbf{\texttt{Tulu3Mixture}} dataset.}\label{fig:mmluprox-Tulu3Mixture-es}\end{figure}
\begin{figure}[h!]\centering\resizebox{\linewidth}{!}{\includegraphics[width=\linewidth]{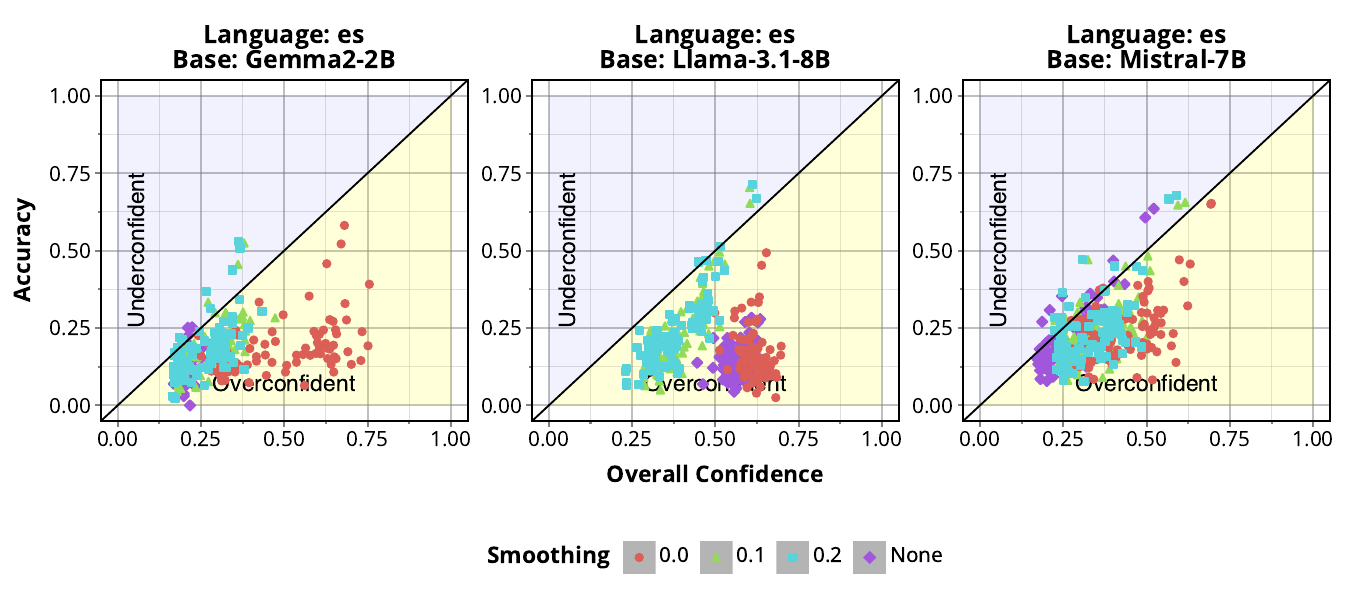}}
\caption{Reliability diagrams for the \textbf{\texttt{MMLU-ProX}} dataset for the \texttt{es} language after instruction-tuning on the \textbf{\texttt{OpenHermes}} dataset.}\label{fig:mmluprox-OpenHermes-es}\end{figure}

\begin{figure}[h!]\centering\resizebox{\linewidth}{!}{\includegraphics[width=\linewidth]{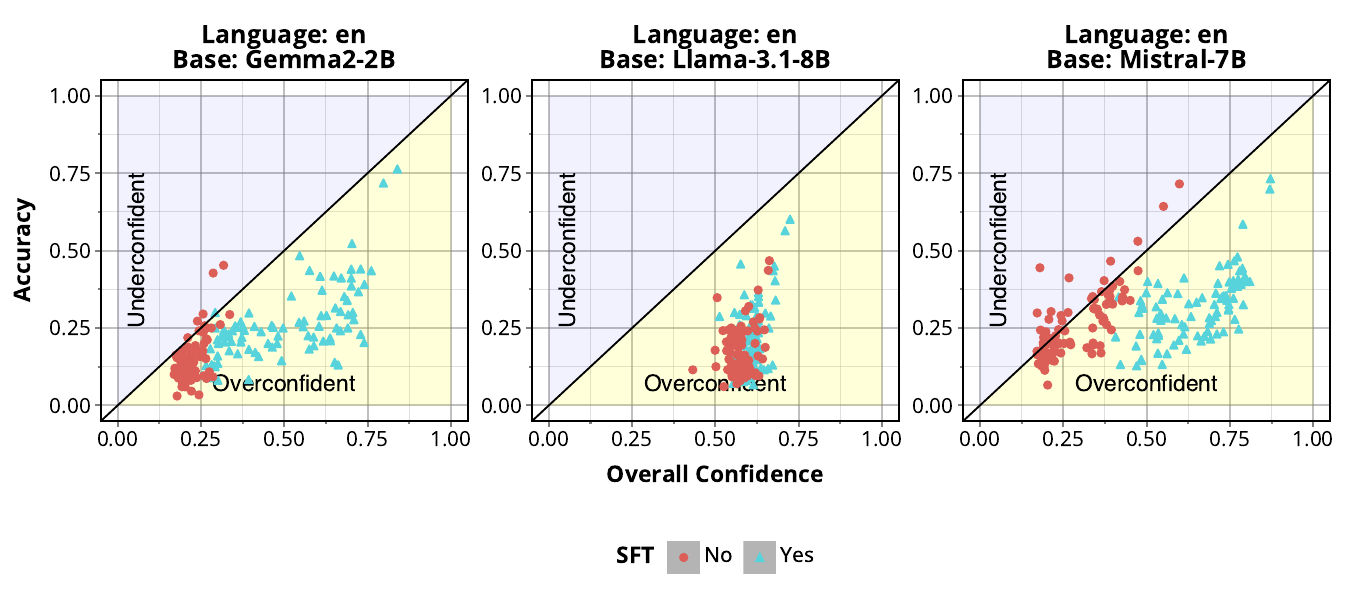}}
\caption{Reliability diagrams for the \textbf{\texttt{MMLU-ProX}} dataset for the \texttt{en} language.}\label{fig:mmluprox-base-en}\end{figure}
\begin{figure}[h!]\centering\resizebox{\linewidth}{!}{\includegraphics[width=\linewidth]{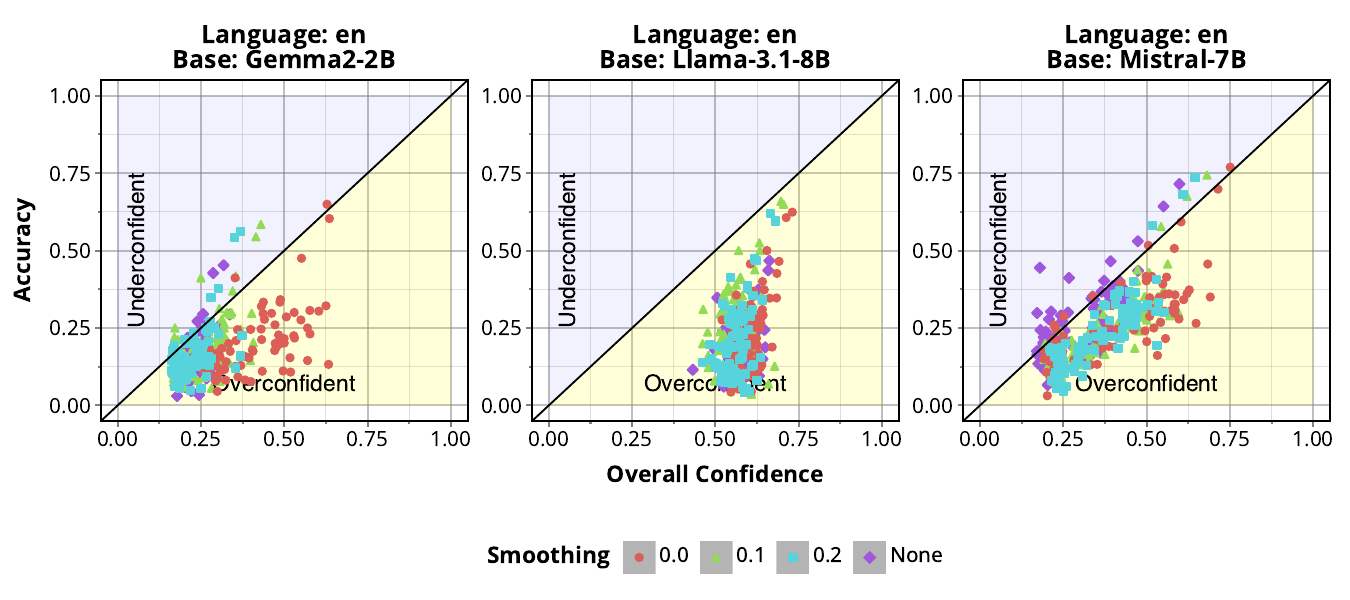}}
\caption{Reliability diagrams for the \textbf{\texttt{MMLU-ProX}} dataset for the \texttt{en} language after instruction-tuning on the \textbf{\texttt{Tulu3Mixture}} dataset.}\label{fig:mmluprox-Tulu3Mixture-en}\end{figure}
\begin{figure}[h!]\centering\resizebox{\linewidth}{!}{\includegraphics[width=\linewidth]{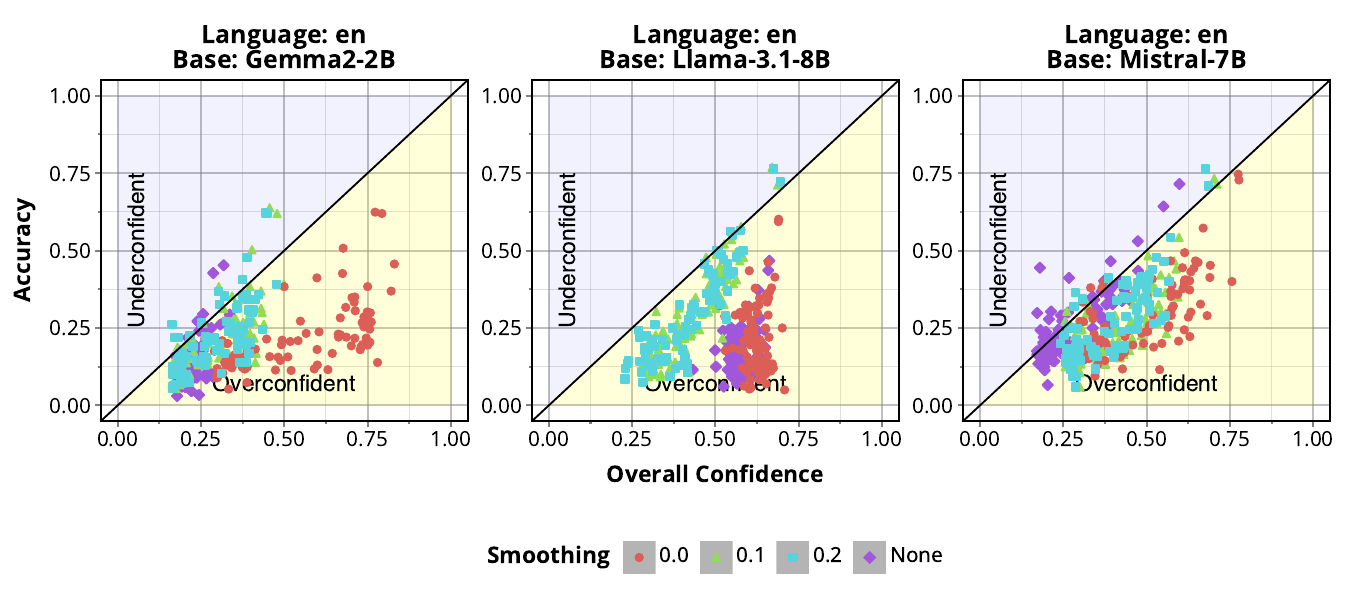}}
\caption{Reliability diagrams for the \textbf{\texttt{MMLU-ProX}} dataset for the \texttt{en} language after instruction-tuning on the \textbf{\texttt{OpenHermes}} dataset.}\label{fig:mmluprox-OpenHermes-en}\end{figure}

\begin{figure}[h!]\centering\resizebox{\linewidth}{!}{\includegraphics[width=\linewidth]{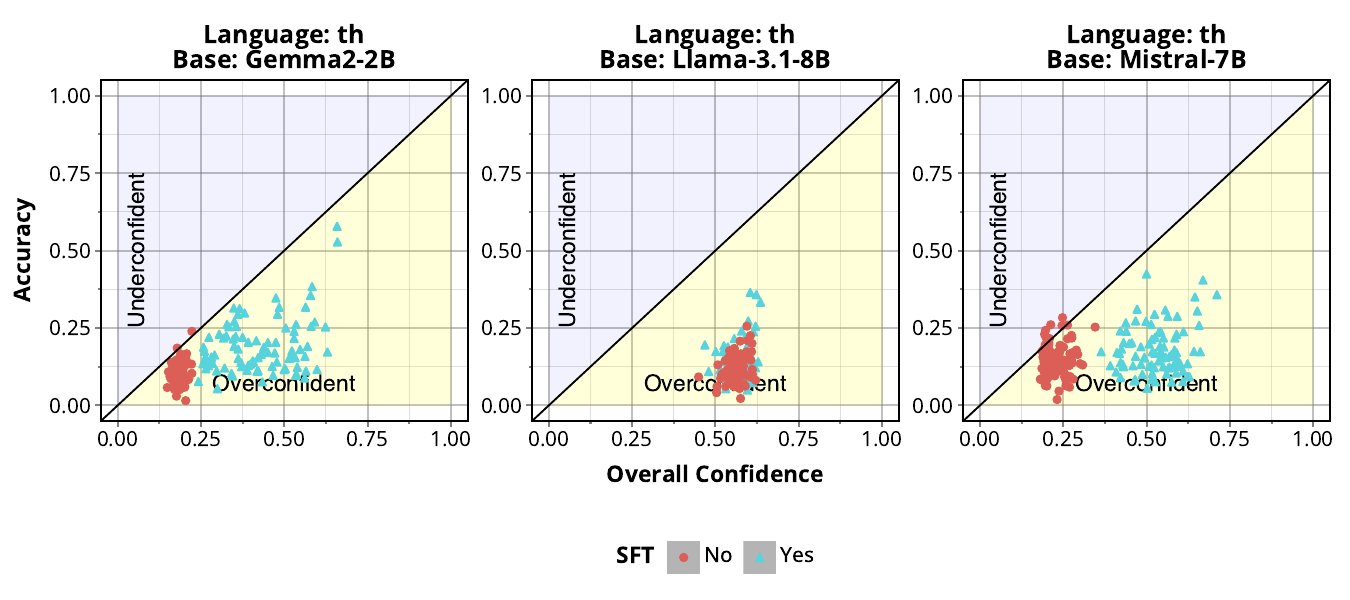}}
\caption{Reliability diagrams for the \textbf{\texttt{MMLU-ProX}} dataset for the \texttt{th} language.}\label{fig:mmluprox-base-th}\end{figure}
\begin{figure}[h!]\centering\resizebox{\linewidth}{!}{\includegraphics[width=\linewidth]{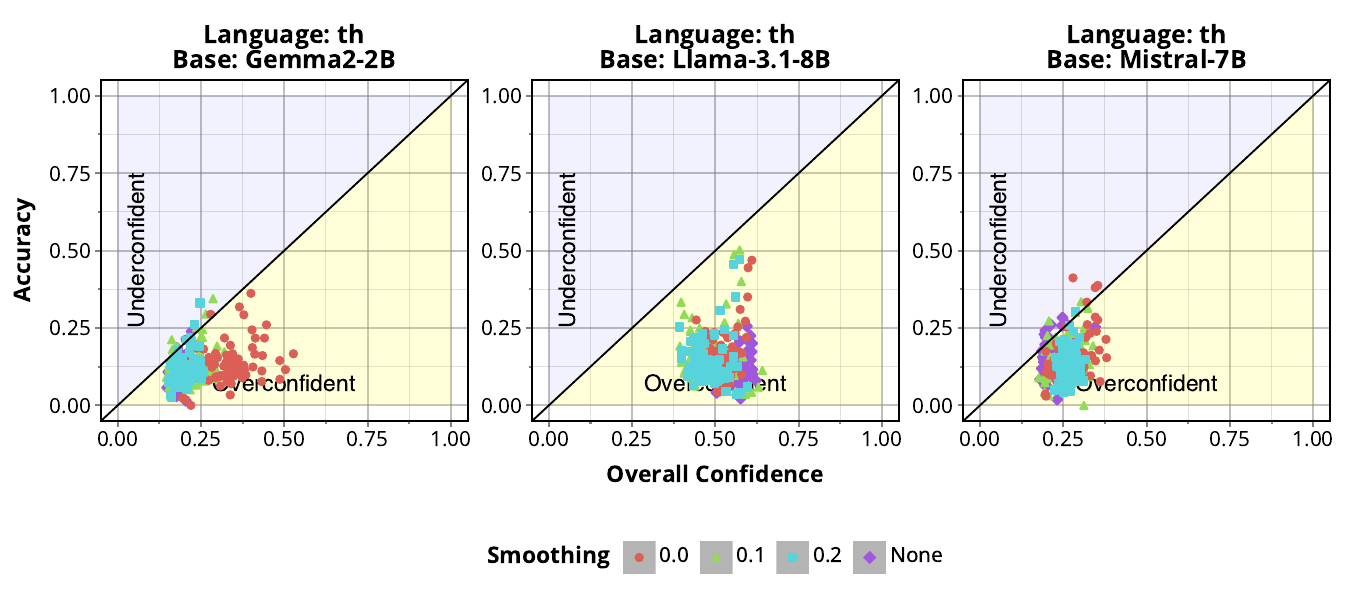}}
\caption{Reliability diagrams for the \textbf{\texttt{MMLU-ProX}} dataset for the \texttt{th} language after instruction-tuning on the \textbf{\texttt{Tulu3Mixture}} dataset.}\label{fig:mmluprox-Tulu3Mixture-th}\end{figure}
\begin{figure}[h!]\centering\resizebox{\linewidth}{!}{\includegraphics[width=\linewidth]{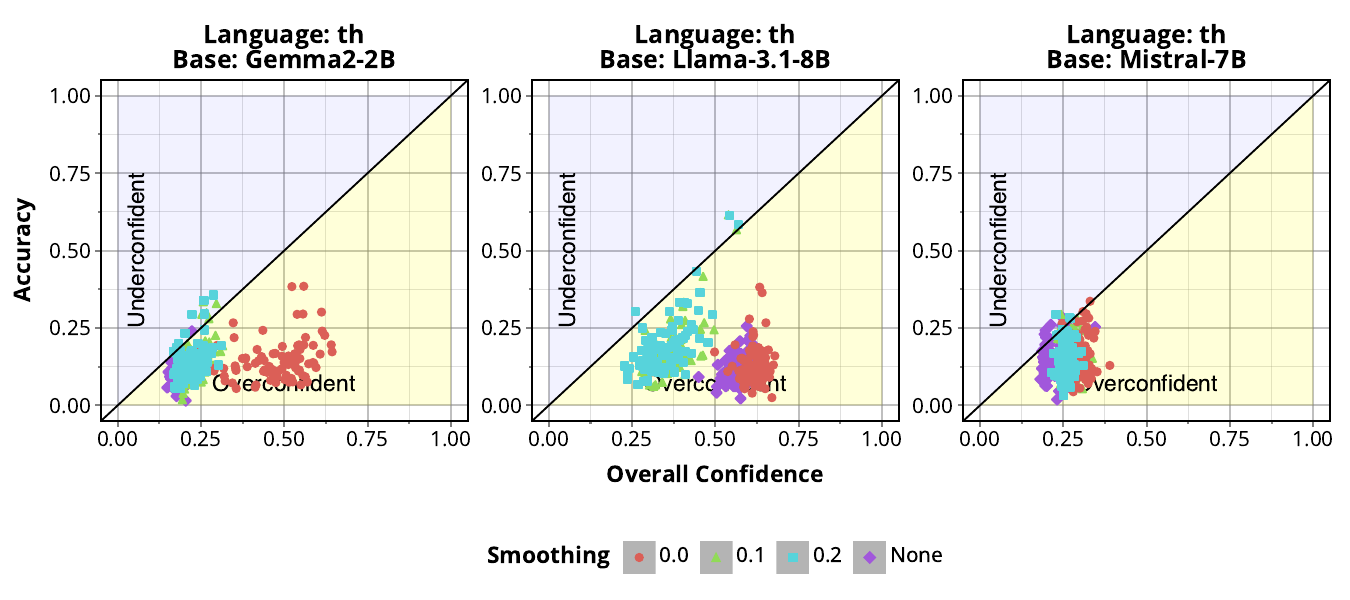}}
\caption{Reliability diagrams for the \textbf{\texttt{MMLU-ProX}} dataset for the \texttt{th} language after instruction-tuning on the \textbf{\texttt{OpenHermes}} dataset.}\label{fig:mmluprox-OpenHermes-th}\end{figure}

\begin{figure}[h!]\centering\resizebox{\linewidth}{!}{\includegraphics[width=\linewidth]{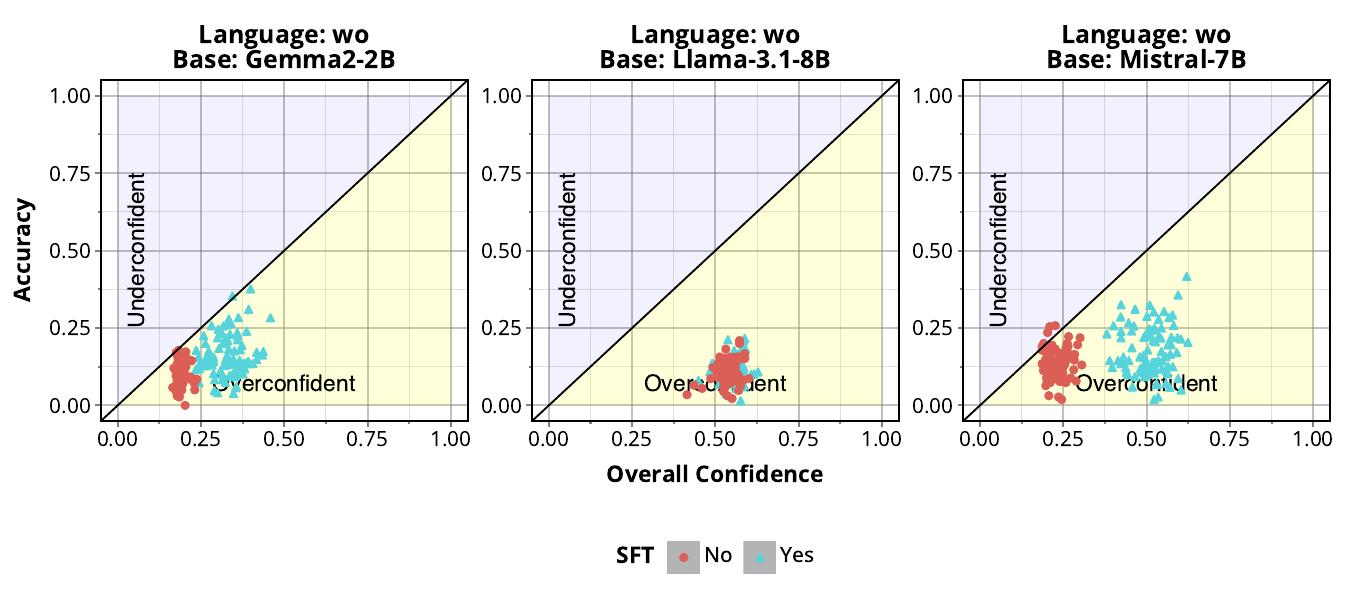}}
\caption{Reliability diagrams for the \textbf{\texttt{MMLU-ProX}} dataset for the \texttt{wo} language.}\label{fig:mmluprox-base-wo}\end{figure}
\begin{figure}[h!]\centering\resizebox{\linewidth}{!}{\includegraphics[width=\linewidth]{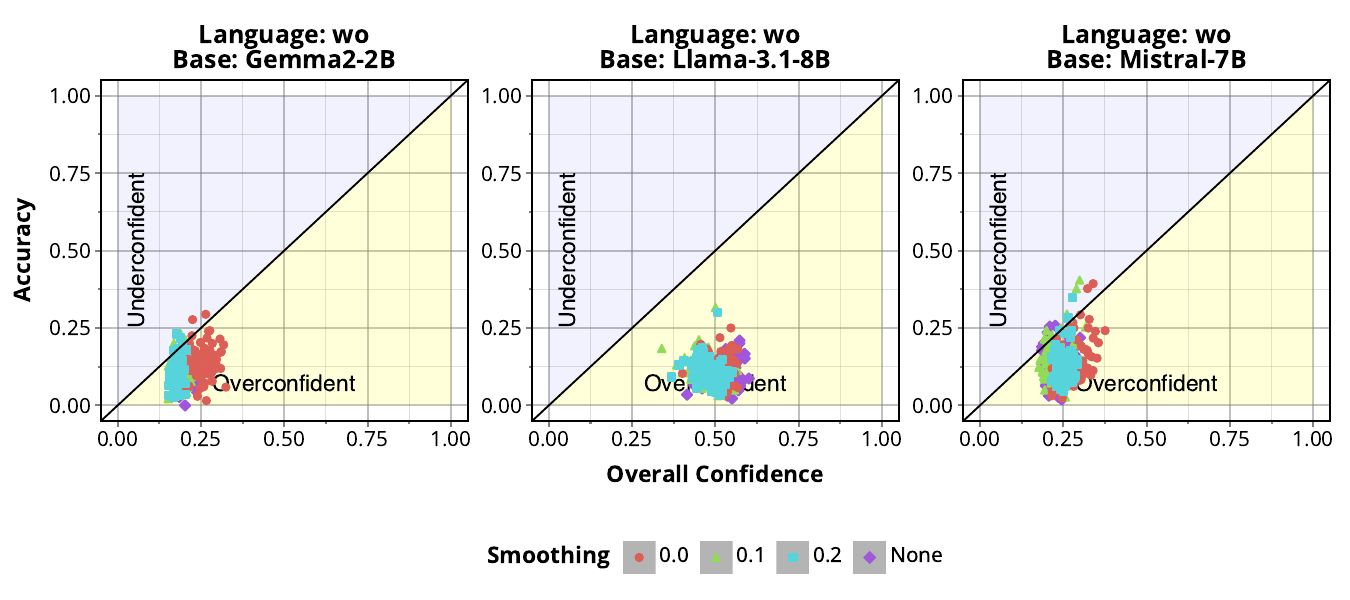}}
\caption{Reliability diagrams for the \textbf{\texttt{MMLU-ProX}} dataset for the \texttt{wo} language after instruction-tuning on the \textbf{\texttt{Tulu3Mixture}} dataset.}\label{fig:mmluprox-Tulu3Mixture-wo}\end{figure}
\begin{figure}[h!]\centering\resizebox{\linewidth}{!}{\includegraphics[width=\linewidth]{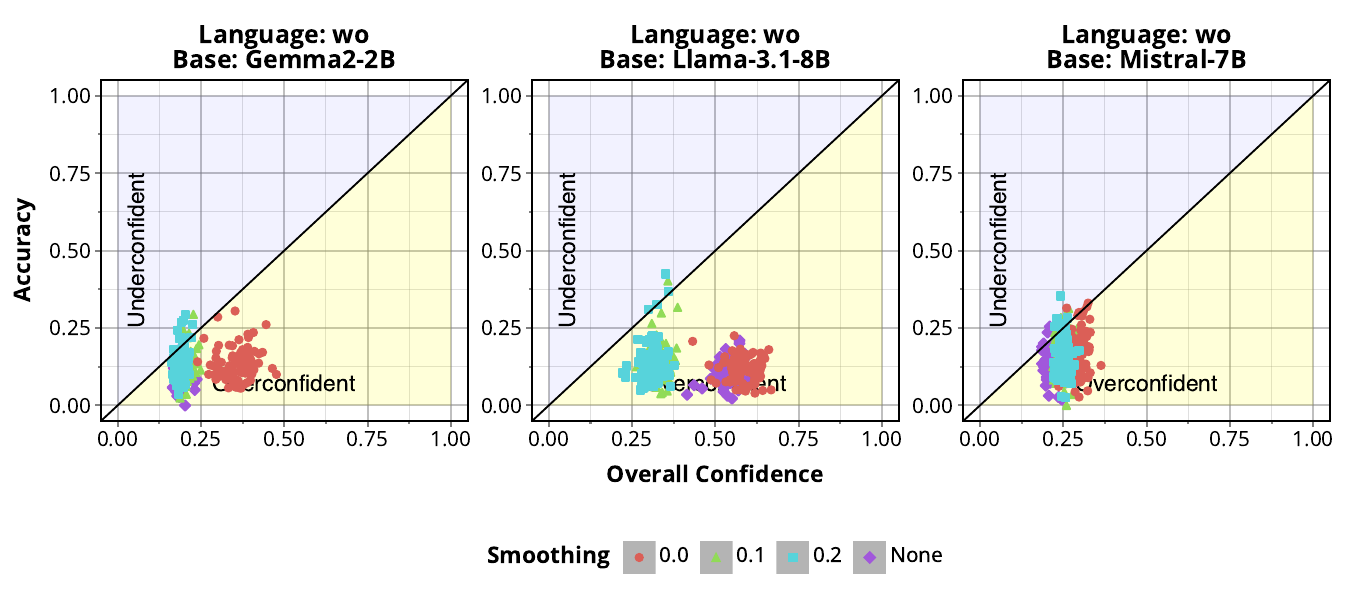}}
\caption{Reliability diagrams for the \textbf{\texttt{MMLU-ProX}} dataset for the \texttt{wo} language after instruction-tuning on the \textbf{\texttt{OpenHermes}} dataset.}\label{fig:mmluprox-OpenHermes-wo}\end{figure}

\end{document}